\documentclass{article}

\PassOptionsToPackage{numbers, compress}{natbib}

\usepackage[final]{neurips_2022}

\usepackage[utf8]{inputenc} 
\usepackage[T1]{fontenc}    
\usepackage{hyperref}       
\usepackage{url}            
\usepackage{booktabs}       
\usepackage{amsfonts}       
\usepackage{nicefrac}       
\usepackage{microtype}      
\usepackage{xcolor}         
 \usepackage{setspace}
\usepackage{bm}
\usepackage{amsmath}
\usepackage{amssymb}
\usepackage{subfigure}
\usepackage{bm}
\usepackage{multicol}
\usepackage{graphicx}
\usepackage{xcolor}
\usepackage{hyperref}
\usepackage{float}
\usepackage{natbib}
\usepackage{listings}
\usepackage{empheq}
\newcommand*\widefbox[1]{\fbox{\hspace{2em}#1\hspace{2em}}}

\usepackage[linesnumbered,ruled]{algorithm2e}

\usepackage{enumitem}
\setitemize{itemsep=2pt,topsep=0pt,parsep=0pt,partopsep=0pt,leftmargin=*}

\setenumerate{itemsep=2pt,topsep=0pt,parsep=0pt,partopsep=0pt,leftmargin=*}

\def\x{\bm x}

\def\k{\bm k}

\def\r{\bm r}
\def\y{\mathbf y}

\def\r{\mathbf r}
\def\w{\bm w}
\def\A{\bm A}

\def\B{\bm B}
\def\C{\bm C}
\def\Y{\bm Y}
\def\b{\bm b}
\def\Q{\bm Q}
\def\G{\bm G}

\def\K{\bm K}
\def\H{\bm H}
\def\X{\bm X}

\def\u{\bm u}

\def\I{\mathbf I}

\def\k{\bm k}
\def\a{\bm a}
\def\v{\bm v}
\def\z{\bm z}
\def\W{\bm W}
\def\f{\bm f}
\def\h{\bm h}

\def\r{\bm r}

\def\y{\bm y}
\def\C{\bm C}
\def\D{\bm D}
\def\g{\bm g}
\def\j{\bm j}

\title{Self-Consistent Dynamical Field Theory of Kernel Evolution in Wide Neural Networks}

\author{%
  Blake Bordelon 
    \  \& \  
    Cengiz Pehlevan \\
  John Paulson School of Engineering and Applied Sciences, Center for Brain Science
  \\
  Harvard University
  \\
  Cambridge MA, 02138 \\
  \texttt{blake\_bordelon@g.harvard.edu}, \texttt{cpehlevan@g.harvard.edu}  \\
}

\definecolor{codegreen}{rgb}{0,0.6,0}
\definecolor{codegray}{rgb}{0.5,0.5,0.5}
\definecolor{codepurple}{rgb}{0.58,0,0.82}
\definecolor{backcolour}{rgb}{0.95,0.95,0.92}
\lstdefinestyle{mystyle}{
    backgroundcolor=\color{backcolour},   
    commentstyle=\color{codegreen},
    keywordstyle=\color{magenta},
    numberstyle=\tiny\color{codegray},
    stringstyle=\color{codepurple},
    basicstyle=\ttfamily\footnotesize,
    breakatwhitespace=false,         
    breaklines=true,                 
    captionpos=b,                    
    keepspaces=true,                 
    numbers=left,                    
    numbersep=5pt,                  
    showspaces=false,                
    showstringspaces=false,
    showtabs=false,                  
    tabsize=2
}

\begin{document}

\maketitle

\begin{abstract}
  We analyze feature learning in infinite-width neural networks trained with gradient flow through a self-consistent dynamical field theory. We construct a collection of deterministic dynamical order parameters which are inner-product kernels for hidden unit activations and gradients in each layer at pairs of time points, providing a reduced description of network activity through training. These kernel order parameters collectively define the hidden layer activation distribution, the evolution of the neural tangent kernel, and consequently output predictions. We show that the field theory derivation recovers the recursive stochastic process of infinite-width feature learning networks obtained from Yang \& Hu with Tensor Programs \cite{yang2021tensor}. For deep linear networks, these kernels satisfy a set of algebraic matrix equations. For nonlinear networks, we provide an alternating sampling procedure to self-consistently solve for the kernel order parameters. 
  We provide comparisons of the self-consistent solution to various approximation schemes including the static NTK approximation, gradient independence assumption, and leading order perturbation theory, showing that each of these approximations can break down in regimes where general self-consistent solutions still provide an accurate description. Lastly, we provide experiments in more realistic settings which demonstrate that the loss and kernel dynamics of CNNs at fixed feature learning strength is preserved across different widths on a CIFAR classification task.
\end{abstract}

\section{Introduction}

Deep learning has emerged as a successful paradigm for solving challenging machine learning and computational problems across a variety of domains \cite{goodfellow2016deep, lecun2015deep}. However, theoretical understanding of the training and generalization of modern deep learning methods lags behind current practice. Ideally, a theory of deep learning would be analytically tractable, efficiently computable, capable of predicting network performance and internal features that the network learns, and interpretable through a reduced description involving desirably initialization-independent quantities.

Several recent theoretical advances have fruitfully considered the idealization of \textit{wide neural networks}, where the number of hidden units in each layer is taken to be large. Under certain parameterization, Bayesian neural networks and gradient descent trained networks converge to gaussian processes (NNGPs) \citep{ neal2012bayesian,lee2018deep,matthews_gp_deep} and neural tangent kernel (NTK) machines \citep{jacot, lee2019wide, arora2019exact} in their respective infinite-width limits. These limits provide both analytic tractability as well as detailed training and generalization analysis \citep{du2019gradient, bordelon, canatar2021spectral, cohen2021learning, jacot2020kernel, loureiro2021learning, simon2021neural, allen2019convergence}. However, in this limit, with these parameterizations, data representations are fixed and do not adapt to data, termed the \textit{lazy regime} of NN training, to contrast it from the \textit{rich regime} where NNs  significantly alter their internal features while fitting the data \citep{chizat2019lazy, geiger2020disentangling}. The fact that the representation of data is fixed renders these kernel-based theories incapable of explaining feature learning, an ingredient which is crucial to the success of deep learning in practice \citep{brown2020language, he2016deep}. 
Thus, alternative theories capable of modeling feature learning dynamics are needed. 

Recently developed alternative parameterizations such as the mean field \cite{mei2018mean} and the $\mu P$ \cite{yang2021tensor} parameterizations allow feature learning in infinite-width NNs trained with gradient descent. Using the Tensor Programs framework, Yang \& Hu identified a stochastic process that describes the evolution of preactivation features in infinite-width $\mu P$ NNs \cite{yang2021tensor}. In this work, we study an equivalent parameterization to $\mu P$ with self-consistent dynamical mean field theory (DMFT) and recover the stochastic process description of infinite NNs using this alternative technique. In the same large width scaling, we include a scalar parameter $\gamma_0$ that allows smooth interpolation between lazy and rich behavior \cite{chizat2019lazy}. We provide a new computational procedure to sample this stochastic process and demonstrate its predictive power for wide NNs. 

Our novel contributions in this paper are the following:
\begin{enumerate}
    \item We develop a path integral formulation of gradient flow dynamics in infinite-width networks in the feature learning regime. Our parameterization includes a scalar parameter $\gamma_0$ to allow interpolation between rich and lazy regimes and comparison to perturbative methods. 
    \item Using a stationary action argument, we identify a set of saddle point equations that the kernels satisfy at infinite-width, relating the stochastic processes that define hidden activation evolution to the kernels and vice versa. We show that our saddle point equations recover at $\gamma_0 = 1$, from an alternative method, the same stochastic process obtained previously with Tensor Programs \cite{yang2021tensor}.
    \item We develop a polynomial-time numerical procedure to solve the saddle point equations for deep networks. In numerical experiments, we demonstrate that solutions to these self-consistency equations are predictive of network training at a variety of feature learning strengths, widths and depths. We provide comparisons of our theory to various approximate methods, such as perturbation theory. 
\end{enumerate}

\subsection{Related Works}
A natural extension to the lazy NTK/NNGP limit that allows the study of feature learning is to calculate finite width corrections to the infinite-width limit. Finite width corrections to Bayesian inference in wide networks have been obtained with various perturbative \cite{aitchison2020bigger,yaida2020non, zavatone2021asymptotics, naveh2021predicting, roberts2021principles, hanin2022correlation, segadlo_nngp_field} and self-consistent techniques \cite{naveh2021self, seroussi2021separation, li2021statistical, zavatone2021depth}. In the gradient descent based setting, leading order corrections to the NTK dynamics have been analyzed to study finite width effects \cite{huang2020dynamics,dyer2019asymptotics, andreassen2020asymptotics, roberts2021principles}. These methods give approximate corrections which are accurate provided the strength of feature learning is small. In very rich feature learning regimes, however, the leading order corrections can give incorrect predictions \cite{zavatone2022contrasting,lewkowycz2020large}. 

Another approach to study feature learning 
is to alter NN parameterization in gradient-based learning to allow significant feature evolution even at infinite-width, the \textit{mean field} limit \cite{mei2018mean, araujo2019mean}. 
Works on mean field NNs have yielded formal loss convergence results \cite{chizat2018global, rotskoff2018trainability} and shown equivalences of gradient flow dynamics to a partial differential equation (PDE) \cite{mei2019mean, nguyen2019mean,  fang2021modeling}.

Our results are most closely related to a set of recent works which studied infinite-width NNs trained with gradient descent (GD) using the Tensor Programs (TP) framework \cite{yang2021tensor}. We show that our discrete time field theory at unit feature learning strength $\gamma_0 = 1$ recovers the stochastic process which was derived from TP. The stochastic process derived from TP has provided insights into practical issues in NN training such as hyper-parameter search \cite{yang2021tuning}. Computing the exact infinite-width limit of GD has exponential time requirements \cite{yang2021tensor}, which we show can be circumvented with an alternating sampling procedure. A projected variant of GD training has provided an infinite-width theory that could be scaled to realistic datasets like CIFAR-10 \cite{yang2022efficient}. Inspired by Chizat and Bach's work on mechanisms of lazy and rich training \cite{chizat2019lazy}, our theory interpolates between lazy and rich behavior in the mean field limit for varying $\gamma_0$ and allows comparison of DMFT to perturbative analysis near small $\gamma_0$. Further, our derivation of a DMFT action allows the possibility of pursuing finite width effects.  


Our theory is inspired by self-consistent dynamical mean field theory (DMFT) from statistical physics \cite{martin1973statistical, de1978dynamics,  sompolinsky1981dynamic, sompolinsky1982relaxational, arous1995large, arous1997symmetric, ben2006cugliandolo}. This framework has been utilized in the theory of random recurrent networks \cite{crisanti2018path, sompolinsky1988chaos, Helias_2020, molgedey1992suppressing, samuelides2007random,rajan2010stimulus}, tensor PCA \cite{mannelli2019passed, mannelli2020marvels}, phase retrieval \cite{mignacco2021stochasticity}, and high-dimensional linear classifiers \cite{agoritsas2018out,mignacco2020dynamical, celentano2021high, mignacco2021effective}, but has yet to be developed for deep feature learning. By developing a self-consistent DMFT of deep NNs, we gain insight into how features evolve in the rich regime of network training, while retaining many pleasant analytic properties of the infinite-width limit.

\section{Problem Setup and Definitions}
Our theory applies to infinite-width networks, both fully-connected and convolutional. For notational ease we will relegate convolutional results to later sections. For input $\x_\mu \in \mathbb{R}^D$, we define the hidden \textit{pre-activation} vectors $\h^\ell \in \mathbb{R}^{N}$ for layers $\ell \in \{1,...,L\}$ as
\begin{align}
    f_\mu = \frac{1}{\gamma \sqrt{N}} \w^L \cdot \phi(\h^L_\mu) \  , \quad  \h^{\ell + 1}_{\mu} = \frac{1}{\sqrt{N}} \W^{\ell}\phi(\h^\ell_\mu)  \ , \quad \h^1_\mu = \frac{1}{\sqrt{D}} \W^{0} \x_\mu, \label{eq:ffn}
\end{align}
where $\bm\theta = \text{Vec}\{\W^0,...,\w^L\}$ are the trainable parameters of the network and $\phi$ is a twice differentiable activation function. Inspired by previous works on the mechanisms of lazy gradient based training, the parameter $\gamma$ will control the laziness or richness of the training dynamics \cite{chizat2019lazy, geiger2020disentangling, yang2021tensor, mei2019mean}. Each of the trainable parameters are initialized as Gaussian random variables with unit variance $W^\ell_{ij} \sim \mathcal{N}(0,1)$. They evolve under gradient flow $\frac{d}{dt} \bm\theta = - \gamma^2 \nabla_{\bm\theta} \mathcal{L}$. The choice of learning rate $\gamma^2$ causes $\frac{d}{dt} \mathcal{L}|_{t=0}$ to be independent of $\gamma$. To characterize the evolution of weights, we introduce backpropagation variables $\g_\mu^\ell = \gamma \sqrt{N} \frac{\partial f_\mu}{\partial \h^\ell_\mu} = \dot\phi(\h^\ell_\mu) \odot \z^\ell_\mu$, where $\z^\ell_\mu = \frac{1}{\sqrt N} \W^{\ell \top} \g^{\ell+1}_\mu$ is the \textit{pre-gradient} signal.\looseness = -1 

The relevant dynamical objects to characterize feature learning are feature and gradient kernels for each hidden layer $\ell \in \{1,...,L\}$, defined as
\begin{align}
    \Phi^{\ell}_{\mu\alpha}(t,s) = \frac{1}{N} \phi(\h^{\ell}_\mu(t)) \cdot \phi(\h^{\ell}_\alpha(s)) \ , \quad G_{\mu\alpha}^{\ell}(t,s) = \frac{1}{N} \g^{\ell}_\mu(t) \cdot \g^{\ell}_\alpha(s) .
\end{align}

 From the kernels $\{\Phi^\ell,G^\ell \}_{\ell=1}^L$, we can compute the \textit{Neural Tangent Kernel} $K^{NTK}_{\mu\alpha}(t,s) = \nabla_{\theta} f_{\mu}(t)\cdot\nabla_{\theta} f_{\alpha}(s) =\sum_{\ell=0}^{L} G^{\ell+1}_{\mu\alpha}(t,s) \Phi^\ell_{\mu\alpha}(t,s),$ \cite{jacot} and the dynamics of the network function $f_\mu$
\begin{align}
    \frac{d}{dt} f_\mu(t) &= \sum_{\alpha=1}^P K^{NTK}_{\mu\alpha}(t,t) \Delta_\alpha(t) \  , \quad \Delta_\mu(t) = - \frac{\partial }{\partial f_\mu} \mathcal{L}|_{f_\mu(t)},
\end{align}
where we define base cases $G_{\mu\alpha}^{L+1}(t,s) = 1, \Phi^0_{\mu\alpha}(t,s) =K^x_{\mu\alpha}= \frac 1D \x_{\mu}\cdot \x_{\alpha}$. We note that the above formula holds for any data point $\mu$ which may or may not be in the set of $P$ training examples. The above expressions demonstrate that knowledge of the temporal trajectory of the NTK on the $t=s$ diagonal gives the temporal trajectory of the network predictions $f_\mu(t)$. 

Following prior works on infinite-width networks \cite{mei2018mean, yang2021tensor, chizat2018global, geiger2020disentangling}, we study the mean field limit
\begin{align}
    N, \gamma \to \infty \ ,\quad \gamma_0 = \frac{\gamma}{\sqrt N} = \mathcal{O}_N(1) 
\end{align}
As we demonstrate in the Appendix \ref{app:dmft_derivation} and \ref{app:equiv_par}, this is the only $N$-scaling which allows feature learning as $N \to \infty$. The $\gamma_0 = 0$ limit recovers the static NTK limit \cite{jacot}. We discuss other scalings and parameterizations in Appendix \ref{app:equiv_par}, relating our work to the $\mu P$-parameterization and TP analysis of \citep{yang2021tensor}, showing they have identical feature dynamics in the infinite-width limit. We also analyze the effect of different hidden layer widths and initialization variances in the Appendix \ref{app:vary_width_init}. We focus on equal widths and NTK parameterization (as in eq. \eqref{eq:ffn}) in the main text to reduce complexity. 

\section{Self-consistent DMFT}

Next, we derive our self-consistent DMFT in a limit where $t, P = \mathcal{O}_N(1)$. Our goal is to build a description of training dynamics purely based on representations, and independent of weights. Studying feature learning at infinite-width enjoys several analytical properties:
\begin{itemize}
    \item The kernel order parameters $\Phi^\ell,G^\ell$ concentrate over random initializations but are dynamical, allowing flexible adaptation of features to the task structure.
    \item In each layer $\ell$, each neuron's preactivation $h_i^\ell$ and pregradient $z^\ell_i$ become i.i.d. draws from a distribution characterized by a set of order parameters $\{ \Phi^\ell,G^\ell,A^{\ell},B^\ell\}$. 
    \item The kernels are defined as self-consistent averages (denoted by $\left< \right>$) over this distribution of neurons in each layer $\Phi^\ell_{\mu\alpha}(t,s) = \left< \phi(h_\mu^\ell(t)) \phi(h_\alpha^\ell(s)) \right>$ and $G_{\mu\alpha}^\ell(t,s) = \left< g_\mu^\ell(t)  g_\alpha^\ell(s) \right>$.
\end{itemize}
The next section derives these facts from a path-integral formulation of gradient flow dynamics.

\subsection{Path Integral Construction}
Gradient flow after a random initialization of weights defines a high dimensional stochastic  process over initalizations for variables $\{\h,\g\}$. Therefore, we will utilize DMFT formalism to obtain a reduced description of network activity during training. For a simplified derivation of the DMFT for the two-layer ($L=1$) case, see \ref{app:two_layer_warmup_derivation}. Generally, we separate the contribution on each forward/backward pass between the initial condition and gradient updates to weight matrix $\W^\ell$, defining new stochastic variables $\bm\chi^\ell ,\bm\xi^\ell \in \mathbb{R}^{N}$ as
\begin{align}
    \bm\chi_\mu^{\ell+1}(t) = \frac{1}{\sqrt N} \W^{\ell}(0) \phi(\h^{\ell}_\mu(t)) \ , \quad \bm\xi^\ell_\mu(t) = \frac{1}{\sqrt N} \W^{\ell}(0)^\top \g^{\ell+1}_\mu(t).
\end{align}
We let $Z$ represent the moment generating functional (MGF) for these stochastic fields 
\begin{align}
    Z[\{\j^\ell,\v^{\ell} \}] = \left< \exp\left(  \sum_{\ell,\mu}\int_0^\infty dt \left[  \j_\mu^\ell(t) \cdot \bm\chi_\mu^\ell(t) + \v^\ell_{\mu}(t) \cdot \bm\xi^\ell_\mu(t) \right] \right) \right>_{\{\W^0(0),...\w^L(0)\}}, \nonumber
\end{align}
which requires, by construction the normalization condition $Z[\{\bm 0,\bm0 \}] = 1$. We enforce our definition of $\bm\chi,\bm\xi$ using an integral representation of the delta-function. Thus for each sample $\mu \in [P]$ and each time $t \in \mathbb{R}_+$, we multiply $Z$ by 
\begin{align}
    1 = \int_{\mathbb{R}^N} \int_{\mathbb{R}^N}  \frac{d\bm\chi^{\ell+1}_\mu(t)d\hat{\bm\chi}^{\ell+1}_\mu(t)}{(2\pi)^N} \exp\left(i \hat{\bm\chi}_\mu^{\ell+1}(t) \cdot \left[\bm\chi_\mu^{\ell+1}(t) - \frac{1}{\sqrt N} \W^{\ell}(0) \phi(\h^{\ell}_\mu(t)) \right]  \right),
\end{align}
for $\bm\chi$ and the respective expression for $\bm\xi$. After making such substitutions, we perform integration over initial Gaussian weight matrices to arrive at an integral expression for $Z$, which we derive in the appendix \ref{app:order_params_action}. We show that $Z$ can be described by set of order-parameters $\{ \Phi^\ell ,\hat\Phi^\ell , G^\ell, \hat G^\ell, A^\ell, B^\ell\}$
\begin{align}
    Z[\{\j^\ell,\v^{\ell} \}] \propto \int \prod_{\ell\mu\alpha ts}& d\Phi_{\mu\alpha}^\ell(t,s) d\hat{\Phi}^\ell_{\mu\alpha}(t,s) dG^\ell_{\mu\alpha}(t,s)d\hat{G}^\ell_{\mu\alpha}(t,s) dA^{\ell}_{\mu\alpha}(t,s) dB^\ell_{\mu\alpha}(t,s) 
    \\
    &\times\exp\left({N S[\{\Phi,\hat\Phi,G,\hat G,A,B,j,v\}]}\right), \nonumber
    \\
    S = \sum_{\ell \mu\alpha} \int_0^\infty dt \int_0^\infty ds &\left[ \Phi_{\mu\alpha}^\ell(t,s) \hat{\Phi}_{\mu\alpha}^\ell(t,s) + G^\ell_{\mu\alpha}(t,s)\hat{G}^\ell_{\mu\alpha}(t,s) - A^{\ell}_{\mu\alpha}(t,s) B^\ell_{\mu\alpha}(t,s) \right] \nonumber
    \\
    &+  \ln \mathcal Z[\{\Phi,\hat\Phi,G,\hat G,A,B, j, v\}], 
\end{align}
where $S$ is the DMFT action and $\mathcal Z$ is a single-site MGF, which defines the distribution of fields $\{\chi^\ell,\xi^\ell\}$ over the neural population in each layer. The kernels $A$ and $B$ are related to the correlations between feedforward and feedback signals in the network. We provide a detailed formula for $\mathcal Z$ in the Appendix \ref{app:order_params_action} and show that it factorizes over different layers $\mathcal Z = \prod_{\ell=1}^L \mathcal{Z}_\ell$. 

\subsection{Deriving the DMFT Equations from the Path Integral Saddle Point}
As $N \to \infty$, the moment-generating function $Z$ is exponentially dominated by the saddle point of $S$. The equations that define this saddle point also define our DMFT. We thus identify the kernels that render $S$ locally stationary ($\delta S = 0$). The most important equations are those which define $\{\Phi^\ell,G^\ell\}$
\begin{align}
    \frac{\delta S}{\delta \hat \Phi^\ell_{\mu\alpha}(t,s)}  &= {\Phi}_{\mu\alpha}^\ell(t,s) + \frac{1}{\mathcal Z} \frac{\delta \mathcal Z}{\delta \hat \Phi^\ell_{\mu\alpha}(t,s)} = {\Phi}^\ell_{\mu\alpha}(t,s) - \left< \phi(h_\mu^\ell(t)) \phi(h_\alpha^\ell(s))  \right> = 0, \nonumber
    \\
    \frac{\delta S}{\delta \hat G^\ell_{\mu\alpha}(t,s)}  &= G_{\mu\alpha}^\ell(t,s) + \frac{1}{\mathcal Z} \frac{\delta \mathcal Z}{\delta \hat{G}_{\mu\alpha}^\ell(t,s)} = G^\ell_{\mu\alpha}(t,s) - \left< g_\mu^\ell(t) g_\alpha^\ell(s) \right> = 0,
\end{align}
where $\left< \right>$ denotes an average over the stochastic process induced by $\mathcal Z$, which is defined below
\begin{align}\label{eq:dmft_stoch_process}
     &\{ u_\mu^\ell(t) \}_{\mu\in[P], t\in\mathbb{R}_+} \sim \mathcal{GP}(0,\bm\Phi^{\ell-1} ) \ , \ \{ r_\mu^\ell(t) \}_{\mu\in[P], t\in\mathbb{R}_+} \sim \mathcal{GP}(0,\G^{\ell+1}), \nonumber
    \\
    h_\mu^{\ell}(t) &= u_\mu^\ell(t) + \gamma_0 \int_0^t ds \sum_{\alpha=1}^P \left[ A_{\mu\alpha}^{\ell-1}(t,s) + \Delta_\alpha(s) \Phi^{\ell-1}_{\mu\alpha}(t,s) \right] z_\alpha^\ell(s) \dot\phi(h^\ell_\alpha(s)), \nonumber
    \\
    z_{\mu}^\ell(t) &= r_\mu^\ell(t) + \gamma_0 \int_0^t ds \sum_{\alpha=1}^P \left[ B^{\ell}_{\mu\alpha}(t,s) + \Delta_\alpha(s) G^{\ell+1}_{\mu\alpha}(t,s)  \right] \phi(h_\alpha^\ell(s)),
\end{align}
where we define base cases $\Phi^0_{\mu\alpha}(t,s) = K^x_{\mu\alpha}$ and $G^{L+1}_{\mu\alpha}(t,s) = 1$, $A^0 = B^L = 0$. We see that the fields $\{ h^\ell,z^\ell \}$, which represent the single site preactivations and pre-gradients, are implicit functionals of the mean-zero Gaussian processes $\{u^\ell,r^\ell\}$ which have covariances $\left< u^\ell_{\mu}(t) u^\ell_\alpha(s) \right> = \Phi^{\ell-1}_{\mu\alpha}(t,s)$ and $\left< r^\ell_{\mu}(t) r^\ell_\alpha(s) \right> = G^{\ell+1}_{\mu\alpha}(t,s)$. The other saddle point equations give $A_{\mu\alpha}^{\ell}(t,s) = \gamma_0^{-1} \left< \frac{\delta \phi(h^\ell_\mu(t))}{\delta r^\ell_\alpha(s)} \right>, B_{\mu\alpha}^{\ell}(t,s) = \gamma_0^{-1} \left< \frac{\delta g^{\ell+1}_\mu(t)}{\delta u^{\ell+1}_\alpha(s)} \right>$ which arise due to coupling between the feedforward and feedback signals. We note that, in the lazy limit $\gamma_0 \to 0$, the fields approach Gaussian processes $h^\ell \to u^\ell$, $z^\ell \to r^\ell$. Lastly, the final saddle point equations $\frac{\delta S}{\delta \Phi^\ell} =0 ,\frac{\delta S}{\delta G^\ell} = 0$ imply that $\hat\Phi^\ell = \hat G^\ell = 0$. The full set of equations that define the DMFT are given in \ref{app:final_dmft_result}.

This theory is easily extended to more general architectures such as networks with varying widths by layer (App. \ref{app:vary_width_init}), trainable bias parameter (App. \ref{app:train_bias}), multiple (but $\mathcal{O}_N(1)$) output channels (App. \ref{app:multiple_outputs}), convolutional architectures (App. \ref{app:cnn}), networks trained with weight decay (App. \ref{app:weight_decay}), Langevin sampling (App. \ref{app:bayes_langevin}) and momentum (App. \ref{app:momentum}), discrete time training (App. \ref{app:discrete_time}). In Appendix \ref{app:equiv_par}, we discuss parameterizations which give equivalent feature and predictor dynamics and show our derived stochastic process is equivalent to the $\mu P$ scheme of Yang \& Hu \cite{yang2021tensor}.



\section{Solving the Self-Consistent DMFT}

\begin{figure}[t]
    \centering
    \subfigure[Lazy vs Rich Loss Dynamics]{\includegraphics[width=0.32\linewidth]{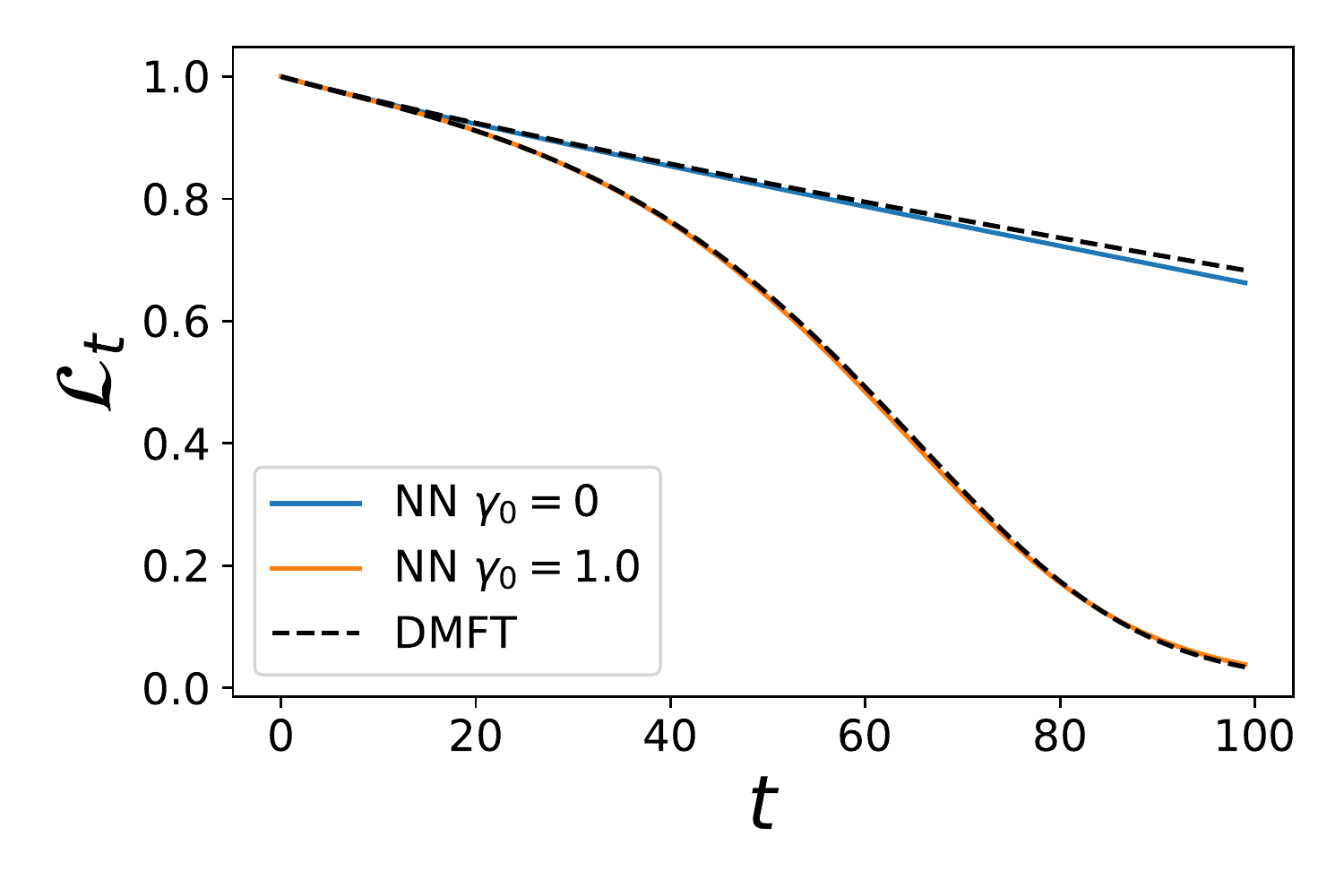}}
    \subfigure[Initial Preactivation Density]{\includegraphics[width=0.32\linewidth]{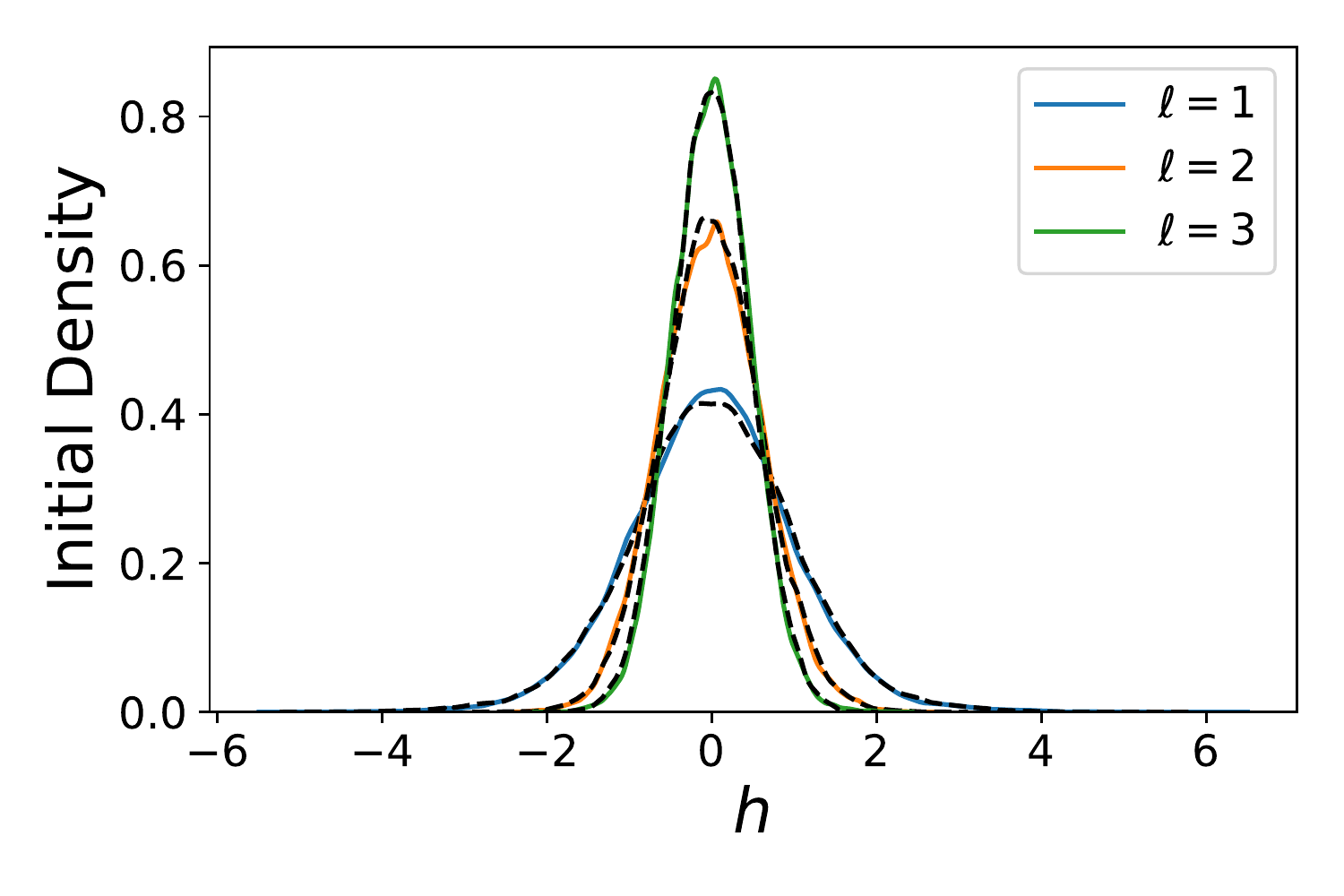}}
    \subfigure[Final Preactivation Density]{\includegraphics[width=0.32\linewidth]{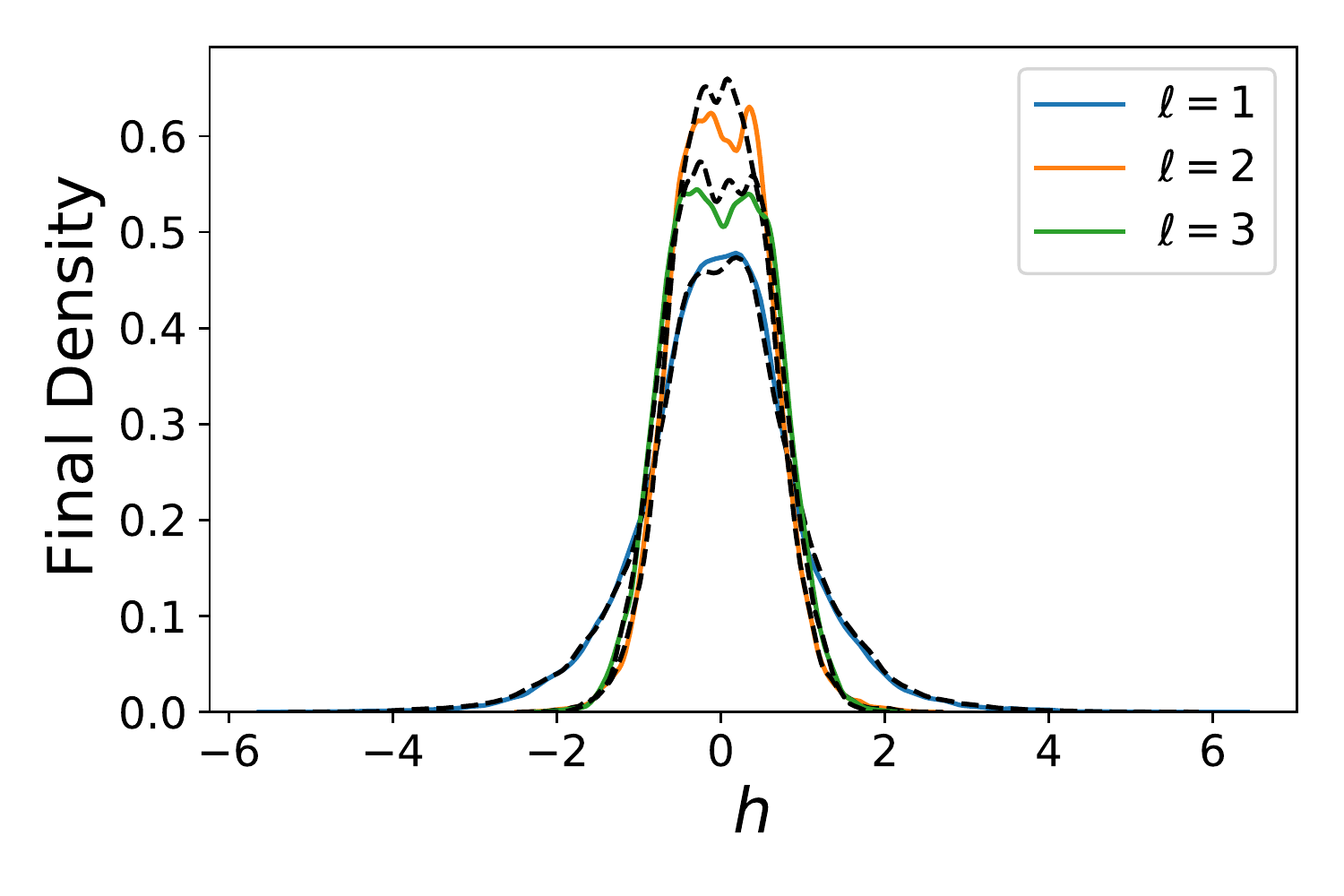}}
    \subfigure[Final $\Phi^\ell$ Kernels $\gamma_0=1$]{\includegraphics[width=0.35\linewidth]{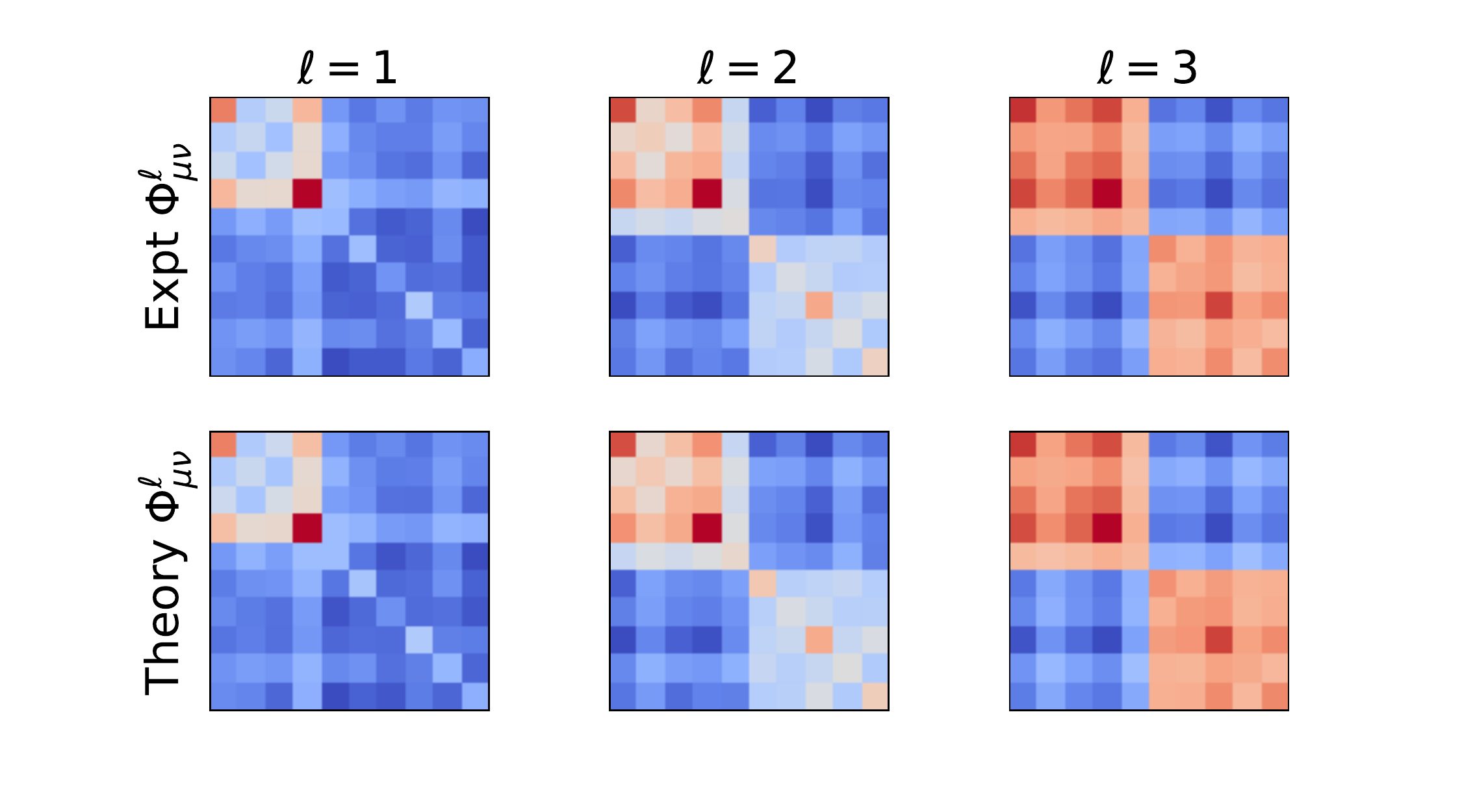}}
    \subfigure[$\Phi^\ell$ Dynamics $\gamma_0=1.0$]{\includegraphics[width=0.35\linewidth]{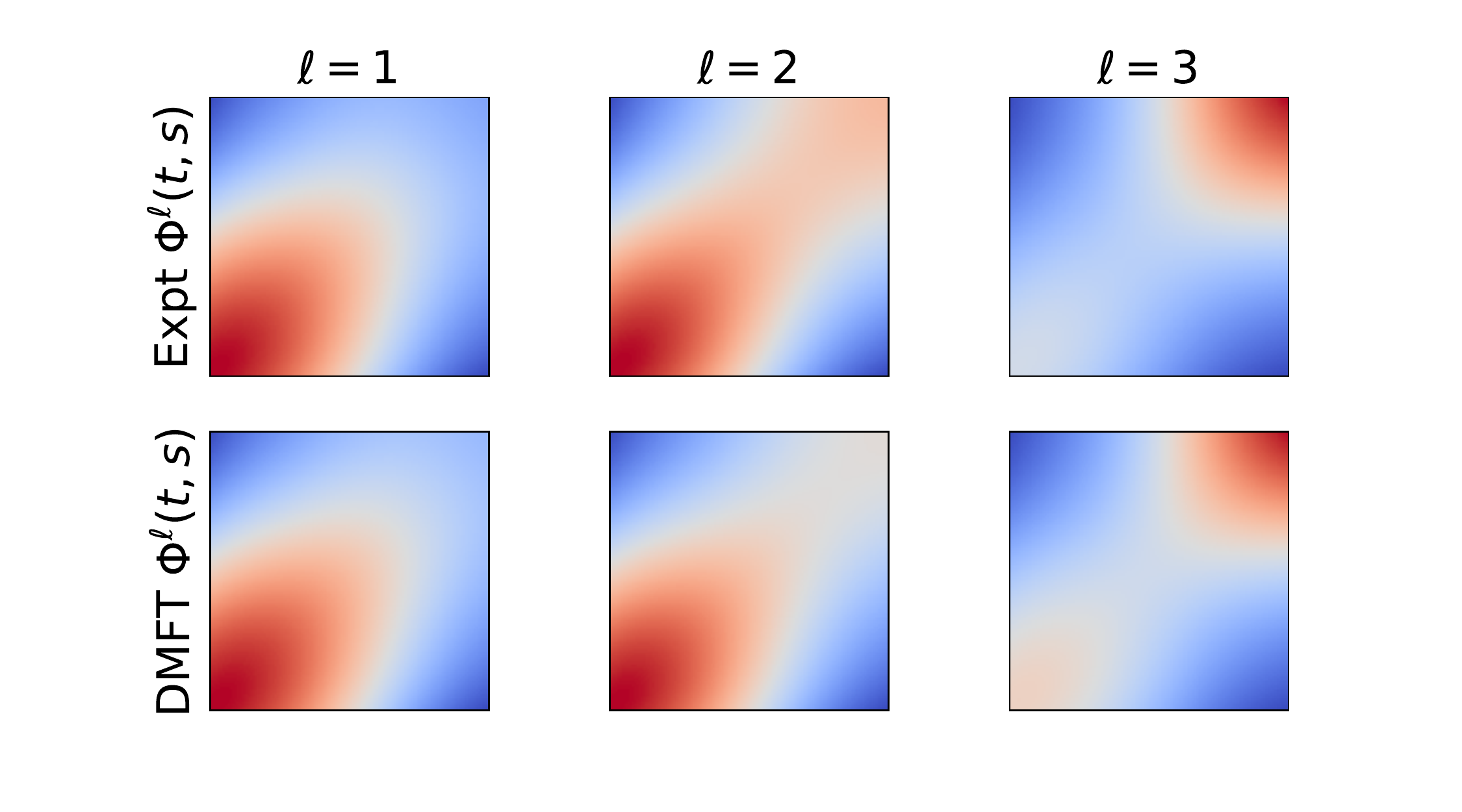}}
    \subfigure[$\Phi^\ell$ Convergence to DMFT ]{\includegraphics[width=0.28\linewidth]{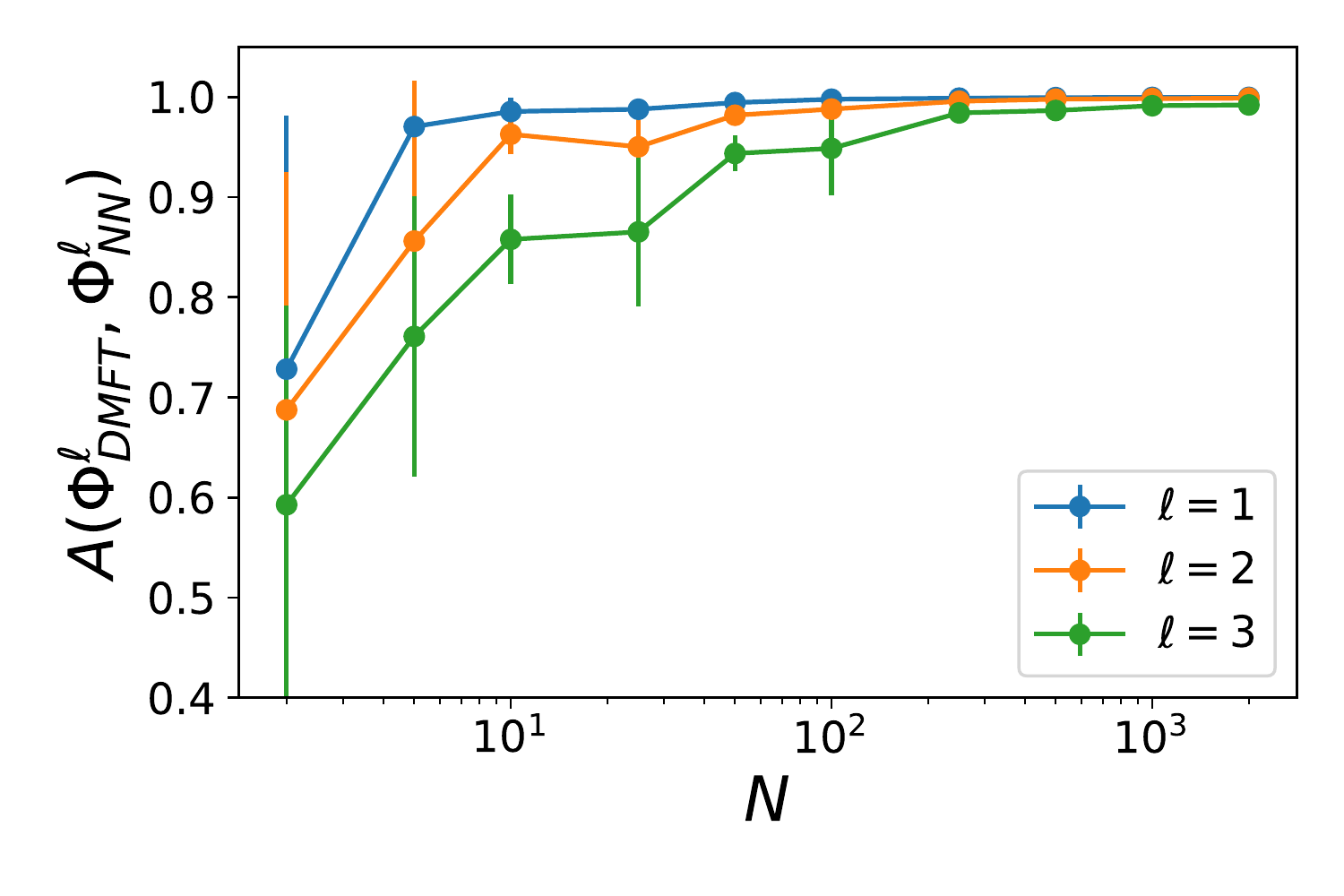}}
    \subfigure[Final $G^\ell$ kernels $\gamma_0 = 1.0$ ]{\includegraphics[width=0.35\linewidth]{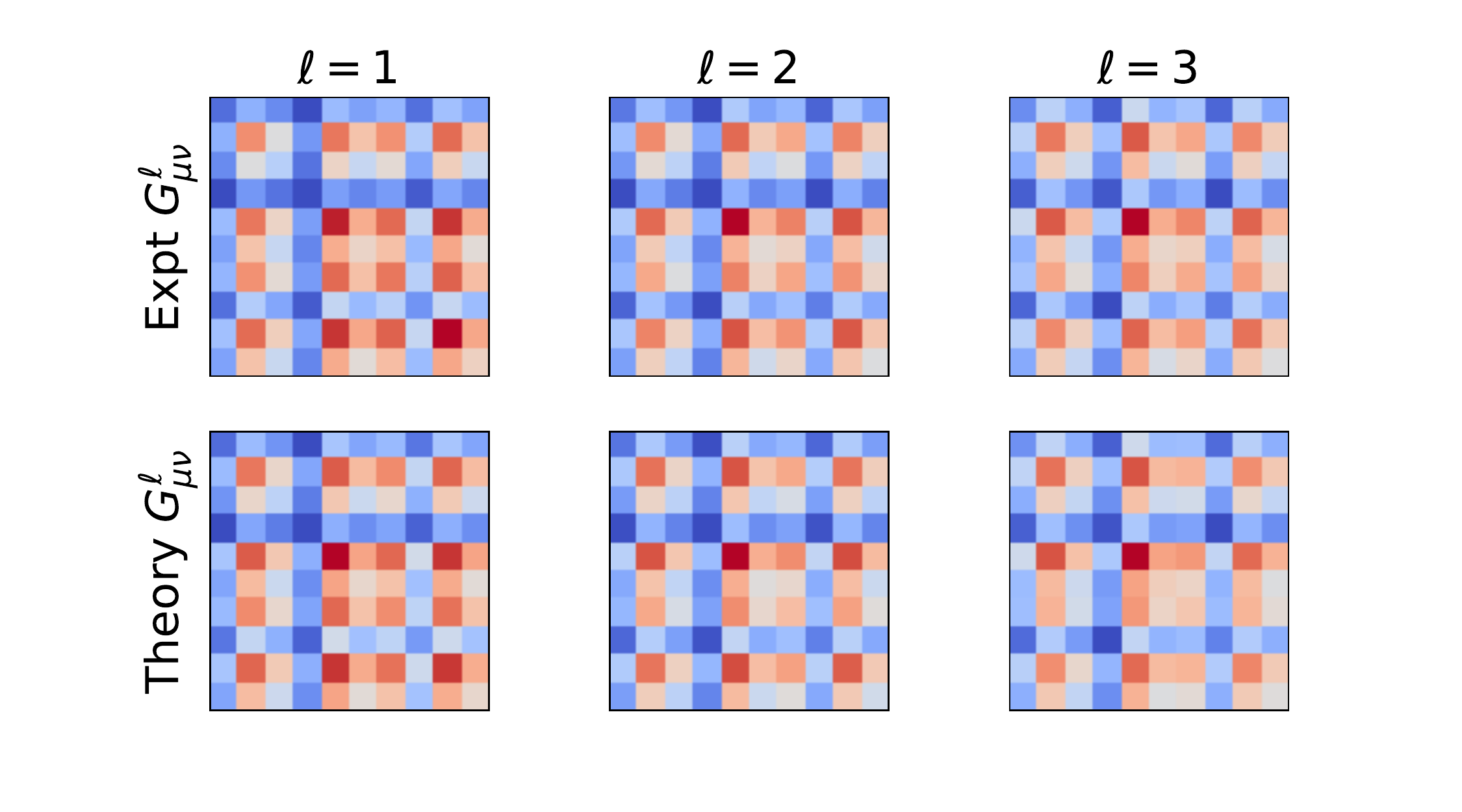}}
    \subfigure[$G^\ell$ Dynamics $\gamma_0=1.0$]{\includegraphics[width=0.35\linewidth]{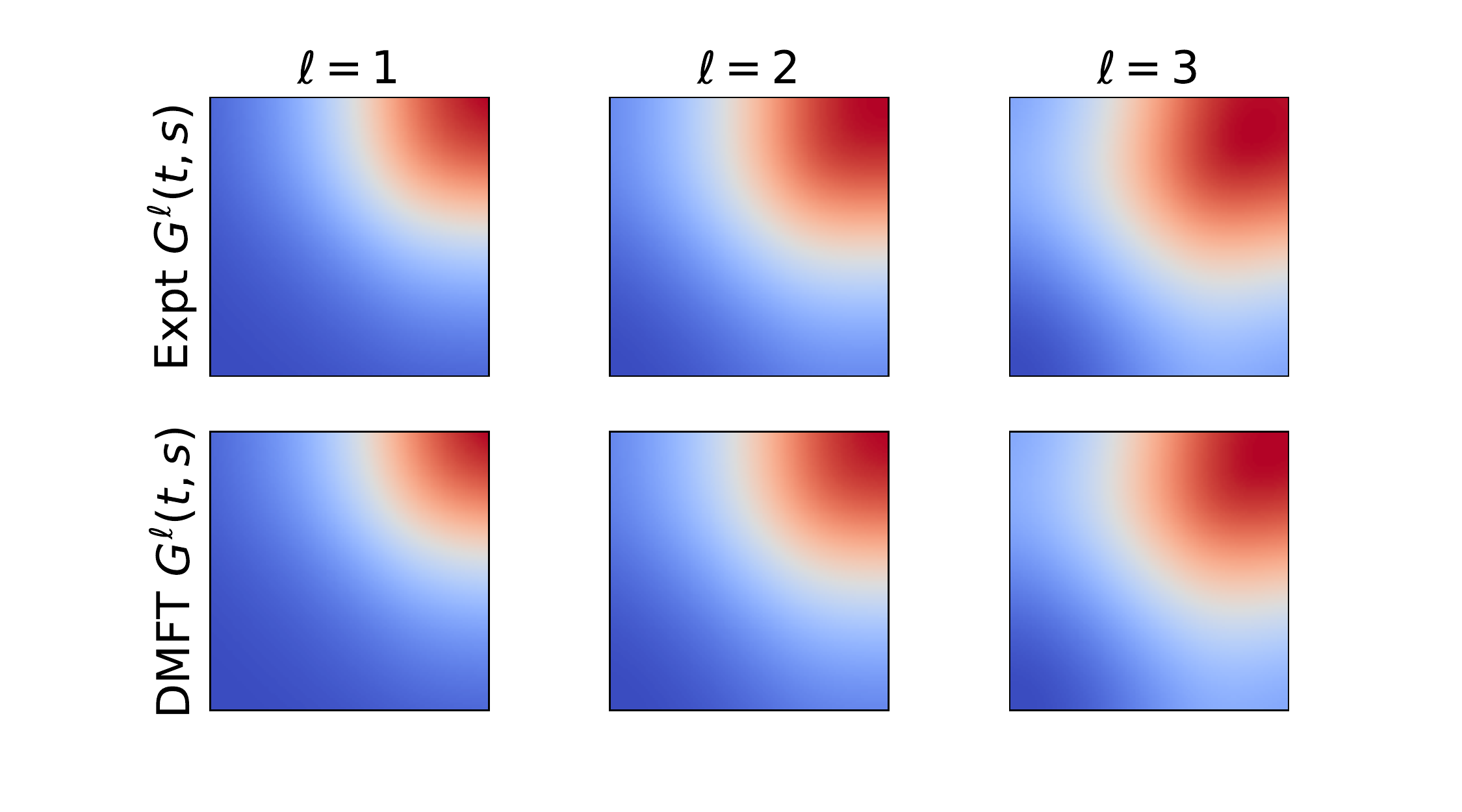}}
    \subfigure[$G^\ell$ Convergence to DMFT]{\includegraphics[width=0.28\linewidth]{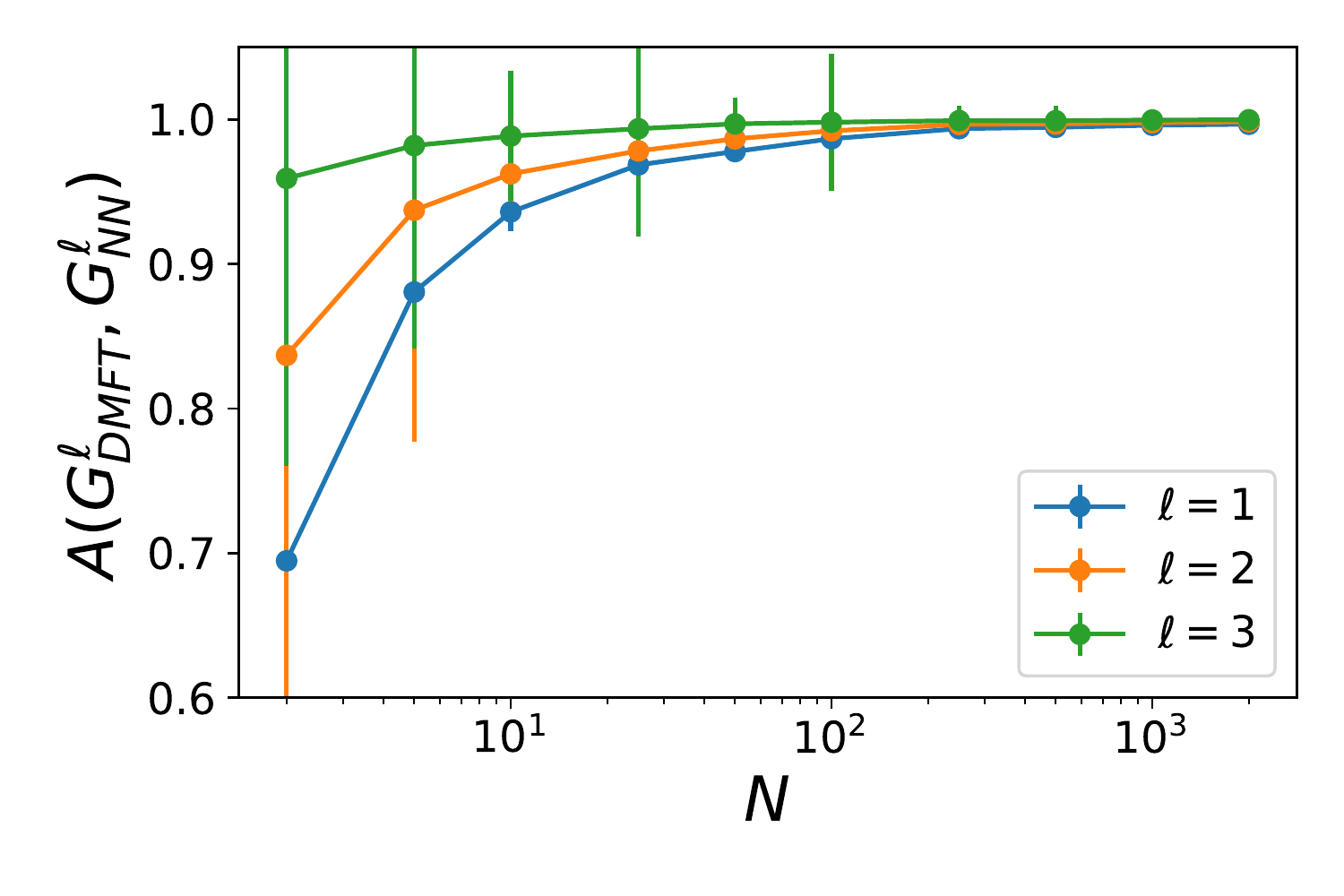}}
    \caption{Neural network feature learning dynamics is captured by self-consistent dynamical mean field theory (DMFT). (a) Training loss curves on a subsample of $P=10$ CIFAR-10 training points in a depth 4 ($L=3$, $N=2500$) tanh network ($\phi(h) = \tanh(h)$) trained with MSE. Increasing $\gamma_0$ accelerates training. (b)-(c) The distribution of preactivations at the beginning and end of training matches predictions of the DMFT. (d) The final $\Phi^\ell$ (at $t=100$) kernel order parameters match the finite width network. (e) The temporal dynamics of the sample-traced kernels $\sum_{\mu} \Phi_{\mu\mu}^{\ell}(t,s)$ matches experiment and reveals rich dynamics across layers. (f) The alignment $A(\bm\Phi^\ell_{DMFT}, \bm\Phi^\ell_{NN})$, defined as cosine similarity, of the kernel $\Phi^\ell_{\mu\alpha}(t,s)$ predicted by theory (DMFT) and width $N$ networks for different $N$ but fixed $\gamma_0 = \gamma/\sqrt{N}$. Errorbars show standard deviation computed over $10$ repeats. Around $N \sim 500$ DMFT begins to show near perfect agreement with the NN. (g)-(i) The same plots but for the gradient kernel $\G^\ell$. Whereas finite width effects for $\bm\Phi^\ell$ are larger at later layers $\ell$ since variance accumulates on the forward pass, fluctuations in $\G^\ell$ are large in early layers. }
    \label{fig:deep_tanh_visual_kernels}
\end{figure}

The saddle point equations obtained from the field theory discussed in the previous section must be solved self-consistently. By this we mean that, given knowledge of the kernels, we can characterize the distribution of $\{h^\ell, z^\ell\}$, and given the distribution of $\{h^\ell,z^\ell\}$, we can compute the kernels \cite{manacorda2020numerical, mignacco2020dynamical}. 
In the Appendix \ref{app:dmft_algorithm}, we provide Algorithm \ref{alg:MC_DMFT}, a numerical procedure based on this idea to efficiently solve for the kernels with an alternating Monte-Carlo strategy. The output of the algorithm are the dynamical kernels $\Phi^\ell_{\mu\alpha}(t,s), G^{\ell}_{\mu\alpha}(t,s), A^\ell_{\mu\alpha}(t,s), B^{\ell}_{\mu\alpha}(t,s)$, from which any network observable can be computed as we discuss in Appendix \ref{app:dmft_derivation}. We provide an example of the solution to the saddle point equations compared to training a finite NN in Figure \ref{fig:deep_tanh_visual_kernels}. We plot $\Phi^\ell, G^\ell$ at the end of training and the sample-trace of these kernels through time. Additionally, we compare the kernels of finite width $N$ network to the DMFT predicted kernels using a cosine-similarity alignment metric $A(\bm\Phi^{DMFT},\bm\Phi^{NN}) = \frac{\text{Tr} \ \bm\Phi^{DMFT} \bm\Phi^{NN}}{|\bm\Phi^{DMFT}||\bm\Phi^{NN}|}$.
Additional examples are in Appendix Figures \ref{fig:two_layer_relu} and Figure \ref{fig:dmft_tanh_depth_4}.

\subsection{Deep Linear Networks: Closed Form Self-Consistent Equations}

\begin{figure}[h]
    \centering
    \subfigure[Deep Linear Loss Dynamics]{\includegraphics[width=0.33\linewidth]{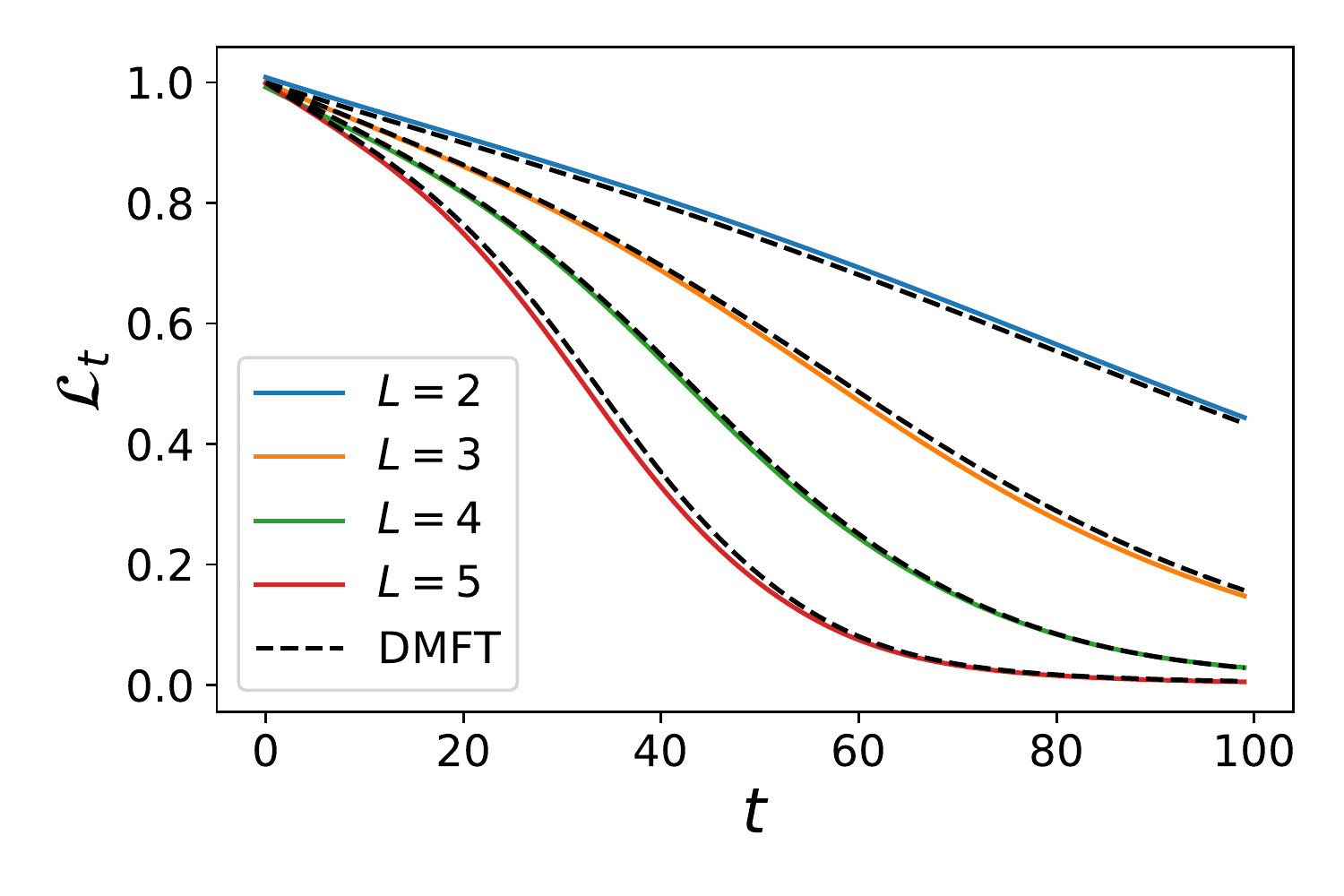}}
    \subfigure[Predicted vs Experimental Final $H^\ell$ Kernels]{\includegraphics[width=0.6\linewidth]{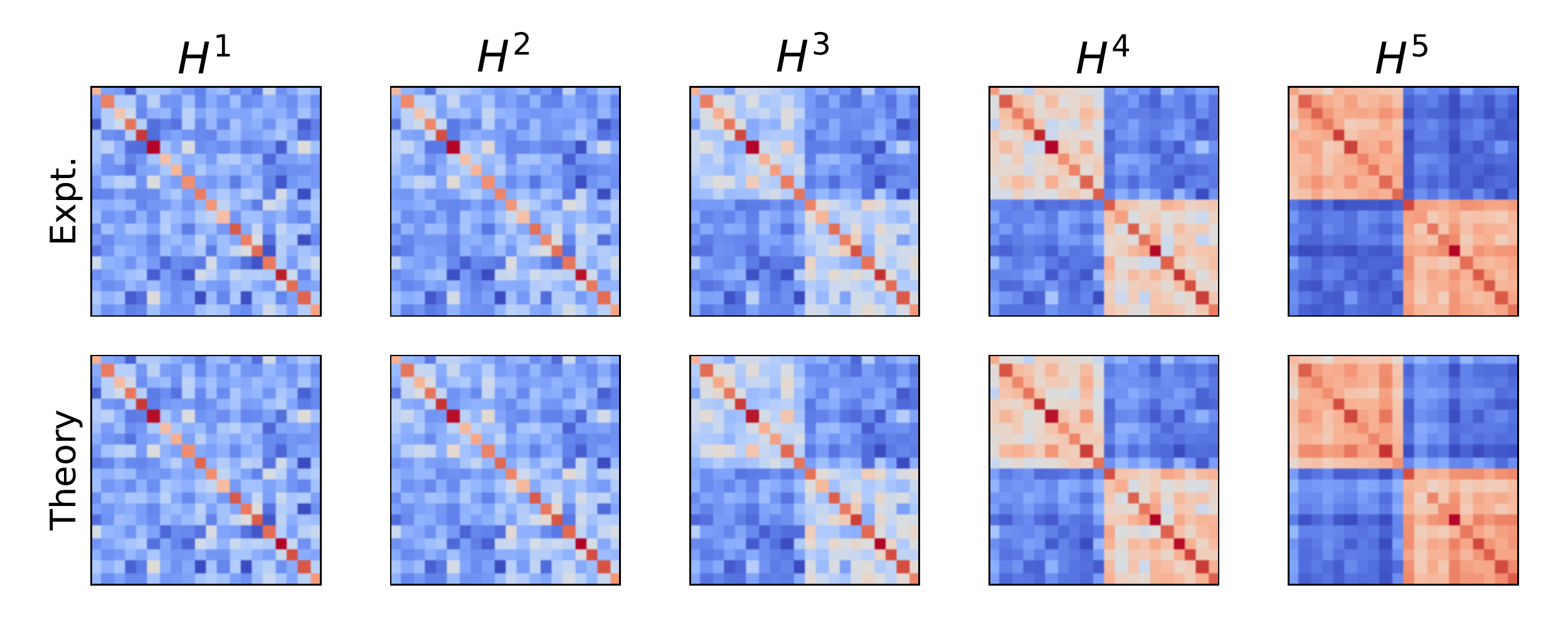}}
    \subfigure[$L$-Dependent Kernel Movement]{\includegraphics[width=0.33\linewidth]{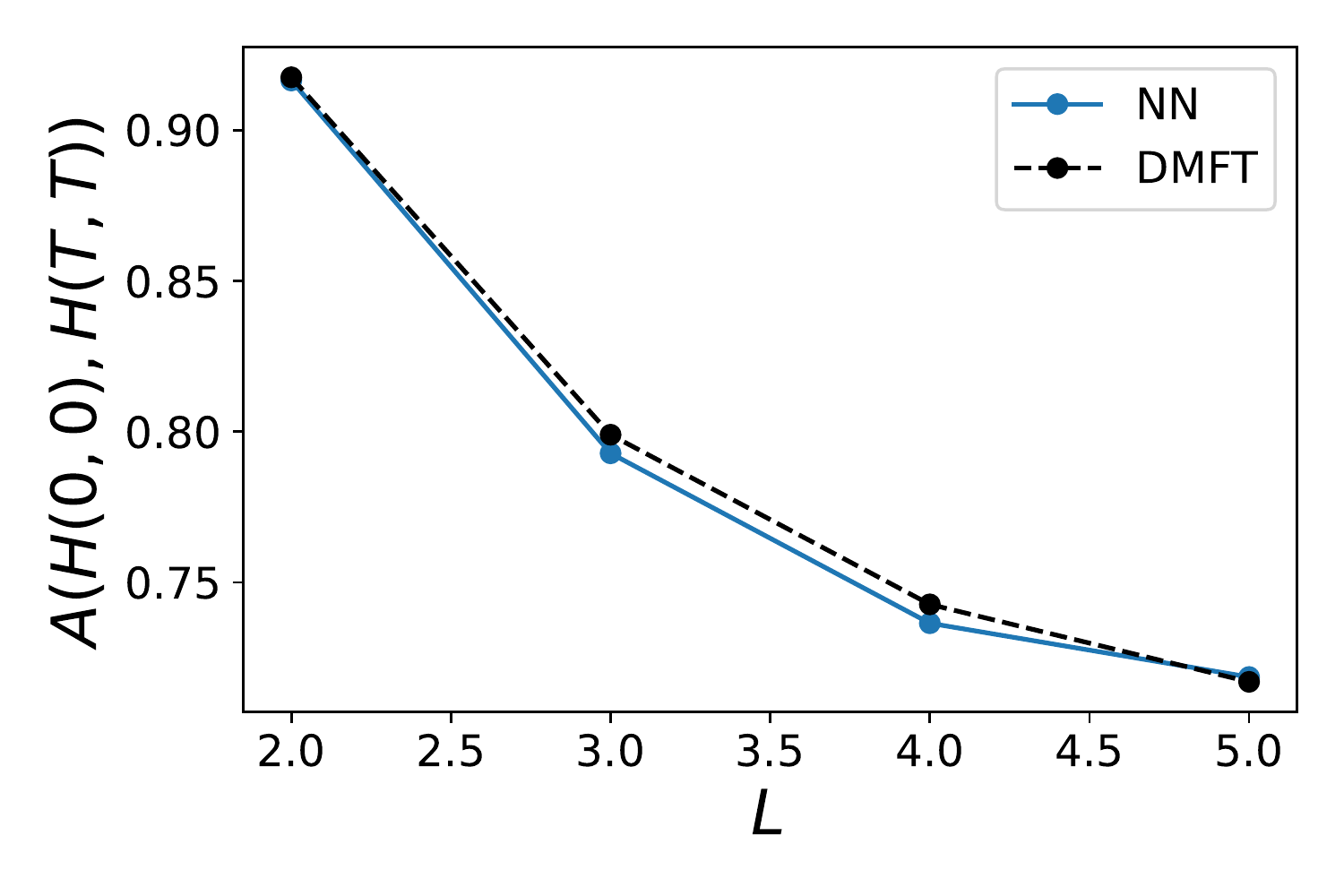}}
    \subfigure[$L=5$ DMFT Temporal Kernels]{\includegraphics[width=0.6\linewidth]{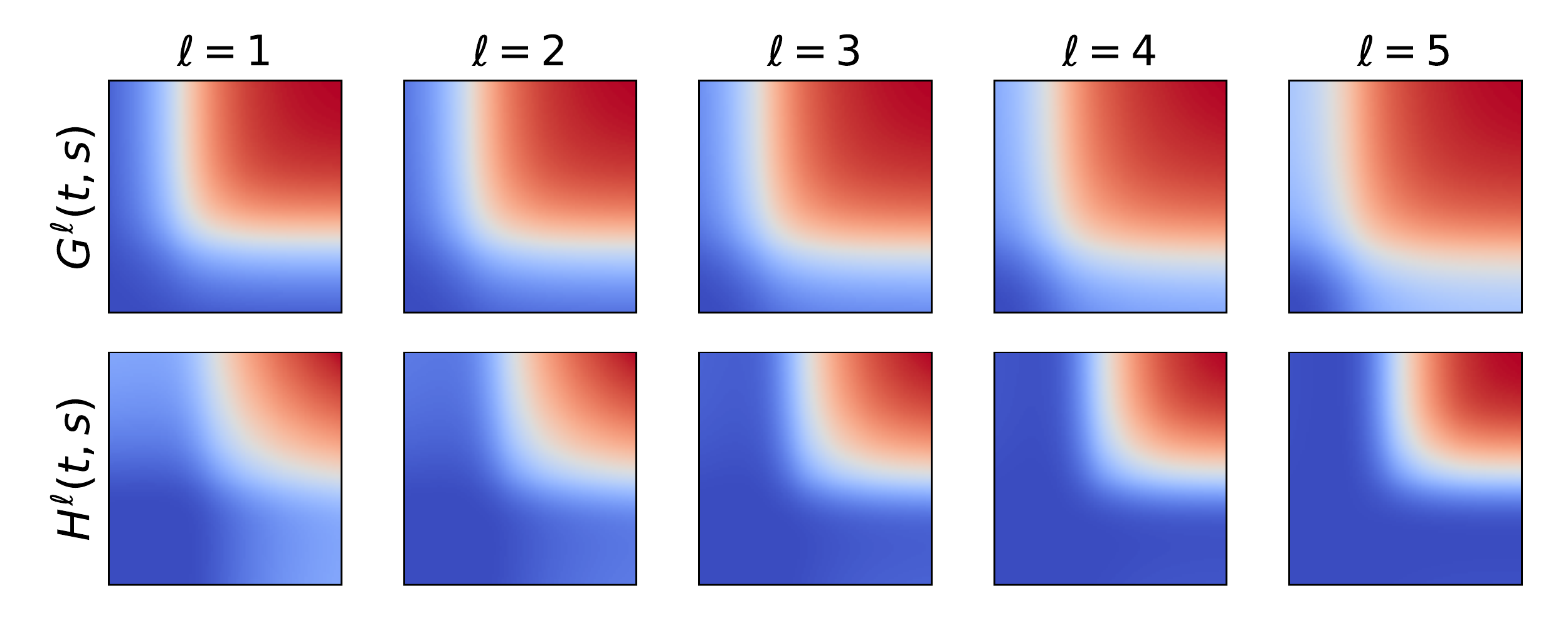}}
    \caption{Deep linear network with the full DMFT. (a) The train loss for NNs of varying $L$. (b) For a $L=5, N=1000$ NN, the kernels $H^\ell$ at the end of training compared to DMFT theory on $P=20$ datapoints. (c) The average displacement of feature kernels for different depth networks at same $\gamma_0$ value. For equal values of $\gamma_0$, deeper networks exhibit larger changes to their features, manifested in lower alignment with their initial $t=0$ kernels $\H$. (d) The solution to the temporal components of the $G^\ell(t,s)$ and $\sum_{\mu}H^\ell_{\mu\mu}(t,s)$ kernels obtained from the self-consistent equations.}
    \label{fig:deep_linear_fig}
\end{figure}

Deep linear networks ($\phi(h)=h$) are of theoretical interest since they are simpler to analyze than nonlinear networks but preserve non-trivial training dynamics and feature learning \cite{fukumizu1998dynamics,saxe2013exact, arora2018a, advani2020high, jacot2021deep, zavatone2021asymptotics, li2021statistical, aitchison2020bigger}. In a deep linear network, we can simplify our saddle point equations to algebraic formulas that close in terms of the kernels $H^\ell_{\mu\alpha}(t,s) = \left< h_\mu^\ell(t) h_\alpha^\ell(s) \right>$, $G^\ell(t,s) = \left< g^\ell(t) g^\ell(s) \right>$ \cite{yang2021tensor}. This is a significant simplification since it allows solution of the saddle point equations without a sampling procedure. 

To describe the result, we first introduce a vectorization notation $\h^\ell = \text{Vec}\{h_\mu^\ell(t) \}_{\mu\in [P], t \in \mathbb{R}_+}$. Likewise we convert kernels $\H^\ell = \text{Mat}\{ H^\ell_{\mu\alpha}(t,s) \}_{\mu,\alpha \in [P], t,s \in \mathbb{R}_+}$ into matrices.  The inner product under this vectorization is defined as $\a \cdot \b = \int_0^\infty dt \sum_{\mu=1}^P a_\mu(t) b_\mu(t)$. In a practical computational implementation, the theory would be evaluated on a grid of $T$ time points with discrete time gradient descent, so these kernels $\H^\ell \in \mathbb{R}^{PT \times PT}$ would indeed be matrices of the appropriate size. The fields $\h^\ell,\g^\ell$ are linear functionals of independent Gaussian processes $\u^\ell,\r^\ell$, giving $(\I - \gamma_0^2 \C^\ell \D^\ell) \h^\ell = \u^\ell + \gamma_0 \C^\ell \r^\ell \ , \ (\I - \gamma_0^2 \D^\ell \C^\ell) \g^\ell = \r^\ell + \gamma_0 \D^\ell \u^\ell$. 
The matrices $\C^\ell$ and $\D^\ell$ are causal integral operators which depend on $\{\A^{\ell-1}, \H^{\ell-1}\}$ and $\{\B^\ell, \G^{\ell+1}\}$ respectively which we define in Appendix \ref{app:linear_theory}. 
The saddle point equations which define the kernels are
\begin{align}
    \H^{\ell} &= \left< \h^\ell \h^{\ell \top}  \right> = (\I - \gamma_0^2 \C^\ell \D^\ell)^{-1} [\H^{\ell-1} + \gamma_0^2 \C^{\ell} \G^{\ell+1} \C^{\ell \top} ]\left[(\I - \gamma_0^2 \C^\ell \D^\ell)^{-1} \right]^\top \nonumber
    \\
    \G^{\ell} &= \left< \g^\ell \g^{\ell \top} \right> = \left( \I - \gamma_0^2 \D^{\ell} \C^\ell \right)^{-1} \left[ \G^{\ell+1} + \gamma^2_0 \D^{\ell} \H^{\ell-1} \D^{\ell \top} \right] \left[\left( \I - \gamma_0^2 \D^{\ell} \C^\ell \right)^{-1} \right]^\top.
\end{align}
Examples of the predictions obtained by solving these systems of equations are provided in Figure \ref{fig:deep_linear_fig}. We see that these DMFT equations describe kernel evolution for networks of a variety of depths and that the change in each layer's kernel increases with the depth of the network.  

Unlike many prior results \cite{fukumizu1998dynamics, saxe2013exact,arora2018a, advani2020high}, our DMFT does not require any restrictions on the structure of the input data but hold for any $\K^x, \y$. However, for whitened data $\K^x = \I$ we show in Appendix \ref{app:two_layer_whitened}, \ref{app:deep_whitened_linear} that our DMFT learning curves interpolate between NTK dynamics and the sigmoidal trajectories of prior works \cite{fukumizu1998dynamics,saxe2013exact} as $\gamma_0$ is increased. For example, in the two layer ($L=1$) linear network with $\K^x = \I$, the dynamics of the error norm $\Delta(t) = ||\bm\Delta(t)||$ takes the form $\frac{\partial}{\partial t} \Delta(t) = - 2 \sqrt{1 + \gamma_0^2 (y - \Delta(t))^2}\Delta(t)$ where $y = ||\y||$. These dynamics give the linear convergence rate of the NTK if $\gamma_0 \to 0$ but approaches logistic dynamics of \cite{saxe2013exact} as $\gamma_0 \to \infty$. Further, $\H(t) = \left< \h^1(t) \h^1(t)^\top  \right> \in \mathbb{R}^{P\times P}$ only grows in the $\y \y^\top$ direction with $H_y(t) = \frac{1}{y^2} \y^\top \H(t) \y = \sqrt{1+ \gamma_0^2 (y-\Delta(t))^2 }$. At the end of training $\H(t) \to \I + \frac{1}{y^2}[\sqrt{1+\gamma_0^2 y^2}-1] \y\y^\top$, recovering the rank one spike which was recently obtained in the small initialization limit \cite{atanasov2022neural}. We show this one dimensional system in Figure \ref{fig:err_H_dynamics_2layer}.

\subsection{Feature Learning with L2 Regularization}

\begin{figure}[H]
    \centering
    \subfigure[Loss for varying $\gamma_0$]{\includegraphics[width=0.38\linewidth]{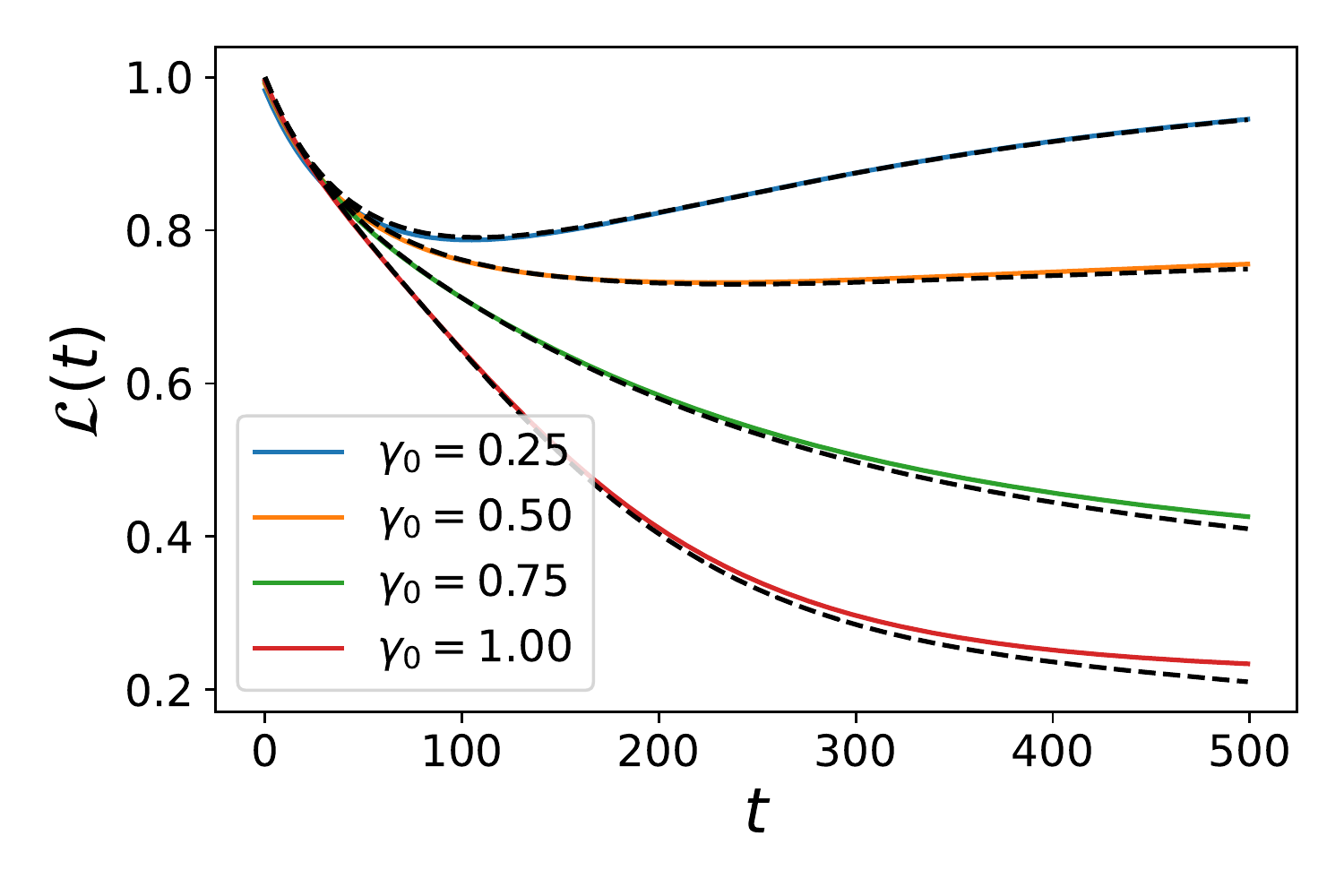}}
    \subfigure[Final $\Phi$ Kernels]{\includegraphics[width=0.45\linewidth]{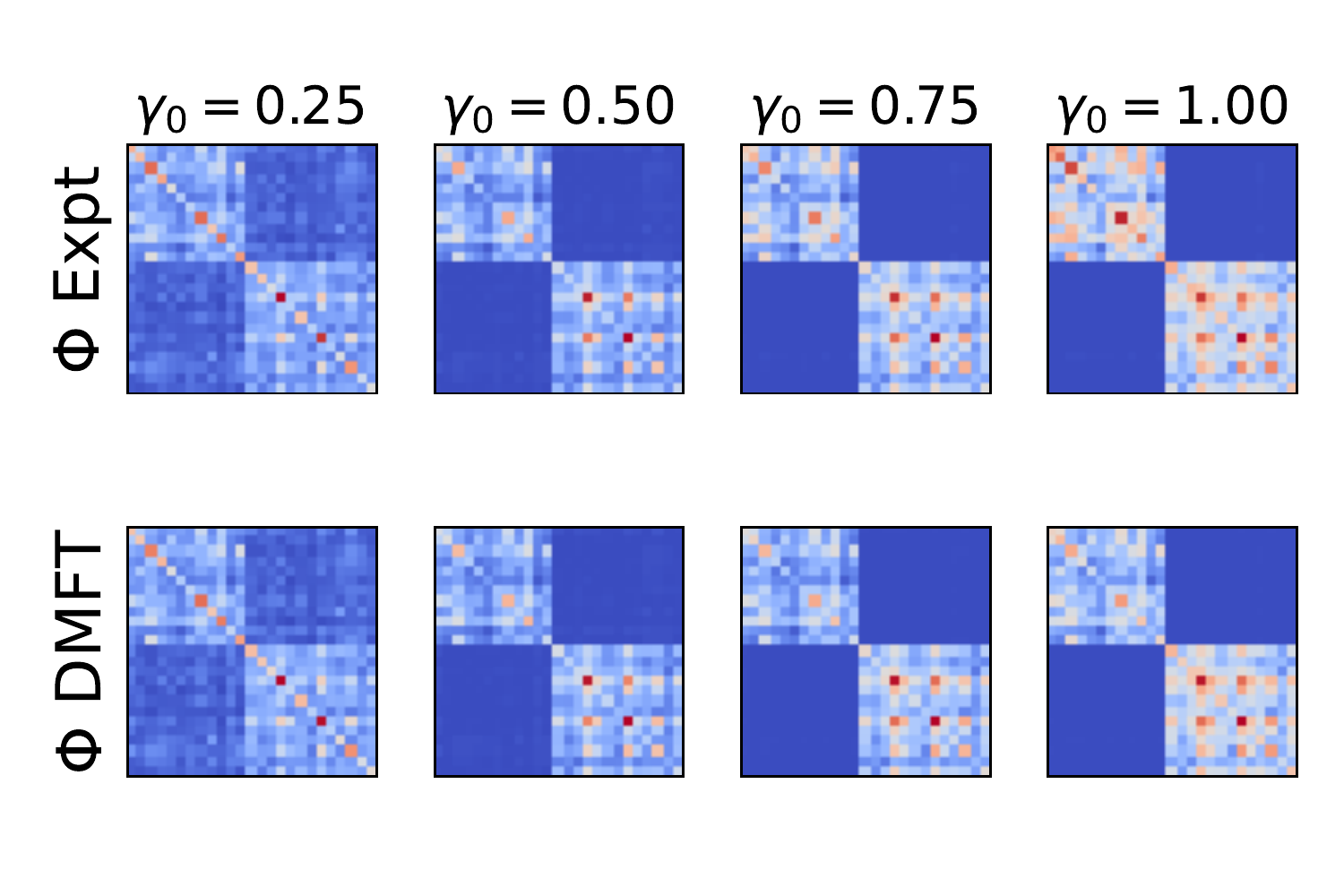}}
    \caption{Width $N=1000$ ReLU networks trained with L2 regularization have nontrivial fixed point in DMFT limit ($\gamma_0 > 0$). (a) Training loss dynamics for a $L=1$ ReLU network with $\lambda = 1$. In $\gamma_0 \to 0$ limit the fixed point is trivial $f = K = 0$. The final loss is a decreasing function of $\gamma_0$. (b) The final kernel is more aligned with target with increasing $\gamma_0$. Networks with homogenous activations enjoy a representer theorem at infinite-width as we show in Appendix \ref{app:weight_decay}.  }
    \label{fig:weight_decay}
\end{figure}

As we show in Appendix \ref{app:weight_decay}, the DMFT can be extended to networks trained with weight decay $\frac{d\bm\theta}{dt} = - \gamma^2 \nabla_{\bm\theta} \mathcal{L} - \lambda \bm\theta$. If neural network is homogenous in its parameters so that $f(c\bm\theta) = c^\kappa f(\bm\theta)$ (examples include networks with linear, ReLU, quadratic activations), then the final network predictor is a kernel regressor with the final NTK $\lim_{t\to\infty} f(\x,t) = \k(\x)^\top [ \bm K +  \lambda \kappa \I]^{-1} \y$ where $K(\x,\x')$ is the \textit{final}-NTK, $[\k(\x)]_{\mu} = K(\x,\x_\mu)$ and $[\K]_{\mu\alpha} = K(\x_\mu,\x_\alpha)$. We note that the effective regularization $\lambda \kappa$ increases with depth $L$. In NTK parameterization, weight decay in infinite width homogenous networks gives a trivial fixed point $K(\x,\x') \to 0$ and consequently a zero predictor $f \to 0$ \cite{lewkowycz2020training}. However, as we show in Figure \ref{fig:weight_decay}, increasing feature learning $\gamma_0$ can prevent convergence to the trivial fixed point, allowing a non-zero fixed point for $K,f$ even at infinite width. The kernel and function dynamics can be predicted with DMFT. The fixed point is a nontrivial function of the hyperparameters $\lambda, \kappa, L, \gamma_0$.

\section{Approximation Schemes} 

We now compare our exact DMFT with approximations of prior works, providing an explanation of when these approximations give accurate predictions and when they break down. 

\subsection{Gradient Independence Ansatz}
We can study the accuracy of the ansatz $ \A^\ell = \B^\ell = 0$, which is equivalent to treating the weight matrices $\W^\ell(0)$ and $\W^\ell(0)^\top$ which appear in forward and backward passes respectively as independent Gaussian matrices. This assumption was utilized in prior works on signal propagation in deep networks in the lazy regime \cite{poole2016exponential,schoenholz2016deep,yang2017mean, yang2019scaling, yang2020tensor2}. A consequence of this approximation is the Gaussianity and statistical independence of $\chi^\ell$ and $\xi^\ell$ (conditional on $\{\bm\Phi^\ell,\G^\ell\}$) in each layer as we show in Appendix \ref{app:grad_indep}. This ansatz works very well near $\gamma_0 \approx 0$ (the static kernel regime) since $\frac{d\h}{d\r}, \frac{d\z}{d\u} \sim \mathcal{O}(\gamma_0)$ or around initialization $t \approx 0$ but begins to fail at larger values of $\gamma_0, t$ (Figure \ref{fig:approx_comparisons}).

\subsection{Perturbation theory in $\gamma_0$ at infinite-width}

In the $\gamma_0 \to 0$ limit, we recover static  kernels, giving linear dynamics identical to the NTK limit \cite{jacot}. Corrections to this lazy limit can be extracted at small but finite $\gamma_0$. This is conceptually similar to recent works which consider perturbation series for the NTK in powers of $1/N$ \cite{dyer2019asymptotics,roberts2021principles,hanin2022correlation} (though not identical, see Appendix \ref{app:perturb_N_dmft} for finite $N$ effects). We expand all observables $q(\gamma_0)$ in a power series in $\gamma_0$, giving $q(\gamma_0) = q^{(0)} + \gamma_0 q^{(1)} + \gamma_0^2 q^{(2)} + ... $ and compute corrections up to $\mathcal O(\gamma_0^2)$. We show that the $\mathcal{O}(\gamma_0)$ and $\mathcal{O}(\gamma_0^3)$ corrections to kernels vanish, giving leading order expansions of the form $\bm\Phi = \bm\Phi^0 + \gamma_0^2 \bm\Phi^2 + \mathcal{O}(\gamma_0^4)$ and $\G = \G^0 + \gamma_0^2 \G^2 + \mathcal{O}(\gamma_0^4)$ (see Appendix \ref{app:pert_nonlin}). Further, we show that the NTK has relative change at leading order which scales linearly with depth $|\Delta K^{NTK}|/|K^{NTK,0}| \sim\mathcal{O}_{\gamma_0, L}(L \gamma_0^2) = \mathcal{O}_{N,\gamma,L}(\frac{\gamma^2 L}{N})$, which is consistent with finite width effective field theory at $\gamma=\mathcal{O}_N(1)$ \cite{naveh2021predicting,roberts2021principles, hanin2022correlation} (Appendix \ref{app:pert_leading_corr}). Further, at the leading order correction, all temporal dependencies are controlled by $P(P+1)$ functions $v_\alpha(t) = \int_0^t ds  \Delta^0_\alpha(s)$ and $v_{\alpha\beta}(t) = \int_0^t ds \Delta^0_\alpha(s) \int_0^s ds' \Delta^0_\beta(s')$, which is consistent with those derived for finite width NNs using a truncation of the Neural Tangent Hierarchy \cite{huang2020dynamics,dyer2019asymptotics, roberts2021principles}. To lighten notation, we focus our main text comparison of our non-perturbative DMFT to perturbation theory in the deep linear case. Full perturbation theory is in Appendix \ref{app:pert_nonlin}.





\begin{figure}[t]
    \centering
    \subfigure[Loss dynamics]{\includegraphics[width=0.4\linewidth]{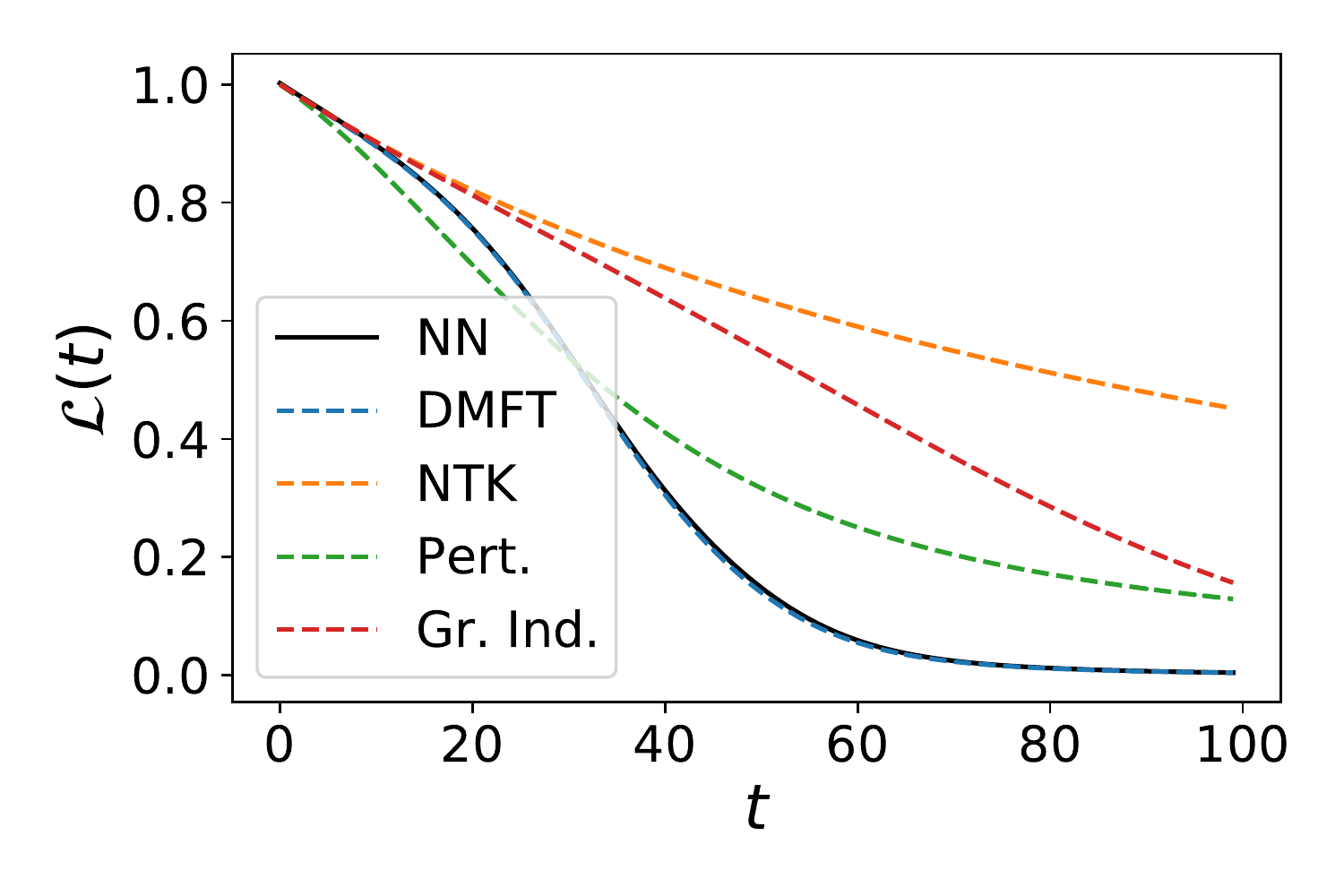}}
    \subfigure[Final $\bm H^\ell$ Kernels $\gamma_0 = 1.5$]{\includegraphics[width=0.42\linewidth]{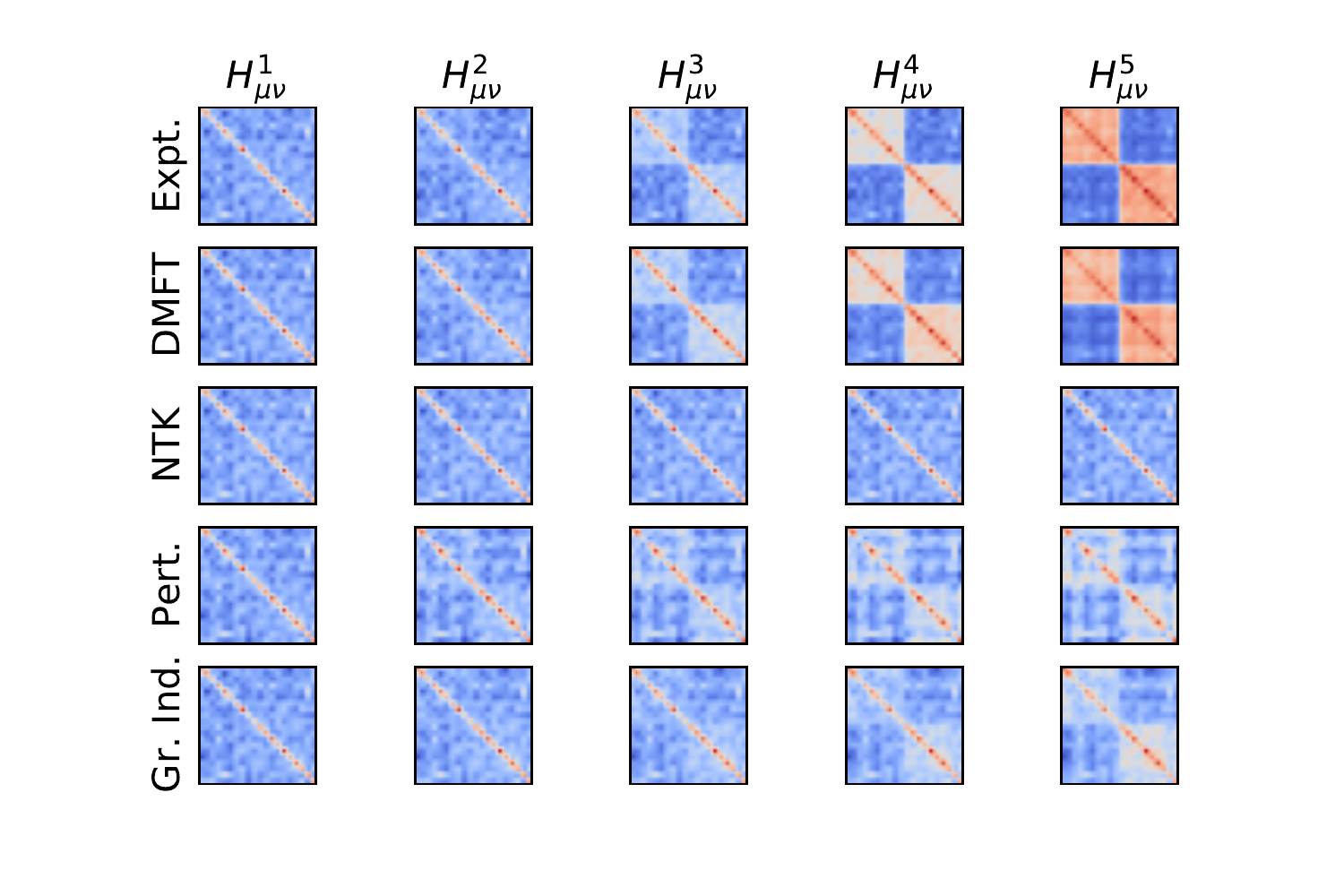}}
    \subfigure[$\bm H^\ell$ Kernel Dynamics $\gamma_0 = 1.5$]{\includegraphics[width=0.42\linewidth]{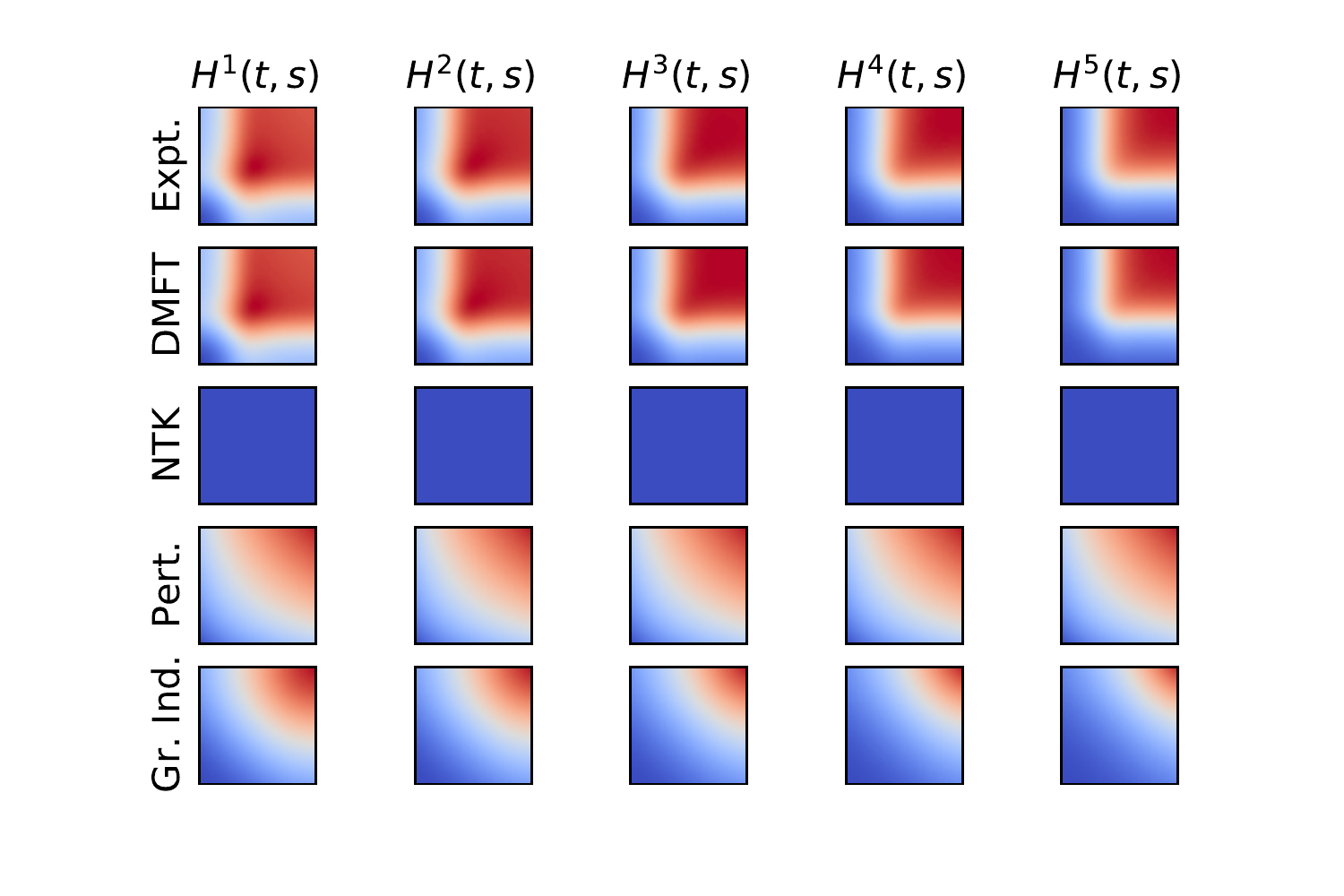}}
    \subfigure[Theory $\H^\ell$ vs NN with $N=1000$]{\includegraphics[width=0.4\linewidth]{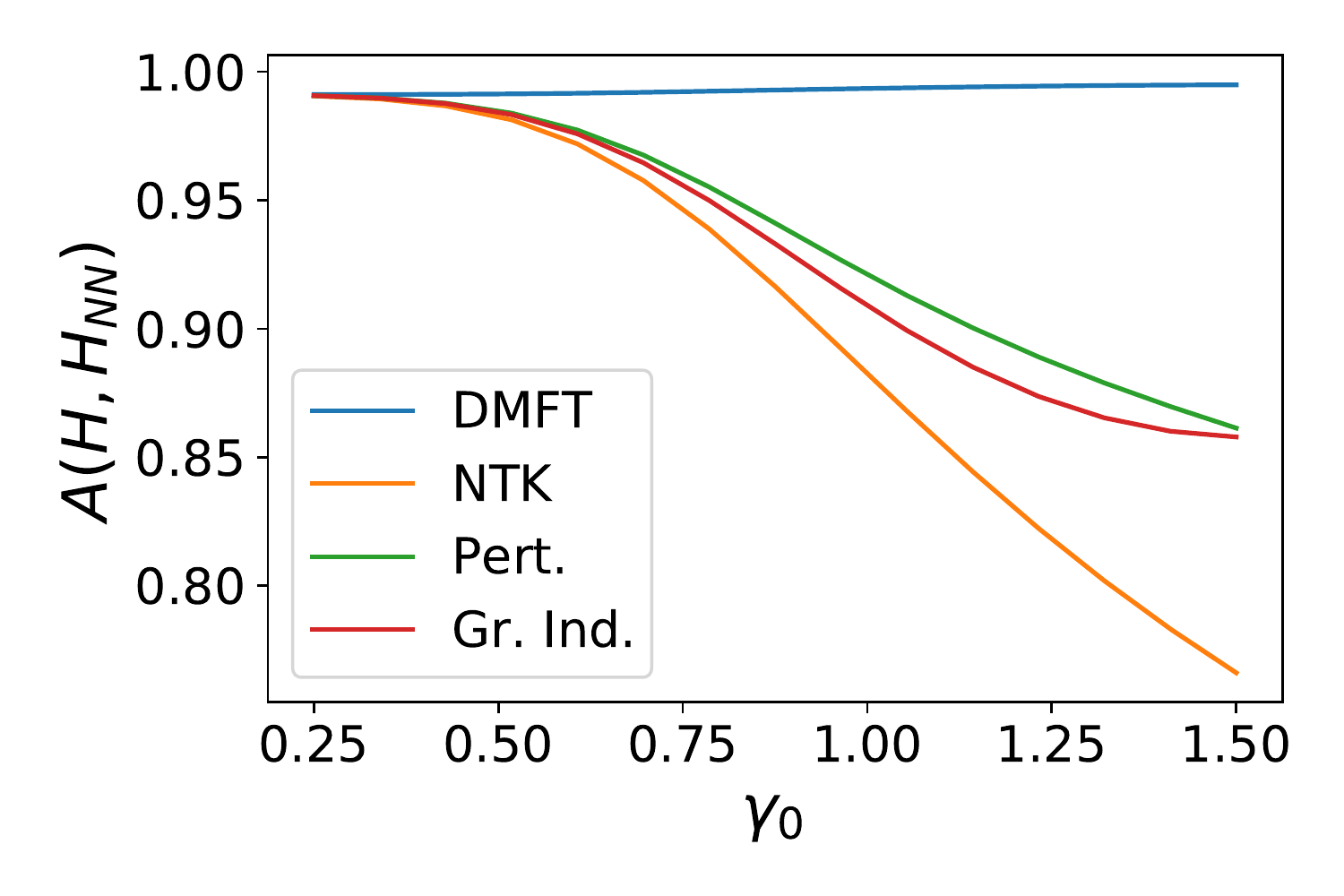}}
    \caption{Comparison of DMFT to various approximation schemes in a $L=5$ hidden layer, width $N=1000$ linear network with $\gamma_0 = 1.0$ and $P=100$. (a) The loss for the various approximations do not track the true trajectory induced by gradient descent in the large $\gamma_0$ regime. (b)-(c) The feature kernels $H^\ell_{\mu\alpha}(t,s)$ across each of the $L=5$ hidden layers for each of the theories is compared to a width $1000$ neural network. Again, we plot the sample-traced dynamics $\sum_{\mu\mu} H^\ell_{\mu\mu}(t,s)$. (d) The alignment of $\H^\ell$ compared to the finite NN $A(\H^\ell, \H^\ell_{NN})$ averaged across $\ell \in \{1,...,5\}$ for varying $\gamma$. The predictions of all of these theories coincide in the $\gamma_0 = 0$ limit but begin to deviate in the feature learning regime. Only the non-perturbative DMFT is accurate over a wide range of $\gamma_0$.  }
    \label{fig:approx_comparisons}
\end{figure}

Using the timescales derived in the previous section, we find that the leading order correction to the kernels in infinite-width deep linear network have the form
\begin{align}
    &K^{NTK}_{\mu\nu}(t,s) = (L+1) K_{\mu\nu}^x + \gamma_0^2 \frac{L(L+1)}{2} K^x_{\mu\nu} \sum_{\alpha\beta} K^x_{\alpha\beta} [v_{\alpha\beta}(t) + v_{\beta\alpha}(s) + v_{\alpha}(t) v_{\beta}(s)]   \nonumber
    \\
    &\quad + \gamma_0^2 \frac{L(L+1)}{2}\left[ \sum_{\alpha \beta} K^x_{\mu\alpha} K^x_{\nu\beta} [ v_{\alpha\beta}(t) + v_{\beta\alpha}(s)] +\sum_{\alpha\beta} K^x_{\mu\alpha} K^x_{\nu\beta} v_{\alpha}(t) v_\beta(s) \right]+ \mathcal{O}(\gamma_0^4).
\end{align}
We see that the relative change in the NTK $|\K^{NTK} - \K^{NTK}(0)|/|\K^{NTK}(0)| \sim \mathcal{O}( \gamma_0^2 L) = \mathcal{O}( \gamma^2 L /N)$, so that large depth $L$ networks exhibit more significant kernel evolution, which agrees with other perturbative studies \cite{dyer2019asymptotics,roberts2021principles, zavatone2021asymptotics} as well as the non-perturbative results in Figure \ref{fig:deep_linear_fig}. However at large $\gamma_0$ and large $L$, this theory begins to break down as we show in Figure \ref{fig:approx_comparisons}.

The DMFT formalism can also be used to extract leading corrections to observables at large but finite width $N$ as we explore in \ref{app:perturb_N_dmft}. When deviating from infinite width, the kernels are no longer deterministic over network initializations. The key observation is that the DMFT action $S$ defines a Gibbs measure over the space of kernel order parameters $\bm k = \text{Vec} \{ \bm\Phi^\ell, \G^\ell, \A^\ell,\B^\ell\}$ with probability density $\frac{1}{Z} \exp\left( N S[\bm k] \right)$ where $Z$ is a normalization constant. Near infinite width, any observable average $\left< O(\bm k) \right> = \frac{1}{Z} \int d\bm k  \exp\left( N S[\bm k] \right) O(\bm k)$ is dominated by order parameters within a $\frac{1}{\sqrt N}$ neighborhood of $\bm k^*$. As a consequence, a perturbative series for $\left< O(\bm k) \right>$ can be obtained from simple averages over Gaussian fluctuations in the kernels $\bm k \sim \mathcal{N}(\bm k^*, -\frac{1}{N} [\nabla^2 S[\bm k^*]]^{-1})$ \cite{segadlo_nngp_field}. The components for $\nabla^2 S[\bm k^*]$ include four point correlations of fields computed over the DMFT distribution. 

\section{Feature Learning Dynamics is Preserved at Fixed $\gamma_0$}

\begin{figure}[t]
    \centering
    \subfigure[Test MSE]{\includegraphics[width=0.32 \linewidth]{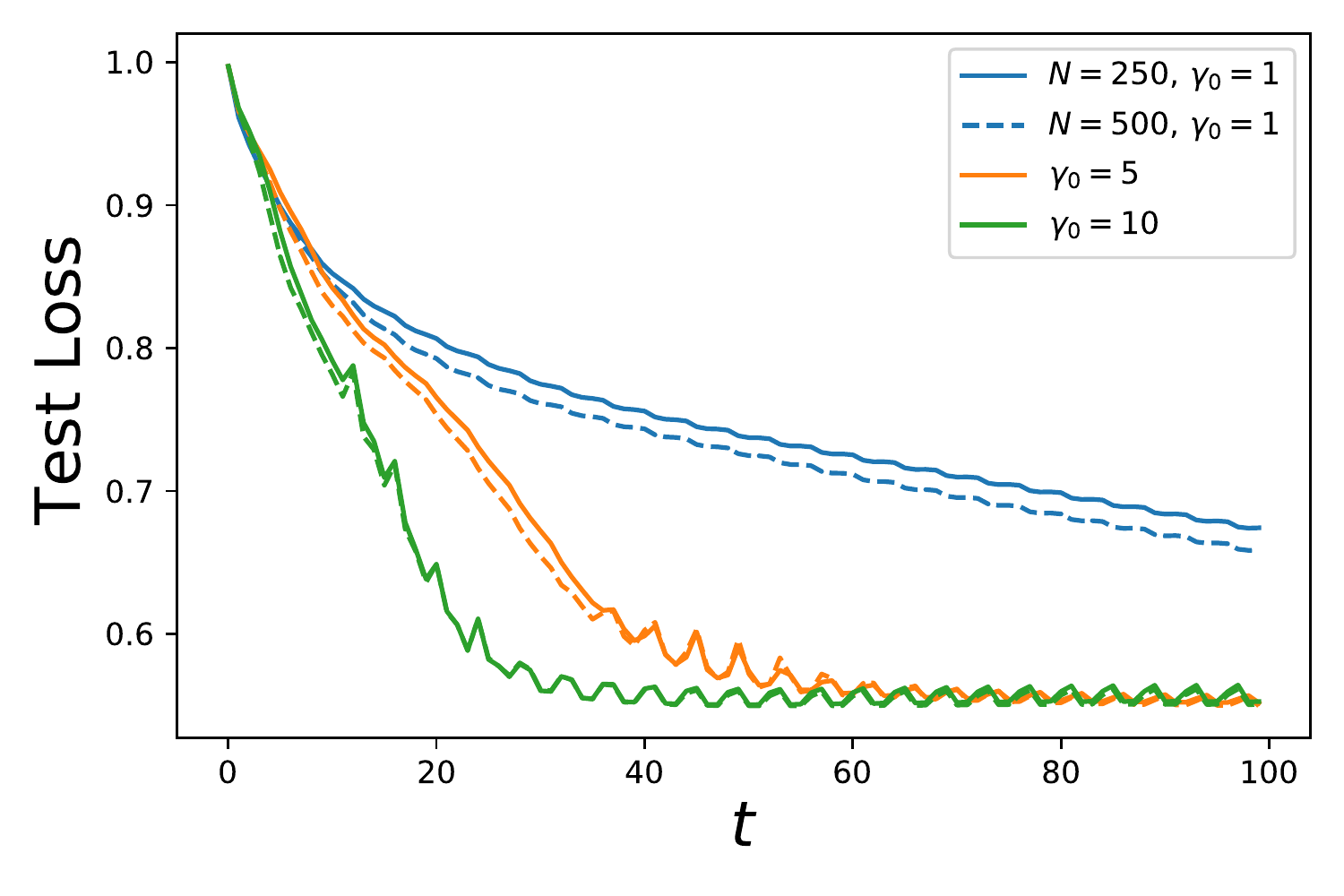}}
    \subfigure[Classification Error]{\includegraphics[width=0.32\linewidth]{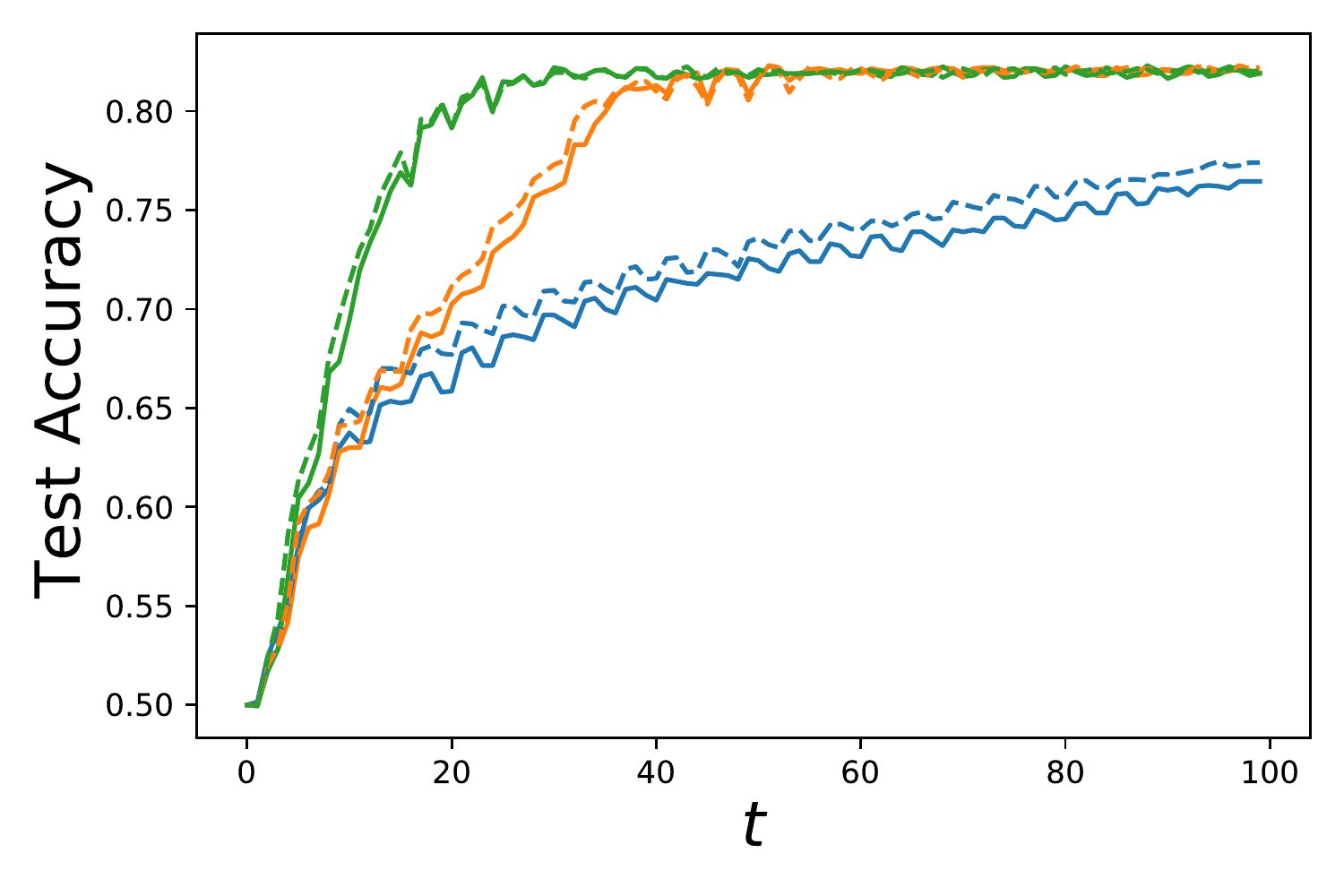}}
    \subfigure[ $A(\Phi^L, y y^\top)$ Dynamics ]{\includegraphics[width=0.32\linewidth]{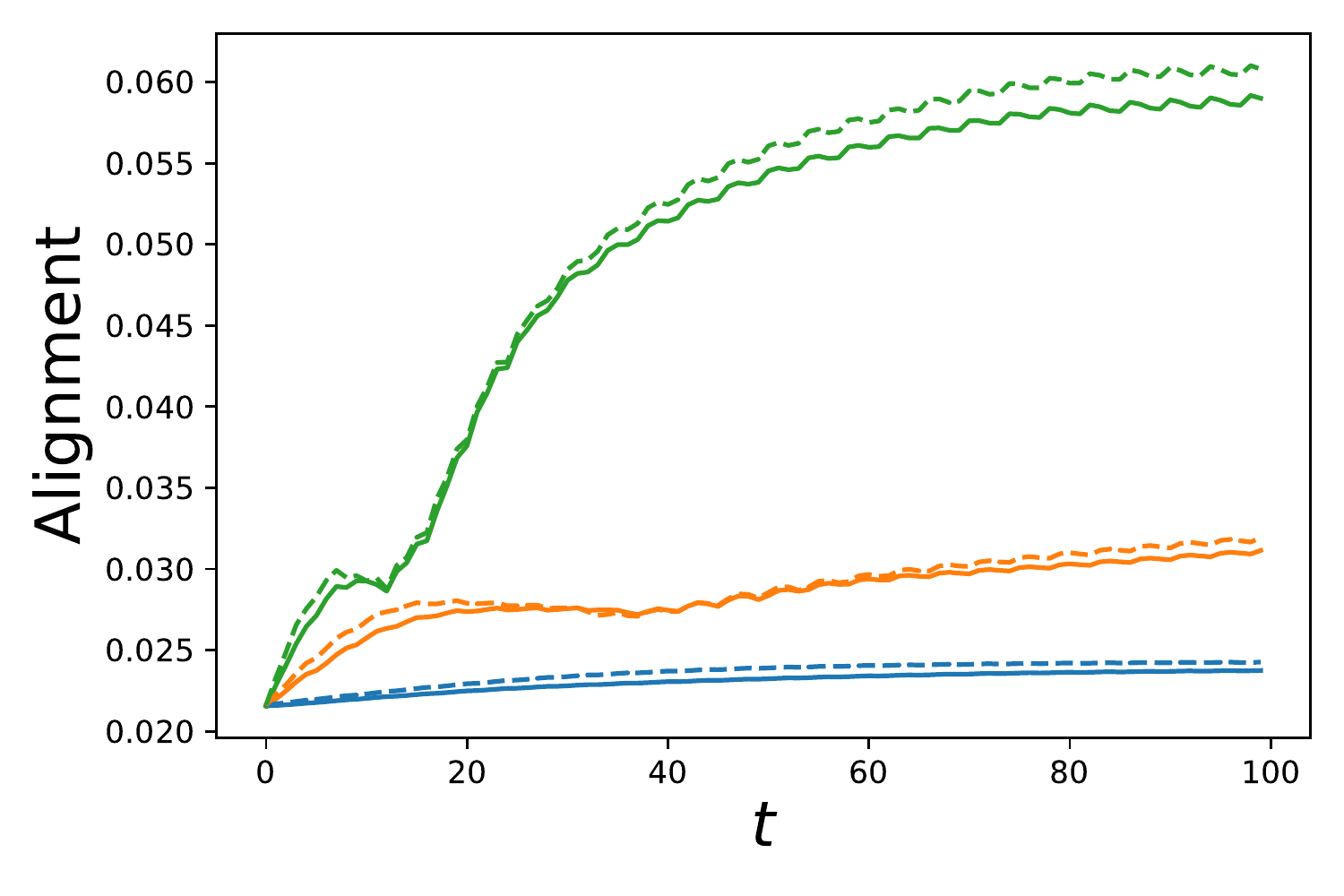}}
    \caption{The dynamics of a depth $5$ ($L=4$ hidden) CNNs trained on first two classes of CIFAR (boat vs plane) exhibit consistency for different channel counts $N \in \{250,500\}$ for fixed $\gamma_0 = \gamma / \sqrt{N}$. (a) We plot the test loss (MSE) and (b) test classification error. Networks with higher $\gamma_0$ train more rapidly. Time is measured in every $100$ update steps. (c) The dynamics of the last layer feature kernel $\Phi^L$, shown as alignment to the target function. As predicted by the DMFT, higher $\gamma_0$ corresponds to more active kernel evolution, evidenced by larger change in the alignment.}
    \label{fig:cifar_gamma_sweep}
\end{figure}

Our DMFT suggests that for networks sufficiently wide for their kernels to concentrate, the dynamics of loss and kernels should be invariant under the rescaling $N \to R N, \gamma \to \gamma / \sqrt{R}$, which keeps $\gamma_0$ fixed. To evaluate how well this idea holds in a realistic deep learning problem, we trained CNNs of varying channel counts $N$ on two-class CIFAR classification \cite{krizhevsky2009learning}. We tracked the dynamics of the loss and the last layer $\Phi^L$ kernel. The results are provided in Figure \ref{fig:cifar_gamma_sweep}. We see that dynamics are largely independent of rescaling as predicted. Further, as expected, larger $\gamma_0$ leads to larger changes in kernel norm and faster alignment to the target function $y$, as was also found in \cite{shan_bordelon}. Consequently, the higher $\gamma_0$ networks train more rapidly. The trend is consistent for width $N = 250$ and $N=500$. 
More details about the experiment can be found in Appendix \ref{app:expt_detail_cnn}.

\section{Discussion}
We provided a unifying DMFT derivation of feature dynamics in infinite networks trained with gradient based optimization. Our theory interpolates between lazy infinite-width behavior of a static NTK in $\gamma_0 \to 0$ and rich feature learning. At $\gamma_0 = 1$, our DMFT construction agrees with the stochastic process derived previously with the Tensor Programs framework \cite{yang2021tensor}.
Our saddle point equations give self-consistency conditions which relate the stochastic fields to the kernels. These equations are exactly solveable in deep linear networks and can be efficiently solved with a numerical method in the nonlinear case. Comparisons with other approximation schemes show that DMFT can be accurate at a much wider range of $\gamma_0$. We believe our framework could be a useful perspective for future theoretical analyses of feature learning and generalization in wide networks. 

\label{sec:discussion}
Though our DMFT is quite general in regards to the data and architecture, the technique is not entirely rigorous and relies on heuristic physics techniques. Our theory holds in the $T,P = \mathcal{O}_N(1)$ and may break down otherwise; other asymptotic regimes (such as $P/N, T/\log(N)=\mathcal{O}_N(1)$, etc) may exhibit phenomena relevant to deep learning practice \cite{li2021statistical,dascoli_dmft_optimal_lr_schedule}. The computational requirements of our method, while smaller than the exponential time complexity for exact solution \cite{yang2021tensor}, are still significant for large $P T$. In Table \ref{tab:compute_comparison}, we compare the time taken for various theories to compute the feature kernels throughout $T$ steps of gradient descent. For a width $N$ network, computation of each forward pass on all $P$ data points takes $\mathcal{O}(P N^2)$ computations. The static NTK requires computation of $\mathcal{O}(P^2)$ entries in the kernel which do not need to be recomputed. However, the DMFT requires matrix multiplications on $PT \times PT$ matrices giving a $\mathcal{O}(P^3 T^3)$ time scaling. Future work could aim to improve the computational overhead of the algorithm, by considering data averaged theories \cite{mignacco2020dynamical} or one pass SGD \cite{yang2021tensor}. Alternative projected versions of gradient descent have also enabled much better computational scaling in evaluation of the theoretical predictions \cite{yang2022efficient}, allowing evaluation on full CIFAR-10. %


\begin{table}[H]
    \centering
    \begin{tabular}{|c|c|c|c|c|c|c|}
    \hline
        Requirements & Width-$N$ NN & Static NTK & Perturbative & Full DMFT  \\ \hline
        Memory for Kernels & $\mathcal O(N^2)$ & $\mathcal O(P^2 )$ & $\mathcal O(P^4 T)$ & $\mathcal O(P^2 T^2)$\\ \hline
        Time for Kernels & $\mathcal O(P  N^2 T)$ & $\mathcal O(P^2)$ & $\mathcal O(P^4 T)$ & $\mathcal O(P^3 T^3)$ \\ \hline 
        Time for Final Outputs & $\mathcal{O}(P N^2 T)$ & $\mathcal{O}(P^3)$ & $\mathcal{O}(P^4)$ & $\mathcal{O}(P^3 T^3)$ \\ \hline
    \end{tabular}
    \caption{Computational requirements to compute kernel dynamics and trained network predictions on $P$ points in a depth $N$ neural network on a grid of $T$ time points trained with $P$ data points for various theories. DMFT is faster and less memory intensive than a width $N$ network only if $N \gg PT$. It is more computationally efficient to compute full DMFT kernels than leading order perturbation theory when $T \ll \sqrt{P}$. The expensive scaling with both samples and time are the cost of a full-batch non-perturbative theory of gradient based feature learning dynamics.}
    \label{tab:compute_comparison}
\end{table}

\begin{ack}
This work was supported by NSF grant DMS-2134157 and an award from the Harvard Data Science Initiative Competitive Research Fund. BB acknowledges additional support from the NSF-Simons Center for Mathematical and Statistical Analysis of Biology at Harvard (award \#1764269) and the Harvard Q-Bio Initiative. 

BB thanks Jacob Zavatone-Veth, Alex Atanasov, Abdulkadir Canatar, and Ben Ruben for comments on this manuscript as well as Greg Yang, Boris Hanin, Yasaman Bahri, and Jascha Sohl-Dickstein for useful discussions.


\end{ack}

\bibliographystyle{unsrt}
\bibliography{mybib}

\section*{Checklist}


\begin{enumerate}

\item For all authors...
\begin{enumerate}
  \item Do the main claims made in the abstract and introduction accurately reflect the paper's contributions and scope?
    \answerYes{As described in the abstract and introduction, we provide a dynamical field theory of deep networks based on kernel evolution.}
  \item Did you describe the limitations of your work?
    \answerYes{We have an explicit limitations as the last paragraph of the paper in Section \ref{sec:discussion}.}
  \item Did you discuss any potential negative societal impacts of your work?
    \answerNA{This work is theoretical and is very unlikely to present negative social impacts.}
  \item Have you read the ethics review guidelines and ensured that your paper conforms to them?
    \answerYes{}
\end{enumerate}

\item If you are including theoretical results...
\begin{enumerate}
  \item Did you state the full set of assumptions of all theoretical results?
    \answerYes{We describe that our theory holds for NN architectures in the infinite-width $N \to \infty$ limit.}
        \item Did you include complete proofs of all theoretical results?
    \answerYes{All claims made in the main text are supported by derivations in the Appendix.}
\end{enumerate}

\item If you ran experiments...
\begin{enumerate}
  \item Did you include the code, data, and instructions needed to reproduce the main experimental results (either in the supplemental material or as a URL)?
    \answerYes{Code to reproduce experimental results is provided in the supplementary material.}
  \item Did you specify all the training details (e.g., data splits, hyperparameters, how they were chosen)?
    \answerYes{We provide details of all experiments in \ref{app:expt_detail}.}
        \item Did you report error bars (e.g., with respect to the random seed after running experiments multiple times)?
    \answerYes{We provided errorbars in the alignment scores of DMFT as a function of width $N$ in Figure \ref{fig:deep_tanh_visual_kernels}. All other runs were over a single wide network, where performance is predicted to concentrate over initialization.}
        \item Did you include the total amount of compute and the type of resources used (e.g., type of GPUs, internal cluster, or cloud provider)?
    \answerYes{We mention our GPU usage in \ref{app:expt_detail_cnn}.}
\end{enumerate}

\item If you are using existing assets (e.g., code, data, models) or curating/releasing new assets...
\begin{enumerate}
  \item If your work uses existing assets, did you cite the creators?
    \answerYes{We cited the creators of Jax, Neural Tangents, and CIFAR-10.}
  \item Did you mention the license of the assets?
    \answerNA{These are all open source provided they are appropriately credited in academic research.}
  \item Did you include any new assets either in the supplemental material or as a URL?
    \answerNA{}
  \item Did you discuss whether and how consent was obtained from people whose data you're using/curating?
    \answerNA{}
  \item Did you discuss whether the data you are using/curating contains personally identifiable information or offensive content?
    \answerNA{}
\end{enumerate}

\item If you used crowdsourcing or conducted research with human subjects...
\begin{enumerate}
  \item Did you include the full text of instructions given to participants and screenshots, if applicable?
    \answerNA{}
  \item Did you describe any potential participant risks, with links to Institutional Review Board (IRB) approvals, if applicable?
    \answerNA{}
  \item Did you include the estimated hourly wage paid to participants and the total amount spent on participant compensation?
    \answerNA{}
\end{enumerate}

\end{enumerate}

\pagebreak
\appendix
\section*{Appendix}

\section{Additional Figures}

\begin{figure}[H]
    \centering
    \subfigure[Loss Dynamics]{\includegraphics[width=0.32\linewidth]{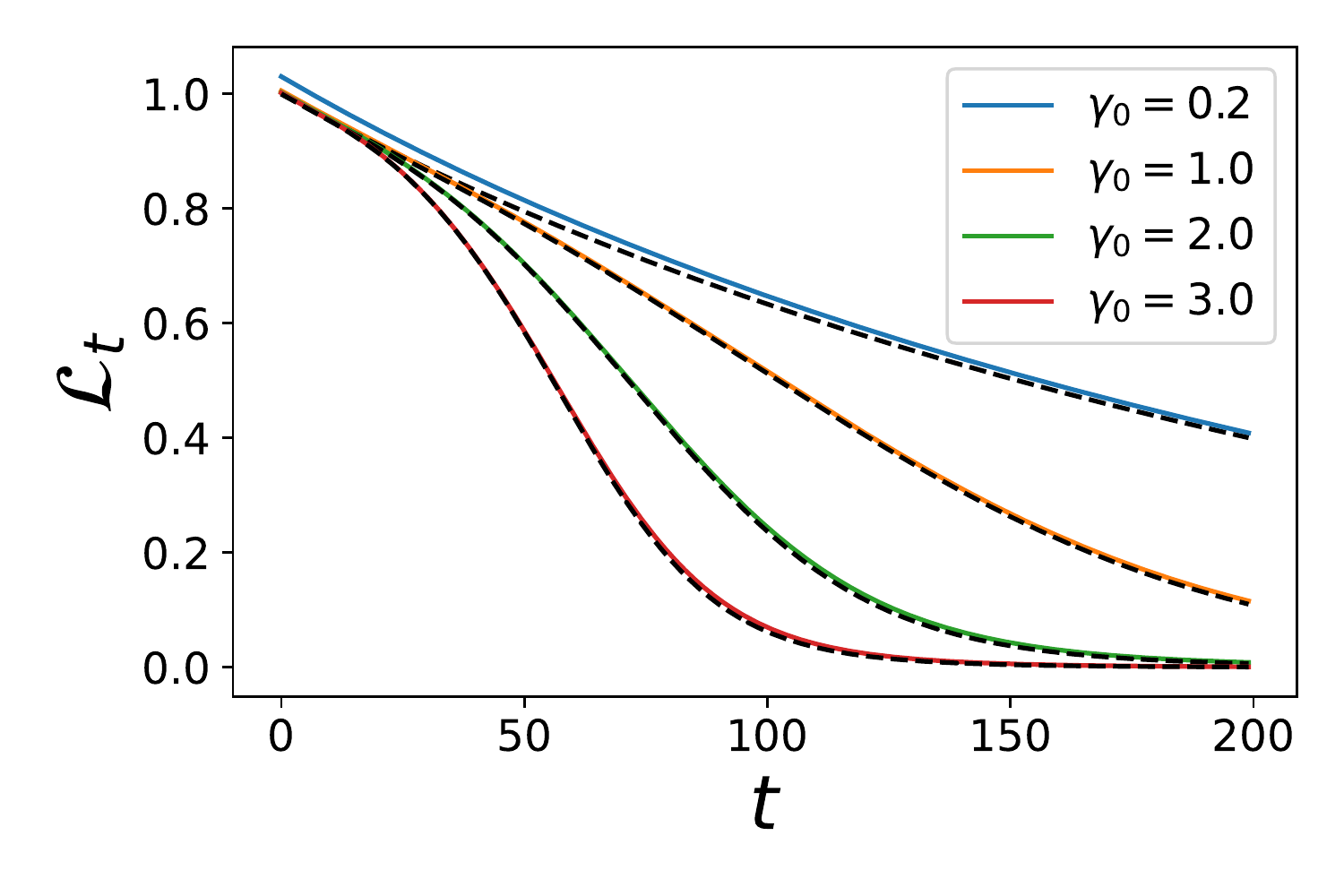}}
    \subfigure[Final $h$ Distribution]{\includegraphics[width=0.32\linewidth]{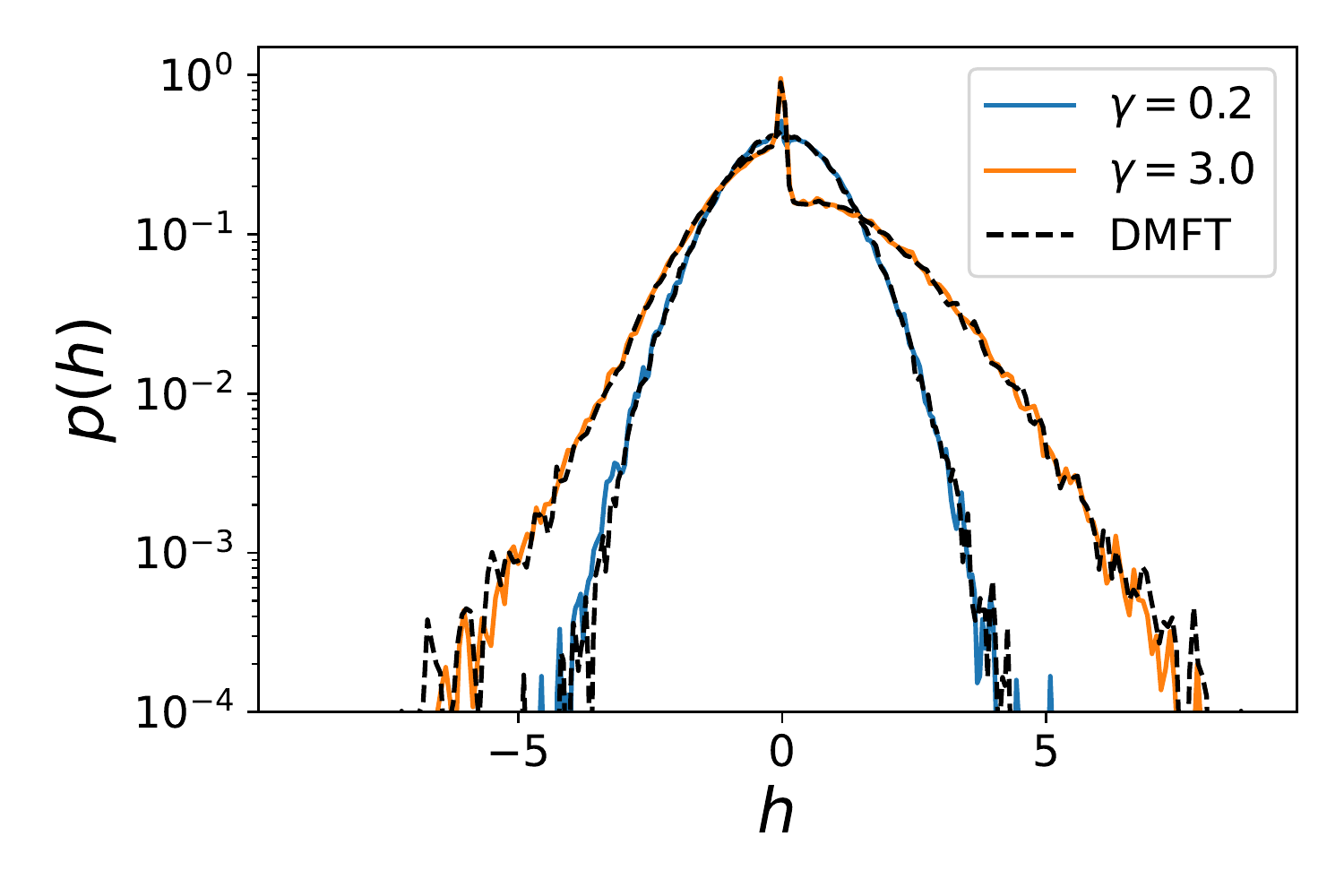}}
    \subfigure[Final $z$ Distribution]{\includegraphics[width=0.32\linewidth]{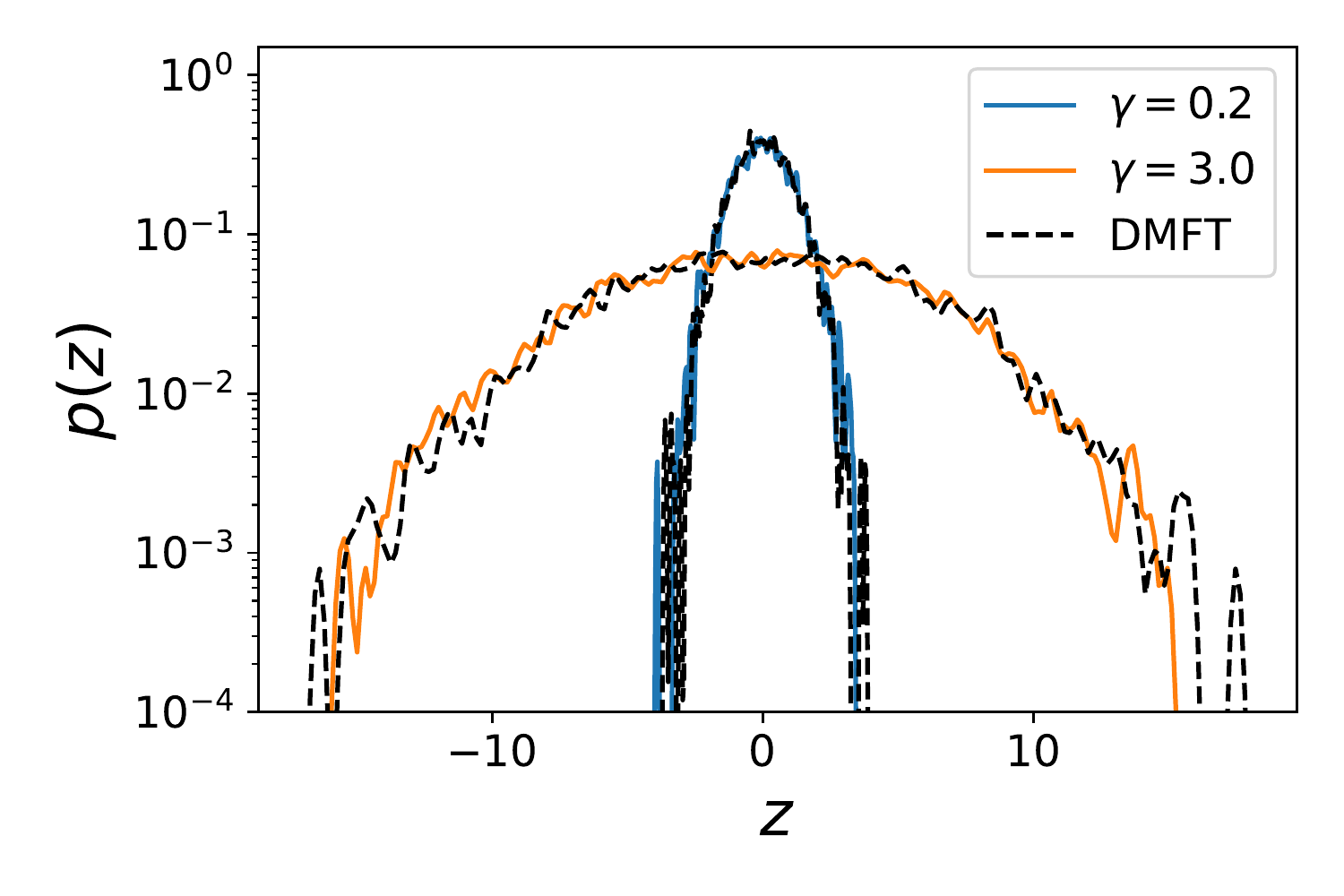}}
    \subfigure[Final $\Phi^1$ Kernels]{\includegraphics[width=0.46\linewidth]{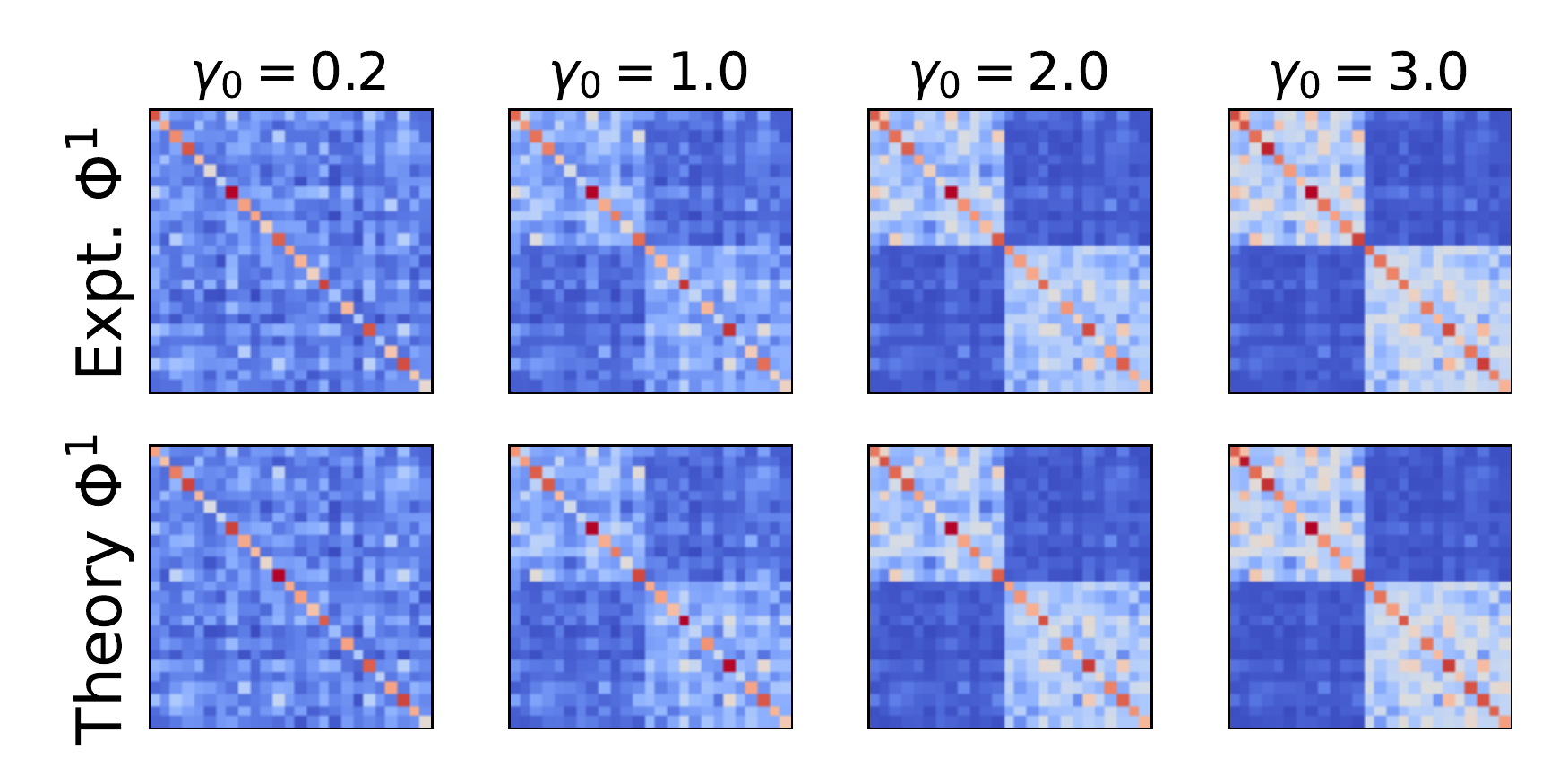}}
    \subfigure[Final $G^1$ Kernels]{\includegraphics[width=0.46\linewidth]{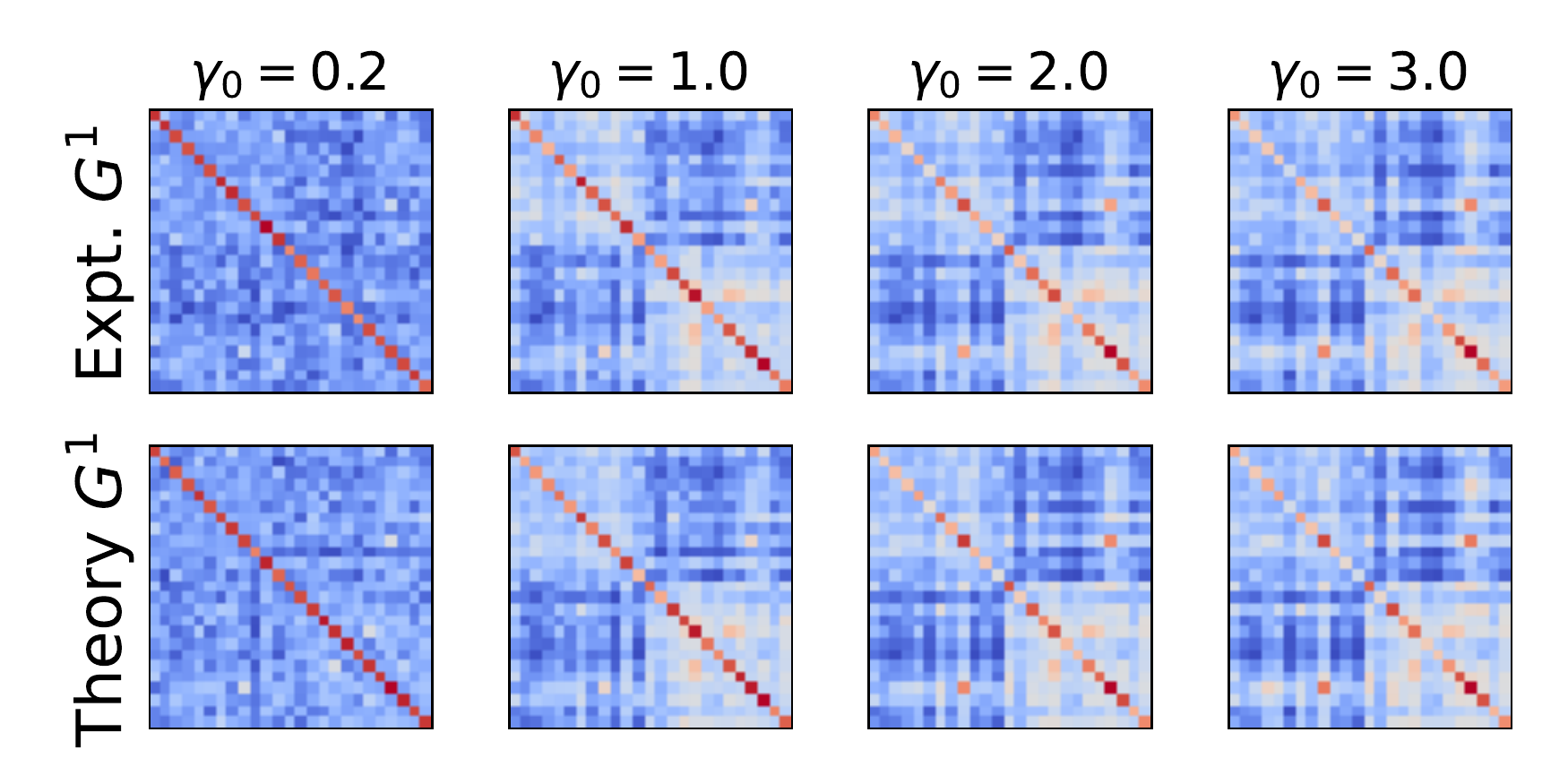}}
    \caption{ Self-consistent DFT reproduces two layer ($L=1$ hidden layer, width $N=2000$) ReLU NN's preactivation density, loss dynamics and learned kernel. (a) The loss is obtained by taking saddle point results for $\Phi,G$ and calculating the NTK's dynamics. The $\gamma_0 \to 0$ limit is governed by a static NTK, while the $\gamma_0 > 0$ network exhibits kernel evolution and accelerated training. (b) We plot the preactivation $h$ distribution for neurons in the hidden layer of the trained NN against the theoretical densities defined by $\mathcal Z[\Phi,G]$. For small $\gamma_0$, the final distribution is approximately Gaussian, but becomes non-Gaussian, asymmetric, and heavy tailed for large $\gamma_0$. The DMFT estimate of the distribution is noisy due to finite sampling error. (c) The pre-gradient distribution $p(z)$ in the trained network has larger final variance for large $\gamma_0$. (d)-(e) The final $\Phi,G$ are accurately predicted by the field theory and exhibit a block structure that increases with $\gamma_0$ due to feature learning. }
    \label{fig:two_layer_relu}
\end{figure}

\begin{figure}[H]
    \centering
    \subfigure[Lazy vs Rich Loss Dynamics]{\includegraphics[width=0.4\linewidth]{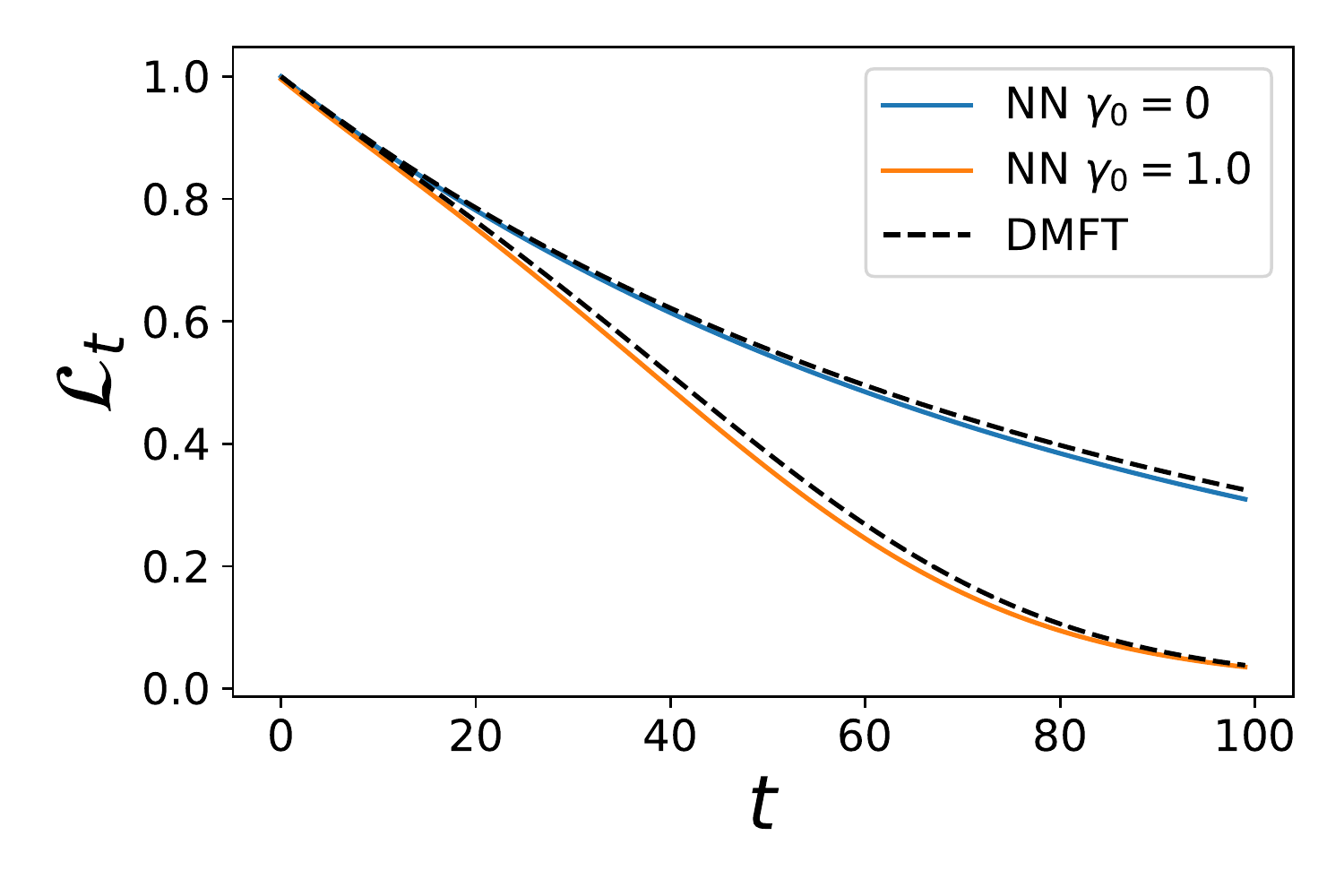}}
    \subfigure[Final $\Phi$ Kernels $\gamma_0 = 1$]{\includegraphics[width=0.4\linewidth]{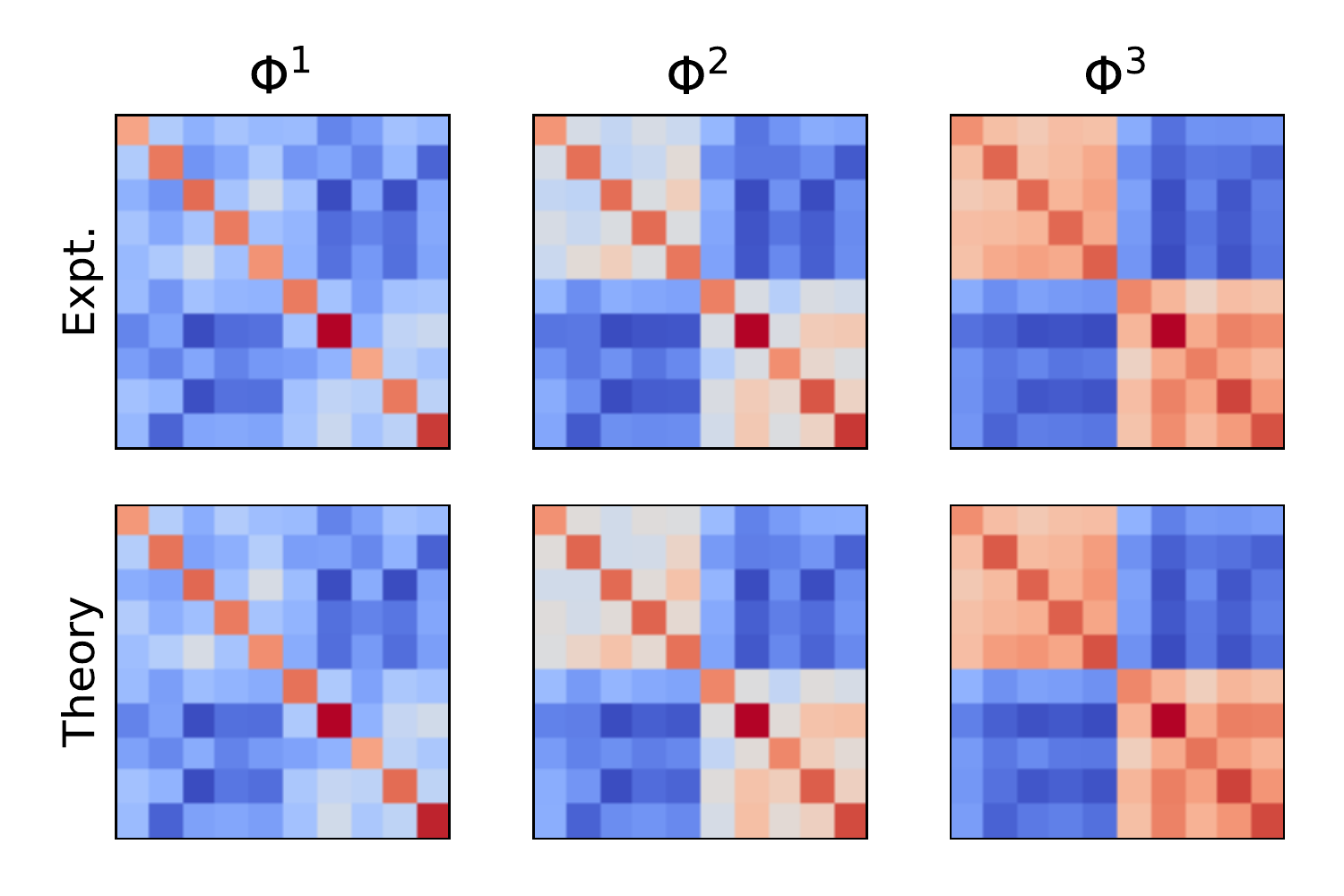}}
    \subfigure[Final $G$ Kernels, $\gamma_0 = 1$]{\includegraphics[width=0.4\linewidth]{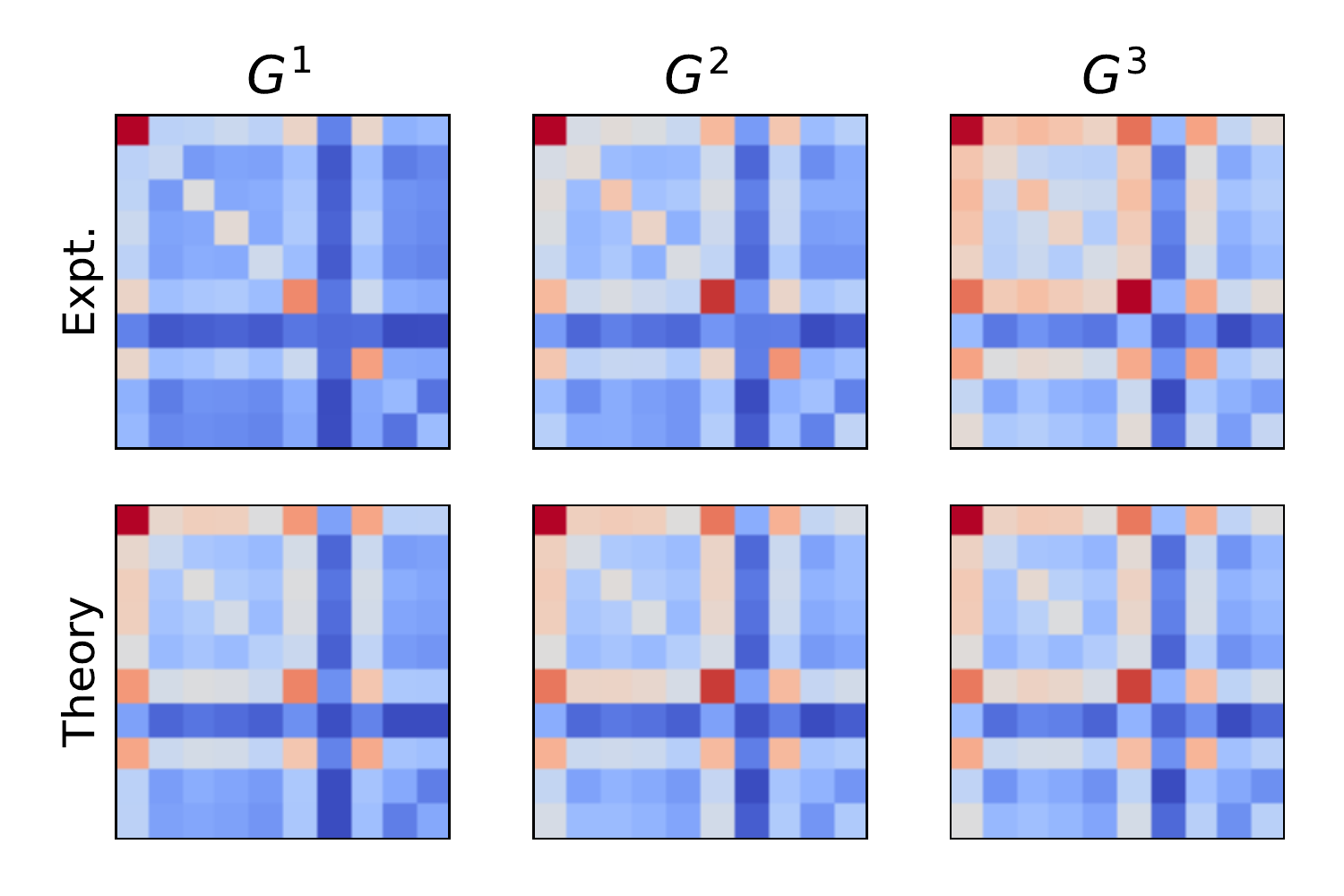}}
    \subfigure[$\Phi,G$ Temporal Dynamics $\gamma_0 = 1$]{\includegraphics[width=0.4\linewidth]{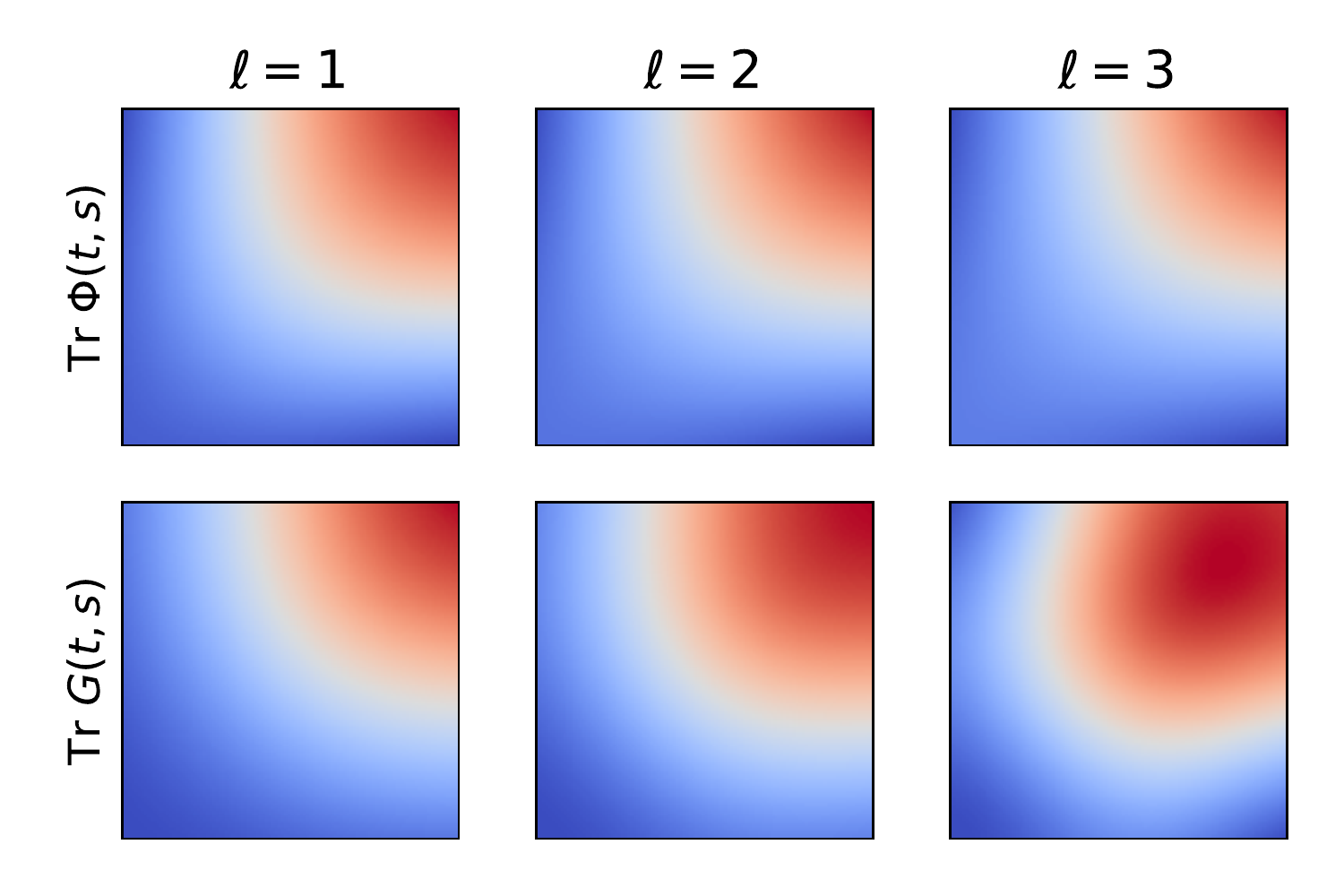}}
    \caption{ Self-consistent DFT reproduces loss dynamics, and kernels through time in a $L=3$ tanh network. (a) The loss when training on synthetic data is obtained by taking saddle point results for $\Phi,G$ and calculating the NTK's dynamics. The $\gamma_0 \to 0$ limit is governed by a static NTK, while the $\gamma_0 > 0$ network exhibits kernel evolution and accelerated training. Solid lines are a $N=2000$ NN and dashed lines are from solving DMFT equations. (b)-(c) The final learned kernels $\Phi$ (b) and $G$ (c) are accurately predicted by the field theory and exhibits block structure due to clustering by class identity. (d) The temporal components of $\Phi,G$ reveals nontrivial dynamical structure. }
    \label{fig:dmft_tanh_depth_4}
\end{figure}

\begin{figure}[H]
    \centering
    \subfigure[Two Layer Error Dynamics]{\includegraphics[width=0.4\linewidth]{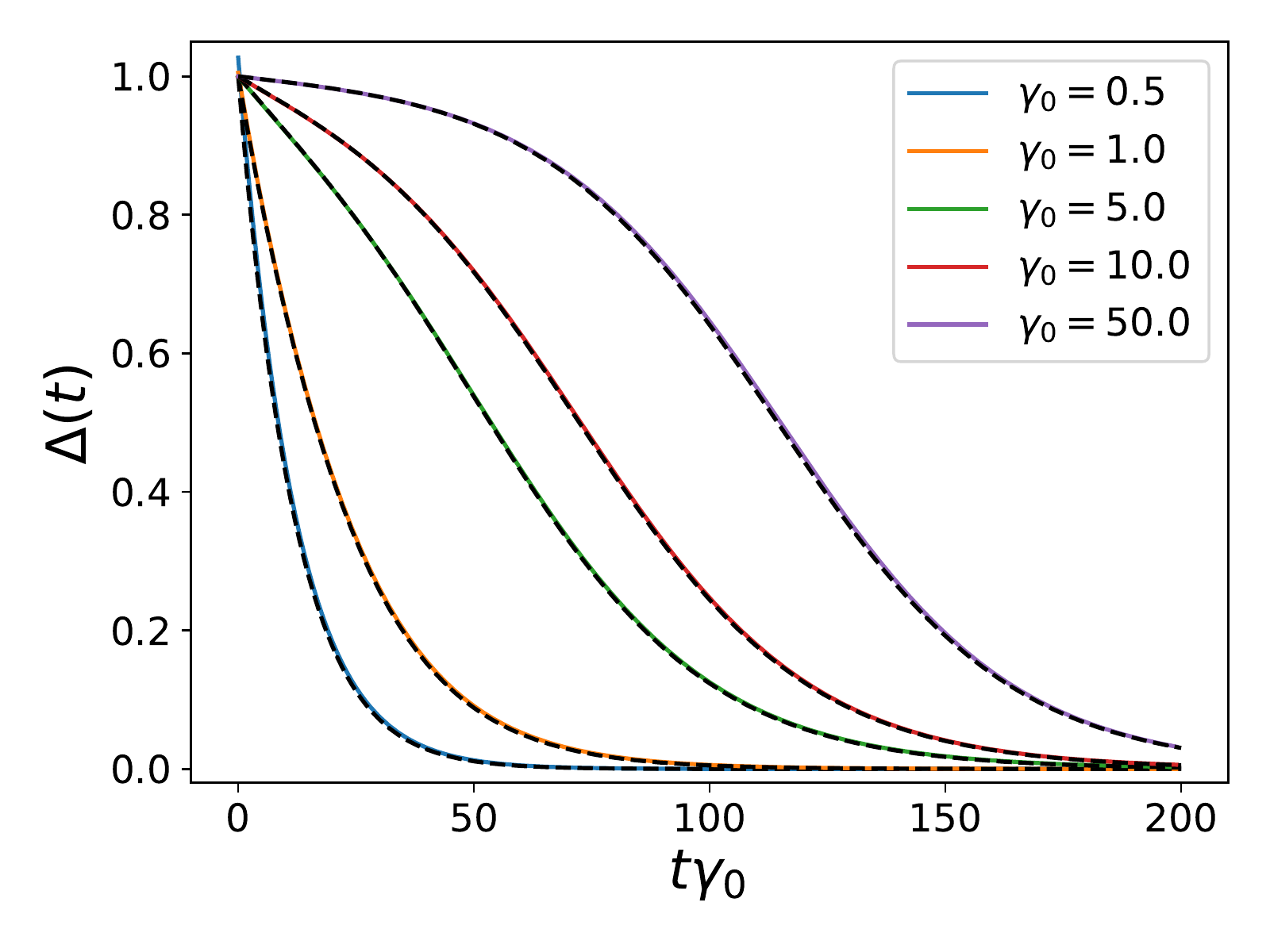}}
    \subfigure[Projection on Target]{\includegraphics[width=0.4\linewidth]{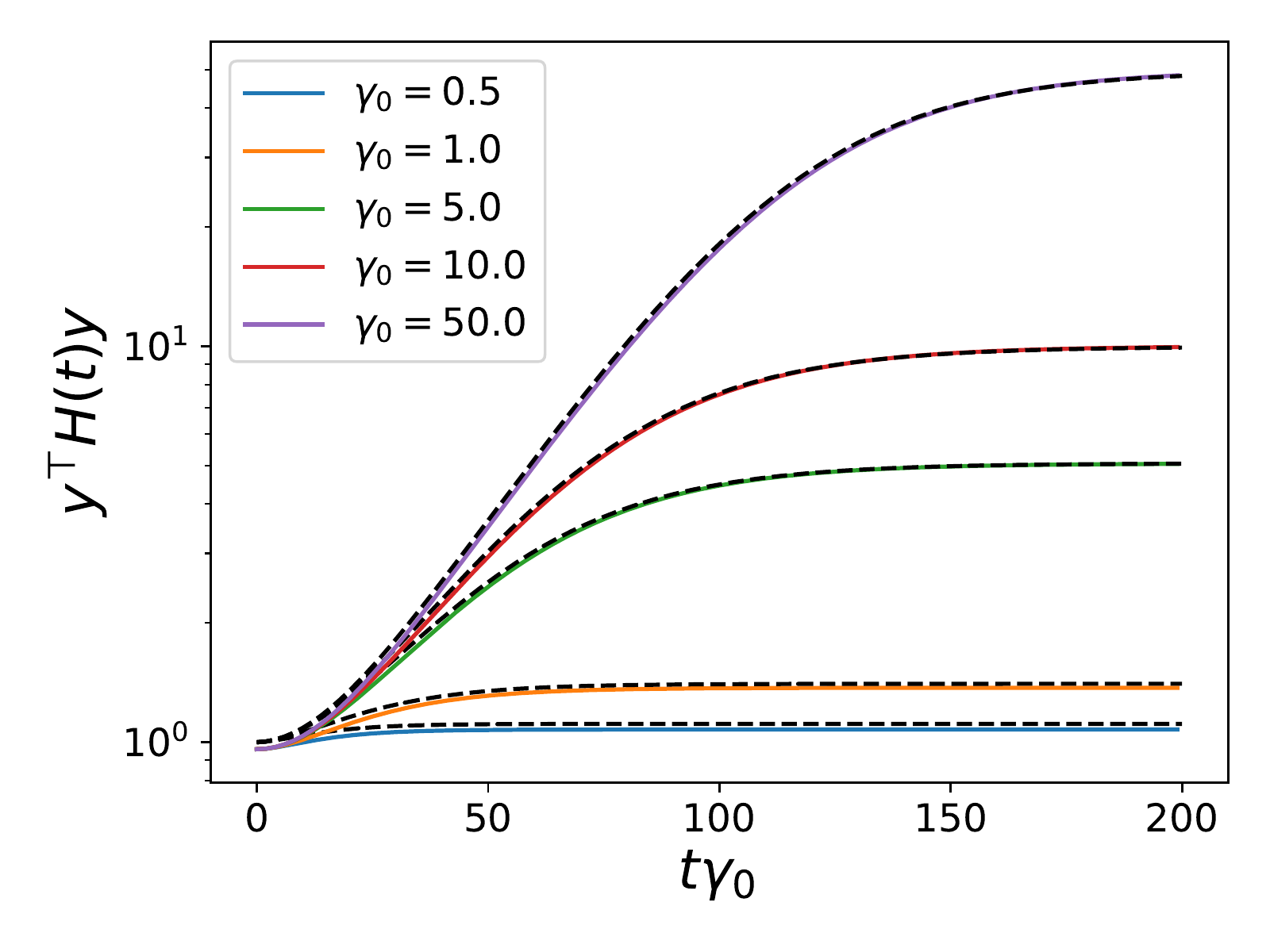}}
    \caption{The error and kernel dynamics obtained by solving a one dimensional ODE system for a depth 2 linear network. (a) The $\Delta(t)$ error dynamics from \ref{app:two_layer_whitened} allows one to solve for $\bm H(t)$ by solving a one dimensional ODE at each value of $\gamma_0$. The learning curves interpolate between exponential convergence at small $\gamma_0$ and logistic sigmoidal trajectories at large $\gamma_0$. (b) The projection of the kernel $\bm H(t)$ along the task relevant subspace $\bm y \in \mathbb{R}^P$. }
    \label{fig:err_H_dynamics_2layer}
\end{figure}

\begin{figure}[H]
    \centering
    \subfigure[Grad. Independence DMFT]{\includegraphics[width=0.32\linewidth]{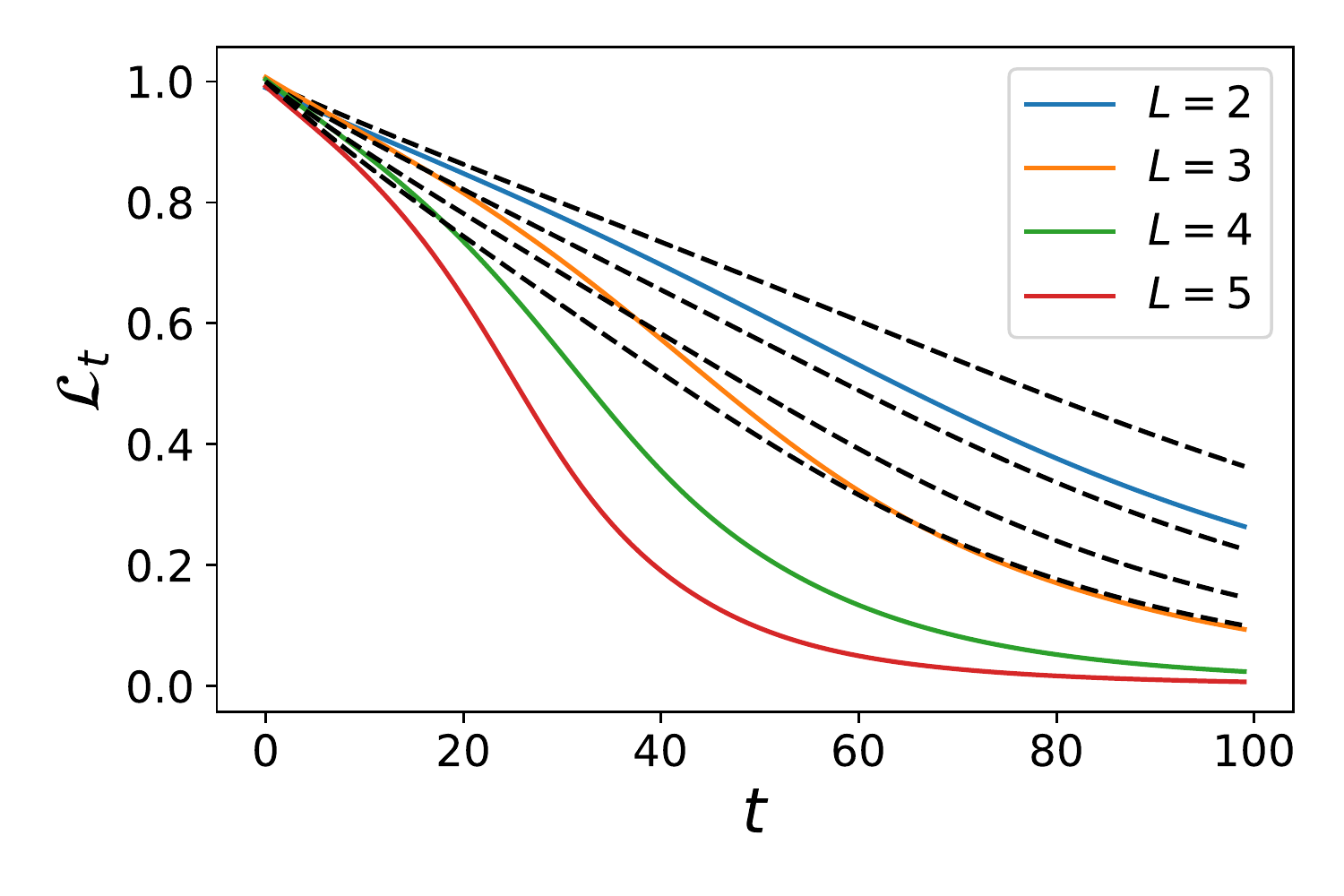}}
    \subfigure[Grad. Independence Predicted Feature Kernels]{\includegraphics[width=0.6\linewidth]{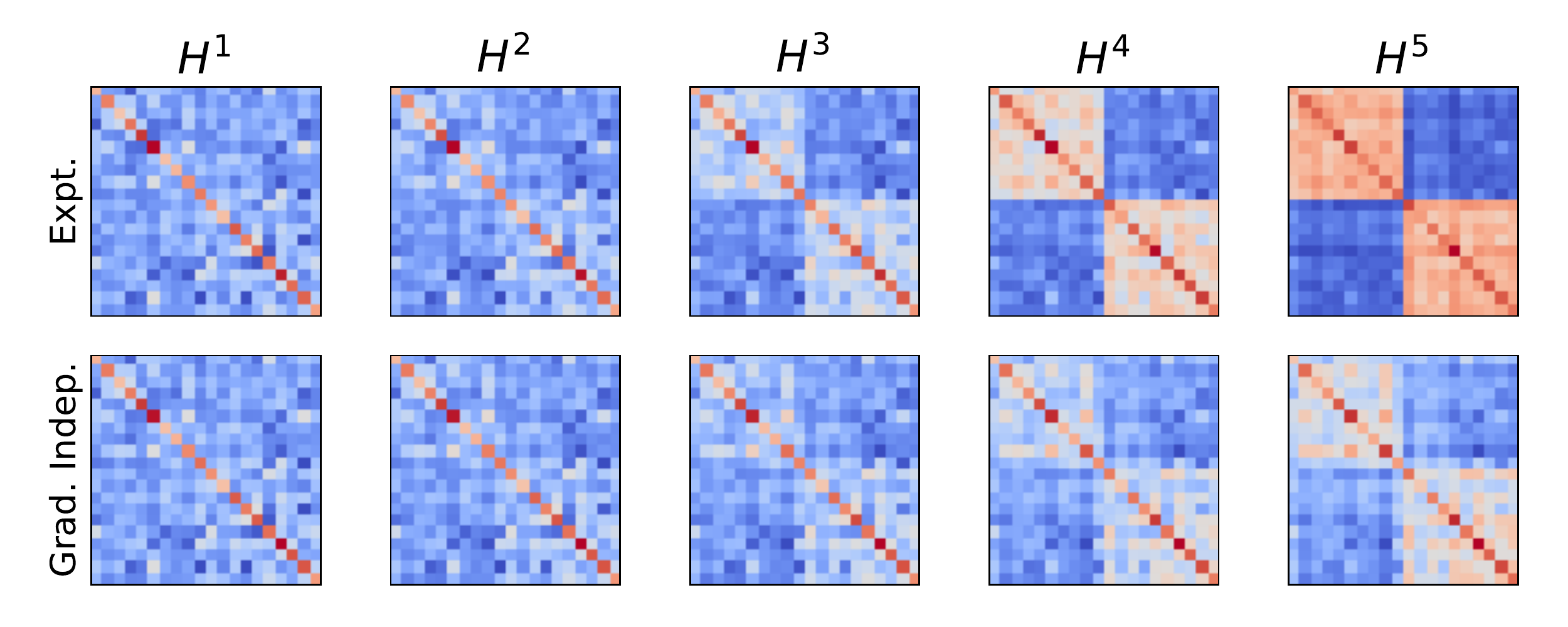}}
    \caption{Gradient independence fails to characterize feature learning dynamics in networks with $L > 1$ and large $\gamma_0$. (a) Loss curves for deep linear networks predicted under gradient independence ansatz for $\gamma_0 = 1.5$. (b) The predicted and experimental feature kernels $\H^\ell$ for the $L=5$ hidden layer network demonstrate that gradient independence underestimates the size of kernel adaptation. }
    \label{fig:grad_indep_fig}
\end{figure}

\begin{figure}[H]
    \centering
    \subfigure[Test MSE Loss]{\includegraphics[width=0.32\linewidth]{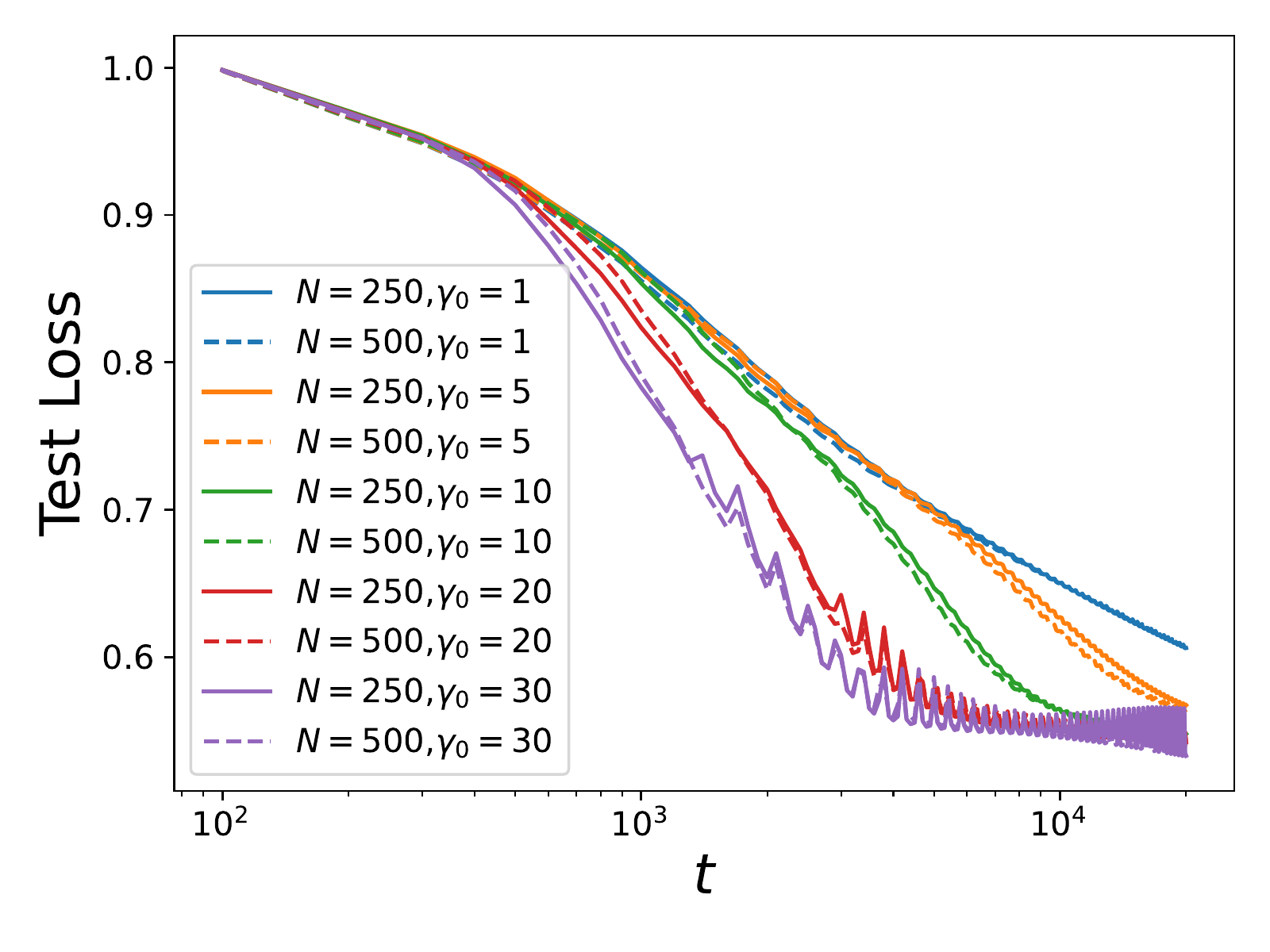}}
    \subfigure[Test Classification Accuracy]{\includegraphics[width=0.32\linewidth]{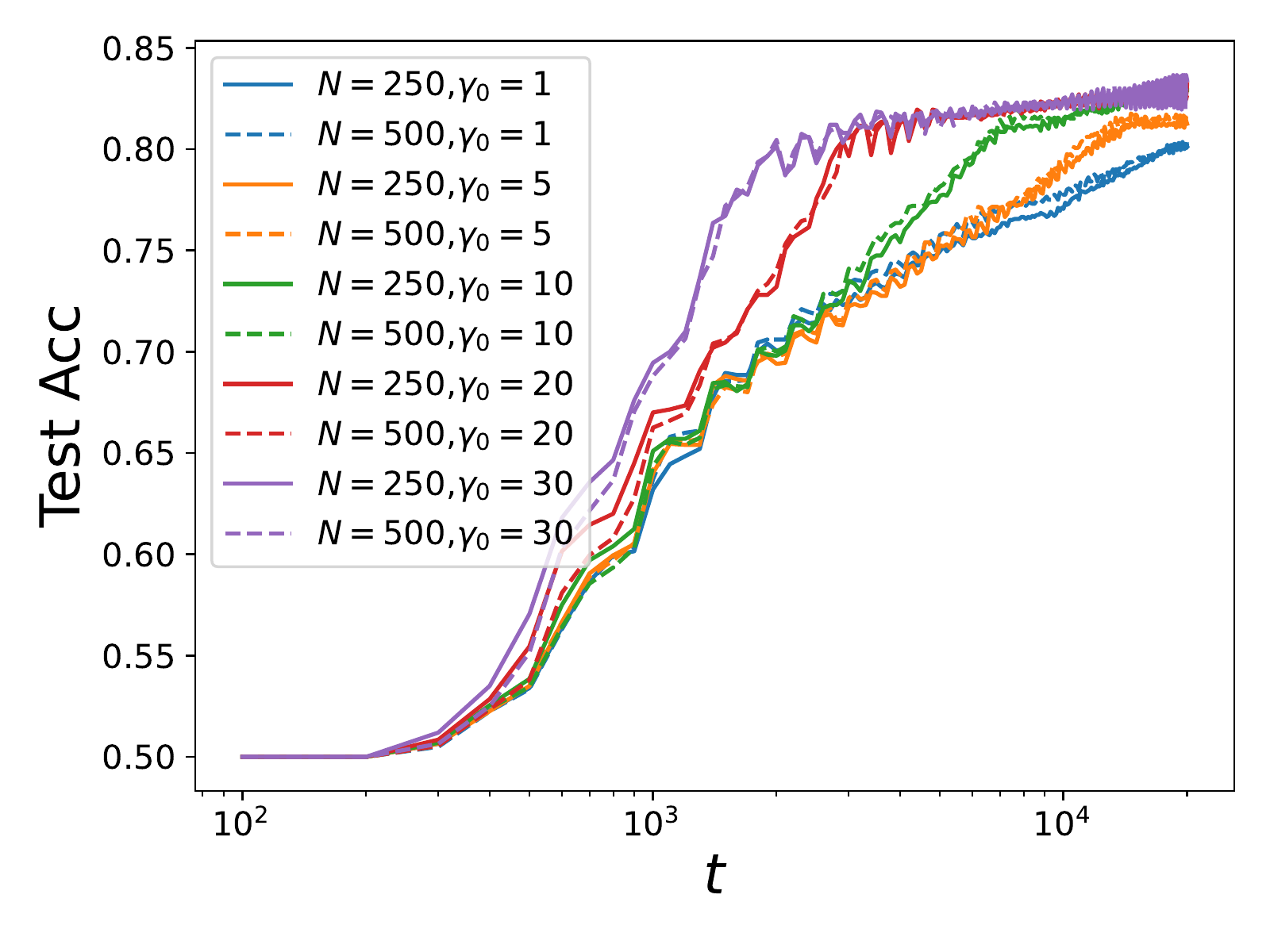}}
    \subfigure[$A(\Phi^L, y y^\top )$ Dynamics]{\includegraphics[width=0.32\linewidth]{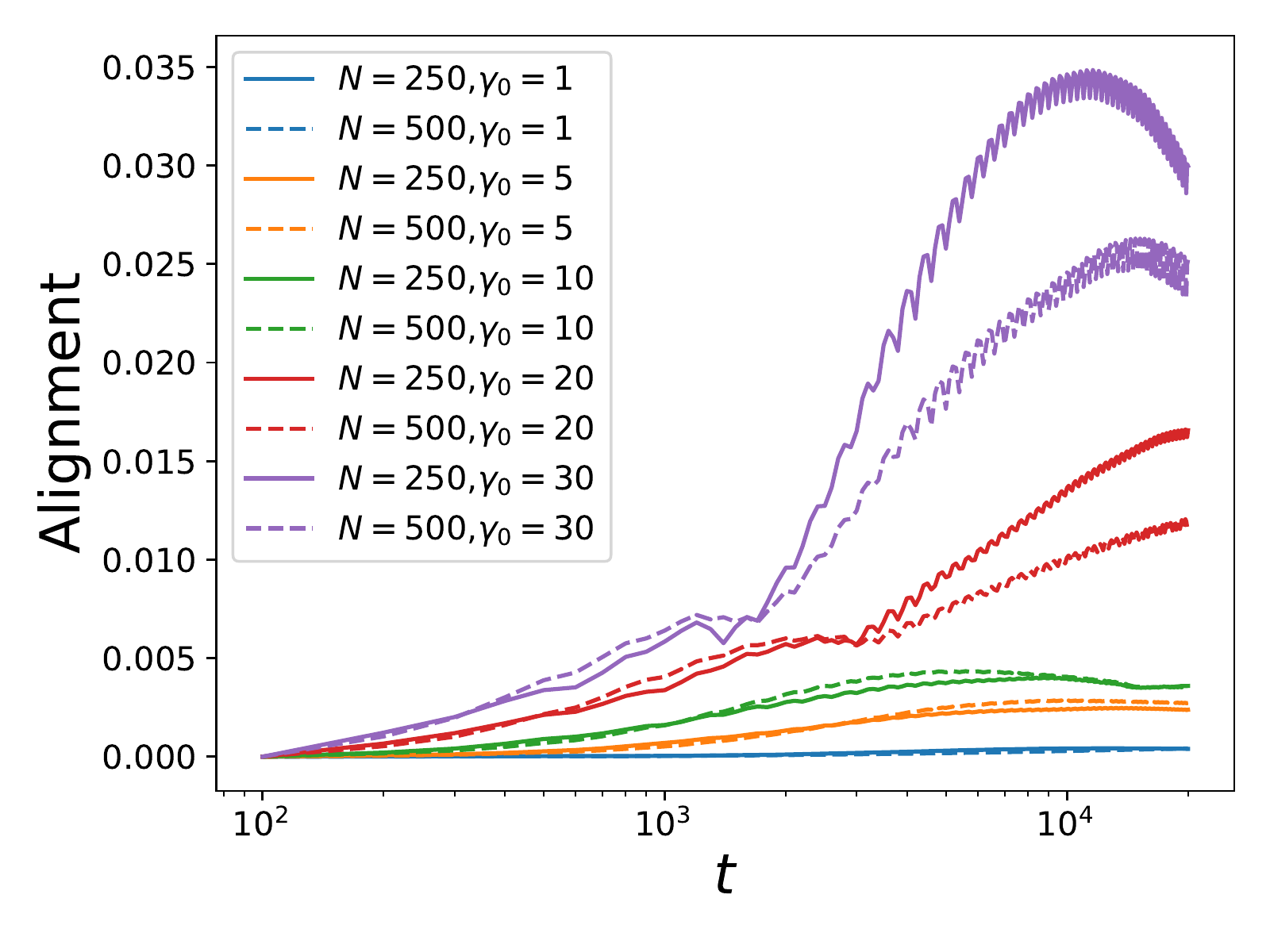}}
    \caption{Repeating the experiment of Figure \ref{fig:cifar_gamma_sweep} with depth 7 ($L=6$ hidden layer) CNN trained on two class CIFAR over a wide range of $\gamma_0$ with $N \in \{250,500\}$. We find consistent agreement of loss and prediction dynamics across widths but finite size effects become more significant when computing feature kernels of deeper layers. We note that, while higher $\gamma_0$ is associated with faster convergence, the final test accuracy for this model is roughly insensitive to choice of $\gamma_0$. 
    }
    \label{fig:my_label}
\end{figure}


\section{Algorithmic Implementation}\label{app:dmft_algorithm}

The alternating sample-and-solve procedure we developed and describe below for nonlinear networks is based on numerical recipes used in the dynamical mean field simulations in computational physics \cite{manacorda2020numerical}. The basic principle is to leverage the fact that, conditional on kernels, we can easily draw samples $\{u_\mu^\ell(t), r_\mu^\ell(t)\}$ from their appropriate GPs. From these sampled fields, we can identify the kernel order parameters by simple estimation of the appropriate moments. 

\begin{algorithm}[H]\label{alg:MC_DMFT}
\caption{Alternating Monte-Carlo Solution to Saddle Point Equations}\label{alg:two}
\KwData{$\K^x, \y$, Initial Guesses $\{ \bm\Phi^\ell ,\G^\ell \}_{\ell=1}^L$, $\{\A^\ell,\B^\ell\}_{\ell=1}^{L-1}$, Sample count $\mathcal S$, Update Speed $\beta$}
\KwResult{Final Kernels $\{ \bm\Phi^\ell ,\G^\ell \}_{\ell=1}^L$, $\{\A^\ell,\B^\ell\}_{\ell=1}^{L-1}$, Network predictions through training $f_\mu(t)$}
$\bm\Phi^0 = \bm K^x \otimes \bm 1 \bm 1^\top$, $\bm\G^{L+1} = \bm 1 \bm 1^\top$ \;
\While{Kernels Not Converged}{
  From $\{ \bm\Phi^\ell, \G^\ell\}$ compute $\K^{NTK}(t,t)$ and solve $\frac{d}{dt} f_\mu(t) = \sum_\alpha \Delta_\alpha(t) K^{NTK}_{\mu\alpha}(t,t)$\;
  $\ell = 1$\;
  \While{$\ell < L+1$}{
  Draw $\mathcal S$ samples $\{ u^\ell_{\mu,n}(t) \}_{n=1}^{\mathcal S} \sim \mathcal{GP}(0,\bm\Phi^{\ell-1})$, $\{ r^\ell_{\mu,n}(t) \}_{n=1}^{\mathcal S} \sim \mathcal{GP}(0,\G^{\ell+1})$\;
  Solve equation \eqref{eq:dmft_stoch_process} for each sample to get $\{h^\ell_{\mu,n}(t),z^\ell_{\mu,n}(t)\}_{n=1}^{\mathcal S}$\;
  Compute new $\bm\Phi^\ell,\G^\ell$ estimates: \\ $\tilde\Phi_{\mu\alpha}^\ell(t,s) = \frac{1}{\mathcal S} \sum_{n \in [\mathcal S]} \phi(h_{\mu,n}^\ell(t) ) \phi(h_{\alpha,n}^\ell(s))$, $\tilde{G}_{\mu\alpha}^{\ell}(t,s) = \frac{1}{\mathcal S} \sum_{n \in [\mathcal S]} g^\ell_{\mu,n}(t) g^\ell_{\alpha,n}(s) $ \;
  Solve for Jacobians on each sample $\frac{\partial \phi(\h_{n}^\ell)}{\partial \r^{\ell \top}_n}, \frac{\partial \g_n^\ell}{\partial \u^{\ell\top}_n}$ \;
  Compute new $\A^\ell,\B^{\ell-1}$ estimates:  \\
  $\tilde{\A}^\ell = \frac{1}{\mathcal S} \sum_{n \in [\mathcal S]} \frac{\partial \phi(\h_{n}^\ell)}{\partial \r^{\ell \top}_n}  \ , \tilde{\B}^{\ell-1} = \frac{1}{\mathcal S} \sum_{n \in [\mathcal S]} \frac{\partial \g_n^\ell}{\partial \u^{\ell\top}_n}$ \;
  $\ell \gets \ell+1$\;
  }
  $\ell = 1$\;
  \While{$\ell < L+1$}{
  Update feature kernels: $\bm\Phi^\ell \gets (1-\beta) \bm\Phi^\ell + \beta \tilde{\bm\Phi}^\ell$, $\G^\ell \gets (1-\beta) \G^\ell + \beta \tilde{\G}^\ell$ \;
  \If{$\ell < L$}{
    Update $\A^{\ell} \gets (1-\beta) \A^\ell + \beta \tilde{\A}^{\ell}, \B^\ell \gets (1-\beta)\B^\ell + \beta \tilde{\B}^\ell$
  }
  $\ell \gets \ell+1$
  }
}
\Return $\{ \bm\Phi^\ell, \G^\ell \}_{\ell=1}^L, \{\A^\ell,\B^\ell\}_{\ell=1}^{L-1}, \{f_\mu(t)\}_{\mu=1}^P$
\end{algorithm}
The parameter $\beta$ controls recency weighting of the samples obtained at each iteration. If $\beta= 1$, then the rank of the kernel estimates is limited to the number of samples $\mathcal S$ used in a single iteration, but with $\beta < 1$ smaller sample sizes $\mathcal S$ can be used to still obtain accurate results. We used $\beta = 0.6$ in our deep network experiments. Convergence is usually achieved in around $\sim 15$ steps for a depth 4 ($L=3$ hidden layer) network such as the one in Figure \ref{fig:deep_tanh_visual_kernels} and \ref{fig:dmft_tanh_depth_4}.

\section{Experimental Details}\label{app:expt_detail}

All NN training was performed with Jax gradient descent optimizer \cite{jax2018github} with fixed learning rate. 

\subsection{MLP Experiments}
For the MLP experiments, we performed full batch gradient descent. Networks were initialized with Gaussian weights with unit standard deviation $W^{\ell}_{ij} \sim \mathcal{N}(0,1)$. The learning rate was chosen as $\eta_0 \gamma^2 = \eta_0 \gamma_0^2 N$ for a network of width $N$. The hidden features $\h^\ell_\mu(t) \in \mathbb{R}^N$ were stored throughout training and used to compute the kernels $\Phi^{\ell}_{\mu\alpha}(t,s) = \frac{1}{N} \phi(\h_\mu^\ell(t)) \cdot \phi(\h_\alpha^\ell(s))$. These experiments can be reproduced with provided jupyter notebooks.

\subsection{CNN Experiments on CIFAR-10}\label{app:expt_detail_cnn}

We define a depth $L$ CNN model with ReLU activations and stride $1$, which is implemented as a pytree of parameters in JAX \cite{jax2018github}. We apply global average pooling in the final layer before a dense readout layer. The code to initialize and evaluate the model is provided below.

\begin{lstlisting}[language=Python]
from jax import random, lax
import jax.numpy as jnp

#L: number of hidden layers, N: width
def initialize_cnn(L, N, seed=0):
    key = random.PRNGKey(seed)
    params = [] # creates list of L+1 weights
    params += [ random.normal(key, (3,3,3,N)) ] # HWIO
    for l in range(L-1):
        key,_ = random.split(key)
        params += [random.normal(key, (3,3,N,N))]
    params += [ random.normal(key, (N,))]
    return params

dn=lax.conv_dimension_numbers((1,3,3,3),(3,3,3,1),('NHWC', 'HWIO','NHWC')) # defines which axis used for convolution
nonlin_fn = lambda h: (h>0.0) * h # ReLU activation
def cnn(params,X):
    L = len(params)-1 # number of hidden layers
    N = params[0].shape[-1] # width
    h = lax.conv_general_dilated(X, params[0],(1,1),'SAME', (1,1),(1,1),dn) # h1
    phi = nonlin_fn(h) # phi(h1)
    for i in range(1,L-1):
        h = 1/jnp.sqrt(N) * lax.conv_general_dilated(phi,                         params[i], (1,1),'SAME',                                          (1,1),(1,1),dn) # recurrence for h
        phi = nonlin_fn(h) # phi(h)
    phi = phi.mean(axis = (1,2)) # global average pooling
    w = params[-1]
    f = 1/N * phi @ w # Mean-field parameterization
    return f
\end{lstlisting}

After constructing a CNN model, we train using MSE loss with base learning rate $\eta_0 = 2.0 \times 10^{-4}$, batch size $250$. The learning rate passed to the optimizer is thus $\eta = \eta_0 \gamma^2 = \eta_0 \gamma_0^2 N$. We optimize the loss function which is scaled appropriately as $\ell( \gamma_0^{-1} f, y)$. Throughout training, we compute the last layer's embedding $\phi(\h^L)$ on the test set to calculate the alignment $A(\bm\Phi^L, \y\y^\top)$. Training was performed on 4 NVIDIA GPUs. Training a $L=3$ network of width $500$ takes roughly $1$ hour.

\section{Derivation of Self-Consistent Dynamical Field Theory}\label{app:dmft_derivation}

In this section, we introduce the dynamical field theory setup and saddle point equations. The path integral theory we develop is based on the Martin-Siggia-Rose-De Dominicis-Janssen (MSRDJ) framework \cite{martin1973statistical}, of which a useful review for random recurent networks can be found here \cite{crisanti2018path}. Similar computations can be found in recent works which consider typical behavior in high dimensional classification on random data \cite{agoritsas2018out, mignacco2020dynamical}.

\subsection{Deep Network Field Definitions and Scaling}
As discussed in the main text, we consider the following wide network architecture parameterzied by trainable weights $\bm\theta = \text{Vec}\{ \W^0 , \W^1 , ... \w^L\}$, giving network output $f_\mu$ defined as 
\begin{align}
    f_\mu &= \frac{1}{\gamma} h^{L+1}_\mu   \ , \ h^{L+1}_\mu = \frac{1}{\sqrt{N}} \w_L \cdot \phi(\h_\mu^L) \nonumber
    \\ 
    \h^{\ell+1}_\mu &= \frac{1}{\sqrt{ N}} \W^{\ell} \phi(\h_\mu^{\ell})  \ , \  \h^1_\mu = \frac{1}{\sqrt D} \W^0 \x_\mu 
\end{align}
Using gradient flow with learning rate $\eta$ on cost $\mathcal L = \sum_\mu \ell(f_\mu,y_\mu)$ for loss function, we introduce functions $\Delta_\mu = - \frac{\partial \mathcal L}{\partial f_\mu}$ and $\eta$ for learning rate, gradient flow induces the following dynamics 
\begin{align}
    \frac{d\bm\theta}{dt} = \frac{\eta}{\gamma} \sum_\mu \Delta_\mu \frac{\partial h_{\mu}^{L+1}}{\partial \bm\theta} \ , \ \frac{\partial f_\mu}{\partial t} =  \frac{\eta}{\gamma^2} \sum_{\alpha} \Delta_\alpha K^{NTK}_{\mu\alpha} \ , \ K^{NTK}_{\mu\alpha} = \frac{\partial h^{L+1}_\mu}{\partial \bm\theta} \cdot \frac{\partial h_\alpha^{L+1}}{\partial \bm\theta} 
\end{align}
Since $K_{NTK}$ is $O_{\gamma}(1)$ at initialization, it is clear that to have $O_{\gamma}(1)$ evolution of the network output at initialization we need $\eta = \gamma^2$. With this scaling, we have the following
\begin{align}
    \frac{d\bm\theta}{dt} =  \gamma \sum_\mu \Delta_\mu \frac{\partial h_{\mu}^{L+1}}{\partial \bm\theta} \ , \ \frac{\partial f_\mu}{\partial t} =  \sum_{\alpha} \Delta_\alpha K^{NTK}_{\mu\alpha}
\end{align}
Now, to build a valid field theory, we want to express everything in terms of features $\h_\mu^\ell$ rather than parameters $\bm\theta$ and we will define the following gradient features $\g^{\ell}_\mu = \sqrt{N} \frac{\partial h^{L+1}_\mu }{\partial \h^{\ell}_\mu}$ which admit the recursion and base case
\begin{align}
    \g^{\ell}_\mu &= \sqrt{N} \frac{\partial h^{L+1}_\mu }{\partial \h^{\ell}_\mu} = \left( \frac{\partial \h^{\ell+1}_\mu}{\partial \h^{\ell}_\mu} \right)^\top \left( \sqrt{N} \frac{\partial h^{L+1}_\mu}{\partial \h^{\ell+1}_\mu} \right) = \dot{\phi}(\h^\ell_\mu) \odot  \z_\mu^\ell  \ , \ \z_\mu^\ell= \frac{1}{\sqrt{N}} \W^{\ell \top} \g^{\ell+1}_\mu \nonumber
    \\
    \g^L_\mu &= \dot{\phi}(\h^L_\mu) \odot \w^L 
\end{align}
We define the \textit{pre-gradient} field $\z^\ell_\mu = \frac{1}{\sqrt N} \W^{\ell \top} \g^{\ell+1}_\mu$ so that $\g^\ell_\mu = \dot\phi(\h^\ell_\mu) \odot \z^\ell_\mu(t)$. From these quantities, we can derive the gradients with respect to parameters
\begin{align}
    \frac{\partial h_\mu^{L+1} }{\partial \W^\ell} = \sum_{i=1}^{N} \frac{\partial h_\mu^{L+1}}{\partial h^{\ell+1}_{\mu,i}}  \frac{\partial h_{\mu,i}^{\ell+1}}{\partial \W^{\ell}} = \frac{1}{N} \g^{\ell+1}_\mu \phi(\h_\mu^\ell)^\top
\end{align}
which allows us to compute the NTK in terms of these features
\begin{align}
    K^{NTK}_{\mu\alpha} = \frac{1}{N} \phi(\h^L_\mu) \cdot \phi(\h^L_\alpha) + \sum_{\ell =1}^{L-1} \left( \frac{\g^{\ell+1}_\mu \cdot \g^{\ell+1}_\alpha}{N} \right) \left( \frac{\phi(\h^\ell_\mu) \cdot \phi(\h^\ell_\alpha)}{N} \right) + \frac{\g^1_\mu \cdot\g_\alpha^1}{N} K^x_{\mu\alpha}
\end{align}
where $K^x_{\mu\alpha} =  \frac{\x_\mu \cdot \x_\alpha}{D}$ is the input Grammian. We see that the NTK can be built out of the following primitive kernels
\begin{align}
    \Phi_{\mu\nu}^{\ell} = \frac{1}{N} \phi(\h^\ell_\mu) \cdot\phi(\h^\ell_\nu) \ , \ G_{\mu\nu}^\ell = \frac{1}{N} \g^{\ell}_\mu \cdot \g^{\ell}_\nu
\end{align}
We utilize the parameter space dynamics to express $\W^\ell$ in terms of the $\{\g,\h\}$ fields
\begin{align}
    \W^\ell(t) = \W^\ell(0) + \frac{\gamma}{N} \int_0^t ds \sum_\mu \Delta_\alpha(s) \g_{\mu}^{\ell+1}(s)  \phi(\h^{\ell}_\mu(s))^\top
\end{align}
Using the field recurrences $\h^{\ell+1}_\mu(t) = \frac{1}{\sqrt N} \W^{\ell}(t) \phi(\h^\ell_\mu(t))$ we can derive the following recursive dynamics for the features
\begin{align}
    \h^{\ell+1}_\mu(t) &= \bm\chi_\mu^{\ell+1}(t) + \frac{ \gamma}{\sqrt N} \int_0^t ds \sum_\nu \Delta_\nu \g^{\ell+1}_\nu(t) \Phi_{\nu\mu}^{\ell}(s,t) \nonumber
    \\
    \z^{\ell}_\mu(t) &= \bm\xi_\mu^{\ell}(t) +   \frac{\gamma}{\sqrt N}  \int_0^t ds \sum_\nu \Delta_\nu(s) \phi(\h^\ell_\nu(s)) G^{\ell+1}_{\nu\mu}(s,t)  \ , \ \g^\ell_\mu(t) = \dot\phi(\h^\ell_\mu(t) ) \odot \z^\ell_\mu(t) \nonumber 
    \\
    \frac{\partial f_\mu}{\partial t} &= \sum_\alpha \Delta_\alpha(t) \left[ \Phi_{\mu\alpha}^{L}(t,t) + \sum_{\ell=1}^{L-1} G^{\ell+1}_{\mu\alpha}(t,t) \Phi^{\ell}_{\mu\alpha}(t,t) + G^1_{\mu\alpha}(t,t) K_{\mu\alpha}^x \right]
\end{align}
where we introduced the following random fields $\bm \chi_\mu^\ell(t), \bm \xi_\mu^\ell(t)$ which involve the random initial conditions
\begin{align}
    \bm \chi_\mu^\ell(t) = \frac{1}{\sqrt N} \W^\ell(0) \phi(\h_\mu^\ell(t)) \ , \ \bm\xi_\mu^\ell(t) = \frac{1}{\sqrt N} \W^{\ell}(0)^\top \g^{\ell+1}_\mu(t)
\end{align}
We observe that the dynamics of the hidden features is controlled by the factor $\frac{\gamma}{\sqrt N}$. If $\gamma = O_N(1)$ then we recover static NTK in the limit as $N\to\infty$. However, if $\gamma = O_N(\sqrt N)$ then we obtain $O_N(1)$ evolution of our features and we reach a new rich regime. We choose the scaling $\gamma = \gamma_0 \sqrt N$ for our field theory so that $\gamma_0 > 0$ will give a feature learning network. 

\subsection{Warmup: DMFT for One Hidden Layer NN}\label{app:two_layer_warmup_derivation}

In this section, we provide a warmup problem of a $L=1$ hidden layer network which allows us to illustrate the mechanics of the MSRDJ formalism. A more detailed computation can be found in the next section. Though many of the interesting dynamical aspects of the deep network case are missing in the two layer case, our aim is to show a simple application of the ideas. The fields of interest are $\bm \chi_\mu = \frac{1}{\sqrt D} \W^0(0) \x_\mu$ and $\bm\xi  = \w^1(0)$. Unlike the deeper $L \geq 2$ case, both of these fields are time invariant since $\x_\mu$ does not vary in time. These random fields provide initial conditions for the preactivation and pre-gradient fields $\h_\mu(t), \z(t) \in \mathbb{R}^N$, which evolve according to 
\begin{align}\label{app:eq_two_layer_field_dependence}
    \h_\mu(t) &= \bm\chi_\mu + \gamma_0 \int_0^t ds  \sum_\alpha [\z(s) \odot \dot\phi(\h_\alpha(s))] K^x_{\mu\alpha} \Delta_\alpha(s) \nonumber
    \\
    \z(t) &= \bm\xi + \gamma_0 \int_0^t ds \sum_\alpha \phi(\h_\alpha(s)) \Delta_\alpha(s) .
\end{align}
where the network predictions evolve as $\frac{\partial}{\partial t} f_\mu(t) = \sum_\alpha [\Phi_{\mu\alpha}(t,t) + G_{\mu\alpha}(t,t) K^x_{\mu\alpha} ] \Delta_\alpha(t)$ for kernels $\Phi_{\mu\alpha}(t,t) = \frac{1}{N} \phi(\h_\mu(t)) \cdot \phi(\h_\alpha(t))$ and $G_{\mu\alpha}(t,t) = \frac{1}{N} \g_\mu(t) \cdot \g_\alpha(t)$. At finite $N$, the kernels $\Phi, G$ will depend on the random initial conditions $\bm\chi, \bm\xi$, leading to a predictor $f_\mu$ which varies over initializations. If we can establish that the kernels $\Phi,G$ concentrate at infinite-width $N \to \infty$, then $\Delta_\mu$ are deterministic.  We now study the moment generating function for the fields
\begin{align}
    Z[\{\bm \j_\mu\}_{\mu\in [P]}, \bm v] &= \left<\exp\left( \sum_\mu \bm j_\mu \cdot \bm\chi_\mu + \bm\xi \cdot \v \right) \right>_{\bm\theta_0}.
\end{align}
To perform the average over $\bm\theta_0 = \{\W^0(0),\w^1(0)\}$, we enforce the definition of $\bm\chi_\mu,\bm\xi$ with delta functions
\begin{align}
    1 &= \int d\bm\chi_\mu \delta\left( \bm\chi_\mu - \frac{1}{\sqrt N} \W^0(0) \x_\mu \right)  = \int \frac{d\bm\chi_\mu d\bm{\hat\chi}_\mu}{(2\pi)^N} \exp\left( i \bm{\hat \chi}_\mu \cdot \left( \bm\chi_\mu - \frac{1}{\sqrt D} \W^0(0) \x_\mu \right) \right) \nonumber
    \\
    1 &= \int d\bm\xi \ \delta\left( \bm\xi - \w^1(0) \right)  = \int \frac{d\bm\xi d\bm{\hat\xi}}{(2\pi)^N} \exp\left( i \bm{\hat \xi} \cdot \left( \bm\xi - \w^1(0) \right) \right) .
\end{align}
Though this step may seem redundant in this example, it will be very helpful in the deep network case, so we pursue it for illustration. After mulitplying by these factors of unity and performing the Gaussian integrals, we obtain
\begin{align}
    Z = \int \prod_{\mu} \frac{d\bm\chi d\bm{\hat\chi}}{(2\pi)^N} \frac{d\bm\xi d\bm{\hat\xi}}{(2\pi)^N} \exp\left( - \frac{1}{2} \sum_{\mu\alpha} \bm{\hat\chi}_\mu \cdot \bm{\hat\chi}_\alpha K^x_{\mu\alpha} +  \sum_\mu \bm\chi_\mu \cdot ( i \bm{\hat\chi}_\mu + \bm \j_\mu ) - \frac{1}{2} |\bm{\hat\xi}|^2 + \bm\xi \cdot ( i\bm{\hat{\xi}} + \v)  \right)
\end{align}
We now aim enforce the definitions of the kernel order parameters with delta functions
\begin{align}
    1&= N \int d\Phi_{\mu\alpha}(t,s) \ \delta\left( N \Phi_{\mu\alpha}(t,s) - \phi(\h_\mu(t)) \cdot \phi(\h_\alpha(s)) \right) \nonumber
    \\
    &= \int  \frac{d\Phi_{\mu\alpha}(t,s) d\hat{\Phi}_{\mu\alpha}(t,s)}{2\pi i N^{-1}} \exp\left( N \hat{\Phi}_{\mu\alpha}(t,s)\left( N \Phi_{\mu\alpha}(t,s) - \phi(\h_\mu(t)) \cdot \phi(\h_\alpha(s)) \right)   \right) \nonumber
    \\
    1&= N \int dG_{\mu\alpha}(t,s) \ \delta\left( N G_{\mu\alpha}(t,s) - \g_\mu(t) \cdot \g_\alpha(s) \right) \nonumber
    \\
    &= \int  \frac{dG_{\mu\alpha}(t,s) d\hat{G}_{\mu\alpha}(t,s)}{2\pi i N^{-1}} \exp\left( N \hat{G}_{\mu\alpha}(t,s)\left( N G_{\mu\alpha}(t,s) - \g_\mu(t) \cdot \g_\alpha(s) \right)   \right),
\end{align}
where the fields $\h_\mu(t), \g_\mu(t)$ are regarded as functions of $\{\bm \chi_\mu\}_{\mu},\bm \xi$ (see Equation \eqref{app:eq_two_layer_field_dependence}) and the $\hat{\Phi}, \hat{G}$ integrals run over the imaginary axis $(-i \infty, i \infty)$. After this step, we can write 
\begin{align}
    Z \propto \int \prod_{\mu\alpha ts} d\Phi_{\mu\alpha}(t,s) d\hat{\Phi}_{\mu\alpha}(t,s) dG_{\mu\alpha}(t,s)d\hat{G}_{\mu\alpha}(t,s) \exp\left( N S[\Phi,\hat{\Phi},G,\hat{G}] \right)
\end{align}
where the DMFT action $S[\Phi,\hat{\Phi},G,\hat{G}]$ is $\mathcal{O}_N(1)$ and has the form
\begin{align}
    S[\Phi,\hat{\Phi},G,\hat{G}] = \sum_{\mu\alpha} \int dt ds [ \Phi_{\mu\alpha}(t,s) \hat{\Phi}_{\mu\alpha}(t,s)  + G_{\mu\alpha}(t,s) \hat{G}_{\mu\alpha}(t,s) ] + \frac{1}{N} \sum_{i=1}^N \ln \mathcal Z[j_i, v_i]. 
\end{align}
The single site moment generating function $\mathcal Z[j,v]$ arises from the factorization of the integrals over $N$ different fields in the hidden layer and takes the form
\begin{align}
    \mathcal Z[j,v] =& \int \prod_\mu \frac{d\chi_\mu d\hat{\chi}_\mu}{2\pi} \frac{d\xi d\hat{\xi}}{2\pi} \exp\left( - \frac{1}{2} \sum_{\mu\alpha}\hat{\chi}_\mu \hat{\chi}_\alpha K^x_{\mu\alpha} + (j_\mu + i \hat{\chi}_\mu) \chi_\mu   - \frac{1}{2} \hat{\xi}^2 + (v + i \hat{\xi} ) \xi \right) \nonumber
    \\
    &\times\exp\left( - \int_0^\infty dt \int_0^\infty ds \sum_{\mu\alpha} [ \hat{\Phi}_{\mu\alpha}(t,s) \phi(h_\mu(t)) \phi(h_\alpha(s)) + \hat{G}_{\mu\alpha}(t,s) g_\mu(t) g_\alpha(s) ]  \right)
\end{align}
where, again we must regard $h_\mu(t), g_\mu(t)$ as functions of $\chi,\xi$. The variables in the above are no longer vectors in $\mathbb{R}^N$ but rather are scalars. We can write $\mathcal Z[j,v] = \int \prod_\mu d\chi_\mu d\hat{\chi}_\mu d\xi d\hat{\xi} \exp\left( - \mathcal{H}[\{\chi_\mu,\hat\chi_\mu\},\xi,\hat\xi, j, v] \right)$ where $\mathcal H$ is the logarithm of the integrand above. Since the full MGF takes the form $Z \propto \int d\Phi d\hat\Phi dGd\hat{G} \exp\left( N S[\Phi,\hat\Phi,G,\hat G] \right)$, characterization of the $N \to \infty$ limit requires one to identify the saddle point of $S$, where $\delta S = 0$ for any variation of these 4 order parameters.
\begin{align}
    \frac{\delta S}{\delta \Phi_{\mu\alpha}(t,s)} &= \hat{\Phi}_{\mu\alpha}(t,s) = 0 \ , \ \frac{\delta S}{\delta \hat\Phi_{\mu\alpha}(t,s)} = \Phi_{\mu\alpha}(t,s) - \frac{1}{N} \sum_{i=1}^N \left< \phi(h_{\mu}(t)) \phi(h_{ \alpha}(s)) \right>_{i} =0 \nonumber
    \\
     \frac{\delta S}{\delta G_{\mu\alpha}(t,s)} &= \hat{G}_{\mu\alpha}(t,s) = 0 \ , \ \frac{\delta S}{\delta \hat G_{\mu\alpha}(t,s)} = G_{\mu\alpha}(t,s) - \frac{1}{N} \sum_{i=1}^N \left< g_{\mu}(t) g_{\alpha}(s) \right>_{i} =0
\end{align}
where the $i$-th single site average $\left< \right>_i$ of an observable $O(\chi,\hat\chi,\xi,\hat\xi)$ is defined as
\begin{align}
    \left< O(\chi,\hat\chi,\xi,\hat\xi) \right>_i = \frac{1}{\mathcal Z[j_i,v_i] } \int \prod_\mu d\chi_\mu d\hat{\chi}_\mu d\xi d\hat{\xi} \exp\left( -\mathcal{H}[\{\chi_\mu,\hat\chi_\mu\},\xi,\hat\xi, j_i, v_i]  \right) O(\chi,\hat\chi,\xi,\hat\xi)
\end{align}
Since $\hat{\Phi} = \hat{G} = 0$ the single site MGF reveals that the initial fields are independent Gaussians $\{\chi_{\mu}\} \sim \mathcal{N}(0,\bm K^x)$ and $\xi \sim\mathcal{N}(0,1)$. At zero source $\bm j, \bm v \to 0$, all single site averages $\left< \right>_i$ are equivalent and we may merely write $\Phi_{\mu\alpha}(t,s) = \left< \phi(h_\mu(t))\phi(h_\alpha(s)) \right> \ , \ G_{\mu\alpha}(t,s) = \left< g_\mu(t) g_\alpha(s) \right>$, where $\left< \right>$ is the average over the single site distributions for $\j,\v \to 0$. 
\subsubsection{Final $L=1$ DMFT equations}
Putting all of the saddle point equations together, we arrive at the following DMFT
\begin{align}
    \{ \chi_\mu \}_{\mu \in [P]} &\sim \mathcal{N}(0,\K^x) \ , \ \xi \sim \mathcal{N}(0,1) \nonumber
    \\
    h_\mu(t) &= \chi_\mu + \gamma_0 \int_0^t ds  \sum_\alpha [z(s) \dot\phi(h_\alpha(s))] K^x_{\mu\alpha} \Delta_\alpha(s) \ , \ z(t) = \xi + \gamma_0 \int_0^t ds \sum_\alpha \phi(h_\alpha(s)) \Delta_\alpha(s) \nonumber
    \\
    \Phi_{\mu\alpha}(t,s) &= \left< \phi(h_\mu(t)) \phi(h_\alpha(s)) \right> \ , \ G_{\mu\alpha}(t,s) = \left< g_\mu(t) g_\alpha(s) \right> = \left< z(t) z(s) \dot\phi(h_\mu(t))\dot\phi(h_\alpha(s)) \right> \nonumber
    \\
    \frac{\partial f_\mu}{\partial t} &= \sum_\alpha [\Phi_{\mu\alpha}(t,t) + G_{\mu\alpha}(t,t) K^x_{\mu\alpha}] \Delta_\alpha(t)
\end{align}
We see that for $L=1$ networks, it suffices to solve for the kernels on the time-time diagonal. Further in this two layer case $\chi, \xi$ are independent and do not vary in time. These facts will not hold in general for $L \geq 2$ networks, which requires a more intricate analysis as we show in the next section.

\subsection{Path Integral Formulation for Deep Networks}

As discussed in the main text, we study the distribution over fields by computing the moment generating functional for the stochastic processes $\{ \bm\chi^\ell, \bm\xi^\ell \}_{\ell=1}^L$
\begin{align}
    Z[\{\j^\ell,\v^{\ell} \}] = \left< \exp\left(  \sum_{\ell,\mu}\int_0^\infty dt \left[  \j_\mu^\ell(t) \cdot \bm\chi_\mu^\ell(t) + \v^\ell_{\mu}(t) \cdot \bm\xi^\ell_\mu(t) \right] \right) \right>_{\bm\theta_0 = \text{Vec}\{\W^0(0),...\w^L(0)\}}
\end{align}
Moments of these stochastic fields can be computed through differentiation of $Z$ near zero-source
\begin{align}
    \left< \chi_{\mu_1}^{\ell_1}(t_1) ... \chi_{\mu_n}^{\ell_n}(t_n) \xi_{\mu_1}^{\ell_1}(t_1) ... \xi_{\mu_m}^{\ell_m}(t_m)  \right> = \frac{\delta }{\delta j^{\ell_1}_{\mu_1}(t_1) }...\frac{\delta }{\delta j^{\ell_n}_{\mu_n}(t_n) } \frac{\delta }{\delta v^{\ell_1}_{\mu_1}(t_1) }...\frac{\delta }{\delta v^{\ell_m}_{\mu_m}(t_m) } Z[\{\j^\ell,\v^{\ell} \}]|_{\j=\v=0} .
\end{align}
To perform the average over the initial parameters, we enforce the definition of the fields $\bm\chi^{\ell+1}(t) = \frac{1}{\sqrt N} \W^\ell(0) \phi(\h^{\ell}_\mu(t))$, $\bm\xi^\ell_\mu(t) = \frac{1}{\sqrt N} \W^{\ell}(0)^\top \g^{\ell+1}_\mu(t)$, by inserting the following terms in the definition of $Z[\{\j,\v\}]$ so we may more easily perform the average over weights $\bm\theta_0$. We enforce these definitions with an integral representation of the Dirac-Delta function $1 = \int_{\mathbb{R}} dx \  \delta(x) = \frac{1}{2\pi} \int_{\mathbb{R}} dx \int_{\mathbb R} d\hat{x} \exp\left( i x \hat{x}  \right)$. We note that we are implicitly working in the Ito scheme, where factors of Jacobian determinants are equal to one \cite{crisanti2018path, honkonen2011ito, gardiner1985handbook} (we note that $\h^\ell_\mu(t)$ does not causally depend on $\bm\chi^{\ell+1}_\mu(t)$ and $\g^\ell_\mu(t)$ does not causally depend on $\bm\xi^\ell(t)$). Applying this to fields $\bm\chi,\bm\xi$, we have
\begin{align}
    1 &= \int_{\mathbb{R}^N} \int_{\mathbb{R}^N}  \frac{d\bm\chi^{1}_\mu(t) d\hat{\bm\chi}^{1}_\mu(t)}{(2\pi)^N} \exp\left(i \hat{\bm\chi}_\mu^{1}(t) \cdot \left[\bm\chi_\mu^{1}(t) - \frac{1}{\sqrt D} \W^{\ell}(0) \x_\mu \right]  \right) \nonumber
    \\
    1 &= \int_{\mathbb{R}^N} \int_{\mathbb{R}^N}  \frac{d\bm\chi^{\ell+1}_\mu(t) d\hat{\bm\chi}^{\ell+1}_\mu(t)}{(2\pi)^N} \exp\left(i \hat{\bm\chi}_\mu^{\ell+1}(t) \cdot \left[\bm\chi_\mu^{\ell+1}(t) - \frac{1}{\sqrt N} \W^{\ell}(0) \phi(\h^{\ell}_\mu(t)) \right]  \right) \ , \  \ell \in \{1,...,L-1\} \nonumber
    \\
    1 &= \int_{\mathbb{R}^N} \int_{\mathbb{R}^N}  \frac{d\bm\xi^{L}_\mu(t) d\hat{\bm\xi}^{L}_\mu(t)}{(2\pi)^N} \exp\left(i \hat{\bm\xi}_\mu^{L}(t) \cdot \left[\bm\xi_\mu^{L}(t) - \w^L(0) \right]  \right) \nonumber
    \\
    1 &= \int_{\mathbb{R}^N} \int_{\mathbb{R}^N}  \frac{d\bm\xi^{\ell}_\mu(t) d\hat{\bm\xi}^{\ell}_\mu(t)}{(2\pi)^N} \exp\left(i \hat{\bm\xi}_\mu^{\ell}(t) \cdot \left[\bm\xi_\mu^{\ell}(t) - \frac{1}{\sqrt N} \W^{\ell}(0)^\top \g^{\ell}_\mu(t) \right]  \right) \ , \ \ell \in \{1,...,L-1\}
\end{align}
where $\{ h^\ell, g^\ell \}$ are understood to be stochastic processes which are causally determined by the $\{\chi^\ell,\xi^\ell\}$ fields, in the sense that $h^\ell(t)$ only depends on $\chi^\ell(s)$ for $s < t$. We thus have an expression of the form
\begin{align}
    Z[\{\j^\ell,\v^{\ell} \}] =& \int \prod_{\ell \mu t} \frac{d\bm\chi^{\ell+1}_\mu(t) d\hat{\bm\chi}^{\ell+1}_\mu(t)}{(2\pi)^N}  \prod_{\ell \mu t}  \frac{d\bm\xi^{\ell}_\mu(t) d\hat{\bm\xi}^{\ell}_\mu(t)}{(2\pi)^N} \exp\left(  \sum_{\ell,\mu}\int_0^\infty dt \left[  \j_\mu^\ell(t) \cdot \bm\chi_\mu^\ell(t) + \v^\ell_{\mu}(t) \cdot \bm\xi^\ell_\mu(t) \right] \right) \nonumber 
    \\
    &\times \prod_{\ell=1}^{L-1} \left< \exp\left( -\frac{i}{\sqrt N} \sum_{\mu} \int_0^\infty dt \left[ \hat{\bm\chi}_\mu^{\ell+1}(t)^\top \W^{\ell}(0) \phi(\h^{\ell}_\mu(t)) + \g^{\ell+1}_\mu(t)^\top \W^\ell(0) \hat{\bm\xi}_\mu^\ell(t) \right]  \right) \right>_{\W^\ell(0)} \nonumber 
    \\
    &\times \left< \exp\left( - \frac{i}{\sqrt D} \sum_\mu \int_0^\infty dt \ \hat{\bm \chi}^1_\mu(t)^\top \W^0(0) \x_\mu \right) \right>_{\W^0(0)} \nonumber
    \\
    &\times \left< \exp\left( - i \sum_\mu \int_0^\infty \hat{\bm\xi}_\mu^L(t) \cdot \w^L(0)  \right) \right>_{\w^L(0)} \nonumber
    \\
    &\times \prod_{\ell=1}^L \exp\left( i \sum_\mu \int_0^\infty dt \ \left[  \hat{\bm\chi}_\mu^\ell(t) \cdot \bm\chi_\mu^\ell(t) + \hat{\bm\xi}^\ell_\mu(t) \cdot \bm\xi^\ell_\mu(t) \right]  \right)
\end{align}
Since $\W^\ell(0)$ are all Gaussian random variables, these averages can be performed quite easily yielding

\begin{align}
&\left< \exp\left( - \frac{i}{\sqrt D} \sum_\mu \int_0^\infty dt \hat{\bm \chi}^1_\mu(t)^\top \W^0(0) \x_\mu \right) \right>_{\W^0(0)} = \exp\left( - \frac{1}{2} \int_0^\infty  \int_0^\infty dt ds \sum_{\mu\alpha} \bm{\hat\chi}_\mu^1(t) \cdot \bm{\hat\chi}_\alpha^1(s) K^x_{\mu\alpha} \right) \nonumber
\\
&\left< \exp\left( - i \sum_\mu \int_0^\infty \hat{\bm\xi}_\mu^L(t) \cdot \w^L(0)  \right) \right>_{\w^L(0)} = \exp\left( - \frac{1}{2} \sum_{\mu\alpha} \int_0^\infty \int_0^\infty  dt ds \ \bm{\hat \xi}_\mu^L(t) \cdot \bm{\hat \xi}_{\alpha}^L(s)   \right) \nonumber
\\
&\left< \exp\left( -\frac{i}{\sqrt N} \sum_{\mu} \int_0^\infty dt \left[ \hat{\bm\chi}_\mu^{\ell+1}(t)^\top \W^{\ell}(0) \phi(\h^{\ell}_\mu(t)) + \g^{\ell+1}_\mu(t)^\top \W^\ell(0) \hat{\bm\xi}_\mu^\ell(t) \right]  \right) \right>_{\W^\ell(0)} \nonumber
\\
&= \exp\left( - \frac{1}{2N} \sum_{\mu\alpha} \int_0^\infty  \int_0^\infty dt ds \left[ \bm{\hat\chi}_\mu^{\ell+1}(t) \cdot  \bm{\hat\chi}_\mu^{\ell+1}(t) \phi(\h^\ell_\mu(t)) \cdot \phi(\h^\ell_\alpha(s)) + \bm{\hat\xi}_\mu^\ell(t) \cdot \bm{\hat\xi}_\alpha^\ell(s) \g^{\ell+1}_\mu(t) \cdot \g^{\ell+1}_\alpha(s) \right] \right) \nonumber
\\
&\times \exp\left( - \frac{1}{N} \sum_{\mu\alpha} \int_0^\infty \int_0^\infty dt ds \  \bm{\hat \chi}^{\ell+1}_\mu(t) \cdot \g^{\ell+1}_\alpha(s) \  \phi(\h^\ell_\mu(t)) \cdot \bm{\hat \xi}_\alpha^\ell(s) \right)
\end{align}

\subsection{Order Parameters and Action Definition}\label{app:order_params_action}

We define the following order parameters which we will show concentrate in the $N \to \infty$ limit
\begin{align}
    \Phi_{\mu,\alpha}^\ell(t,s) = \frac{1}{N} \phi(\h_\mu^\ell(t)) \cdot \phi(\h_\alpha^\ell(s)) \ , \ G^{\ell}_{\mu\alpha}(t,s) = \frac{1}{N} \g^\ell_\mu(t) \cdot \g^\ell_\alpha(s) \ , \ A_{\mu\alpha}^\ell(t,s) = - \frac{i}{N} \phi(\h^\ell_\mu(t)) \cdot \bm{\hat \xi}_\alpha^\ell(s) . 
\end{align}
The NTK only depends on $\{\Phi^\ell,G^\ell\}$ so from these order parameters, we can compute the function evolution. The parameter $\A^\ell$ arises from the coupling of the fields across a single layer's initial weight matrix $\W^\ell(0)$. We can again enforce these definitions with integral representations of the Dirac-delta function. For each pair of samples $\mu,\alpha$ and each pair of times $t,s$, we multiply by
\begin{align}
    1 &= \int \int \frac{d\Phi^\ell_{\mu\alpha}(t,s) d\hat{\Phi}^\ell_{\mu\alpha}(t,s) }{2\pi i N^{-1}} \exp\left( N \Phi^\ell_{\mu\alpha}(t,s) \hat{\Phi}^\ell_{\mu\alpha}(t,s) - \hat{\Phi}^\ell_{\mu\alpha}(t,s) \phi(\h_\mu^\ell(t)) \cdot \phi(\h_\alpha^\ell(s)) \right) \nonumber \ell \in \{1,...,L\}
    \\
    1 &= \int \int \frac{dG_{\mu\alpha}(t,s) d\hat{G}_{\mu\alpha}(t,s) }{2\pi i N^{-1}} \exp\left( N G^\ell_{\mu\alpha}(t,s) \hat{G}^\ell_{\mu\alpha}(t,s) - \hat{G}_{\mu\alpha}^\ell(t,s) \g^{\ell}_\mu(t) \cdot \g^\ell_\alpha(s) \right) \ , \  \ell \in \{1,...,L\} \nonumber
    \\
    1 &= \int \int \frac{dA^\ell_{\mu\alpha}(t,s) dB^\ell_{\mu\alpha}(t,s)}{2\pi i N^{-1}} \exp\left( - N A^\ell_{\mu\alpha}(t,s) B^\ell_{\mu\alpha}(t,s) - i B_{\mu\alpha}^\ell(t,s) \phi(\h^\ell_\mu(t)) \cdot \bm{\hat\xi}^\ell_\alpha(s)) \right) \ , \ \ell \in \{1,...,L-1\}
\end{align}

After introducing these order parameters into the definition of the partition function, we have a factorization of the integrals over each of the $N$ sites in each hidden layer. This gives the following partition function
\begin{align}
    Z = \int \prod_{\ell, \mu\alpha, ts} \frac{d\Phi^\ell_{\mu\alpha}(t,s) d\hat{\Phi}^\ell_{\mu\alpha}(t,s) }{2\pi i N^{-1}}& \frac{dG_{\mu\alpha}(t,s) d\hat{G}_{\mu\alpha}(t,s) }{2\pi i N^{-1}} \frac{dA^\ell_{\mu\alpha}(t,s) dB^\ell_{\mu\alpha}(t,s)}{2\pi i N^{-1}} \exp\left( N S[\{ \Phi, \hat\Phi, G,\hat G, A, B \}] \right) \nonumber
    \\
    S[\{ \Phi, \hat\Phi, G,\hat G, A, B \}] =  \sum_{\ell \mu\alpha} \int_0^\infty \int_0^\infty & dt ds \left[ \Phi^\ell_{\mu\alpha}(t,s) \hat{\Phi}^\ell_{\mu\alpha}(t,s) + G^\ell_{\mu\alpha}(t,s) \hat{G}^\ell_{\mu\alpha}(t,s) -   A^\ell_{\mu\alpha}(t,s) B^\ell_{\mu\alpha}(t,s)\right] \nonumber
    \\
    &+ \ln \mathcal{Z}[\{ \Phi, \hat\Phi, G,\hat G, A, B , j, v\}]
\end{align}
We thus see that the action $S$ consists of inner-products between order parameters $\{\Phi,G,A\}$ and their duals $\{\hat\Phi,\hat G, B\}$ as well as a single site MGF $\mathcal{Z}[\{ \Phi, \hat\Phi, G,\hat G, A, B , j, v\}]$, which is defined as 
\begin{align}
    \mathcal Z =& \int \prod_{\ell \mu t} \frac{d\hat{\chi}_\mu^\ell(t) d\chi_\mu^\ell(t)}{2\pi}\frac{d\hat{\xi}_\mu^\ell(t) d\xi_\mu^\ell(t)}{2\pi} \exp\left( \sum_{\ell \mu} \int_0^\infty dt \left[ \left(j^\ell_\mu(t) + i\hat\chi^\ell_\mu(t)\right) \chi_\mu^\ell(t) + \left(v^\ell_\mu(t) + i \hat\xi^\ell_\mu(t)\right) \xi_\mu^\ell(t) \right] \right) \nonumber
    \\
    &\times \exp\left( -\frac{1}{2} \sum_{\mu\alpha}\int_0^\infty dt \int_0^\infty  ds \hat{\chi}_\mu^1(t) \hat{\chi}_\alpha^1(s) K^x_{\mu\alpha} - \frac{1}{2} \sum_{\mu\alpha} \int_0^\infty dt \int_0^\infty ds \hat\xi^L_\mu(t) \hat\xi^L_\alpha(s)  \right) \nonumber
    \\
    &\times \exp\left( - \frac{1}{2} \sum_{\ell=1}^{L-1} \sum_{\mu\alpha}\int_0^\infty dt \int_0^\infty ds \left[ \hat{\chi}_\mu^{\ell+1}(t) \hat\chi_\alpha^{\ell+1}(s) \Phi^\ell_{\mu\alpha}(t,s) + \hat\xi^\ell_\mu(t) \hat\xi^\ell_\alpha(s) G^{\ell+1}_{\mu\alpha}(t,s) \right] \right) \nonumber
    \\
    &\times \exp\left( - \sum_{\ell=1}^L \sum_{\mu\alpha} \int_0^\infty dt \int_0^\infty ds \left[ \phi(h^\ell_\mu(t)) \phi(h_\alpha^\ell(s)) \hat\Phi_{\mu\alpha}^\ell(t,s) + g^{\ell}_\mu(t) g^\ell_\alpha(s) \hat G^\ell_{\mu\alpha}(t,s) \right] \right) \nonumber
    \\
    &\times \exp\left( - i \sum_{\ell=1}^L \sum_{\mu\alpha} \int_0^\infty dt \int_0^\infty ds \left[ \phi(h^\ell_\mu(t)) \hat\xi^\ell_\alpha(s) B^\ell_{\mu\alpha}(t,s) + \hat{\chi}^{\ell+1}_\mu(t) g^{\ell+1}_\alpha(s)  A^\ell_{\mu\alpha}(t,s) \right] \right)
\end{align}

\subsection{Saddle Point Equations}

Since the integrand in the moment generating function $Z$ takes the form $e^{N S[\{\Phi,\hat\Phi,G,\hat G, A, B\}]}$, the $N\to \infty$ limit can be obtained from saddle point integration, also known as the method of steepest descent \cite{bender1999advanced}. This consists in finding order parameters $\{\Phi,\hat\Phi,G,\hat G, A, B\}$ which render the action $S$ locally stationary. Concretely, this leads to the following saddle point equations.
\begin{align}
\frac{\delta S}{\delta \hat \Phi^\ell_{\mu\alpha}(t,s)}  &= {\Phi}_{\mu\alpha}^\ell(t,s) + \frac{1}{\mathcal Z} \frac{\delta \mathcal Z}{\delta \hat \Phi^\ell_{\mu\alpha}(t,s)} = {\Phi}_{\mu\alpha}^\ell(t,s) - \left< \phi(h_\mu^\ell(t)) \phi(h_\alpha^\ell(s))  \right> = 0 \nonumber
\\
\frac{\delta S}{\delta \Phi^\ell_{\mu\alpha}(t,s)} &= \hat\Phi^\ell_{\mu\alpha}(t,s) + \frac{1}{\mathcal Z} \frac{\delta \mathcal Z}{\delta \Phi^\ell_{\mu\alpha}(t,s)} = \hat\Phi^\ell_{\mu\alpha}(t,s) - \frac{1}{2} \left< \hat{\chi}^{\ell+1}_\mu(t)\hat{\chi}^{\ell+1}_\alpha(s)  \right> = 0 \nonumber
\\
\frac{\delta S}{\delta \hat G_{\mu\alpha}^\ell(t,s)}  &= G_{\mu\alpha}^\ell(t,s) + \frac{1}{\mathcal Z} \frac{\delta \mathcal Z}{\delta \hat{G}_{\mu\alpha}^\ell(t,s)} = G^\ell_{\mu\alpha}(t,s) - \left< g_\mu^\ell(t) g_\alpha^\ell(s) \right> = 0 \nonumber
\\
\frac{\delta S}{\delta  G_{\mu\alpha}^\ell(t,s)}  &= \hat G_{\mu\alpha}^\ell(t,s) + \frac{1}{\mathcal Z} \frac{\delta \mathcal Z}{\delta {G}_{\mu\alpha}^\ell(t,s)} = \hat G^\ell_{\mu\alpha}(t,s) - \frac{1}{2} \left< \hat g_\mu^\ell(t) \hat g_\alpha^\ell(s) \right> = 0 \nonumber
\\
\frac{\delta S}{\delta  A^\ell_{\mu\alpha}(t,s)}  &= -B_{\mu\alpha}^\ell(t,s) + \frac{1}{\mathcal Z} \frac{\delta \mathcal Z}{\delta A^\ell_{\mu\alpha}(t,s)} = -B_{\mu\alpha}^\ell(t,s) - i \left< \hat\chi^{\ell+1}_\mu(t) g_\alpha^{\ell+1}(s) \right> = 0 \nonumber
\\
\frac{\delta S}{\delta  B^\ell_{\mu\alpha}(t,s)}  &= -A_{\mu\alpha}^\ell(t,s) + \frac{1}{\mathcal Z} \frac{\delta \mathcal Z}{\delta B^\ell_{\mu\alpha}(t,s)} = -A_{\mu\alpha}^\ell(t,s) - i \left< \phi(h_\mu^\ell(t)) \hat{\xi}^\ell_\alpha(s)  \right> = 0 
\end{align}
We use the notation $\left< \right>$ to denote an average over the self-consistent distribution on fields induced by the single-site moment generating function $\mathcal{Z}$ at the saddle point. Concretely if $\mathcal Z = \int d\chi d\xi d\hat\chi d\hat\xi \exp\left( - \mathcal H[\chi, \xi ,\hat\chi, \hat\xi] \right)$ then the single-site self-consistent average of observable $O([\chi, \xi ,\hat\chi, \hat\xi])$ is defined as
\begin{align}
    \left< O([\chi, \xi ,\hat\chi, \hat\xi]) \right> = \frac{1}{\mathcal Z} \int d\chi d\xi d\hat\chi d\hat\xi \ O([\chi, \xi ,\hat\chi, \hat\xi]) \exp\left( - \mathcal H[\chi, \xi ,\hat\chi, \hat\xi] \right)
\end{align}
To calculate the averages of the dual variables such as $\left< \hat\chi^{\ell+1} \hat\chi^{\ell+1} \right>$, it will be convenient to work with vector and matrix notation. We let $\bm{\chi}^\ell = \text{Vec}\{ \chi_\mu^\ell(t) \}_{\mu\in[P], t\in \mathbb R_+}$ represent the vectorization of the stochastic process over different samples and times and define the dot product between two of these vectors as $\a \cdot \b = \sum_{\mu=1}^P \int_0^\infty  dt \ a_\mu(t) b_\mu(t)$. We also apply this procedure on the kernels so that $\bm\Phi = \text{Mat}\{\Phi_{\mu\alpha}(t,s)\}_{\mu\alpha \in [P], t,s \in \mathbb{R}_+}$. Matrix vector products take the form $[\A \b]_{\mu,t} = \int_0^\infty ds \sum_\alpha A_{\mu\alpha}(t,s) b_\alpha(s)$. We can obtain the behavior of $\left< \bm{\hat \chi}^{\ell+1}_\mu \bm{\hat \chi}^{\ell+1 \top}_\mu \right>$ in terms of primal fields $\{\chi,\xi, h, z\}$ by insertion of a dummy source $\u$ into the effective partition function.

\begin{align}
    \left< \bm{\hat\chi}^{\ell+1} \bm{\hat\chi}^{\ell+1} \right> &= -\frac{\partial^2}{\partial \u \partial \u^\top } \left< \exp\left( i \u \cdot \bm{\hat\chi}^{\ell+1} \right) \right>|_{\u=\bm0} \nonumber
    \\
    &= - \frac{1}{\mathcal Z} \frac{\partial^2}{\partial \u \partial \u^\top } \int d\bm\chi^{\ell+1} ... \exp\left( - \frac{1}{2} \left( \bm\chi^{\ell+1} + \u - \A^\ell \g^{\ell+1} \right)^\top [\bm\Phi^{\ell}]^{-1} \left( \bm\chi^{\ell+1} + \u - \A^\ell \g^{\ell+1} \right) - ... \right) \nonumber 
    \\
    &= [\bm\Phi^{\ell}]^{-1} - [\bm\Phi^{\ell}]^{-1} \left< \left( \bm\chi^{\ell+1} - \A^\ell \g^{\ell+1} \right) \left( \bm\chi^{\ell+1} - \A^\ell \g^{\ell+1} \right)^\top  \right> \left[ \bm\Phi^{\ell} \right]^{-1}
\end{align}
Similarly, we can obtain the equation for $\left< \bm{\hat\xi}^\ell \bm{\hat\xi}^{\ell \top} \right>$ by inserting a dummy source $\r$ and differentiating near zero source
\begin{align}
    \left< \bm{\hat\xi}^{\ell} \bm{\hat\xi}^{\ell} \right> &= -\frac{\partial^2}{\partial \r \partial \r^\top } \left< \exp\left( i \hat{\r} \cdot \bm{\hat\xi}^{\ell} \right) \right>|_{\r=\bm0}  \nonumber
    \\
    &= [\G^{\ell+1}]^{-1} - [\G^{\ell+1}]^{-1} \left< ( \bm\xi^{\ell} - \B^{\ell \top} \bm\phi^\ell ) ( \bm\xi^{\ell} - \B^{\ell \top} \bm\phi^\ell )^\top \right>[\G^{\ell+1}]^{-1}
\end{align}
As we will demonstrate in the next subsection, these correlators must vanish. Lastly, we can calculate the remaining correlators in terms of primal variables
\begin{align}
    -i \left< \hat{\bm\chi}^{\ell+1} \g^{\ell+1,\top} \right> &= \frac{\partial }{\partial \u} \left< \exp\left( -i \hat{\u} \cdot \bm{\hat\chi}^{\ell+1} \right) \g^{\ell+1 \top}\right> = [\bm\Phi^\ell]^{-1} \left< (\bm\chi^{\ell+1} -\A^\ell \g^{\ell+1}) \g^{\ell+1 \top} \right> \nonumber
    \\
    -i \left< \phi(\h^\ell) \bm{\hat\xi}^{\ell\top} \right> &= \frac{\partial }{\partial \r^\top} \left< \phi(\h) \exp\left( -i \r \cdot \bm{\hat\xi}^\ell \right) \right> = \left< \bm\phi(\h^\ell) (\bm\xi^\ell - \B^{\ell\top} \bm\phi(\h^\ell)) \right> [\G^{\ell+1}]^{-1}
\end{align}

\subsection{Single Site Stochastic Process: Hubbard Trick}
To get a better sense of this distribution, we can now simplify the quadratic forms appearing in $\mathcal Z$ using the Hubbard trick \cite{hubbard1959calculation}, which merely relates a Gaussian function to its Fourier transform.
\begin{align}
    \exp\left( - \frac{1}{2} \x^\top \A \x \right) = \int_{\mathbb{R}^d} \frac{d\u}{(2\pi)^{d/2} \sqrt{\det \A}} \exp\left( - \frac{1}{2} \u^\top \A^{-1}\u - i \u \cdot \x \right) = \left< \exp\left( - i \u \cdot \x \right) \right>_{\u \sim \mathcal{N}(0,\A)}
\end{align}
Applying this to the quadratic forms in the single-site MGF $\mathcal{Z}$, we get
\begin{align}
    \exp&\left( -\frac{1}{2} \sum_{\mu\alpha} \int_0^\infty dt \int_0^\infty ds \ \hat\chi_\mu^{1}(t)  \hat\chi_\alpha^{1}(s) K^x_{\mu\alpha} \right) =\left< \exp\left( - i \sum_{\mu} \int_0^\infty dt \ u_\mu^{1}(t) \hat{\chi}_\mu^{\ell+1}(t)  \right) \right>_{\{u^1\} \sim \mathcal{GP}(0,\K^x \otimes \bm 1 \bm1^\top)} \nonumber 
    \\
    \exp&\left( -\frac{1}{2} \sum_{\mu\alpha} \int_0^\infty dt \int_0^\infty ds \ \hat\chi_\mu^{\ell+1}(t)  \hat\chi_\alpha^{\ell+1}(s) \Phi^\ell_{\mu\alpha}(t,s) \right) =\left< \exp\left( - i \sum_{\mu} \int_0^\infty dt \ u_\mu^{\ell+1}(t) \hat{\chi}_\mu^{\ell+1}(t)  \right) \right>_{\{u^\ell\} \sim \mathcal{GP}(0,\bm\Phi^{\ell})} \nonumber
    \\
    \exp&\left( -\frac{1}{2} \sum_{\mu\alpha} \int_0^\infty dt \int_0^\infty ds \ \hat\xi_\mu^{\ell}(t)  \hat\xi_\alpha^{\ell}(s) G^{\ell+1}_{\mu\alpha}(t,s) \right) = \left< \exp\left( - i \sum_{\mu} \int_0^\infty dt \ r_\mu^{\ell}(t) \hat{\xi}_\mu^{\ell}(t) \right) \right>_{\{r^\ell\} \sim \mathcal{GP}(0,\G^{\ell+1})} \nonumber
    \\
    \exp&\left( -\frac{1}{2} \sum_{\mu\alpha} \int_0^\infty dt \int_0^\infty ds \ \hat\xi_\mu^{L}(t)  \hat\xi_\alpha^{L}(s) \right) = \left< \exp\left( - i \sum_{\mu} \int_0^\infty dt \ r_\mu^{L}(t) \hat{\xi}_\mu^{\ell}(t) \right) \right>_{\{r^L\} \sim \mathcal{GP}(0,\bm 1\bm 1^\top)}
\end{align}

Next, we integrate over all $\hat\chi^\ell, \hat\xi^\ell$ variables which yield Dirac-delta functions
\begin{align}
    &\int \prod_{\mu t} \frac{d\hat\chi^\ell_\mu(t)}{2\pi} \exp\left( i \bm{\hat \chi}^\ell \cdot \left[ \bm\chi^\ell - \u^\ell -  \A^{\ell-1}\g^\ell   \right] \right)  = \delta\left(\bm\chi^\ell - \u^\ell - \A^{\ell-1} \g^\ell \right) \nonumber
    \\
    &\int \prod_{\mu t} \frac{d\hat\xi^\ell_\mu(t)}{2\pi} \exp\left( i \bm{\hat \xi}^\ell \cdot \left[ \bm\xi^\ell - \r^\ell -  \B^{\ell \top} \phi(\h^\ell)   \right] \right)  = \delta\left(\bm\xi^\ell - \r^\ell -  \B^{\ell \top} \phi(\h^\ell)\right) 
\end{align}
To remedy the notational asymmetry, we redefine $\bm B^\ell$ as its transpose $\B^\ell \to \B^{\ell \top}$. The presence of these delta-functions in the MGF $\mathcal Z$ indicate the constraints $\u^\ell = \bm\chi^\ell - \A^{\ell-1} \g^\ell$ and $\r^\ell = \bm\xi^\ell -  \B^{\ell} \phi(\h^\ell)$. We can thus return to the $\hat\Phi$ and $\hat G$ saddle point equations and verify that these order parameters vanish
\begin{align}
    \hat{\bm \Phi}^\ell &= - \frac{1}{2} \left< \bm{\hat\chi}^{\ell+1} \bm{\hat\chi}^{\ell+1\top} \right> =  \frac{1}{2} [\bm\Phi^{\ell}]^{-1} \left< \left( \bm\chi^{\ell+1} - \A^\ell \g^{\ell+1} \right) \left( \bm\chi^{\ell+1} - \A^\ell \g^{\ell+1} \right)^\top  \right> \left[ \bm\Phi^{\ell} \right]^{-1} - \frac{1}{2} [\bm\Phi^{\ell}]^{-1}  \nonumber
    \\
    &= \frac{1}{2} [\bm\Phi^{\ell}]^{-1} \left< \u^{\ell+1} \u^{\ell+1 \top}  \right> \left[ \bm\Phi^{\ell} \right]^{-1} - \frac{1}{2} [\bm\Phi^{\ell}]^{-1}  = 0 ,
\end{align}
since $\left< \u^{\ell+1} \u^{\ell+1 \top} \right> = \bm\Phi^{\ell}$. Following an identical argument, $\hat\G^\ell = 0$. After this simplification, the single site MGF takes the form
\begin{align}
    \mathcal Z[\{\j^\ell,\v^\ell\}] = \left< \int \prod_\ell d\bm\chi^\ell d\bm\xi^\ell \delta\left(\bm\chi^\ell - \u^\ell - \A^{\ell-1} \g^\ell \right) \delta\left(\bm\xi^\ell - \r^\ell -  \B^{\ell} \phi(\h^\ell)\right) \exp\left( i \j^\ell \cdot \bm\chi^\ell + i \v^\ell \cdot \bm\xi^\ell \right) \right>_{\{\u^\ell, \r^\ell\}}
\end{align}
The interpretation is thus that $\u^\ell, \r^\ell$ are sampled independently from their respective Gaussian processes and the fields $\bm\chi^\ell$ and $\bm\xi^\ell$ are determined in terms of $\u^\ell, \r^\ell, \h^\ell, \g^\ell$. This means that we can apply Stein's Lemma (integration by parts) \cite{stein1972bound} to simplify the last two saddle point equations
\begin{align}
    \A^\ell = \left< \phi(\h^\ell) \r^{\ell \top} \right> [\G^{\ell+1}]^{-1} = \left<\frac{\partial \phi(\h^\ell)}{\partial \r^{\ell \top}} \right> \ , \ \B^\ell = \left< \g^{\ell+1} \u^{\ell+1} \right> [\bm\Phi^\ell]^{-1} = \left< \frac{\partial \g^{\ell+1}}{\partial \u^{\ell+1\top}} \right>
\end{align}
\subsection{Final DMFT Equations}\label{app:final_dmft_result}
We can now close this stochastic process in terms of preactivations $h^\ell$ and pre-gradients $z^\ell$. To match the formulas provided in the main text, we rescale $A^\ell \to A^\ell /\gamma_0 = \mathcal{O}_{\gamma_0}(1)$ and $B^\ell \to B^\ell /\gamma_0 = \mathcal{O}_{\gamma_0}(1)$, which makes it clear that the non-Gaussian corrections to the $h_\mu^\ell(t), z_\mu^\ell(t)$ fields are $\mathcal{O}(\gamma_0)$. After this rescaling, we have the following complete DMFT equations.

\begin{subequations}
\begin{empheq}[box=\widefbox]{align}
   h_\mu^\ell(t) &= \chi_\mu^\ell(t) + \gamma_0 \int_0^t ds \sum_\alpha \Delta_\alpha(s) \Phi^{\ell-1}_{\mu\alpha}(t,s) z_\alpha(s) \dot\phi(h^\ell_\alpha(s)) \nonumber
    \\
    &= u^\ell_\mu(t) + \gamma_0 \int_0^t ds \sum_\alpha \left[  A^{\ell-1}_{\mu\alpha}(t,s) +  \Delta_\alpha(s)\Phi^{\ell-1}_{\mu\alpha}(t,s) \right] \dot\phi(h_\alpha^\ell(s)) z_\alpha^\ell(s) \nonumber
    \\
    z_\mu^\ell(t) &= \xi^\ell_\mu(t) + \gamma_0 \int_0^t ds \sum_\alpha \Delta_\alpha(s) G^{\ell+1}_{\mu\alpha}(t,s) \phi(h^\ell_\alpha(s)) \nonumber
    \\
    &= r_\mu^\ell(t) + \gamma_0 \int_0^t \sum_\alpha \left[ B^\ell_{\mu\alpha}(t,s) +  \Delta_\alpha(s) G^{\ell+1}_{\mu\alpha}(t,s)  \right] \phi(h_\alpha^\ell(s)) \nonumber
    \\
    \Phi^\ell_{\mu\alpha}(t,s) &= \left< \phi(h_\mu^\ell(t)) \phi(h^\ell_\alpha(s)) \right> \ , \ G^\ell_{\mu\alpha}(t,s) = \left< g^\ell_\mu(t) g^\ell_\alpha(s) \right> \nonumber
    \\ 
    A^\ell_{\mu\alpha}(t,s) &= \gamma_0^{-1} \left< \frac{\delta \phi(h^\ell_\mu(t))}{\delta r^\ell_\alpha(s)}  \right> \ , \ B^\ell_{\mu\alpha}(t,s) = \gamma_0^{-1} \left< \frac{\delta g^{\ell+1}_\mu(t)}{\delta u^{\ell+1}_\alpha(s)}  \right> \nonumber
\end{empheq}
\end{subequations}

The base cases in the above equations are that $A^{0} = B^L = 0$ and $\Phi^0_{\mu\alpha}(t,s) = K^x_{\mu\alpha}$ and $G^{L+1}_{\mu\alpha}(t,s) = 1$. From the above self-consistent equations, one obtains the NTK dynamics and consequently the output predictions of the network with $\frac{\partial f_\mu}{\partial t} = \sum_{\alpha} \Delta_\alpha(t) \left[ \sum_\ell G^{\ell+1}_{\mu\alpha}(t,t) \Phi^\ell_{\mu\alpha}(t,t) \right]$.

\subsection{Varying Network Widths and Initialization Scales}\label{app:vary_width_init}

In this section, we relax the assumption of network widths being equal while taking all widths to infinity at a fixed ratio. This will allow us to analyze the influence of bottlenecks on the dynamics. We let $N^\ell = a_\ell N$ represent the width of layer $\ell$. Without loss of generality, we can choose that $N^L = N$ and proceed by defining order parameters in the usual way
\begin{align}
    \Phi^\ell_{\mu\alpha}(t,s) = \frac{1}{N^\ell} \phi(\h^\ell_\mu(t) ) \cdot\phi(\h^\ell_\alpha(s)) \ , \ G^\ell_{\mu\alpha}(t,s) = \frac{1}{N^\ell} \g^\ell_\mu(t) \cdot \g^\ell_\alpha(s) 
\end{align}
Since $N^L = N$, the variable $\g^L =  \sqrt{N^L} \frac{\partial h^{L+1}}{\partial \h^L} = \w^L \odot \dot\phi(\h^L) = \mathcal{O}_{N,\gamma}(1)$ as desired. We extend this definition to each layer as before $\g^\ell = \sqrt{N^{\ell}} \frac{\partial h^{L+1}}{\partial \h^\ell}$ which again satisfies the recursion
\begin{align}
    \g^{\ell}_\mu(t) = \z^\ell_\mu(t) \odot \dot\phi(\h^\ell_\mu(t)) \ , \ \z^\ell_\mu(t) = \frac{1}{\sqrt{N^{\ell+1}}} \W^{\ell}(t)^\top \g^{\ell+1}_\mu(t)
\end{align}
Now, we need to calculate the dynamics on weights $\W^\ell$
\begin{align}
    \frac{d}{dt} \W^\ell &= \gamma^2 \sum_\mu \Delta_\mu \frac{\partial f_\mu}{\partial \W^\ell} = \gamma^2 \sum_\mu \Delta_\mu \frac{\partial f_\mu}{\partial \h^{\ell+1}_\mu} \cdot \frac{\partial \h^{\ell+1}_\mu}{\partial \W^{\ell}} \nonumber
    \\
    &= \frac{\gamma}{ \sqrt{N^{\ell}} \sqrt{N^{\ell+1}}} \sum_\mu \Delta_\mu \g_\mu^{\ell+1} \phi(\h_\mu^\ell)^\top 
\end{align}
Using our definition of the kernels and the $\h,\z$ fields
\begin{align}
    \h^{\ell}_\mu(t) &= \bm\chi^{\ell}_\mu(t) + \frac{\gamma}{\sqrt{N^{\ell}}} \sum_\alpha \int_0^t ds \ \Delta_\alpha(s) \g^{\ell}_{\alpha}(s) \Phi^{\ell-1}_{\mu\alpha}(t,s) \nonumber
    \\
    \z^\ell_\mu(t) &= \bm\xi^\ell_\mu(t) + \frac{\gamma}{\sqrt{N^{\ell}}} \sum_\alpha \int_0^t ds \ \Delta_\alpha(s) \phi(\h^\ell_\alpha(s)) G^{\ell+1}_{\mu\alpha}(t,s)
\end{align}
We also find the usual formula for the NTK
\begin{align}
    K^{NTK}_{\mu\alpha} = \gamma^2 \sum_\ell \text{Tr} \left[\frac{\partial f_\mu}{\partial \W^\ell} \right]^\top \frac{\partial f_\alpha}{\partial \W^\ell} = \Phi^L_{\mu\alpha} + \sum_{\ell=1}^{L-1} G^{\ell+1}_{\mu\alpha} \Phi^{\ell}_{\mu\alpha} +  G^1_{\mu\alpha} K^x_{\mu\alpha}
\end{align}

Now, as before, we need to consider the distribution of $\bm\chi,\bm\xi$ fields. We assume $W^\ell_{ij}(0) \sim \mathcal{N}(0,\sigma^2_\ell)$. This requires computing integrals like
\begin{align}
    &\left< \exp\left( i \sum_{\mu} \int_0^\infty dt \left[ \bm{\hat{\chi}}^{\ell+1}_{\mu}(t)^\top \W^{\ell}(0) \phi(\h^\ell_\mu(t)) /\sqrt{N^\ell} +  \g^{\ell+1}_\mu(t)^\top \W^\ell(0) \bm{\hat\xi}^\ell_\mu(t) /\sqrt{N^{\ell+1}} \right]  \right) \right>_{\W^\ell(0)} \nonumber
    \\
    &= \exp\left( - \frac{\sigma^2_\ell}{2} \sum_{\mu\alpha} \int_0^\infty dt \int_0^\infty ds \left[ \bm{\hat\chi}^{\ell+1}_\mu(t) \cdot \bm{\hat\chi}^{\ell+1}_\mu(t) \Phi^{\ell}_{\mu\alpha}(t,s) + \bm{\hat\xi}^{\ell}_\mu(t) \cdot \bm{\hat\xi}^{\ell}_\mu(t) G^{\ell+1}_{\mu\alpha}(t,s)   \right] \right) \nonumber
    \\
    &\times \exp\left( - i \sigma^2_\ell \sqrt{\frac{a_{\ell}}{a_{\ell+1}}} \sum_{\mu\alpha}\int_0^\infty dt \int_0^\infty ds A^\ell_{\mu\alpha}(t,s) \bm\chi^{\ell+1}_\mu(t) \cdot \g^{\ell+1}_\alpha(s) \right)
\end{align}
where $\A^{\ell}_{\mu\alpha}(t,s) = -\frac{i}{N^{\ell}} \bm\phi(\h^\ell_{\mu}(t)) \cdot \bm{\hat\xi}^{\ell}_\alpha(s)$. 
The action thus takes the form
\begin{align}
    S = \sum_\ell a_\ell \text{Tr} \left[ \bm{\hat\Phi}^{\ell \top} \bm\Phi^\ell + \G^{\ell \top} \hat{\G}^\ell - \A^{\ell \top} \B^\ell \right] + \sum_\ell a_\ell \ln \mathcal{Z}_\ell
\end{align}
where the zero-source MGF for layer $\ell$ has the form
\begin{align}
    \mathcal{Z}_\ell = \int \prod_{\mu t}\frac{d\chi_\mu^\ell(t) d\hat{\chi}_\mu^\ell(t) }{2\pi} \frac{d\xi_\mu^\ell(t) d\hat{\xi}_\mu^\ell(t)}{2\pi} &\exp\left( - \phi(\h^\ell)^\top \bm{\hat\Phi}^\ell \phi(\h^\ell) - \g^{\ell \top} \hat{\G}^\ell \g^\ell   + i \bm{\chi}^\ell \cdot  \bm{\hat{\chi}}^\ell + i \bm{\xi}^\ell \cdot \bm{\hat{\xi}}^\ell  \right) \nonumber
    \\
    &\exp\left( - \frac{\sigma^2_{\ell-1}}{2} \bm{\hat{\chi}}^\ell \bm\Phi^{\ell-1} \bm{\hat\chi}^\ell - \frac{\sigma^2_\ell}{2}  \bm{\hat{\xi}}^\ell \G^{\ell+1} \bm{\hat \xi}^\ell \right) \nonumber
    \\
    &\exp\left( - i \sigma^2_{\ell-1} \sqrt{\frac{a_{\ell-1}}{a_{\ell}}} \bm{\hat{\chi}}^{\ell} \A^{\ell-1} \g^\ell - i  \phi(\h^\ell)^\top \B^{\ell}\bm{\hat{\xi}}^{\ell} \right) 
\end{align}
The saddle point equations give
\begin{align}
    \bm\Phi^\ell &=  \left< \phi(\h^\ell) \phi(\h^\ell)^\top \right> \ , \ \G^\ell = \left< \g^\ell \g^{\ell \top} \right> \nonumber
    \\
     \A^\ell &= -i \left< \phi(\h^\ell) \bm{\hat{\xi}}^{\ell\top} \right>  =  \left< \frac{\partial \phi(\h^\ell)}{\partial \r^{\ell\top}} \right> \nonumber
     \\  
    a_\ell \B^\ell &= -i a_{\ell+1} \sigma^2_\ell \sqrt{\frac{a_\ell}{a_{\ell+1}}}  \left< \bm{\hat{\chi}}^{\ell+1} \g^{\ell+1,\top} \right> \implies  \B^\ell = \sigma^2_\ell \sqrt{\frac{a_{\ell+1}}{a_{\ell}}}  \left< \frac{\partial \g^{\ell+1 \top}}{\partial \u^{\ell+1}} \right>
\end{align}
where $\u^\ell \sim \mathcal{GP}(0,\sigma^2_{\ell-1} \bm\Phi^{\ell-1}) , \r^{\ell} \sim \mathcal{GP}(0,\sigma^2_\ell \G^{\ell+1})$. We redefine $\B^\ell \to \frac{1}{\sigma^2_\ell} \sqrt{\frac{a_\ell}{a_{\ell+1}}}\B^\ell$. To take the $N\to\infty$ limit of the field dynamics, again use $\gamma_{0} = \gamma / \sqrt{N} = O_{N}(1)$. The field equations take the form
\begin{align}
    h^\ell_\mu(t) &= u_\mu^\ell(t) + \int_0^\infty \sum_{\alpha=1}^P \left[ \sigma^2_{\ell-1} \sqrt{\frac{a_{\ell-1}}{a_{\ell}}} A_{\mu\alpha}^{\ell-1}(t,s) +  \frac{\gamma_0}{\sqrt{a_\ell}}  \Theta(t-s) \Phi^{\ell-1}_{\mu\alpha}(t,s)  \right] \dot\phi(h^\ell_\alpha(s)) z^\ell_\alpha(s) \nonumber
    \\
    z^\ell_\mu(t) &= r_\mu^\ell(t) +  \int_0^\infty \sum_{\alpha=1}^P \left[\sigma^2_\ell \sqrt{\frac{a_{\ell+1}}{a_\ell}} B_{\mu\alpha}^{\ell}(t,s) + \frac{\gamma_0}{\sqrt{a_\ell}} \Theta(t-s) G^{\ell+1}_{\mu\alpha}(t,s) \right] \phi(h^\ell_\alpha(s)) 
\end{align}
We thus find that the evolution of the scalar fields in a given layer is set by the parameter $\gamma_{0}/\sqrt{a_\ell}$, indicating that relatively wider layers evolve less and contribute less of a change to the overall NTK. This definition for $\A^\ell,\B^\ell$ is non-ideal to extract intuition about bottlenecks since $\A^{\ell-1} \sim \mathcal{O}\left( \frac{\gamma_0}{\sqrt{a_{\ell-1}}} \right)$ and $\B^{\ell} \sim \mathcal{O}\left( \frac{\gamma_0}{\sqrt{a_{\ell+1}}} \right)$. To remedy this, we redefine $\tilde{\A}^\ell = \frac{\sqrt{a_\ell}}{\gamma_0} \A^\ell, \tilde{\B}^\ell = \frac{\sqrt{a_{\ell+1}}}{\gamma_0} \B^\ell$. With this choice, we have

\begin{align}
    h^\ell_\mu(t) &= u_\mu^\ell(t) + \frac{\gamma_0}{\sqrt{a_\ell}} \int_0^\infty \sum_{\alpha=1}^P \left[ \sigma^2_{\ell-1}  \tilde{A}_{\mu\alpha}^{\ell-1}(t,s) + \Theta(t-s) \Phi^{\ell-1}_{\mu\alpha}(t,s)  \right] \dot\phi(h^\ell_\alpha(s)) z^\ell_\alpha(s) \nonumber
    \\
    z^\ell_\mu(t) &= r_\mu^\ell(t) +  \frac{\gamma_0}{\sqrt{a_\ell}} \int_0^\infty \sum_{\alpha=1}^P \left[\sigma^2_\ell  \tilde{B}_{\mu\alpha}^{\ell}(t,s) +  \Theta(t-s) G^{\ell+1}_{\mu\alpha}(t,s) \right] \phi(h^\ell_\alpha(s)) 
\end{align}
where $\tilde{A}^{\ell-1}, \tilde{B}^\ell$ do not have a leading order scaling with $a_{\ell-1}$ or $a_{\ell+1}$ respectively. Under this change of variables, it is now apparent that a very wide layer $\ell$, where $\frac{\gamma_0}{\sqrt{a_\ell}} \ll 1$ is small, the fields $h^\ell, z^\ell$ become well approximated by the Gaussian processes $u^\ell, r^\ell$, albeit with evolving covariances $\bm\Phi^{\ell-1}, \G^{\ell+1}$ respectively. In a realistic CNN architecture where the number of channels increases across layers, this result would predict that more feature learning and deviations from Gaussianity to occur in the early layers and the later layers to be well approximated as Gaussian fields $u^\ell, r^\ell$ with temporally evolving covariances for $\ell \sim L$. We leave evaluation of this prediction to future work.

\section{Two Layer Networks}\label{app:two_layer}

In a two layer network, there are no $\A$ or $\B$ order parameters, so the fields $\chi^1$ and $\xi^1$ are always independent. Further, $\chi^1$ and $\xi^1$ are both constant throughout training dynamics. Thus we can obtain differential rather than integral equations for the stochastic fields $h^1, z^1$ which are
\begin{align}
    \frac{\partial }{\partial t} h^1_\mu(t) &= \gamma_0 \sum_{\alpha=1}^P \Delta_\alpha(t) K^x_{\mu\alpha} \dot\phi(h_\alpha^1(t)) z^{1}(t)  \ , \ \frac{\partial }{\partial t} z^1(t) =  \gamma_0 \sum_{\alpha=1}^P \Delta_\alpha(t) \phi(h_\alpha^1(t))  \nonumber 
    \\
    \Phi^1_{\mu\alpha}(t) &= \left< \phi(h^1_\mu(t)) \phi(h^1_\alpha(t)) \right> 
    \ , \ G^1_{\mu\alpha}(t) = \left< z(t)^2 \dot\phi(h^1_\mu(t))  \dot\phi(h^1_\alpha(t)) \right> \nonumber
    \\
    \frac{\partial}{\partial t} \Delta_\mu(t) &= -\sum_{\alpha=1}^P \left[ G^1_{\mu\alpha}(t) K^x_{\mu\alpha} + \Phi^1_{\mu\alpha}(t) \right] \Delta_\alpha(t)
\end{align}
where the average is taken over the random initial conditions $\h^1(0) \sim \mathcal{N}(0,\K^x)$ and $\z^1(0) \sim \mathcal{N}(0,\bm 1 \bm 1^\top)$. An example of the two layer theory for a ReLU network can be found in Appendix Figure \ref{fig:two_layer_relu}. In this two layer setting, a drift PDE can be obtained for the joint density of preactivations and feedback fields $p(\h,z;t)$
\begin{align}
    \frac{\partial}{\partial t} p( \h, z, t) &=  - p(\h,z,t) z(t) \sum_{\mu} \Delta_\mu(t) K^x_{\mu\mu} \ddot\phi(h_\mu(t))  \nonumber
    \\
    &-\gamma_0 \sum_{\mu \alpha} K^x_{\mu\alpha} \Delta_\alpha \dot\phi(h_\alpha(t)) z(t) \frac{\partial p(\h, \z, t)}{\partial h_\mu} - \gamma_0 \sum_{\mu\alpha} \Delta_\alpha \phi(h_\alpha) \frac{\partial p(\h,\z,t)}{\partial z_\mu}  \nonumber
    \\
    \frac{\partial}{\partial t} \Delta_\mu(t) &= -\sum_{\alpha=1}^P \left[ G^1_{\mu\alpha}(t) K^x_{\mu\alpha} + \Phi^1_{\mu\alpha}(t) \right] \Delta_\alpha(t) \nonumber 
    \\
    \Phi^1_{\mu\alpha}(t) &= \left< \phi(h^1_\mu(t)) \phi(h^1_\alpha(t)) \right> 
    \ , \ G^1_{\mu\alpha}(t) = \left< z^1(t)^2 \dot\phi(h^1_\mu(t))  \dot\phi(h^1_\alpha(t)) \right>
    ,
\end{align}
which is a zero-diffusion feature space version of the PDE derived in the original two layer mean field limit of neural networks \cite{mei2018mean, mei2019mean, nguyen2019mean}.

\section{Deep Linear Networks}\label{app:linear_theory}

In the deep linear case, the $g^\ell_\mu(t)$ fields are independent of sample index $\mu$. We introduce the kernel $H^\ell_{\mu\alpha}(t,s) = \left< h^\ell_\mu(t) h^\ell_\alpha(s) \right>$. The field equations are 
\begin{align}
    h^\ell_\mu(t) &= u^\ell_\mu(t) + \gamma_0 \int_0^\infty \sum_{\alpha=1}^P  \left[ A^{\ell-1}_{\mu\alpha}(t,s) + \Theta(t-s) H^{\ell-1}_{\mu\alpha}(t,s) \right] \Delta_\alpha(s) g^{\ell}(s) \nonumber
    \\
    g^\ell(t) &= r^\ell(t) + \gamma_0 \int_0^\infty \sum_{\alpha=1}^P [ B^{\ell}_{\alpha}(t,s) + \gamma_0 \Theta(t-s) G^{\ell+1}(t,s) ] \Delta_\alpha(s) h^\ell_{\alpha}(s)
\end{align}

Or in vector notation $\h^{\ell} = \u^{\ell} + \gamma_0 \C^{\ell}\g^{\ell}$ and $\g^{\ell} = \r^{\ell} + \gamma_0 \D^{\ell} \h^{\ell}$ where
\begin{align}
    C^{\ell}_{\mu}(t,s) = \sum_{\alpha=1}^P [ A_{\mu\alpha}^{\ell-1}(t,s) + \Theta(t-s) H^{\ell-1}_{\mu\alpha}(t,s) ] \Delta_\alpha(s) \ , \ D^{\ell}_{\mu}(t,s) = [B^{\ell}_{\mu}(t,s) + \Theta(t-s) G^{\ell+1}(t,s)] \Delta_\mu(s)
\end{align}
Using the formulas which define the fields, we have
\begin{align}
    \h^{\ell} &= \u^\ell + \gamma_0 \C^{\ell} \r^\ell + \gamma_0^2 \C^\ell \D^\ell \h^\ell \implies \h^\ell = (\I - \gamma_0^2 \C^\ell \D^\ell)^{-1} [\u^\ell + \gamma_0 \C^\ell \r^\ell ] \nonumber
    \\
    \g^{\ell} &= \r^\ell + \gamma_0 \D^\ell \u^\ell + \gamma_0^2 \D^\ell \C^\ell \g^\ell \implies \g^\ell = (\I - \gamma_0^2 \D^\ell \C^\ell)^{-1} [ \r^\ell + \gamma_0 \D^\ell \u^\ell] 
\end{align}
The saddle point equations can thus be written as
\begin{align}\label{eq_app:HG_linear}
    \H^{\ell} &= \left< \h^\ell \h^{\ell \top}  \right> = (\I - \gamma_0^2 \C^\ell \D^\ell)^{-1} [\H^{\ell-1} + \gamma_0^2 \C^{\ell} \G^{\ell+1} \C^{\ell \top} ]\left[(\I - \gamma_0^2 \C^\ell \D^\ell)^\top\right]^{-1} \nonumber
    \\
    \G^{\ell} &= \left< \g^\ell \g^{\ell \top} \right> = \left( \I - \gamma_0^2 \D^{\ell} \C^\ell \right)^{-1} \left[ \G^{\ell+1} + \gamma^2_0 \D^{\ell} \H^{\ell-1} \D^{\ell \top} \right] \left[\left( \I - \gamma_0^2 \D^{\ell} \C^\ell \right)^\top \right]^{-1} \nonumber
    \\
    \A^\ell &= (\I - \gamma_0^2 \C^{\ell} \D^{\ell})^{-1} \C^{\ell} \ , \ \B^{\ell-1} = (\I - \gamma_0^2 \D^\ell \C^\ell)^{-1} \D^\ell
\end{align}
We solve these equations by repeatedly updating $\H^\ell, \G^\ell$, using Equation \eqref{eq_app:HG_linear} and the current estimate of $\C^\ell, \D^\ell$. We then use the new $\H^\ell, \G^\ell$ to recompute $\K^{NTK}$ and $\bm\Delta(t)$, calculating $\C^\ell, \D^\ell$ and then recomputing $\H^\ell, \G^\ell$. This procedure usually converges in $\sim 5-10$ steps.

\subsection{Two Layer Linear Network}

As we saw in Appendix \ref{app:two_layer}, the field dynamics simplify considerably in the two layer case, allowing description of all fields in terms of differential equations. In a two layer linear network, we let $\h(t) \in \mathbb{R}^P$ represent the hidden activation field and $g(t) \in \mathbb{R}$ represent the gradient
\begin{align}
    \frac{\partial }{ \partial t} \h(t) = \gamma_0 g(t) \K^x \bm\Delta(t)
    \ , \ \frac{\partial}{\partial t} g(t) = \gamma_0 \bm\Delta(t) \cdot \h(t)
\end{align}
The kernels $\H(t) = \left< \h(t) \h(t)^\top \right>$ and $G(t) = \left< g(t)^2  \right>$ thus evolve as
\begin{align}
    \frac{\partial}{\partial t} \H(t) &= \gamma_0 \K^x \bm\Delta \left< g(t) \h(t)^\top \right> + \gamma_0  \left< g(t) \h(t) \right>  \bm\Delta^\top \K^x \nonumber
    \\
    \frac{\partial}{\partial t} G(t) &= 2 \gamma_0 \left< g(t) \h(t) \right> \cdot \bm\Delta(t)
\end{align}
It is easy to verify that the network predictions on the $P$ training points are $\f(t) = \y - \bm\Delta(t) = \frac{1}{\gamma_0} \left< g(t) \h(t) \right> \in \mathbb{R}^P$. Thus the dynamics of $\H(t), G(t)$ and $\bm\Delta(t)$ close
\begin{align}
    \frac{\partial}{\partial t} \H(t) &= \gamma_0^2 \K^x \bm\Delta (\y - \bm\Delta)^\top + \gamma_0^2 (\y-\bm\Delta)  \bm\Delta^\top \K^x \nonumber
    \\
    \frac{\partial}{\partial t} G(t) &= 2 \gamma_0^2 (\y - \bm\Delta) \cdot \bm\Delta(t) \nonumber
    \\
    \frac{\partial}{\partial t} \bm\Delta(t) &= - [\H(t) + G(t) \K^x ] \bm\Delta(t)
\end{align}
where the initial conditions are $\H(0) = \bm I$, $G(0)=1$ and $\bm\Delta(0) = \y$. These equations hold for any choice of data $\K^x, \y$.

\subsubsection{Whitened Data in Two Layer Linear}\label{app:two_layer_whitened}

For input data which is whitened where $\K^x = \I$, then the dynamics can be simplified even further, recovering the sigmoidal curves very similar to those obtained under a special initialization \cite{fukumizu1998dynamics, saxe2013exact, advani2020high, atanasov2022neural}. In this case we note that the error signal always evolves in the $\y$ direction, $\bm\Delta(t) =  \Delta(t) \frac{\y}{|\y|}$, and that $\H$ only evolves in a rank one direction $\y \y^\top$ direction as well. Let $\frac{1}{|\y|^2} \y^\top \H(t) \y = H_y(t)$. Let $y = |\y|$ represent the norm of the target vector, then the relevant scalar dynamics are
\begin{align}
    \frac{\partial }{\partial t} H_y(t) &= 2 \gamma_0^2 \Delta(t) (y - \Delta(t))  
    \ , \ \frac{\partial}{\partial t} G(t) = 2\gamma_0^2 \Delta(t) (y-\Delta(t)) \nonumber
    \\
    \frac{\partial}{\partial t} \Delta(t) &= - [H_y(t) + G(t) ] \Delta(t)
\end{align}
Now note that, at initialization $H_y(0) = G(0) = 1$ and that $ \frac{\partial }{\partial t} H_y(t) = \frac{\partial}{\partial t} G(t)$. Thus, we have an automatic balancing condition $H_y(t) = G(t)$ for all $t \in \mathbb{R}_+$ and the dynamics reduce to two variables
\begin{align}
    \frac{\partial }{\partial t} H_y(t) &= 2 \gamma_0^2 \Delta(t) (y - \Delta(t)) 
    \ , \ 
    \frac{\partial}{\partial t} \Delta(t) = - 2 H_y(t) \Delta(t)
\end{align}
We note that this system obeys a conservation law which constrains $(H_y, y-\Delta)$ to a hyperbola 
\begin{align}
    \frac{1}{2} \frac{\partial}{\partial t} \left[ H_y^2 - \gamma_0^2 (y-\Delta(t))^2  \right] = 2\gamma_0^2 H_y \Delta(y-\Delta) - 2\gamma_0^2 H_y \Delta(y-\Delta) = 0
\end{align}
This conservation law implies that $H_y(0)^2 = 1 = \lim_{t\to\infty} H_y(t)^2 - \gamma_0^2 y^2$ or that the final kernel has the form $\lim_{t\to\infty} \H(t) = \frac{1}{y^2}[\sqrt{1 + \gamma_0^2 y^2 } -1 ] \y \y^\top + \I$. The result that the final kernel becomes a rank one spike in the direction of the target function was also obtained in finite width networks in the limit of small initialization \cite{atanasov2022neural} and also from a normative toy model of feature learning \cite{shan_bordelon}. We can use the conservation law above $1 = H_y(t)^2 - \gamma_0^2 (\Delta(t) - y)^2$ to simplify the dynamics to a one dimensional system
\begin{align}
    \frac{\partial}{\partial t} \Delta(t) = - 2\sqrt{1 + \gamma_0^2 (\Delta(t)-y)^2 } \ \Delta(t) \implies \frac{\partial}{\partial t} f = 2 \sqrt{1+\gamma_0^2 f^2 } (y - f)
\end{align}
where $f = y-\Delta$. We see that increasing $\gamma_0$ provides strict acceleration in the learning dynamics, illustrating the training benefits of feature evolution. Since this system is separable, we can solve for the time it takes for the network output norm to reach output level $f$
\begin{align}\label{eq:analytic_timescale_whitened}
    2t = \int_0^{f} \frac{ds}{(y-s) \sqrt{1+\gamma_0^2 s^2 }} &= \frac{1}{\sqrt{1+\gamma_0^2 y^2}} \tanh^{-1}\left( \frac{1 + \gamma_0^2 y f}{\sqrt{1+\gamma_0^2 y^2} \sqrt{1+\gamma_0^2 f^2 }} \right) \nonumber
    \\
    &-\frac{1}{\sqrt{1+\gamma_0^2 y^2}} \tanh^{-1}\left( \frac{1}{\sqrt{1+\gamma_0^2 y^2}} \right)
\end{align}
The NTK limit can be obtained by taking $\gamma_0 \to 0$ which gives
\begin{align}
    \frac{\partial }{\partial t} \Delta(t) \sim - 2\Delta(t) \implies \Delta(t) \sim e^{-2t}
\end{align}
which recovers the usual convergence rate of a linear model. The right hand side of Equation \eqref{eq:analytic_timescale_whitened} has a perturbation series in $\gamma_0^2$ which converges in the disk $\gamma_0 < \frac{1}{y}$. The other limit of interest is the $\gamma_0 \to \infty$ limit where
\begin{align}
    \frac{d}{dt} \Delta(t) \sim - 2\gamma_0 (y - \Delta(t)) \Delta(t) 
\end{align}
which recovers the logistic growth observed in the initialization scheme of prior works \cite{fukumizu1998dynamics, saxe2013exact}. The timescale $\tau$ required to learn is only $\tau \sim \frac{1}{\gamma_0} \ll 1$, which is much smaller than the $O_{\gamma_0}(1)$ time to learn predicted from the small $\gamma_0$ expansion. We note that the above leading order asymptotic behavior at large $\gamma_0$ considers the DMFT initial condition $\Delta(0) = y$ as an unstable fixed point. For realistic learning curves, one would need to stipulate some alternative initial condition such as $\Delta = y-\epsilon$ for some small $\epsilon > 0$ in order to have nontrivial leading order dynamics. 


\subsection{Deep Linear Whitened Data}\label{app:deep_whitened_linear}

In this section, we examine the role of depth when linear networks are trained on whitened data. As in the two layer case, all hidden kernels $\H^\ell(t,s)$ need only be tracked in the one dimensional task relevant subspace along the vector $\y$. We let $\Delta(t) = \frac{1}{y}\y \cdot \bm\Delta(t)$ and let $h_y(t) = \frac{1}{y} \h^\ell(t) \cdot \y$. We have 
\begin{align}
    h^{\ell}_y(t) &= u_y^\ell(t) + \gamma_0 \int_0^\infty ds \ C^\ell(t,s) g^\ell(s) \ , \ C^{\ell}(t,s)= A_y^{\ell-1}(t,s) + \Theta(t-s) H^{\ell-1}_y(t,s) \Delta(s) \nonumber 
    \\
    g^\ell(t) &= r^\ell(t) + \gamma_0 \int_0^\infty ds \ D^\ell(t,s) h_y^\ell(s) \ , \ D^{\ell}(t,s) = B_y^{\ell-1}(t,s) + \Theta(t-s) G^{\ell+1}(t,s) \Delta(s)
\end{align}
Lastly we have the simple evolution equation for the scalar error $\Delta(t)$
\begin{align}
    \frac{\partial \Delta(t)}{\partial t} = - \sum_{\ell=0}^L G^{\ell+1}(t,t) H_y^{\ell}(t,t) \Delta(t) \implies \Delta(t) = \exp\left( - \int_0^t ds \sum_{\ell=0}^L G^{\ell+1}(s,s) H^\ell_y(s,s) \right) y
\end{align}
Vectorizing we find the following equations for the time $\times$ time matrix order parameters $\h^\ell = \u^\ell + \gamma_0 \C^\ell \g^\ell \ , \ \g^\ell = \r^\ell + \gamma_0 \D^\ell \h^\ell$, we can solve for the response functions $\A^{\ell} = \left( \I - \gamma_0^2 \C^\ell \D^\ell \right)^{-1} \C^{\ell}$ and $\B^\ell = \left( \I - \gamma_0^2 \D^\ell \C^\ell \right)^{-1} \D^{\ell}$. This formulation has the advantage that it no longer has any sample-size dependence: arbitrary sample sizes can be considered with no computational cost.

\section{Convolutional Networks with Infinite Channels}\label{app:cnn}

The DMFT described in this work can be extended to CNNs with infinitely many channels, much in the same way that infinite CNNs have a well defined kernel limit \cite{novak2018bayesian, yang2019tensor}. We let $W^{\ell}_{ij,\mathfrak{a}}$ represent the value of the filter at spatial displacement $\mathfrak{a}$ from the center of the filter, which maps relates activity at channel $j$ of layer $\ell$ to channel $i$ of layer $\ell+1$. The fields $h_{\mu,i,\mathfrak{a}}^{\ell}$ are defined recursively as
\begin{align}
    h^{\ell+1}_{\mu,i,\mathfrak{a}} = \frac{1}{\sqrt N} \sum_{j=1}^N \sum_{\mathfrak{b} \in \mathcal S^\ell} W^{\ell}_{ij,\mathfrak{b}} \phi(h_{\mu,j, \mathfrak{a+b}}^\ell) \ , \ i \in \{ 1, ... , N\}
\end{align}
where $\mathcal{S}^\ell$ is the spatial receptive field at layer $\ell$. For example, a $(2k+1) \times (2k+1)$ convolution will have $\mathcal{S}^\ell = \{ (i,j) \in \mathbb{Z}^2 : -k\leq i \leq k, -k \leq j \leq k \}$. The output function is obtained from the last layer is defined as $f_\mu = \frac{1}{\gamma_0 N} \sum_{i=1}^N w_{i,\mathfrak{a}}^{L} \phi(h^L_{\mu,i,\mathfrak{a}})$. The gradient fields have the same definition as before $\g^\ell_{\mu,\mathfrak{a}} = \gamma_0 N \frac{\partial f_\mu}{\partial \h^\ell_{\mu,\mathfrak{a}}}$, which as before enjoy the following recursion from the chain rule
\begin{align}
    \g^\ell_{\mu,\mathfrak{a}} = \gamma_0 N \sum_{\mathfrak{b}}  \frac{\partial f_\mu}{\partial \h^{\ell+1}_{\mu,\mathfrak{b}}} \cdot \frac{\partial \h^{\ell+1}_{\mu,\mathfrak{b}}}{\partial \h^\ell_{\mu,\mathfrak{a}}} = \dot\phi(\h^\ell_{\mu,\mathfrak{a}}) \odot\left[ \frac{1}{\sqrt N} \sum_{j=1}^N \sum_{\mathfrak{b} \in \mathcal{S}^{\ell}} \W^{\ell \top}_{\mathfrak{b}} \g^{\ell+1}_{\mu, \mathfrak{a-b}} \right]
\end{align}
The dynamics of each set of filters $\{\W^\ell_{\mathfrak{b}} \}$ can therefore be written in terms of the features $\h^\ell_{\mathfrak{a}}, \g^\ell_{\mathfrak{a}}$
\begin{align}
    \frac{d}{dt} \W^\ell_{\mathfrak{b}} = \frac{\gamma_0}{\sqrt N} \sum_{\mu, \mathfrak{a}} \Delta_\mu \g^{\ell+1}_{\mu,\mathfrak{a}} \phi(\h^{\ell}_{\mu,\mathfrak{a+b}})^\top .
\end{align}

The feature space description of the forward and backward pass relations is
\begin{align}
    \h^{\ell+1}_{\mu,\mathfrak{a}}(t) =  \bm\chi^{\ell+1}_{\mu,\mathfrak{a}}(t) + \gamma_0 \int_0^t ds \sum_{\alpha \mathfrak{b,c}}  \Delta_\alpha(s)  \Phi^{\ell}_{\mu\alpha,\mathfrak{a+b},\mathfrak{c+b}}(t,s) \g^{\ell+1}_{\alpha,\mathfrak{c}}(s) \nonumber
    \\
    \z^{\ell}_{\mu,\mathfrak{a}}(t) = \bm\xi^\ell_{\mu\mathfrak{a}}(t) + \gamma_0 \int_0^t ds \sum_{\alpha \mathfrak{b,c}} \Delta_\alpha(s)  G^{\ell+1}_{\mu\alpha,\mathfrak{a-b},\mathfrak{c-b}}(t,s) \phi(\h^{\ell}_{\alpha,\mathfrak{c}})
\end{align}
where $\bm\chi^{\ell+1}_{\mu,\mathfrak{a}}(t) = \frac{1}{\sqrt N} \W^\ell(0) \phi(\h^\ell_{\mu \mathfrak{a}}(t))$. The order parameters for this network architecture are
\begin{align}
    \Phi^{\ell}_{\mu\alpha,\mathfrak{ab}}(t,s) = \frac{1}{N} \phi(\h^\ell_{\mu \mathfrak{a}}(t)) \cdot \phi(\h^\ell_{\alpha \mathfrak{b}}(s)) \ , \  G^{\ell}_{\mu\alpha,\mathfrak{ab}}(t,s) = \frac{1}{N} \g^\ell_{\mu \mathfrak{a}}(t) \cdot \g^\ell_{\alpha \mathfrak{b}}(s) 
\end{align}
These two order parameters per layer collectively define the neural tangent kernel. Following the computation in \ref{app:dmft_derivation}, we obtain the following field theory in the $N \to \infty$ limit:
\begin{align}
    &\{ u^\ell_{\mu \mathfrak{a}}(t) \} \sim \mathcal{GP}(0,\bm\Phi^{\ell-1}) \ , \ \{ r^\ell_{\mu \mathfrak{a}}(t) \} \sim \mathcal{GP}(0,\G^{\ell+1}) \nonumber
    \\
    h^\ell_{\mu \mathfrak{a}}(t) &= u^\ell_{\mu \mathfrak{a}}(t) + \gamma_0 \int_0^t ds \sum_{\alpha,\mathfrak{b}}  A^{\ell-1}_{\mu\alpha,\mathfrak{ab}}(t,s) \dot\phi(h^\ell_{\alpha \mathfrak{b}}(s)) z^\ell_{\alpha \mathfrak{b}}(s) \nonumber
    \\
    &+ \gamma_0 \int_0^t ds \sum_{\alpha \mathfrak{b,c}} \Delta_\alpha(s) \Phi^{\ell-1}_{\mu\alpha,\mathfrak{a+b},\mathfrak{c+b}} \dot\phi(h^\ell_{\alpha \mathfrak{c}}(s)) z^\ell_{\alpha \mathfrak{c}}(s) \nonumber
    \\
    z^\ell_{\mu \mathfrak{a}}(t) &= r^\ell_{\mu \mathfrak{a}}(t) + \gamma_0 \int_0^\infty ds \sum_{\alpha,\mathfrak{b}} B^{\ell}_{\mu\alpha,\mathfrak{ab}}(t,s)  \phi(h^\ell_{\alpha \mathfrak{b}}(s))\nonumber
    \\
    &+ \gamma_0 \int_0^t ds \sum_{\alpha \mathfrak{b,c}} \Delta_\alpha(s) G^{\ell+1}_{\mu\alpha,\mathfrak{a-b},\mathfrak{c-b}} \phi(h^\ell_{\alpha \mathfrak{c}}(s)) \nonumber
    \\
    &\Phi^{\ell}_{\mu\alpha,\mathfrak{ab}}(t,s) = \left< \phi(h^\ell_{\mu \mathfrak{a}}(t))  \phi(h^\ell_{\alpha \mathfrak{b}}(s)) \right> \ , \  G^{\ell}_{\mu\alpha,\mathfrak{ab}}(t,s) =\left< g^\ell_{\mu \mathfrak{a}}(t)  g^\ell_{\alpha \mathfrak{b}}(s) \right> \nonumber
    \\
    &A^\ell_{\mu\alpha,\mathfrak{ab}}(t,s) = \frac{1}{\gamma_0} \left< \frac{\delta \phi(h^\ell_{\mu\mathfrak{a}}(t))}{\delta r^\ell_{\alpha \mathfrak{b}}(s)} \right> \ , \ B^\ell_{\mu\alpha,\mathfrak{ab}}(t,s) = \frac{1}{\gamma_0}\left< \frac{\delta g^{\ell+1}_{\mu\mathfrak{a}}(t)}{\delta u^{\ell+1}_{\alpha \mathfrak{b}}(s)} \right> 
\end{align}
We see that this field theory essentially multiples the number of sample indices by the number of spatial indices $P \to P |\mathcal S|$. Thus the time complexity of evaluation of this theory scales very poorly as $\mathcal{O}(P^3 |\mathcal S|^3 T^3)$, rendering DMFT solutions very computationally intensive.

\section{Trainable Bias Parameter}\label{app:train_bias}

If we include a bias $\b^\ell(t) \in \mathbb{R}^{N}$ in our trainable model, so that
\begin{align}
    \h^{\ell+1}_\mu(t) = \frac{1}{\sqrt N} \W^{\ell}(t) \phi(\h^\ell_\mu(t)) + \b^\ell(t) 
\end{align}
then the dynamics on $\b^\ell(t)$ induced by gradient flow is
\begin{align}
    \frac{d}{dt} \b^\ell(t) &= \gamma^2 \sum_{\alpha} \Delta_\alpha(t) \frac{\partial f_\alpha}{\partial b^\ell} = \frac{\gamma}{\sqrt N} \sum_\alpha \Delta_\alpha(t) \g^{\ell+1}_\alpha(t) = \gamma_0 \sum_\alpha \Delta_\alpha(t) \g_\alpha^\ell(t) 
\end{align}
Assuming that $b_i^\ell(0) \sim \mathcal{N}(0,1)$, the dynamics of the DMFT becomes
\begin{align}
    &\{ u^\ell \} \sim \mathcal{GP}(0,\bm\Phi^{\ell-1} + \bm 1 \bm 1^\top ) \ , \ \{ r^\ell \} \sim \mathcal{GP}(0, \G^{\ell+1}) \nonumber
    \\
    h^\ell_\mu(t) &= u^\ell_\mu(t) + \gamma_0 \int_0^\infty ds \sum_{\alpha} [ A^{\ell-1}_{\mu\alpha}(t,s) + \Theta(t-s) \Delta_\alpha(s) \Phi^{\ell-1}_{\mu\alpha}(t,s) ] g_\alpha^\ell(s) + \gamma_0 \int_0^t  ds \sum_{\alpha} \Delta_\alpha(s) g^\ell_\alpha(s) \nonumber
    \\
    z^\ell_\mu(t) &= r^\ell_\mu(t) + \gamma_0 \int_0^\infty ds \sum_\alpha [ B^\ell_{\mu\alpha}(t,s) + \Theta(t-s) \Delta_\alpha(s) G^{\ell+1}_{\mu\alpha}(t,s)] \phi(h^\ell_\alpha(s))
\end{align}

\section{Multiple Output Channels}\label{app:multiple_outputs}
We now consider network outputs on $C = \mathcal{O}_N(1)$ classes. The prediction for a data point $\mu \in [P] $ at time $t \in \mathbb{R}_+$ is $\f_\mu(t) \in \mathbb{R}^C$. 
As before, we define the error signal as $\bm \Delta_{\mu} = - \frac{\partial }{\partial \f_{\mu} } \ell(\f_\mu,\y_\mu) \in \mathbb{R}^C$. For any pair of data points $\mu,\alpha$ the NTK is a $C \times C$ matrix $\K^{NTK}_{\mu\alpha} \in \mathbb{R}^{C \times C}$ with entries $K_{\mu\alpha,cc'}^{NTK} = \frac{\partial f_c(\x_\mu)}{\partial \bm\theta} \cdot \frac{\partial f_{c'}(\x_\alpha)}{\partial \bm\theta}$. From these matrices, we can compute the evolution of the predictions in the network.
\begin{align}
    \frac{d}{dt} \f_\mu = \sum_{\alpha=1}^P \K^{NTK}_{\mu\alpha} \bm\Delta_\alpha 
\end{align}
In this case, we have matrices for the backprop features $\g^{\ell} = \gamma \sqrt{N} \frac{\partial \f^\top}{\partial \h^{\ell}} \in \mathbb{R}^{N \times C}$. These satisfy the usual recursion 
\begin{align}
    \g^\ell = \gamma \sqrt{N} \frac{\partial \f^\top}{\partial \h^{\ell}} =  \gamma \sqrt{N} \left( \frac{\partial \h^{\ell+1}}{\partial \h^\ell} \right)^\top \frac{\partial \f^\top}{\partial \h^{\ell+1}} = \left[\dot\phi(\h^{\ell}) \bm 1^\top \right] \odot \left [ \frac{1}{\sqrt N} \W^{\ell \top} \g^{\ell+1} \right]
\end{align}
We can now compute the NTK for samples $\mu,\alpha$
\begin{align}
    \K^{NTK}_{\mu\alpha} &= \sum_{\ell} \frac{\partial \f(\x_\mu)}{\partial \W^\ell} \cdot \frac{\partial \f(\x_\alpha)}{\partial \W^\ell} \nonumber
    \\
    &=  \Phi^L_{\mu\alpha} \ \I + \sum_{\ell=1}^{L-1} \G^{\ell+1}_{\mu\alpha} \Phi^{\ell}_{\mu\alpha} + \G^1_{\mu\alpha} K^x_{\mu\alpha}
\end{align}
where $\G_{\mu\alpha}^{\ell} = \frac{1}{N} \g_\mu^{\ell \top} \g_\alpha^\ell \in \mathbb{R}^{C \times C}$ and $\Phi_{\mu\alpha}^\ell = \frac{1}{N} \phi(\h^\ell_\mu) \cdot \phi(\h^\ell_{\alpha}) \in \mathbb{R}$. Next we introduce kernels $\A^{\ell}_{\mu\alpha}(t,s) \in\mathbb{R}^C$ and $\B^{\ell}_{\mu\alpha}(t,s) \in \mathbb{R}^C$ which are defined in the usual way. The corresponding field theory has the form
\begin{align}
    h^\ell_\mu(t) &= \chi^\ell_\mu(t) + \gamma_0 \int_0^{\infty} ds \sum_{\alpha=1}^P \left[ \A_{\mu\alpha}^{\ell-1}(t,s) + \Theta(t-s) \bm\Delta_\alpha(s) \Phi^{\ell-1}_{\mu\alpha}(t,s)  \right] \cdot \g^{\ell}_{\alpha}(s) \in \mathbb{R} \nonumber
    \\
    \z^\ell_\mu(t) &= \bm\xi^\ell_\mu(t) + \gamma_0 \int_0^\infty ds \sum_{\alpha =1}^P \left[ \B^{\ell}_{\mu\alpha}(t,s) + \Theta(t-s) \G^{\ell+1}_{\mu\alpha} \bm\Delta_\alpha(s)  \right] \phi(h^\ell_\mu(t)) \in \mathbb{R}^C \nonumber
    \\
    \g^{\ell}_\mu(t) &= \dot\phi(h^\ell_\mu(t)) \z^\ell_{\mu}(t) \in \mathbb{R}^C
\end{align}
From these fields, the saddle point equations define the kernels as 
\begin{align}
    \Phi^\ell_{\mu\alpha}(t,s) &= \left< \phi(h^\ell_\mu(t)) \phi(h^\ell_\alpha(s)) \right> \in \mathbb{R} \ , \ \G^{\ell}_{\mu\alpha}(t,s) = \left< \g^{\ell}_\mu(t) \g^{\ell}_\alpha(s)^\top \right> \in \mathbb{R}^{C \times C} \nonumber
    \\
    \A^{\ell}_{\mu\alpha}(t,s) &= \frac{1}{\gamma_0} \left< \frac{\delta \phi(h^\ell_\mu(t))}{\delta \r^\ell_\alpha(s)} \right> \in \mathbb{R}^C \ , \ \B^{\ell}_{\mu\alpha}(t,s) = \frac{1}{\gamma_0} \left< \frac{\delta \g^\ell_\mu(t)}{\delta u^\ell_\alpha(s)} \right> \in \mathbb{R}^C .
\end{align}
This allows studying the multi-class structure of learned representations.



\section{Weight Decay in Deep Homogenous Networks}\label{app:weight_decay}

If we train with weight decay, $\frac{d}{dt}\bm\theta = - \gamma^2 \nabla_{\bm\theta} \mathcal{L} - \lambda \bm\theta$, in a $\kappa$-degree homogenous network ($f(c\bm\theta) = c^\kappa f(\bm\theta)$), then the prediction dynamics satisfy
\begin{align}
    \frac{d}{dt} f(\x, t) = \sum_{\alpha} \Delta_\alpha(t) K^{NTK}_{\mu\alpha}(\x,\x_\alpha, t) - \lambda \kappa f(\x,t) \nonumber \ , \ 
\end{align}
This holds by the following identity $\frac{\partial}{\partial c} f(c \bm\theta)= \frac{\partial }{\partial c} c^\kappa f(\bm\theta)$, which when evaluated at $c = 1$ gives $\frac{\partial}{\partial \bm\theta} f(\bm\theta) \cdot \bm\theta = \kappa f(\bm\theta)$. This identity was utilized in a prior work which studied L2 regularization in the lazy regime \cite{lewkowycz2020training}. For a $L$-hidden layer ReLU network $\phi(h) = \max(0,h)$, the degree is $\kappa = L+1$, while rectified power law nonlinearities $\phi(h) = \max(0,h)^q$ give degrees $\kappa = \frac{q^{L+1}-1}{q-1}$. We note that the fixed point of the function dynamics above gives a representer theorem with the final NTK
\begin{align}
    f(\x) = \bm k(\x)^\top \left[ \bm K + \lambda \kappa \bm I \right]^{-1} \y
\end{align}
where $[\bm k(x)]_{\mu} = \lim_{t\to\infty }K(\x,\x_\mu, t)$ and $K_{\mu\alpha} = \lim_{t \to \infty} K(\x_\mu,\x_\alpha,t)$. The prior work of Lewkowycz et al \cite{lewkowycz2020training} considered NTK parameterization $\gamma_0 = 0$. In this limit, the kernel (and consequently output function) decay to zero at large time, but if $\gamma_0 > 0$, then the network converges to a nontrivial fixed point as $t \to \infty$. In the DMFT limit we can determine the final kernel by solving the following field dynamics
\begin{align}
    h_\mu^{\ell}(t) &= e^{-\lambda t} \chi_\mu^\ell(t) + \gamma_0 \int_0^t ds \ e^{-\lambda(t-s)} \sum_{\alpha=1}^P \Delta_{\alpha}(s) g_\alpha^\ell(s) \Phi^{\ell-1}_{\mu\alpha}(t,s)  \nonumber
    \\
    z_\mu^{\ell}(t) &= e^{-\lambda t} \xi_\mu^\ell(t) + \gamma_0 \int_0^t ds \ e^{-\lambda(t-s)} \sum_{\alpha=1}^P \Delta_{\alpha}(s) \phi(h_\alpha^\ell(s)) G^{\ell+1}_{\mu\alpha}(t,s).
\end{align}
We see that the contribution from initial conditions is exponentially suppressed at large time $t$ while the second term contributes most when the system has equilibrated. We provide an example of the weight decay DMFT showing its validity in a two layer ReLU network in Figure \ref{fig:weight_decay}.

\section{Bayesian/Langevin Trained Mean Field Networks}\label{app:bayes_langevin}

Rather than studying exact gradient flow, many works have considered Langevin dynamics (gradient flow with white noise process on the weights) of neural network training \cite{zavatone2021asymptotics, li2021statistical, aitchison_feature_learn_bayes, naveh2021self, seroussi2021separation}. This setting is of special theoretical interest since the distribution of parameters converges at long times to a Gibbs equilibrium distribution which has a Bayesian interpretation \cite{neal2012bayesian, lee2018deep, aitchison_feature_learn_bayes}. The relevant Langevin equation for our mean field gradient flow is
\begin{align}
    d\bm\theta(t) = -\gamma^2 \nabla L(\bm\theta(t)) dt  - \lambda \beta^{-1} \bm\theta(t) dt + \sqrt{2 \beta^{-1}} d\bm\epsilon(t) ,
\end{align}
where $\lambda$ is a ridge penalty which controls the scale of parameters, and $d\bm\epsilon(t)$ is a Brownian motion term which has covariance structure $\left<  d\bm\epsilon(t) d\bm\epsilon(t')^\top \right> = \delta(t-t') \I$. The parameter $\beta$, known as the inverse temperature controls the scale of the random Gaussian noise injected into this stochastic process. The dynamical treatment of the $\beta \to \infty$ limit will coincide with our usual DMFT while the $\beta \ll \infty$ will exhibit a nontrivial balance between the usual DMFT feature updates and the random Langevin noise. At late times, such a system will equilibrate to its Gibbs distribution.

\subsection{Dynamical Analysis}
In this section we analyze the dynamical mean field theory for these Langevin dynamics. First we note that the effect of regularization can be handled with a simple integrating factor 
\begin{align}
    d[ \W^\ell(t) e^{\frac{\lambda t}{\beta}} ] &= e^{\frac{\lambda}{\beta} t}\left[ \frac{\gamma_0}{\sqrt N} \sum_\mu \Delta_\mu(t) \g^{\ell+1}_\mu(t) \phi(\h^\ell_\mu(t))^\top  \right] dt + \sqrt{2\beta^{-1}} e^{\frac{\lambda t}{\beta}} d\bm\epsilon^\ell(t) .
\end{align}
where $d\bm\epsilon(t) \in \mathbb{R}^{N \times N}$ is the Gaussian noise for layer $\ell$ at time $t$. It is straightforward to verify by Ito's lemma that, under mean field parameterization, the fluctuations in $f's$ dynamics due to Brownian motion are $\frac{\partial f}{\partial \bm\theta} \cdot d\bm\epsilon(t) \sim \mathcal{O}(N^{-1/2})$ and are thus negligible in the $N \to \infty$ limit. Thus the evolution of the network function takes the form
\begin{align}
    \frac{\partial f_\mu(t) }{\partial t} &= \sum_\alpha \Delta_\alpha(t) K_{\mu\alpha}(t,t) - \lambda \beta^{-1} \bm\theta(t) \cdot \nabla_{\bm\theta} f_\mu(t) + \frac{1}{\beta} \text{Tr} \nabla^2_{\bm\theta} f_\mu(t) \nonumber
\end{align}
We can express both of these parameter contractions in feature space provided we introduce the new features $r_{i,\mu}^{\ell}(t) = \frac{\partial g_{i,\mu}^{\ell}}{\partial h_{i,\mu}^{\ell}}$ which are necessary to compute Hessian terms like $\frac{\partial^2 f}{\partial W_{ij}^\ell \partial W^\ell_{ij}} = N^{-3/2} \frac{\partial }{\partial W^\ell_{ij}} [ g_i^{\ell+1} \phi(h_j^\ell)]= N^{-2} \ r_i \ \phi(h_j^\ell)^2$ in each layer. This gives the following evolution 
\begin{align}
    \frac{\partial f_\mu(t)}{\partial t} = \sum_\alpha \Delta_\alpha(t) K_{\mu\alpha}(t,t) - \lambda \beta^{-1} \sum_{\ell} \left< z_\mu^{\ell}(t) \phi(h_\mu^\ell(t)) \right> + \beta^{-1} \sum_\ell \left< r^{\ell+1}_{\mu}(t) \right> \left< \phi(h^\ell_\mu(t))^2  \right>
\end{align}

As before, we compute the next layer field $\h^{\ell+1}$ in terms of $\bm\chi^{\ell+1}$ and $\z^{\ell}$ in terms of $\bm \xi^{\ell}$ 
\begin{align}
    \h^{\ell+1}_\mu(t) &= e^{-\frac{\lambda}{\beta} t} \bm\chi_\mu^{\ell+1}(t) + \int_0^t  \ e^{- \frac{\lambda}{\beta}(t-s) } \left[ ds \frac{\gamma_0}{N}  \sum_\alpha \Delta_\alpha(s) \g^{\ell+1}_\alpha(s) \phi(\h^\ell_\alpha(s))^\top + \sqrt{\frac{2}{\beta N} } d\bm\epsilon^\ell(s)  \right] \phi(\h^\ell_\mu(t)) \nonumber 
    \\
    \z^{\ell+1}_\mu(t) &= e^{-\frac{\lambda}{\beta} t} \bm\xi_\mu^{\ell+1}(t) + \int_0^t \ e^{- \frac{\lambda}{\beta}(t-s) } \left[ ds  \frac{\gamma_0}{N}  \sum_\alpha \Delta_\alpha(s) \g^{\ell+1}_\alpha(s) \phi(\h^\ell_\alpha(s))^\top + \sqrt{\frac{2}{\beta N} } d\bm\epsilon^\ell(s)  \right]^\top \g^{\ell+1}_\mu(t) \nonumber 
\end{align}
The dependence on the initial condition through $\bm\chi,\bm\xi$ is suppressed at long times due the regularization factor $e^{-\frac{\lambda}{\beta} t}$, while the Brownian motion and gradient updates will survive in the $t\to\infty$ limit. In addition to the usual $\{\bm\chi^\ell,\bm\xi^\ell\}$ fields which arise from the initial condition, we see that $\h^\ell(t), \z^\ell(t)$ also depend on the following fields which arise from the integrated Brownian motion
\begin{align}
    \bm\chi^{\epsilon,\ell}_{\mu}(t) = \sqrt{\frac{2}{\beta N}} \int_0^\infty ds \ e^{-\frac{\lambda}{\beta}(t-s)} \Theta(t-s) d\bm\epsilon^{\ell}(s) \phi(\h^\ell_\mu(t)) \nonumber
    \\
    \bm\xi^{\epsilon,\ell}_\mu(t) = \sqrt{\frac{2}{\beta N}} \int_0^\infty ds \ e^{-\frac{\lambda}{\beta}(t-s)} \Theta(t-s) d\bm\epsilon^{\ell}(s)^\top \g^{\ell+1}_\mu(t)
\end{align}
Our aim is now to compute the moment generating function for the $\{\bm\chi,\bm\xi,\bm\chi^{\epsilon},\bm\xi^{\epsilon}\}$ fields which causally determine $\{\h,\z\}$. This MGF has the form
\begin{align}
    Z = \left< \exp\left( \sum_{\ell\mu} \int_0^\infty \left[ \j_\mu^\ell(t) \cdot \bm\chi_\mu^\ell(t) + \v^\ell_\mu(t) \cdot \bm\xi^\ell_\mu(t) + \j_\mu^{\epsilon,\ell}(t) \cdot \bm\chi_\mu^{\epsilon,\ell}(t) + \v^{\epsilon,\ell}_\mu(t) \cdot \bm\xi^{\epsilon \ell}_\mu(t) \right] \right) \right>_{\bm\theta_0, \bm\epsilon(t)}
\end{align}
We insert Dirac-delta functions in the usual way to enforce the definitions of $\bm\chi,\bm\xi,\bm\chi^{\epsilon},\bm\xi^{\epsilon}$ and then average over $\bm\theta_0 , \bm\epsilon(t)$. These averages can be performed separately with the $\bm\theta_0$ average giving the identical terms as derived in previous sections. We focus on the average over Brownian disorder
\begin{align}
&\ln \left<\exp\left( i \sqrt{2} \beta^{-1/2} N^{-1/2} \sum_{\mu} \int_0^\infty dt \ \text{Tr}\left[ \hat{\bm\chi}^{\epsilon,\ell+1}_\mu(t) \phi(\h^\ell_\mu(t))^\top + \g^{\ell+1}_{\mu}(t) \hat{\bm\xi}^{\epsilon,\ell}_\mu(t)^\top \right] \int e^{-\frac{\lambda}{\beta}(t-s)} \Theta(t-s) d\bm\epsilon(s) \right) \right>_{\bm\epsilon(t)} \nonumber
\\
&=  - \frac{1}{\beta N} \int_0^\infty ds \left| \int dt \ \Theta(t-s) e^{-\frac{\lambda}{\beta}(t-s)} \sum_{\mu} \left[\hat{\bm\chi}^{\epsilon,\ell+1}_\mu(t) \phi(\h^\ell_\mu(t))^\top + \g^{\ell+1}_{\mu}(t) \hat{\bm\xi}^{\epsilon,\ell}_\mu(t)^\top \right] \nonumber  \right|^2  
\\
&=- \frac{1}{\beta} \int_0^\infty ds \int_0^\infty dt \int_0^\infty dt' \Theta(t-s) \Theta(t'-s)e^{-\frac{\lambda}{\beta}(t-s+t'-s)} \nonumber
\\
&\times \sum_{\mu\alpha} \left[ \bm{\hat{\chi}}^{\epsilon,\ell+1}_\mu(t) \cdot \bm{\hat{\chi}}^{\epsilon,\ell+1}_\alpha(t') \Phi^\ell_{\mu\alpha}(t,t') +   \bm{\hat{\xi}}^{\epsilon,\ell}_\mu(t) \cdot \bm{\hat{\xi}}^{\epsilon,\ell}_\alpha(t') G^{\ell+1}_{\mu\alpha}(t,t') + 2 i \bm{\hat\chi}^{\epsilon,\ell+1}_{\mu}(t) \cdot \g^{\ell+1}_\alpha(t') A^{\epsilon,\ell}_{\mu\alpha}(t,t')  \right]\nonumber
\\
&= - \frac{1}{2\lambda} \int_0^\infty dt \int_0^\infty dt' \exp\left( - \frac{\lambda}{\beta}(t+t') \right) \left[ e^{2 \frac{\lambda}{\beta} \min\{t,t'\}} - 1 \right] \nonumber
\\
&\times \sum_{\mu\alpha} \left[ \bm{\hat{\chi}}^{\epsilon,\ell+1}_\mu(t) \cdot \bm{\hat{\chi}}^{\epsilon,\ell+1}_\alpha(t') \Phi^\ell_{\mu\alpha}(t,t') +   \bm{\hat{\xi}}^{\epsilon,\ell}_\mu(t) \cdot \bm{\hat{\xi}}^{\epsilon,\ell}_\alpha(t') G^{\ell+1}_{\mu\alpha}(t,t') + 2 i \bm{\hat\chi}^{\epsilon,\ell+1}_{\mu}(t) \cdot \g^{\ell+1}_\alpha(t') A^{\epsilon,\ell}_{\mu\alpha}(t,t')  \right]
\end{align}
where we introduced the order parameter $i A^{\epsilon,\ell}_{\mu\alpha}(t,t') = \frac{1}{N} \phi(\h^{\ell}_\mu(t)) \cdot \bm{\hat\xi}^{\epsilon,\ell}_\alpha(s)$. We will use the shorthand for the temporal prefactor in the above 
$C_{\lambda,\beta}(t,t') = \frac{1}{\lambda} \exp\left( - \frac{\lambda}{\beta}(t+t') \right) \left[ e^{2 \frac{\lambda}{\beta} \min\{t,t'\}} - 1 \right] \sim_{t,t' \to \infty} \frac{1}{\lambda} \exp\left( - \frac{\lambda}{\beta} |t-t'| \right)$. We insert a Lagrange multiplier $B^{\epsilon,\ell}$ to enforce the definition of $A^{\epsilon,\ell}$. After 
\begin{align}
    Z \propto& \int d\Phi_{\mu\alpha}^\ell(t,s) d\hat{\Phi}_{\mu\alpha}^\ell(t,s)  dG_{\mu\alpha}^\ell(t,s) d\hat{G}_{\mu\alpha}^\ell(t,s) dA_{\mu\alpha}^\ell(t,s) dB_{\mu\alpha}^\ell(t,s) dA_{\mu\alpha}^{\epsilon \ell}(t,s)dB_{\mu\alpha}^{\epsilon \ell}(t,s) \nonumber \\
    &\times \exp\left( N S[\Phi,\hat\Phi,G,\hat{G},A,B,A^{\epsilon},B^{\epsilon}] \right)
\end{align}
The order parameters can be determined by the saddle point equations. These equations for $\Phi,\hat\Phi,G,\hat{G},A,B$ are the same as before. The new equations are
\begin{align}
    \frac{\delta S}{\delta A^{\epsilon,\ell}_{\mu\alpha}(t,s)} &= -B^{\epsilon,\ell}_{\mu\alpha}(t,s) - i C_{\lambda,\beta}(t,s) \left< {\hat\chi}_\mu^{\epsilon,\ell+1}(t) g^{\ell+1}_\alpha(s)  \right> = 0 \nonumber
    \\
    \frac{\delta S}{\delta B^{\epsilon,\ell}_{\mu\alpha}(t,s)} &= -A^{\epsilon,\ell}_{\mu\alpha}(t,s) - i C_{\lambda,\beta}(t,s) \left< \phi(h^{\ell}_\mu(t)) \hat{\xi}^{\epsilon,\ell}_\alpha(s)  \right> = 0 
\end{align}
Using the fact that $\bm\Phi^\ell,G^\ell$ concentrate, we can use the Hubbard trick to linearize the quadratic terms in $\hat{\chi}^{\epsilon}$ and $\hat{\xi}^{\epsilon}$.
\begin{align}
    &\exp\left( - \frac{1}{2} \int_0^\infty dt \int_0^\infty ds \ C_{\lambda,\beta}(t,s) \sum_{\mu\alpha} \hat{\chi}_\mu^{\epsilon,\ell+1}(t) \hat{\chi}_\alpha^{\epsilon,\ell+1}(s) \Phi^{\ell}_{\mu\alpha}(t,s) \right) \nonumber
    \\
    &= \left<  \exp\left( - i \sum_\mu \int_0^\infty dt \ u^{\epsilon,\ell+1}_\mu(t) \hat{\chi}^{\epsilon,\ell+1}_\mu(t)  \right) \right>_{u^{\epsilon,\ell+1}_\mu(t) \sim \mathcal{GP}(0,C \odot \Phi^\ell) } 
    \\
    &\exp\left( - \frac{1}{2} \int_0^\infty dt \int_0^\infty ds \ C_{\lambda,\beta}(t,s) \sum_{\mu\alpha} \hat{\xi}_\mu^{\epsilon,\ell}(t) \hat{\xi}_\alpha^{\epsilon,\ell}(s) G^{\ell+1}_{\mu\alpha}(t,s) \right) \nonumber
    \\
    &= \left<  \exp\left( - i \sum_\mu \int_0^\infty dt \ r^{\epsilon,\ell}_\mu(t) \hat{\xi}^{\epsilon,\ell}_\mu(t)  \right) \right>_{r^{\epsilon,\ell}_\mu(t) \sim \mathcal{GP}(0,C \odot G^{\ell+1}) } 
\end{align}
Using the vectorization notation, we find the interpretation that $\bm\chi^{\epsilon,\ell} $ and $\bm\xi^{\epsilon,\ell}$ decouple as
\begin{align}
    \bm\chi^{\epsilon,\ell+1} &= \u^{\epsilon,\ell+1} + \A^{\epsilon,\ell+1} \g^{\ell+1} \ , \ \bm\xi^{\epsilon,\ell} = \r^{\epsilon,\ell} + \B^{\epsilon, \ell \top} \phi(\h^\ell)
    \\
    \A^{\epsilon,\ell+1} &= \C_{\lambda,\beta} \odot \left<  \frac{\partial \phi(\h^{\ell})}{\partial \r^{\epsilon,\ell}} \right> \ , \ \B^{\epsilon \ell} = \C_{\lambda,\beta} \odot \left< \frac{\partial \g^{\ell+1}}{\partial \u^{\ell+1}} \right>^\top
\end{align}
As before, we make the substitutions $\B \to \gamma_0^{-1} {\B}^\top$ and $\A \to \gamma_0^{-1} \A$ and arrive at the final DMFT equations
\begin{align}
    &\{ u^{\ell}_{\mu}(t) \} \sim \mathcal{GP}(0,\Phi^{\ell-1}) \ , \ \{ r^\ell_\mu(t) \} \sim \mathcal{GP}(0,G^{\ell+1}) \nonumber
    \\
    &\{ u^{\epsilon,\ell}_{\mu}(t) \} \sim \mathcal{GP}(0, C_{\lambda,\beta} \odot \Phi^{\ell-1}) \ , \ \{ r^{\epsilon,\ell}_\mu(t) \} \sim \mathcal{GP}(0, C_{\lambda,\beta} \odot G^{\ell+1}) \nonumber
    \\
    h^{\ell}_\mu(t) &= e^{-\frac{\lambda}{\beta} t} \left[u^{\ell}_\mu(t) + \gamma_0 \int_0^t ds \sum_{\alpha}  A^{\ell-1}_{\mu\alpha}(t,s) g^{\ell}_{\alpha}(s) \right] \nonumber
    \\
    &+ u^{\epsilon,\ell}_\mu(t) + \gamma_0 \int_0^t ds  \   \sum_{\alpha} \left[ A^{\epsilon,\ell-1}_{\mu\alpha}(t,s) + e^{-\frac{\lambda}{\beta} (t-s)} \Delta_\alpha(s) \Phi^{\ell-1}_{\mu\alpha}(t,s) \right] g^{\ell}_\alpha(s) \nonumber
    \\
    z^{\ell}_\mu(t) &= e^{-\frac{\lambda}{\beta} t} \left[r^{\ell}_\mu(t) + \gamma_0 \int_0^t ds \sum_{\alpha}  B^{\ell}_{\mu\alpha}(t,s) g^{\ell}_{\alpha}(s) \right]  \nonumber
    \\
    &+ r^{\epsilon,\ell}_\mu(t) + \gamma_0 \int_0^t ds  \ \sum_{\alpha} \left[ B^{\epsilon,\ell}_{\mu\alpha}(t,s) + e^{-\frac{\lambda}{\beta} (t-s)}   \Delta_\alpha(s) G^{\ell+1}_{\mu\alpha}(t,s) \right] \phi(h^{\ell}_\alpha(s)) \nonumber
    \\
\end{align}
where the kernels are defined in the usual way. As expected, the contributions from the initial conditions $\chi^\ell, \xi^\ell$ are exponentially suppressed at late time whereas the contributions from the Brownian disorder $\chi^{\epsilon,\ell},\xi^{\epsilon,\ell}$ persist at late time. 

\subsection{Weak Feature Learning, Long Time Limit}
In the weak feature learning $\gamma_0 \to 0$ and long time $t \to \infty$ limit, the preactivation fields equilibrate to Gaussian processes $h^{\ell}_\mu(t) \sim u^{\epsilon,\ell}_\mu(t), z^\ell_\mu(t) \sim r^{\epsilon,\ell}_\alpha(t)$, which have respective covariances $H^\ell_{\mu\alpha}(t,s) = \left< h^{\ell}_\mu(t) h^{\ell}_\alpha(s) \right> = C_{\lambda,\beta}(t,s) \Phi^{\ell-1}_{\mu\alpha}(t,s),  Z^{\ell}_{\mu\alpha}(t,s) = \left< z^\ell_\mu(t) z^\ell_\alpha(s) \right> = C_{\lambda,\beta}(t,s) G^{\ell+1}_{\mu\alpha}(t,s)$. In this long time limit, the feature kernels will be time translation invariant eg $\Phi^{\ell}_{\mu\alpha}(t,s)= \Phi^{\ell}_{\mu\alpha}(|t-s|)$. Letting $\tau = |t-s|$ and $C_{\lambda,\beta}(\tau ) = \frac{1}{\lambda} \exp\left(-\frac{\lambda}{\beta} \tau \right)$, we have the following recurrence for $H^{\ell},\Phi^{\ell}$
\begin{align}
    H^1_{\mu\alpha}(\tau) &= C_{\lambda,\beta}(\tau) K^x_{\mu\alpha} \ , \ \Phi^{1}_{\mu\alpha}(\tau) = \left< \phi(h) \phi(h') \right>_{h,h' \sim \mathcal{N}\left(0, \H^1
    \right)}  \ , \ \H^1 = \begin{bmatrix} H^1_{\mu\mu}(0) & H^1_{\mu\alpha}(\tau) \\
    H^1_{\mu\alpha}(\tau) & H^1_{\alpha\alpha}(0) \end{bmatrix} \nonumber
    \\
    H^{\ell+1}_{\mu\alpha}(\tau) &=C_{\lambda,\beta}(\tau) \Phi^\ell_{\mu\alpha}(\tau) \ , \ \Phi^{\ell+1}_{\mu\alpha}(t,s) = \left< \phi(h) \phi(h') \right>_{h,h' \sim \mathcal{N}\left(0, \H^{\ell+1}
    \right)}  \nonumber
    \\ \H^{\ell+1} &= \begin{bmatrix} H^{\ell+1}_{\mu\mu}(0) & H^{\ell+1}_{\mu\alpha}(\tau) \\
    H^{\ell+1}_{\mu\alpha}(\tau) & H^{\ell+1}_{\alpha\alpha}(0) \end{bmatrix}
\end{align}
Similarly, we can obtain $Z^{\ell}$ and $G^{\ell}$ in a backward pass recursion
\begin{align}
    Z^L_{\mu\alpha}(\tau) &= C_{\lambda,\beta}(\tau) \ , \ G^{L}_{\mu\alpha}(\tau) = \dot\Phi^L_{\mu\alpha}(\tau) Z^L_{\mu\alpha}(\tau) \ , \ \dot\Phi^{L}_{\mu\alpha}(\tau) = \left< \dot\phi(h) \dot\phi(h') \right>_{h,h'\sim \mathcal{N}(0,\H^L)} \nonumber
    \\
    Z^{\ell}_{\mu\alpha}(\tau) &= C_{\lambda,\beta}(\tau) G^{\ell+1}_{\mu\alpha}(\tau) \ , \ G^{\ell}_{\mu\alpha}(\tau) = \dot\Phi^\ell_{\mu\alpha}(\tau) Z^{\ell}_{\mu\alpha}(\tau) \ , \ \dot\Phi^{\ell}_{\mu\alpha}(\tau) = \left< \dot\phi(h) \dot\phi(h') \right>_{h,h'\sim \mathcal{N}(0,\H^\ell)}
\end{align}
On the temporal diagonal $\tau = 0$, these equations give the usual recursions used to compute the NNGP kernels at initialization \cite{lee2018deep}, though with initialization variance $C_{\lambda,\beta}(0) = \lambda^{-1}$, set by the weight decay term in the Langevin dynamics. This indicates that the long time Langevin dynamics at $\gamma_0 \to 0$ simply rescales the Gaussian weight variance based on $\lambda$. It would be interesting to explore fluctuation dissipation relationships at finite $\gamma_0$ within this framework which we leave to future work.
%

\subsection{Equilibrium Analysis}
The Langevin dynamics at finite $N$ converges (possibly in a time extensive in $N$) to an equilibrium distribution with several interesting properties, as was recently studied by Aitchison et al \cite{aitchison_feature_learn_bayes} and implicitly by Seroussi et al \cite{seroussi2021separation} in a large sample size limit. This setting differs from the previous section where first $N \to \infty$ limit is taken, followed by a $t \to \infty$ limit in the DMFT. This, section, on the other hand studies for any $N$, the $t \to \infty$ limiting equilibrium distribution. This equilibrated distribution is then analyzed in the $N \to \infty$ limit. The relationship between these two orders of limits remains an open problem. 
The equilibrium distribution over parameters $p(\bm\theta|\mathcal D) \propto \exp\left( - \beta \gamma^2 L(\bm\theta) - \frac{\lambda}{2} |\bm\theta|^2 \right)$ can be viewed as a Bayes posterior with log-likelihood $- \beta \gamma^2 L(\bm\theta)$ and a Gaussian prior with scale $\lambda^{-1/2}$. In the mean field limit with $\gamma = \sqrt{N} \gamma_0$, we can express the density over pre-activations $\h^\ell$ and the output predictions $f$. This gives
\begin{align}
    p(\f|\mathcal D) &\propto \exp\left( - N \gamma_0^2 \beta \sum_\mu \ell(f_\mu,y_\mu)  \right) \nonumber
    \\
    &\times \int d\h^\ell_\mu \left< \prod_{\mu} \delta\left( f_\mu - \frac{1}{N \gamma_0} \w^L \cdot \phi(\h^L_\mu)  \right) \prod_{\mu\ell} \delta\left( \h^{\ell+1}_\mu - \frac{1}{\sqrt N} \W^\ell \phi(\h^\ell_\mu) \right) \right>_{\bm\theta \sim \mathcal{N}(0,\lambda^{-1} \I)} \nonumber
    \\
    \propto &\int \prod_\mu d\hat{f}_\mu \prod_{\ell \mu\alpha} d\Phi^\ell_{\mu\alpha} d\hat{\Phi}^\ell_{\mu\alpha} \exp\left( - N \gamma_0^2 \beta \sum_\mu \ell(f_\mu,y_\mu) -  N \gamma_0 \sum_\mu \hat{f}_\mu f_\mu + \frac{N}{2\lambda} \sum_{\mu\alpha} \hat{f}_\mu \Phi^L_{\mu\alpha} \hat{f}_\alpha  \right) \nonumber
    \\
    &\exp\left( \frac{N}{2} \sum_{\ell \mu\alpha} \Phi^{\ell}_{\mu\alpha} \hat{\Phi}^\ell_{\mu\alpha} + N \sum_{\ell} \ln \mathcal Z[\Phi^{\ell-1},\hat{\Phi}^\ell]  \right) \nonumber
    \\
    &\mathcal Z[\Phi^{\ell-1},\hat{\Phi}^\ell] = \int \prod_{\mu } \frac{dh_\mu d\hat{h}_\mu}{2\pi} \exp\left( - \frac{1}{2\lambda} \sum_{\mu\alpha} \hat{h}_\mu \Phi^{\ell-1}_{\mu\alpha} \hat{h}_\alpha - \frac{1}{2} \sum_{\mu\alpha} \phi(h_\mu) \hat{\Phi}^\ell_{\mu\alpha} \phi(h_\alpha) + i\sum_\mu \hat{h}_\mu h_\mu  \right) 
\end{align}

We see that $p(\f|\mathcal D) \propto \int d\Phi d\hat{\Phi} \exp\left( N S[\Phi,\hat{\Phi}] \right)$ where
\begin{align}
    S = - \gamma_0^2 \beta \sum_\mu \ell(f_\mu,y_\mu) - \gamma_0 \sum_\mu \hat{f}_\mu f_\mu + \frac{1}{2 \lambda} \sum_{\mu\alpha} \hat{f}_\mu \Phi^L_{\mu\alpha} \hat{f}_\alpha + \frac{1}{2} \sum_{\ell \mu\alpha} \Phi^{\ell}_{\mu\alpha} \hat{\Phi}^\ell_{\mu\alpha} + \sum_\ell \ln\mathcal Z[\Phi^{\ell-1},\hat{\Phi}^\ell]
\end{align}
Thus the predictions $f_\mu$ become non-random in this $N \to \infty$ limit and can be determined from the saddle point equations as in \cite{aitchison_feature_learn_bayes}. Again, letting $\Delta_\mu = - \frac{\partial}{\partial f_\mu} \ell(f_\mu,y_\mu)$, we find
\begin{align}
    \frac{\partial S}{\partial f_\mu} &= \gamma_0 \hat{f}_\mu - \gamma_0^2 \beta \Delta_\mu = 0  \ , \ \frac{\partial S}{\partial \hat{f}_\mu} = - \gamma_0 f_\mu + \frac{1}{\lambda} \sum_{\alpha} \Phi^{L}_{\mu\alpha} \hat{f}_\alpha = 0 \nonumber
    \\
    \frac{\partial S}{\partial \Phi^L_{\mu\alpha}} &= \frac{1}{2\lambda} \hat{f}_\mu \hat{f}_\alpha + \frac{1}{2} \hat{\Phi}^{L}_{\mu\alpha} = 0  \ , \ \frac{\partial S}{\partial \hat{\Phi}^L_{\mu\alpha}} = \frac{1}{2} \Phi^{\ell}_{\mu\alpha} - \frac{1}{2} \left< \phi(h^L_\mu) \phi(h^L_\alpha) \right>  = 0 \nonumber 
    \\
    \frac{\partial S}{\partial \Phi^\ell_{\mu\alpha}} &= - \frac{1}{2} \left< \hat{h}^{\ell+1}_{\mu} \hat{h}_\alpha^{\ell+1} \right> + \frac{1}{2} \hat{\Phi}^\ell_{\mu\alpha} = 0 \ , \ \frac{\partial S}{\partial \hat{\Phi}^\ell_{\mu\alpha}} = \frac{1}{2} \Phi^\ell_{\mu\alpha} - \frac{1}{2} \left< \phi(h^\ell_\mu)\phi(h^\ell_\alpha) \right> = 0
\end{align}
which implies that $f_\mu$ at the fixed point satisfies the following equations
\begin{align}
    f_\mu = \frac{ \beta}{\lambda} \sum_\alpha \Phi^L_{\mu\alpha} \Delta_\alpha \ , \ \Delta_\alpha = - \frac{\partial \ell(f_\alpha,y_\alpha)}{\partial f_\alpha}.
\end{align}
The last layer's dual kernel has the form $\hat{\Phi}^L_{\mu\alpha} = - \frac{\gamma_0^2 \beta^2}{2\lambda} \Delta_\mu \Delta_\alpha$, which we see vanishes as feature learning strength is taken to zero $\gamma_0 \to 0$, while for non-negligible $\gamma_0$, we see that the last layer features are non-Gaussian. We thus see that the moment generating function for the last layer field has the form
\begin{align}
    \mathcal Z[\Phi^{L-1},\hat{\Phi}^L] = \int \prod_{\mu } \frac{dh_\mu d\hat{h}_\mu}{2\pi} \exp\left( - \frac{1}{2\lambda} \sum_{\mu\alpha} \hat{h}_\mu \hat{h}_\alpha \Phi^{L-1}_{\mu\alpha} + \frac{\gamma_0^2 \beta^2}{2\lambda} \left[\sum_{\mu} \Delta_\mu \phi(h_\mu) \right]^2 + i \sum_\mu \hat{h}_\mu h_\mu \right)
\end{align}
In the $\gamma_0 \to 0$ limit, the non-Gaussian component of this density vanishes. Now that we have this form, we can compute $\Phi^L$ conditional on $\Phi^{L-1}$. Next, we calculate $\hat{\Phi}^{L-1}_{\mu\alpha} = \left< \hat{h}^L_\mu \hat{h}^L_\alpha \right>$, giving
\begin{align}
    \bm{\hat{\Phi}}^{L-1} = \lambda [\bm\Phi^{L-1}]^{-1} - \lambda^2 [\bm\Phi^{L-1}]^{-1} \left< \h^L \h^{L\top} \right>  [\bm\Phi^{L-1}]^{-1} 
\end{align}
Again, we note that in the $\gamma_0 \to 0$ limit, since $\left< \h^L \h^L \right> \sim \lambda^{-1} \bm\Phi^{L-1}$, so that $\hat{\bm\Phi}^{L-1} = 0$, implying that the $h^{L-1}$ fields are also Gaussian in this $\gamma_0 \to 0$ limit. 
For arbitrary $\gamma_0$, this recursive argument can be completed going backwards using 
\begin{align}
    \bm\Phi^{\ell} = \left< \phi(\h^\ell)\phi(\h^\ell)^\top \right> \ , \ 
    \bm{\hat{\Phi}}^{\ell-1} = \lambda [\bm\Phi^{\ell-1}]^{-1} - \lambda^2 [\bm\Phi^{\ell-1}]^{-1} \left< \h^\ell \h^{\ell\top} \right>  [\bm\Phi^{\ell-1}]^{-1}   
\end{align}
For deep linear networks, the distributions are all Gaussian, allowing one to close algebraically, the saddle point equations for $\Phi,\hat{\Phi}$ \cite{aitchison_feature_learn_bayes}.

\section{Momentum Dynamics}\label{app:momentum}

Standard gradient descent often converges slowly and requires careful tuning of learning rate. Momentum, in contrast can, be stable under a wider range of learning rates and can benefit from acceleration on certain problems \cite{nesterov1983method,nesterov2006cubic,goh2017why, muehlebach2021optimization}. In this section we show that our field theory is still valid when training with momentum; simply altering the field definitions appropriately gives the infinite-width feature learning behavior.

Momentum uses a low-pass filtered version of the gradients to update the weights. A continuous limt of momentum dynamics on the trainable parameters $\{\W^\ell \}$ would give the following differential equations.
\begin{align}
    \frac{\partial }{\partial t} \W^\ell(t) &= \Q^\ell(t) \nonumber
    \\
    \tau \frac{d}{dt} \Q^{\ell}(t) &= - \Q^{\ell} +  \frac{\gamma}{N} \sum_\mu \Delta_\mu(t) \g^{\ell+1}_\mu(t) \phi(\h^{\ell}_\mu(t))^\top 
\end{align}
We write the expression this way so that the small time constant $\tau \to 0$ limit corresponds to classic gradient descent. Integrating out the $\Q^\ell(t)$ variable, this gives the following weight dynamics
\begin{align}
    \W^{\ell}(t) = \W^{\ell}(0) + \frac{\gamma}{N \tau} \int_0^t dt' \int_0^{t'} dt'' e^{-(t'-t'') / \tau} \sum_{\mu} \Delta_\mu(t'') \g^{\ell+1}_{\mu}(t'') \phi(\h^{\ell}_\mu(t''))^\top
\end{align}
which implies the following field evolution
\begin{align}
    h^{\ell+1}_\mu(t) &= \chi^{\ell+1}_{\mu}(t) + \frac{\gamma_0}{\tau} \int_0^t dt' \int_0^{t'} dt'' e^{-(t'-t'')/\tau} \sum_{\alpha} \Delta_\alpha(t'') g^{\ell+1}_{\alpha}(t'') \Phi^{\ell}_{\mu\alpha}(t,t'') \nonumber \\
    z^{\ell}_{\mu}(t) &= \xi^\ell_\mu(t) + \frac{\gamma_0}{\tau}  \int_0^t dt' \int_0^{t'} dt'' e^{-(t'-t'')/\tau} \sum_{\alpha} dt'' \Delta_\alpha(t'') \phi(h^{\ell}_\alpha(t'')) G^{\ell+1}_{\mu\alpha}(t,t'')  
\end{align}
We see that in the $\tau \to 0$ limit, the $t''$ integral is dominated by the contribution at $t'' \sim t'$ recovering usual gradient descent dynamics. For $\tau \gg 0$, we see that the integral accumulates additional contributions from the past values of fields and kernels.

\section{Discrete Time}\label{app:discrete_time}

Our model can also be accommodated in discrete time, though we lose the NTK as a key player in the theory (note that $\frac{d}{dt} f_\mu = \frac{df_\mu}{d\theta} \cdot \frac{d\theta}{dt} = \sum_{\alpha} \Delta_\alpha K_{\mu\alpha}^{NTK}$ requires a continuous time limit of the gradient descent dynamics). For a discrete time analysis we let $t \in \mathbb{N}$ and define our network function as
\begin{align}
    f_\mu(t) &= \frac{1}{N\gamma_0} \w^L(t) \cdot \phi(\h^L_\mu(t)) = \frac{1}{N \gamma_0}\left[ \w^L(0) + \gamma_0 \sum_{s=0}^{t-1} \sum_{\alpha} \Delta_\alpha(s) \phi(\h^L_\alpha(s)) \right] \cdot \phi(\h^L_\mu(t)) \nonumber
    \\
    &= \frac{1}{N \gamma_0} \w^L(0) \cdot \phi(\h^L_\mu(t)) + \sum_{\alpha} \sum_{s<t} \Delta_\alpha(s) \Phi^L_{\mu\alpha}(t,s)
\end{align}
We treat $f_\mu(t)$ as a potentially random variable and insert
\begin{align}
    1 = \int \frac{d\hat f_\mu(t) df_\mu(t)}{2\pi N^{-1}} \exp\left( i \hat{f}_\mu(t)\left[ N f_\mu(t) - \frac{1}{\gamma_0}\w^L(0) \cdot \phi(\h^L(t)) - N \sum_{\alpha} \sum_{s<t} \Delta_\alpha(s) \Phi^L_{\mu\alpha}(t,s)  \right] \right)
\end{align}
Noting that $\w^L(0)$ is involved in the definition of both $f_\mu(t)$ and $\bm\xi_\mu^L(t)$, we see that the average over $\w^L(0)$ now takes the form
\begin{align}
    \left< \exp\left( i \sum_{\mu t} [\hat{\bm\xi}^L_{\mu}(t) + \gamma_0^{-1} \hat{f}_\mu(t) \phi(\h^L_\mu(t))] \cdot \w^L(0) \right) \right>_{\w^L(0)} = &\exp\left( -\frac{1}{2} \sum_{\mu t \alpha s} \hat{\bm\xi}^L_{\mu}(t) \cdot \hat{\bm\xi}^L_\alpha(s) \right) \nonumber
    \\
    &\exp\left( - \frac{N}{2 \gamma_0^2 } \sum_{\mu\alpha ts} \hat{f}_\mu(t) \hat{f}_\alpha(s) \Phi^L_{\mu\alpha}(t,s)  \right) \nonumber
    \\
    &\exp\left( - \frac{1}{\gamma_0} \sum_{\mu\alpha ts} \hat{f}_\mu(t) \phi(\h_\mu^L(t)) \cdot \hat{\bm\xi}^L_\alpha(s) \right)
\end{align}
We extend our definition as before $i A^L_{\mu\alpha}(t,s) = \frac{1}{N \gamma_0} \phi(\h^L_\mu(t)) \cdot \bm\xi^L_{\alpha}(s)$. Proceeding with the calculation as usual, we find that 
\begin{align}
    Z &\propto \int df_\mu(t) d\hat{f}_\mu(t) d\Phi^\ell ... dB^\ell \exp\left( N S[\{ f, \hat f, \Phi^\ell , \hat\Phi^\ell, ... , A^\ell,B^\ell \}] \right) \nonumber
    \\
    S &= i \sum_{\mu t} \hat{f}_\mu(t) f_\mu(t) - \frac{1}{2 \gamma_0^2} \sum_{\mu\alpha ts} \hat{f}_\mu(t) \hat{f}_\alpha(s) \Phi^L_{\mu\alpha}(t,s) - i \sum_{\mu\alpha ts} \hat{f}_\mu(t) A^L_{\mu\alpha}(t,s) - i\sum_{\mu t \alpha s} \hat{f}_\mu(t) [ \Theta(t-s) \Delta_\alpha(s) \Phi^L_{\mu\alpha}(t,s) ] \nonumber
    \\
    &+ \sum_{\ell \mu\alpha ts} [ \Phi_{\mu\alpha}^\ell \hat{\Phi}^\ell_{\mu\alpha}(t,s) + G^\ell_{\mu\alpha}(t,s)\hat{G}_{\mu\alpha}(t,s) - A^\ell_{\mu\alpha}(t,s) B^\ell_{\mu\alpha}(t,s) ] \nonumber
    \\
    &+ \ln \mathcal{Z}[\{ \Phi^\ell , \hat\Phi^\ell, ... , A^\ell,B^\ell \}]
\end{align}
The saddle point equations can now be analyzed. In addition to the usual order parameters, we note that $f,\hat{f}$ also generate saddle point equations
\begin{align}
    \frac{\partial S}{\partial f_\mu(t)} &= i \hat{f}_\mu(t) = 0 \nonumber
    \\
    \frac{\partial S}{\partial i\hat{f}_\mu(t)} &= f_\mu(t) + \frac{1}{\gamma_0^2} \sum_{\alpha s} \Phi^L_{\mu\alpha}(t,s) (i\hat{f}_\alpha(s)) - \sum_{\alpha s} A^L_{\mu\alpha}(t,s) - \sum_{\alpha s} \Theta(t-s) \Delta_\alpha(s) \Phi^L_{\mu\alpha}(t,s)
\end{align}
We also obtain saddle point equations for the new $A^L, B^L$ order parameters.
\begin{align}
    \frac{\partial S}{\partial A^L_{\mu\alpha}(t,s)} &= - B^L_{\mu\alpha}(t,s) - i\hat{f}_\mu(t) = 0
    \\
    \frac{\partial S}{\partial B^L_{\mu\alpha}(t,s)} &= - A^L_{\mu\alpha}(t,s) + i \gamma_0^{-1} \left< \phi(h^L_{\mu}(t)) \hat{\xi}^L_\alpha(s) \right> = 0
\end{align}
which implies $B^L_{\mu\alpha}(t,s) = 0$ and $A^L = \gamma_0^{-1} \left< \frac{\phi(h^L_{\mu}(t))}{\partial r^L_\alpha(s)}  \right>$. This gives the following DMFT
\begin{align}
    f_{\mu}(t) &= \sum_{s<t} \sum_\alpha  \Phi^{L}_{\mu\alpha}(t,s) \Delta_\alpha(s) + \sum_{\alpha s} A^L_{\mu\alpha}(t,s) \nonumber
    \\
    \u^\ell &\sim \mathcal{N}(0, \bm\Phi^{\ell-1}) \ , \ \r^\ell \sim \mathcal{N}(0,\G^{\ell+1}) \nonumber
    \\
    h^{\ell}_\mu(t) &= u^\ell_\mu(t) +   \gamma_0 \sum_{\alpha s} [ A^{\ell-1}_{\mu\alpha}(t,s) + \Theta(t-s)\Delta_\alpha(s)  \Phi^{\ell-1}_{\mu\alpha}(t,s) ]g^{\ell}_{\alpha}(s)\nonumber
    \\
    z^\ell_{\mu}(t) &= r^\ell_\mu(t) +  \gamma_0 \sum_{\alpha s} [B^\ell_{\mu\alpha}(t,s) + \Theta(t-s) \Delta_\alpha(s) G^{\ell+1}_{\mu\alpha}(t,s) ]\phi(h^\ell_\alpha(s))\nonumber
    \\
    \Phi^\ell_{\mu\alpha}(t,s) &= \left< \phi(h^\ell_\mu(t)) \phi(h^\ell_\alpha(s))\right> \ , \ G^\ell_{\mu\alpha}(t,s) = \left< g^\ell_\mu(t) g^\ell_{\alpha}(s) \right> \nonumber
    \\
    A^\ell_{\mu\alpha}(t,s) &= \gamma_0^{-1} \left< \frac{\partial \phi(h^\ell_\mu(t))}{\partial u^\ell_{\alpha}(s)} \right> \ , \ B^\ell_{\mu\alpha}(t,s) = \gamma_0^{-1} \left< \frac{\partial g_{\mu}^{\ell+1}(t)}{\partial r^{\ell+1}_\alpha(s)} \right> .
\end{align}
We leave it to future work to verify that a continuous time limit of the above DMFT recovers function evolution governed by the NTK.

\section{Equivalent Parameterizations}\label{app:equiv_par}

In this section, we show the equivalence of our parameterization scheme with many alternatives including the $\mu P$ parameterization of Yang \cite{yang2021tensor}. We also compare the derived stochastic processes obtained with DMFT and Tensor Programs in Appendix Section \ref{app:dmft_tp_equiv}. Following Yang we use a modified variant of abc parameterization (we assume one  which defines the features $\h^{\ell+1} = N^{-a_\ell} \W^{\ell} \phi(\h^\ell)$ with $W^{\ell}_{ij} \sim \mathcal{N}(0,N^{-b_\ell})$ and $\eta = \gamma^2 N^{-c}$). Lastly, we will take $\gamma = \gamma_0 N^d$ and find that only $d = \frac{1}{2}$ will allow feature learning
\begin{align}
    &\h^1 = D^{-a_0} \W^0 \x_\mu \ , \ W_{ij}^0 \sim \mathcal{N}(0, D^{-b_0}) \nonumber
    \\
    &\h^{\ell+1 } = N^{-a_\ell} \W^{\ell} \phi(\h^\ell) \ , \ W^{\ell}_{ij} \sim \mathcal{N}(0, N^{-b}) \nonumber
    \\
    &f = \frac{1}{\gamma} h^{L+1} \ , \ h^{L+1} = N^{-a_L} \w^L \cdot \phi(\h^{L}) \ , w^L_i \sim \mathcal{N}(0,N^{-b}) \nonumber
    \\
    &\gamma = \gamma_0 N^{d} \ , \ \gamma_0 = \mathcal{O}_N(1)
\end{align}

We will now derive constraints on $(a, b, c, d)$ which give desired large width behavior. We will identify a one-dimensional family of parameterizations which satisfy three desiderata of network training 1. finite preactivations, 2. learning in finite time, 3. feature learning. 

\subsection{Predictions Evolve in $\mathcal{O}_N(1)$ time}
As before we let the NTK be the matrix which defines network prediction dynamics $\partial_t f_\mu = \sum_\alpha K^{NTK}_{\mu\alpha} \Delta_\alpha$. We demand that this matrix be $\mathcal O_N(1)$ so that the network predictions have $\mathcal{O}_N(1)$ evolution
\begin{align}
    K^{NTK}_{\mu\alpha} &= \gamma^2 N^{-c} \sum_{\ell} \frac{\partial f_\mu}{\partial \W^\ell} \cdot \frac{\partial f_\alpha}{\partial \W^\ell} \nonumber
    \\
    &= N^{-c}\left[  \frac{\phi(\h_\mu^L) \cdot \phi(\h_\alpha^L)}{N^{2a_L}} + \sum_{\ell} \frac{\partial h^{L+1}_\mu}{\partial \h^{\ell+1}_\mu} \cdot \frac{\partial h^{L+1}_\alpha}{\partial \h^{\ell+1}_\alpha}  \frac{\phi(\h^\ell_\mu) \cdot\phi(\h^\ell_\alpha) }{N^{2a_\ell}} + \frac{\partial h^{L+1}_\mu}{\partial \h^{1}_\mu} \cdot \frac{\partial h^{L+1}_\alpha}{\partial \h^{1}_\alpha} \frac{\x_\mu\cdot\x_\alpha}{D^{2a_0}} \right] \nonumber
    \\
    &= N^{-c} \left[ N^{1-2a_L} \Phi^{L}_{\mu\alpha} + \sum_{\ell} N^{1-2 a_\ell} \frac{\partial h^{L+1}_\mu}{\partial \h^{\ell+1}_\mu} \cdot \frac{\partial h^{L+1}_\alpha}{\partial \h^{\ell+1}_\alpha} \Phi^\ell_{\mu\alpha} + D^{1-2a_0} \frac{\partial h^{L+1}_\mu}{\partial \h^{1}_\mu} \cdot \frac{\partial h^{L+1}_\alpha}{\partial \h^{1}_\alpha} \K^x \right] 
\end{align}
where we used the usual definition of the kernels $\Phi^\ell = \frac{1}{N} \phi(\h^\ell) \cdot \phi(\h^\ell)$ which are $\mathcal{O}_N(1)$ provided each neuron's preactivation $h_i^\ell=\mathcal{O}_N(1)$. We see that the choice $a_\ell = \frac{1}{2}$ recovers the parameterization discussed in Appendix \ref{app:dmft_derivation}. Further to have $\mathcal{O}_N(1)$ evolution of the output predictions $f_\mu$ we need $K^{NTK} = \mathcal{O}_N(1)$.  Now, to enforce the $\mathcal O_{N}(1)$ evolution of predictions we demand
\begin{align}
    N^{1-c - 2 a_\ell} &= \mathcal{O}_N(1) \ , \  \ell \in \{1, ..., L\} \nonumber
    \\
    N^{-c} D^{1-2a_0} &= \mathcal{O}_N(1)
\end{align}
If, on the other hand, we take $D \sim \mathcal{O}_N(N)$, then this simply demands the constraint that $c = 2a_\ell-1$ for all $\ell \in \{0,...,L\}$.  
\subsection{Fields Are $\mathcal{O}_N(1)$}
Having fields which are $\mathcal{O}_N(1)$ can be ensured at initialization provided that
\begin{align}
    \left< h_i^{\ell+1} h_j^{\ell+1} \right> &= N^{-2a_\ell} \sum_{k,k'} \left<W_{ik}^\ell(0) W_{jk'}^\ell(0) \right> \phi(h^\ell_k) \phi(h^\ell_{k'}) \nonumber
    \\
    &= \delta_{ij} N^{1 -2a_\ell - b_\ell} \Phi^{\ell} = \mathcal{O}_N(1)
\end{align}
which implies that $2a_\ell + b_\ell = 1$. Again we see that $a_\ell = \frac{1}{2}, b_\ell = 0$ works, but this is not the only possible scaling. Alternatively standard parameterization $a_\ell = 0, b_\ell = 1$ will also preserve the $\mathcal{O}_N(1)$ scale of the features. We next need to analyze the scale of the feature gradients $\frac{\partial h^{L+1}}{\partial \h^{\ell}}$. We start with the last layer 
\begin{align}
    &\frac{\partial h^{\ell+1}}{\partial \h^L} = N^{-a_L} \w^L \odot \dot\phi(\h^L) \nonumber
    \\
    \implies &\frac{\partial h^{\ell+1}}{\partial \h^L} \cdot \frac{\partial h^{\ell+1}}{\partial \h^L} = \mathcal{O}_N(N^{1-2 a_\ell - b_\ell} )
\end{align}
Since we already demanded that $2a_L + b_L = 1$, this inner product will be $\mathcal{O}_N(1)$. Now we will see whether it remains $\mathcal{O}_N(1)$ under its recursion
\begin{align}
    \frac{\partial h^{L+1}}{\partial \h^{\ell} }= \left(\frac{\partial \h^{\ell+1}}{\partial \h^{\ell} } \right)^\top \frac{\partial h^{L+1}}{\partial \h^{\ell+1}} = \dot\phi(\h^\ell) \odot \left[ N^{-a_\ell} \W^{\ell}(0)^\top \frac{\partial h^{L+1}}{\partial \h^\ell} \right]
\end{align}
Now, letting $\g^\ell = \sqrt{N} \frac{\partial h^{L+1}}{\partial \h^\ell}$ and $\z^\ell = N^{-a_\ell} \W^{\ell}(0)^\top \g^{\ell+1}$ we have
\begin{align}
    \left< z_i z_j \right> = \delta_{ij} N^{1 -2 a_\ell - b_\ell} G^{\ell+1}
\end{align}
which is indeed $\mathcal{O}_N(1)$ as desired provided that $2a + b = 1$. 

\subsection{$\mathcal{O}_N(1)$ Feature Evolution}
Now, we desire that the fields $h_i, z_i$ all evolve by an $\mathcal{O}_N(1)$ amount during network training, which is equivalent to stable feature learning. The update equation for $\W^\ell$ and $\h^\ell$ give
\begin{align}
    \frac{d}{dt} \W^\ell &= \gamma N^{-c-a_\ell} \sum_\mu \Delta_\mu \frac{\partial h^{L+1}}{\partial \h^{\ell+1}_\mu} \bm\phi_\mu^{\ell \top} = \gamma_0  N^{d-c-a_\ell-\frac{1}{2}} \sum_\mu \Delta_\mu \g^{\ell+1}_\mu \bm\phi_\mu^{\ell \top} \nonumber
    \\
    \h^{\ell+1}_\mu(t) &= \bm\chi^{\ell+1}_\mu(t) + \gamma_0 N^{d-c-2a_\ell+\frac{1}{2}} \sum_\alpha \int_0^t ds \Delta_\alpha(s) \g_\alpha^{\ell+1}(s) \Phi_{\mu\alpha}^{\ell}(t,s)
\end{align}
where we used $\gamma = \gamma_0 N^d$. This equation implies that $d - c -2a_\ell + \frac{1}{2} = 0$ is necessary and sufficient for $\mathcal{O}_N(1)$ feature evolution.

\subsection{Putting Constraints Together}
We now let $\gamma = \gamma_0 N^d$. We see that the set of parameterizations which yield $\mathcal{O}(1)$ feature evolution are those for which 
\begin{enumerate}
    \item Features $h,z$ are $\mathcal{O}_N(1) \implies 2a_\ell + b_\ell = 1$
    \item Outputs predictions evolve in $\mathcal{O}_N(1)$ time $\implies c + 2a_\ell = 1$
    \item Features $h,z $ have $\mathcal{O}_N(1)$ evolution $\implies d = c + 2a_\ell - \frac{1}{2} = \frac{1}{2}$.
\end{enumerate}
We see that the parameterization discussed in Appendix \ref{app:dmft_derivation} satisfies these with $d = \frac{1}{2}, a_\ell = \frac{1}{2}, b_\ell = 0, c=0$.  The quite general requirement for feature learning that $d = \frac{1}{2}$ indicates that $\gamma = \gamma_0 \sqrt{N}$ for any choice of $a_\ell, b_\ell, c$. The set of parameterizations which meet these three requirements is one dimensional with $d = \frac{1}{2}$, and $(a_\ell,b_\ell,c_\ell) \in \{ (a_\ell,1-2a_\ell, 1-2a_\ell) : a_\ell \in \mathbb{R} \}$. However, in the next section, we show that if one demands $\mathcal{O}_N(1)$ learning rate, then the parameterization is unique and is the $\mu P$ parameterization of Yang and Hu \cite{yang2021tensor}.

\subsection{$\mathcal{O}_N(1)$ Learning Rate}
We are also interested in a parameterization for which we can have $\mathcal O(1)$ learning rate which are those for which $\gamma^2 N^{-c} = \mathcal{O}_N(N^{2d-c}) = \mathcal{O}_N(1) \implies c = 2d = 1$. Under this constraint, $a_\ell = 0$ and $b_\ell = 1$, which corresponds to standard parameterization, modified by $\gamma = \gamma_0 \sqrt{N}$ in the last layer. In a computational algorithm, the learning rate would be $\eta = \gamma^2 N^{-c} = \gamma_0^2$. This is equivalent to the $\mu P$ parameterization of Yang and Hu \cite{yang2021tensor}.

\subsection{Equivalence of DMFT at $\gamma_0=1$ and Tensor Programs derived Stochastic Process}\label{app:dmft_tp_equiv}
Now that we have established that the parameterization we consider here (modified NTK parameterization) is equivalent to $\mu P$, (modified standard parameterization), we will now demonstrate that the stochastic process which we obtained through a stationary action principle applied to our DMFT action $S$ is equivalent to the stochastic process derived from the Tensor Programs framework of Yang \cite{yang2019tensor, yang2021tensor}. Using the notation from Appendix H of Yang and Hu \cite{yang2021tensor}, they give the following evolution equations for the preactivations in a hidden layer in one pass SGD
\begin{align}
    Z^{h_t} &= \hat{Z}^{W x_t} + \dot{Z}^{W x_t} - \sum_{s=0}^{t-1} \chi_s Z^{dh_s} \mathbb{E}[ Z^{x_s} Z^{x_t} ]  \nonumber
    \\
    Z^{dx_t} &= \hat{Z}^{W^\top dh_t} + \dot{Z}^{W^\top dh_t} - \sum_{s=0}^{t-1} \chi_s Z^{x_s} \mathbb{E}[ Z^{dh_t} Z^{dh_s} ] 
\end{align}
where $\hat{Z}^{W x_t}$ is mean zero Gaussian variable with covariance $\mathbb{E}[Z^{x_t} Z^{x_s}]$ and $\hat{Z}^{W^\top dh_t}$ is mean zero Gaussian with covariance $\mathbb{E}[Z^{dh_t} Z^{dh_s}]$. We can switch to the notation of this work by making the substitutions $Z^{h_t} \to h(t)$, $\hat{Z}^{W x_t} \to u(t)$, $\chi_s \to - \Delta(s)$,  $\dot{Z}^{Wx} \to \sum_{s} \Delta(s) A(t,s)$ and $\mathbb{E}[ Z^{x_s} Z^{x_t} ] \to \Phi(t,s)$, and so on. A summary of the full set of notational substitutions between this work and TP are summarized in Table \ref{tab:tp_notation_table}.
\begin{table}[h]
    \centering
    \begin{tabular}{|c|c|c|c|c|c|c|c|c|}
    \hline
        DMFT &  $h(t)$ & $\chi(t)$ & $g(t)$ & $\xi(t)$ & $\Phi^\ell(t,s)$ & $G^\ell(t,s)$ & $A^{\ell}(t,s), B^{\ell}(t,s)$ & $\Delta(t)$  \\ \hline
        TP & $Z^{h_t}$ & $Z^{Wx_t}$ & $Z^{dx_t}$ & $Z^{W^\top dh_t}$ & $\mathbb{E}[Z^{x_t} Z^{x_s}]$ & $\mathbb{E}[Z^{dh_t} Z^{dh_s}]$ & $\theta_{ts}$ & $-\chi_t$ \\  \hline
    \end{tabular}
    \caption{A dictionary relating the notation of the Tensor Programs (TP) framework \cite{yang2021tensor} and this work.}
    \label{tab:tp_notation_table}
\end{table}
\\
After these substitutions are made, we see that the equations above match the one-pass SGD version of the DMFT Equations in Appendix \ref{app:discrete_time}. A similar identification can be made for the backward pass. This shows that both Tensor Programs and DMFT, though alternative derivations, give identical descriptions of the stochastic processes induced by random initializations + GD in infinite neural networks.

\section{Gradient Independence}\label{app:grad_indep}

The gradient independence approximation treats the random initial weight matrix $\W^{\ell}(0)$ as a \textit{independently sampled Gaussian matrix} when used in the backward pass. We let this second matrix be $\tilde\W^{\ell}(0)$. As before, we have $\bm\chi^{\ell+1} = \frac{1}{\sqrt N} \W^{\ell}(0) \bm\phi(\h^{\ell})$, however we now define $\bm\xi^{\ell} = \frac{1}{\sqrt N} \tilde\W^{\ell}(0)^\top \g^{\ell+1}$. Now, when computing the moment generating function $Z$, the integrals over $\W^\ell(0)$ and $\tilde\W^\ell(0)$ factorize
\begin{align}
    &\left< \exp\left( \frac{i}{\sqrt N} \int_0^\infty dt \left[ \sum_{\mu} \hat{\bm\chi}^{\ell+1}(t) \W^{\ell}(0) \phi(\h^\ell_\mu(t)) + \g^{\ell+1}_\mu(t)^\top \tilde\W^{\ell}(0) \bm\xi^{\ell}_\mu(t) \right] \right) \right> \nonumber
    \\
    &= \exp\left( - \frac{1}{2} \sum_{\mu\alpha} \int_0^\infty dt' \int_0^\infty ds' \left[ \hat{\bm\chi}^{\ell+1}_\mu(t) \cdot \hat{\bm\chi}^{\ell+1}_\alpha(s) \bm\Phi^{\ell}_{\mu\alpha}(t,s) + \hat{\bm\xi}^\ell_\mu(t) \cdot \hat{\bm\xi}^\ell_\alpha(s) G^{\ell+1}_{\mu\alpha}(t,s) \right] \right).
\end{align}
We see that in this field theory, the fields $\chi, \xi$ are all independent Gaussian processes $\{ \chi^{\ell+1}_{\mu}(t) \} \sim \mathcal{GP}(0, \bm\Phi^\ell)$ and $\{ \xi^{\ell}_\mu(t) \} \sim \mathcal{GP}(0,\G^{\ell+1})$. This corresponds to making the assumption that $\A^\ell = \B^\ell = 0$ so that $\chi = u$ and $\xi = r$ within the full DMFT.

\section{Perturbation Theory}

\subsection{Small $\gamma_0$ Expansion}
In this section we analyze the leading corrections in a small $\gamma_0$ expansion of our DMFT theory. All fields are expanded in power series in $\gamma_0$.
\begin{align}
     h^{\ell}_\mu(t) - u^\ell_{\mu}(t) &= \sum_{n=1}^\infty \gamma_0^n h^{\ell, (n)}_{\mu}(t) \nonumber 
    \\
     z^{\ell}_\mu(t) - r^\ell_\mu(t) &= \sum_{n=1}^\infty \gamma_0^n z^{\ell, (n)}_{\mu}(t)
\end{align}
Our goal is to calculate all corrections to the kernels up to $\mathcal{O}(\gamma_0^3)$ to show that the leading correction is $\mathcal{O}(\gamma_0^2)$ and the subleading correction is $\mathcal{O}(\gamma_0^4)$. It will again be convenient to utilize the vector notation defined in \ref{app:dmft_derivation}.

We note that unlike other works on perturbation theory in wide networks, we do not attempt to characterize fluctuation effects in the kernels due to finite width, but rather operate in a regime where the kernels are concentrating and their variance is negligible. For a more thorough discussion of perturbative field theory in finite width networks, see \cite{ roberts2021principles, hanin2022correlation, dyer2019asymptotics}.

\subsubsection{Linear Network}

The kernels in deep linear networks can be expanded in powers of $\gamma_0^2$ giving a leading order correction of size $\mathcal O(\gamma_0^2)$ and can be computed explicitly from the closed saddle point equations. 
We use the symmetrizer $\{\X,\Y \}_{sym} = \X \Y + \Y^\top \X^\top$ as shorthand. The leading order behavior of $\C^{\ell} \sim \C^{(0)} + \mathcal O(\gamma_0^2) \ , \ \D^{\ell} \sim \D^{(0)} + \mathcal O(\gamma_0^2), \H^{\ell,0} = \H^{(0)} = \K^x \otimes \bm 1 \bm 1^\top, \G^{\ell,(0)} = \G^{(0)} = \bm 1\bm1^\top$ is independent of layer index so we find the following leading order corrections
\begin{align}
\H^\ell &\sim \H^{(0)} + \ell \gamma_0^2 \left( \{ \C^{(0)} \D^{(0)}, \H^{(0)} \}_{sym} + \C^{(0)} \bm 1 \bm 1^\top \C^{(0)\top} \right) +  \mathcal{O}(\gamma_0^4)  \nonumber
    \\
    \G^\ell &\sim \bm 1 \bm 1^\top + (L+1-\ell) \gamma_0^2 ( \{ \D^{(0)} \C^{(0)} , \bm 1 \bm 1^\top \}_{sym} + \D^{(0)} \H^{(0)} [\D^{(0)}]^\top ) + \mathcal{O}(\gamma_0^4) \nonumber
    \\
    \K^{NTK} &\sim L \H^0 + \gamma_0^2 \frac{L(L+1)}{2} \left( \{ \C^{(0)} \D^{(0)}, \K^x \}_{sym} + \C^{(0)} \bm 1 \bm 1^\top \C^{(0)\top} \right) \nonumber
    \\
    &+ \gamma_0^2 \frac{L(L+1)}{2}  \K^x \otimes ( \{ \D^{(0)} \C^{(0)} , \bm 1 \bm 1^\top \}_{sym} + \D^{(0)} \H^{(0)} [\D^{(0)}]^\top ) + O(\gamma_0^4)
\end{align}

Note that $[\C^0 \g]_{\mu t} = \int_0^t dt' \sum_{\beta} H^0_{\mu\beta}(t,t') \Delta_\beta(t') g(t') =\sum_\beta K^x_{\mu\beta} \int_0^t dt' \Delta_\beta(t') g(t')$ and note that $[\D \h]_{t} = \int_0^t dt' G^0(t,t') \sum_\alpha \Delta_\alpha(t') h_\alpha(t') = \sum_\alpha \int_0^t dt' \Delta_\alpha(t') h_\alpha(t')$. 

\begin{align}
    H^\ell_{\mu\nu}(t,s) &= K^x_{\mu\nu} \nonumber
    \\
    &+ \ell \gamma_0^2 \sum_{\alpha \beta} K^x_{\mu\alpha} K^x_{\nu\beta} \int_0^t dt' \Delta_\alpha(t') \int_0^{t'} dt'' \Delta_\beta(t'') + ((\mu,t) \leftrightarrow (\nu,s)) \nonumber \\
    &+ \ell \gamma_0^2 \sum_{\alpha\beta} K^x_{\mu\alpha} K^x_{\nu\beta} \left[ \int_0^t dt' \Delta_\alpha(t') \right]\left[ \int_0^s ds' \Delta_\beta(s') \right] \nonumber
    \\
    G^{\ell}(t,s) &= 1 + \gamma_0^2 (L+1-\ell) \sum_{\alpha\beta} K^x_{\alpha\beta} \int_0^t dt' \Delta_\alpha(t') \int_0^{t'} dt'' \Delta_\beta(t'') + ( t \leftrightarrow s) \nonumber
    \\
    &+ \gamma_0^2 (L+1-\ell) \sum_{\alpha\beta} K^x_{\mu\alpha} \left[\int_0^t dt' \Delta_\alpha(t') \right] \left[\int_0^s ds' \Delta_\alpha(s') \right]
\end{align}

We can simplify the notation by introducing functions $v_{\alpha}(t) = \int_0^t \Delta_\alpha(t')$ and $v_{\alpha\beta}(t) = \int_0^t dt' \Delta_\alpha(t') \int_0^{t'} dt'' \Delta_\beta(t'')$. 
\begin{align}
    H^\ell_{\mu\nu}(t,s) &= K^x_{\mu\nu} + \ell \gamma_0^2 \sum_{\alpha \beta} K^x_{\mu\alpha} K^x_{\nu\beta} [ v_{\alpha\beta}(t) + v_{\beta\alpha}(s)] + \ell \gamma_0^2 \sum_{\alpha\beta} K^x_{\mu\alpha} K^x_{\nu\beta} v_{\alpha}(t) v_\beta(s) \nonumber
    \\
    G^{\ell}(t,s) &= 1 + \gamma_0^2 (L+1-\ell) \sum_{\alpha\beta} K^x_{\alpha\beta} [v_{\alpha\beta}(t) + v_{\beta\alpha}(s) + v_{\alpha}(t) v_{\beta}(s)]
\end{align}
Using the fact that
\begin{align}
    K^{NTK}_{\mu\alpha}(t,s) &= \sum_{\ell=0}^L G^{\ell+1}(t,s) H^{\ell}_{\mu\alpha}(t,s) \nonumber
    \\
    &\sim (L+1) K^x_{\mu\alpha} + \gamma_0^2  \sum_{\ell=1}^L H^{\ell,2}_{\mu\alpha}(t,s) + \gamma_0^2 \sum_{\ell=1}^L G^{\ell,2}(t,s) K^x_{\mu\alpha}  + \mathcal{O}(\gamma_0^4)
\end{align}
and utilizing the identity $\sum_{\ell=1}^L \ell = \frac{1}{2} L(L+1)$, we recover the result provided in the main text.

\subsection{Nonlinear Perturbation Theory}\label{app:pert_nonlin}

We start with the formula which implicitly defines $\h,\z$
\begin{align}
    \h^{\ell} &= \u^\ell + \gamma_0 \C^{\ell} [\dot\phi(\h^\ell) \odot \z^\ell] \ , \ \z^\ell = \r^\ell + \gamma_0 \D^\ell \phi(\h^\ell)
\end{align}
We proceed under the assumption of a power series in $\gamma_0$ 
\begin{align}
    \h^\ell - \u^\ell &= \gamma_0  \h^{\ell,1} + \gamma_0^2 \h^{\ell,2} + ... \nonumber
    \\  
    \z^\ell - \r^\ell &= \gamma_0  \z^{\ell,1} + \gamma_0^2 \z^{\ell,2} + ... \nonumber
    \\
    \bm\Phi^\ell -\bm\Phi^{\ell,0} &=  \gamma_0 \bm\Phi^{\ell,1} + \gamma_0^2 \bm\Phi^{\ell,2} + ... \nonumber
    \\
    \G^\ell -\G^{\ell,0}  &=  \gamma_0 \G^{\ell,1} + \gamma_0^2 \G^{\ell,2} + ... \nonumber
    \\
    \C^\ell -\C^{\ell,0}  &= \gamma_0 \C^{\ell,1} + \gamma_0^2 \C^{\ell,2} + ... \nonumber
    \\
    \D^\ell -\D^{\ell,0} &= \gamma_0 \D^{\ell,1} + \gamma_0^2 \D^{\ell,2} + ...
\end{align}
Expanding both sides of the implicit equation for $\z^\ell$ we have
\begin{align}
    \gamma_0  \z^{\ell,1} + \gamma_0^2 \z^{\ell,2} + ... =& \gamma_0 \D^{\ell,0} \phi(\u^\ell) \nonumber \\
    &+ \gamma_0^2 \left[ \D^{\ell,0} \dot\phi(\u) \odot \h^{\ell,1} + \D^{\ell,1} \phi(\u)  \right] \nonumber
    \\
    &+ \gamma_0^3 \left[ \D^{\ell,0} \dot\phi(\u) \odot \h^{\ell,2} + \D^{\ell,0} \ddot\phi(\u)\odot [\h^{\ell,1}]^2 + \D^{\ell,1} \dot\phi(\u) \odot \h^{\ell,1}  + \D^{\ell,2} \phi(\u) \right] \nonumber
    \\
    &+\mathcal{O}(\gamma_0^4)
\end{align}
Performing a similar exercise for $\h^\ell$, we get the following first three leading terms for $\z^\ell, \h^\ell$, we find
\begin{align}
    &\z^{\ell,1} = \D^{\ell,0} \phi(\u)  \nonumber
    \\
    &\z^{\ell,2} =  \D^{\ell,0} \dot\phi(\u) \odot \h^{\ell,1} + \D^{\ell,1} \phi(\u) \nonumber
    \\
    &\z^{\ell,3} = \D^{\ell,0}\left[ \frac{1}{2} \ddot\phi(\u) \odot [\h^{\ell,1}]^2 + \dot\phi(\u) \odot \h^{\ell,2} \right] + \D^{\ell,1}[\dot\phi(\u) \odot \h^{\ell,1}] + \D^{\ell,2} \phi(\u) \nonumber
    \\
    &\h^{\ell,1} =   \C^{\ell,0} \g^{\ell,0} = \C^{\ell,0} [ \dot\phi(\u) \odot \r] \nonumber
    \\
    &\h^{\ell,2} = \C^{\ell,1} \g^{\ell,1} + \C^{\ell,0} \g^{\ell,2} \nonumber
    \\
    &= \C^{\ell,0} \left[\dot\phi(\u) \z^{\ell,1} + \ddot\phi(\u) \h^{\ell,1} \r \right] + \C^{\ell,1} \left[\dot\phi(\u) \z^{\ell,2} + \ddot\phi(\u) \h^{\ell,1} \z^{\ell,1} + \frac{1}{2} \dddot\phi(\u) [\h^{\ell,1}]^2 \r + \ddot\phi(\u) \h^{\ell,2} \r \right] \nonumber
    \\
    &\h^{\ell,3} = \C^{\ell,0} \g^{\ell,2} + \C^{\ell,1} \g^{\ell,1} + \C^{\ell,2}
    \g^{\ell,0} \nonumber
    \\
    &= \C^{\ell,0} \left[ \dot\phi(\u) \z^{\ell,2} + \ddot\phi(\u) \h^{\ell,1} \z^{\ell,1} + \ddot\phi(\u) \h^{\ell,2} \r + \frac{1}{2} \dddot\phi(\u) [\h^{\ell,1}]^2 \r  \right] \nonumber
    \\
    &+ \C^{\ell,1}\left[ \dot\phi(\u) \z^{\ell,1} + \ddot\phi(\u) \h^{\ell,1} \r \right] + \C^{\ell,2} \left[ \dot\phi(\u) \r \right]
\end{align}
As will become apparent soon, it is crucially important to identify the dependence of each of these terms on $\r$. We note that $z^{\ell,1}$ does not depend on $r$ and $h^{\ell,1}$ is linear in $r$. In the next section, we use this fact to show that $\Phi^{\ell,1}=0$ and $G^{\ell,1} = 0$. These conditions imply that $C^{\ell,0}$ and $D^{\ell,1} = 0$. As a consequence, $z^{\ell,2}$ is linear in $r$ and $h^{\ell,2}$ only contains even powers of $r$. Lastly, this implies that $z^{\ell,3}$ only contains even powers of $r$ and $h^{\ell,3}$ contains only odd powers of $r$. 

\subsubsection{Leading Corrections to $\Phi^{1}$ Kernel is $\mathcal{O}(\gamma_0^2)$}
We start in the first layer where $\u^1 \sim \mathcal{GP}(0,\bm K^x \otimes \bm 1 \bm 1^\top)$ (note that this is $\mathcal{O}_{\gamma_0}(1)$) and compute the expansion of $\Phi^{1}$ in $\gamma_0$
\begin{align}
    \bm\Phi^{1} =& \left< \phi(\h^1) \phi(\h^1)^\top \right> = \left< \phi(\u^1) \phi(\u^1)^\top \right> \nonumber 
    \\
    &+\gamma_0 \left< \left[ \dot\phi(\u^1) \h^{1,1} \right] \phi(\u^1)^\top \right> + \gamma_0 \left< \phi(\u^1) \left[\dot\phi(\u^1) \h^{1,1}\right]^\top \right> \nonumber \\
    &+ \gamma_0^2 \left<  \left[ \dot\phi(\u^1) \h^{1,1} \right] \left[\dot\phi(\u^1) \h^{1,1}\right]^\top \right> \nonumber
    \\
    &+ \frac{\gamma_0^2}{2} \left<  \left[ \ddot\phi(\u^1) \h^{1,2} \right] \phi(\u^1) \right> + \frac{\gamma_0^2}{2} \left<  \left[ \ddot\phi(\u^1) \h^{1,2} \right] \phi(\u^1) \right> \nonumber
    \\
    &+ \gamma_0^3 \left< \left[\dot\phi(\u^1) \h^{1,3} + \ddot\phi(\u) \h^{1,1} \h^{1,2} + \frac{1}{6} \dddot\phi(\u) (\h^{1,1})^3 \right] \phi(\u)^\top \right> \nonumber
    \\
    &+ \gamma_0^3 \left< \phi(\u) \left[\dot\phi(\u^1) \h^{1,3} + \ddot\phi(\u) \h^{1,1} \h^{1,2} + \frac{1}{6} \dddot\phi(\u) (\h^{1,1})^3\right]^\top \right> \nonumber
    \\
    &+ \gamma_0^3 \left< \left[ \dot\phi(\u) \h^{1,2} + \frac{1}{2} \ddot\phi(\u) (\h^{1,1})^2 \right] \left[\dot\phi(\u) \h^{1,1} \right]^\top \right> \nonumber \\
    &+ \gamma_0^3 \left<  \left[\dot\phi(\u) \h^{1,1} \right] \left[ \dot\phi(\u) \h^{1,2} + \frac{1}{2} \ddot\phi(\u) (\h^{1,1})^2 \right]^\top \right> \nonumber
    \\
    &+ \mathcal{O}(\gamma_0^4)
\end{align}
where powers and multiplications of vectors are taken elementwise. Now, note that, as promised, the terms linear in $\gamma_0$ vanish since $\h^{1,1}$ is linear the Gaussian random variable $\r^1$, which is a mean zero and independent of $\u^1$ so an average like $\left< \r^{1} F(\u^{1}) \right> =\left< \r^{1,0} \right>  \left<F(\u^{1}) \right> = 0$ must vanish for any function $F$. Thus we see that $\bm\Phi^{\ell}$'s leading correction is $\mathcal{O}(\gamma_0^2)$. 

We also obtain, by a similar argument, that the cubic $\mathcal{O}(\gamma_0^3)$ term vanishes. To see this, note that $\h^{1,3}$ only contains odd powers of $\r^1$. Next, $\h^{1,1} \h^{1,2}$ contains only odd powers of $\r$, and $(\h^{1,1})^3$ is cubic in $\r$. Since all odd moments of a mean-zero Gaussian vanish, all averages of these terms over $\r$ annihilate, causing the $\gamma_0^3$ terms to vanish. Thus $\bm\Phi^{1} = \bm\Phi^{1,0} + \gamma_0^2 \bm\Phi^{1,2} + \mathcal{O}(\gamma_0^4)$.

\subsection{Forward Pass Induction for $\Phi^\ell$}

We now assume the inductive hypothesis that for some $\ell \in \{1,...,L-1\}$ that
\begin{align}
    \bm\Phi^{\ell} = \bm\Phi^{\ell,0} + \gamma_0^2 \bm\Phi^{\ell,2} + \mathcal{O}(\gamma_0^4) 
\end{align}
and we will show that this will imply that the next layer must have a similar expansion $\bm\Phi^{\ell+1} = \bm\Phi^{\ell+1,0} + \gamma_0^2 \bm\Phi^{\ell+1,2} + \mathcal{O}(\gamma_0^4)$. First, we note that $\u^{\ell+1} \sim \mathcal{GP}(0, \bm\Phi^{\ell,0} + \gamma_0^2 \bm\Phi^{\ell,2} + ...)$. As before, we compute the leading terms in the expansion of $\bm\Phi^{\ell+1}$ 
\begin{align}
    \bm\Phi^{\ell+1} =& \left< \phi(\h^{\ell+1}) \phi(\h^{\ell+1})^\top \right> \nonumber
    \\
    =& \left< \phi(\u^{\ell+1}) \phi(\u^{\ell+1}) \right> + \gamma_0^2 \left<  \left[ \dot\phi(\u^{\ell+1}) \h^{\ell+1,1} \right] \left[\dot\phi(\u^{\ell+1}) \h^{\ell+1,1}\right]^\top \right> \nonumber
    \\
    &+ \frac{\gamma_0^2}{2} \left<  \left[ \ddot\phi(\u^{\ell+1}) \h^{\ell+1,2} \right] \phi(\u^{\ell+1})^\top \right> + \frac{\gamma_0^2}{2} \left<\phi(\u^{\ell+1})  \left[ \ddot\phi(\u^{\ell+1}) \h^{\ell+1,2} \right]^\top  \right> + \mathcal{O}(\gamma_0^4) 
\end{align}
where, as before the $\gamma_0$ and $\gamma_0^3$ terms vanish by the fact that odd moments of $\r^{\ell+1}$ vanish. Now, note that all averages are performed over $\u^{\ell+1} \sim \mathcal{GP}(0,\bm\Phi^{\ell,0} + \gamma_0^2 \bm\Phi^{\ell,2} + ...)$, which depends on the perturbed kernel of the previous layer. How can we calculate the contribution of the correction which is due to the previous layer's kernel movement? This can be obtained easily from the following identity. Let $F(\u,\r)$ be an arbitrary observable which depends on Gaussian fields $\u$ and $\r$ which have covariances $\bm\Phi^{\ell,0} + \gamma_0^2 \bm\Phi^{\ell,2} + \mathcal{O}(\gamma_0^4)$ and $\bm G^{\ell+2,0} + \gamma_0^2 \bm G^{\ell+2,2} + \mathcal{O}(\gamma_0^3)$ (note this only requires that the linear in $\gamma_0$ terms of $G$ vanish which is easy to verify). Then 
\begin{align}
    \left< F(\u,\r) \right>_{\u,\r} &= \int d\k d\u d\v d\r F(\u,\r)  \exp\left( - \frac{1}{2} \k^\top [\bm\Phi^{\ell,0} + \gamma_0^2 \bm\Phi^{\ell,2}+...] \k + i \k\cdot \u  \right) \nonumber
    \\
    &\exp\left(- \frac{1}{2} \v^\top [\G^{\ell+2,0} + \gamma_0^2 \G^{\ell+2,2}+...] \v + i \v \cdot \r \right)
    \\
    &\sim \left< F(\u,\r) \right>_{\u_0 \r_0} \nonumber
    \\
    &+ \frac{\gamma_0^2}{2} \text{Tr} \left[ \bm\Phi^{\ell-1,2} \left< \frac{\partial^2}{\partial \u \partial \u^\top} f(\u,\r) \right>_{\u_0 \r_0} \right] \nonumber
    \\
    &+ \frac{\gamma_0^2}{2} \text{Tr} \left[ \G^{\ell+1,2} \left< \frac{\partial^2}{\partial \r \partial \r^\top} f(\u,\r) \right>_{\u_0 \r_0} \right] + \mathcal{O}(\gamma_0^3)
\end{align}
where $\u_0 \sim \mathcal{GP}(0,\bm\Phi^{\ell,0}) ,  \r_0 \sim \mathcal{N}(0,\G^{\ell+2,0})$. Thus, the leading order behavior of $\bm\Phi^{\ell+1}$ can easily be obtained in terms of averages over the original unperturbed covariances
\begin{align}
    \bm\Phi^{\ell+1} &= \left< \phi(\u_0) \phi(\u_0)^\top \right>_{\u_0} + \frac{\gamma_0^2}{2} \text{Tr}\left[ \bm\Phi^{\ell,2} \left< \frac{\partial^2}{\partial \u_0 \partial \u_0^\top} \phi(\u_0) \phi(\u_0)^\top \right>_{\u_0} \right] \nonumber
    \\
    &+ \frac{\gamma_0^2}{2} \frac{\partial^2}{\partial \gamma_0^2}|_{\gamma_0 =0} \left< \phi(\h(\u_0,\r_0,\gamma_0)) \phi(\h(\u_0,\r_0,\gamma_0)) \right>_{\u_0, \r_0} + \mathcal{O}(\gamma_0^4) ,
\end{align}
where the trace is taken against the Hessian indices and the indices on $\bm\Phi^{\ell,2}$. This gives us the desired result by induction that for all $\ell \in \{1,...,L\}$, we have $\bm\Phi^{\ell} = \bm\Phi^{\ell,0} + \gamma_0^2 \bm\Phi^{\ell,2} + \mathcal{O}(\gamma_0^4)$. We see that $\Phi^\ell$ accumulates corrections from the previous layers' corrections through the forward pass recursion.

\subsection{Leading Corrections to $G^L$ Kernel is $\mathcal{O}(\gamma_0^2)$}
The analogous argument for $\G^{L}$ now can be provided. First note that $\r^L$ is independent of $\u^L$ and of $\gamma_0$. Thus we can find that $\G^{L}$ has no linear-in-$\gamma_0$ term in its expansion since
\begin{align}
    \G^{L,1} &= \left< [\dot\phi(\u^L) \r^L] \left[ \dot\phi(\u^L) \z^{L,1} + \ddot\phi(\u^L) \h^{L,1} \r^L \right] \right> +  \left< [\dot\phi(\u^L) \r^L] \left[ \dot\phi(\u^L) \z^{L,1} + \ddot\phi(\u^L) \h^{L,1} \r^L \right] \right> =0 \nonumber
\end{align}
each term contains only odd powers of $\r^L$ and odd moments of Gaussian variables vanish. After much more work, one can verify that $\G^{L,3}$ also must vanish since all terms contain odd powers of $\r$.
\begin{align}
    \G^{L,3} =& \left< \g^{L,3} \g^{L,0 \top} \right> +\left< \g^{L,0} \g^{L,3 \top} \right> + \left< \g^{L,2} \g^{L,1 \top} \right> + \left< \g^{L,1} \g^{L,2\top} \right>
\end{align}
First, note that $\g^{L,0}$ is linear in $\r$. Next, note that $\g^{L,1}$ only depends on even powers of $\r$ since $\g^{L,1} = \dot\phi(\u) \z^{L,1} + \ddot\phi(\u) \h^{L,1} \r$. Next, we have
\begin{align}
    \g^{L,2} = \dot\phi(\u) \z^{L,2} + \ddot\phi(\u)[\h^{L,2} \r + \h^{L,1} \z^{L,1}] + \frac{1}{2} \dddot\phi(\u) [\h^{L,1}]^2 
\end{align}
which only depends on odd powers of $\r$. Lastly, we have $\g^{L,3}$ 
\begin{align}
    \g^{L, 3} &= \dot\phi(\u) \z^{L,3} + \ddot\phi(\u)[ \h^{L,3} \r + \h^{L,2} \z^{L,1} + \h^{L,1} \z^{L,2} ] \nonumber
    \\
    &+\frac{1}{2} \dddot\phi(\u)[ 2 \h^{L,1} \h^{L,2} \r + [\h^{L,1}]^2 \z^{L,1} ]
    + \frac{1}{6} \phi^{(4)}(\u) [\h^{L,1}]^3 \r 
\end{align}
which we see only contains even powers of $\r$. Thus $\g^{L,3} \g^{L,0}$ will be odd in $\r$. Looking at the expansion for $\G^{L,3}$, we see that all terms are odd in $\r$ and so the averages vanish under the Gaussian integrals.

\subsection{Backward Pass Recursion for $G^\ell$}

We can derive a similar recursion on the backward pass for $\G^{\ell}$'s leading order corrections. Using the same idea from the previous section, we find the following expressions
\begin{align}
    \G^{\ell} =& \left< \left[ \dot\phi(\u_0) \r_0 \right] \left[ \dot\phi(\u_0) \r_0 \right]^\top \right>_{\u_0, \r_0} + \frac{\gamma_0^2}{2} \left< \dot\phi(\u_0) \dot\phi(\u_0) \right>_{\u_0} \odot \G^{\ell+1,2} \nonumber
    \\
    &+ \frac{\gamma_0^2}{2} \frac{\partial^2}{\partial \gamma_0^2 }|_{\gamma_0 = 0} \left< \left[ \dot\phi(\h(\u_0,\r_0,\gamma_0)) \r_0 \right] \left[ \dot\phi(\h(\u_0,\r_0,\gamma_0)) \r_0 \right]^\top \right>_{\u_0, \r_0} + \mathcal{O}(\gamma_0^4) \nonumber
\end{align}
This time, we see that $\G^\ell$ accumulates corrections from succeeding layers through the backward pass recursion.

\subsection{Form of the Leading Corrections}\label{app:pert_leading_corr}

We can expand the $\h^{\ell}$ and $\z^\ell$ fields around $\u^{\ell,0},\r^{\ell,0}$ to find the leading order corrections to each feature kernel
\begin{align}
    \bm\Phi^{\ell,2} =& \frac{1}{2} \frac{\partial^2}{\partial\gamma_0^2}|_{\gamma_0= 0} \left< \phi(\h^\ell(\u_0,\r_0,\gamma_0)) \phi(\h^\ell(\u_0,\r_0,\gamma_0))^\top \right>_{\u_0,\r_0 } \nonumber
    \\
    &+ \frac{1}{2} \text{Tr} \left[ \bm\Phi^{\ell-1,2} \left< \frac{\partial^2}{\partial \u_0 \partial \u_0^\top} \left[\phi(\u_0) \phi(\u_0)^\top\right] \right>_{\u_0 } \right] 
\end{align}
The first term requires additional expansion to extract the corrections in $\gamma_0^2$
\begin{align}
    \phi(\u + \gamma_0 \C^\ell \g^\ell) &\sim \phi(\u) + \gamma_0 \dot\phi(\u) \odot [\C^\ell \g^\ell] + \frac{\gamma_0^2}{2} \ddot\phi(\u) \odot [ \C^\ell \g^\ell ]^2 \nonumber
    \\
    &\sim \phi(\u) + \gamma_0 \dot\phi(\u) \odot [\C^{\ell,0} \g^{\ell,0}] + \gamma_0^2 \dot\phi(\u) \odot [\C^{\ell,0} \g^{\ell,1}] + \frac{\gamma_0^2}{2} \ddot\phi(\u) \odot [ \C^{\ell,0} \g^{\ell,0} ]^2 \nonumber
    \\
    \dot\phi(\h^\ell)\odot \z^\ell &\sim \dot\phi(\u)\odot \r + \gamma_0 \ddot\phi(\u) \odot [\C^{\ell,0}  \g^{\ell,0}] \odot \r + \gamma_0 \dot\phi(\u) \odot [\D^{\ell,0} \phi(\u)] + \mathcal{O}(\gamma_0^2) \nonumber
    \\
    C^{\ell,0}_{\mu\alpha}(t,s) &= A^{\ell-1,1}_{\mu\alpha}(t,s) + \Theta(t-s) \Delta_\alpha^0(s) \Phi^{\ell-1,0}_{\mu\alpha}(t,s) \nonumber 
    \\
    D^{\ell,0}_{\mu\alpha}(t,s) &= B^{\ell,1}_{\mu\alpha}(t,s) + \Theta(t-s) \Delta_\alpha^0(s) \Phi^{\ell-1,0}_{\mu\alpha}(t,s)
\end{align}
where we used the fact that $\C^{\ell,1} = 0$ which follows from the fact that $\Phi^{\ell-1,1} = 0$, and $\Delta^{\ell,1} = 0$. Now, expanding out term by term
\begin{align}
\bm\Phi^{\ell} =& \bm\Phi^{\ell,0} + \gamma_0^2 \left< [\dot\phi(\u)\odot (\C^{\ell,0} \g^{\ell,0}) ] [\dot\phi(\u)\odot (\C^{\ell,0} \g^{\ell,0}) ]^\top \right> \nonumber
    \\
    &+\gamma_0^2 \left< \left[\dot\phi(\u) \odot (\C^{\ell,0} [\ddot\phi(\u) \odot [\C^{\ell,0} \g^{\ell,0}] \odot \r]) \right]  \phi(\u)^\top \right> + \text{transpose}\nonumber
    \\
    &+\gamma_0^2 \left< \left[\dot\phi(\u) \odot (\C^{\ell,0} [\dot\phi(\u) \odot [\D^{\ell,0} \phi(\u)]]) \right]  \phi(\u)^\top \right> + \text{transpose}\nonumber
    \\
    &+\frac{\gamma_0^2}{2} \left< \left[ \ddot\phi(\u) \odot [ \C^{\ell,0} \g^{\ell,0} ]^2 \right] \phi(\u)^\top \right> + \text{transpose} \nonumber
    \\
    &+ \frac{\gamma_0^2}{2} \text{Tr} \left[ \bm\Phi^{\ell-1,2} \left< \frac{\partial^2}{\partial \u \partial \u^\top} \left[\phi(\u) \phi(\u)^\top\right] \right>_{\u \sim \mathcal{GP}(0,\bm\Phi^{\ell-1,0}) } \right] + \mathcal{O}(\gamma_0^4)
\end{align}
We see that the corrections for the $\Phi^\ell$ kernels accumulate on the forward pass through the final term so $\Phi^{\ell,2} \sim \mathcal{O}(\ell)$. Now we will perform the same analysis for $\G^{\ell}$.
\begin{align}
    \G^{\ell} =& \left< \g^\ell(\u,\r) \g^\ell(\u,\r)^\top \right>_{\u \sim \mathcal{GP}(0,\bm\Phi^{\ell-1,0}) \r\sim\mathcal{GP}(0,\bm\G^{\ell+1,0}) } \nonumber
    \\
    &+ \frac{\gamma_0^2}{2} \text{Tr} \left[ \G^{\ell+1,2} \left< \frac{\partial^2}{\partial \r \partial \r^\top} \left[(\dot\phi(\u) \odot \r) (\dot\phi(\u)\odot \r)^\top \right] \right>_{\u \sim \mathcal{GP}(0,\bm\Phi^{\ell-1,0}) \r\sim\mathcal{GP}(0,\bm\G^{\ell+1,0})} \right] + \mathcal{O}(\gamma_0^4) \nonumber
    \\
    =& \left< \g^\ell(\u,\r) \g^\ell(\u,\r)^\top \right>_{\u \sim \mathcal{GP}(0,\bm\Phi^{\ell-1,0}) \r\sim\mathcal{GP}(0,\bm\G^{\ell+1,0}) } \nonumber
    \\
    &+ \frac{\gamma_0^2}{2}  \G^{\ell+1,2} \odot \left< \dot\phi(\u) \dot\phi(\u) \right>_{\u \sim \mathcal{GP}(0,\bm\Phi^{\ell-1,0})} + \mathcal{O}(\gamma_0^4)
\end{align}
We see that, through the second term, the $\G^\ell$ kernels accumulate on the backward pass so that $\G^{\ell,2} \sim \mathcal{O}(L+1-\ell)$.
As before the difficult term is the first expression which requires a full expansion of $\g^{\ell}$ to second order
\begin{align}
    \g^{\ell} \sim& \dot\phi(\u) \odot \r + \gamma_0 \dot\phi(\u) \odot [\D^{\ell,0} \phi(\u) + \gamma_0 \D^{\ell,0} \dot\phi(\u) \C^{\ell,0} \g^{\ell,0} ] \nonumber
    \\
    &+ \gamma_0 \ddot\phi(\u) [\C^{\ell,0} \g^{\ell,0} + \gamma_0 \C^{\ell,0} \g^{\ell,1}] \odot \r 
\end{align}
From these terms we find
\begin{align}
    \G^{\ell} =& \G^{\ell,0} + \gamma_0^2 \left<[\dot\phi(\u) \odot (\D^{\ell,0} \phi(\u))]  [\dot\phi(\u) \odot (\D^{\ell,0} \phi(\u))]^\top \right>\nonumber
    \\
    &+\gamma_0^2 \left< [\ddot\phi(\u) (\C^{\ell,0} \g^{\ell,0})] [\ddot\phi(\u) (\C^{\ell,0} \g^{\ell,0})]^\top \right> \nonumber
    \\
    &+\gamma_0^2 \left< \left[\dot\phi(\u) \odot \left(\D^{\ell,0} \dot\phi(\u) \C^{\ell,0} \g^{\ell,0}\right)\right] \g^{\ell,0} \right> + \text{transpose} \nonumber
    \\
    &+\gamma_0^2 \left<  [ \ddot\phi(\u) \odot \C^{\ell,0} (\ddot\phi(\u)\odot \C^{\ell,0}\g^{\ell,0})] \g^{\ell,0} \right> + \text{transpose} \nonumber
    \\
    &+ \frac{\gamma_0^2}{2}  \G^{\ell+1,2} \odot \left< \dot\phi(\u) \dot\phi(\u) \right>_{\u \sim \mathcal{GP}(0,\bm\Phi^{\ell-1,0})} + \mathcal{O}(\gamma_0^4)
\end{align}
Now the correction to the NTK has the form
\begin{align}
    \K^{NTK,2} = \bm\Phi^{L,2} + \sum_{\ell=1}^{L-1} \G^{\ell,0} \bm\Phi^{\ell,2} + \sum_{\ell=1}^{L-1} \G^{\ell,2} \bm\Phi^{\ell,0} + \G^{1,2} \odot ( \K^x \otimes \bm 1 \bm 1^\top) 
\end{align}
Since each $\Phi^{\ell,2}, G^{L+1-\ell,2} \sim \mathcal{O}(\ell)$, each of the two sums from $\ell \in \{1,...,L-1\}$ gives a depth scaling of the form $\sim \sum_{\ell=1}^{L-1} \ell = \frac{L(L-1)}{2}$. Since the original NTK has scale $\K^{NTK,0} \sim \mathcal{O}(L)$, the relative change in the kernel is $\frac{|\K^2|}{|\K^0|} = \mathcal{O}(\gamma_0^2 L)$. In a finite width $N$, network, our definition $\gamma = \gamma_0 \sqrt{N}$ would indicate that a width $N$ network would have corrections of scale $\gamma_0^2 L = \frac{\gamma^2 L }{N }$ in the NTK regime where $\gamma = \mathcal O_N(1)$ provided the network is sufficiently wide to disregard initialization dependent fluctuations in the kernels. 

\subsection{Perturbation Theory in Width $N$ (Finite Size Corrections)}\label{app:perturb_N_dmft}

Finite size corrections to the DMFT can also be obtained within our field theoretic framework. Let $\bm k = \text{Vec}\{ \bm \Phi^\ell, \hat{\bm{\Phi}}^{\ell}, \bm G^\ell, \hat{\G}^\ell, \bm A^\ell, \bm B^\ell \}$ denote the collection of kernel order parameters of the DMFT. To simplify the subsequent discussion, we redefine the DMFT action to be its negation $S \to - S$. The DMFT action $S[\bm k]$ defines a Gibbs measure over order parameters $\bm k$, where observables $O(\bm k)$ have averages which can be computed as
\begin{align}
    \left< O(\bm k) \right> = \frac{\int d\bm k \exp\left( - N S[\bm k] \right) O(\bm k) }{\int d\bm k \exp\left( - N S[\bm k] \right)}
\end{align}

The infinite-width DMFT is characterized by the set of saddle point equations which are $\nabla_{\bm k} S[\bm k]|_{\bm k = \bm k^*} = 0$. Let the saddle point be $\bm k^*$. To identify corrections to the observable average $\left< O(\bm k) \right>$ due to finite size, we now Taylor expand $S$ around $\bm k^*$
\begin{align}
    S[\bm k] =& S[\bm k^*] + \frac{1}{2} (\bm k - \bm k^*) \nabla^2_{\bm k} S[\bm k]|_{\bm k = \bm k^*} (\bm k - \bm k^*) \nonumber
    \\
    &+ \frac{1}{6} \sum_{ijl} (k_i - k_i^*)  (k_j - k_j^*)  (k_l - k_l^*) \frac{\partial^3 S}{\partial k_i \partial k_j \partial k_l} + ...
\end{align}
The linear component vanishes at the saddle point since $\nabla_{\bm k} S[\bm k]|_{\bm k = \bm k^*} = 0$. Our observable average is thus
\begin{align}
    \left< O(\bm k) \right> = \frac{\int d\bm k \exp\left( - \frac{N}{2} (\bm k-\bm k^*) \nabla^2 S[\bm k^*] (\bm k-\bm k^*) + ... \right) O(\bm k) }{\int d\bm k \exp\left( - \frac{N}{2} (\bm k-\bm k^*) \nabla^2 S[\bm k^*] (\bm k-\bm k^*) + ... \right)}
    \\
    = \frac{\int d\bm \delta \exp\left( - \frac{1}{2} \bm\delta \nabla^2 S[\bm k^*] \bm\delta - U(\bm\delta) \right) O(\bm k^* + N^{-1/2} \bm\delta) }{\int d\bm \delta \exp\left( - \frac{1}{2} \bm\delta \nabla^2 S[\bm k^*] \bm\delta - U(\bm\delta) \right)}
\end{align}
where we made the change of variables $\bm\delta = \sqrt{N}(\k - \k^*)$. The function $U(\bm\delta)$ contains all higher order terms (cubic and higher) in the Taylor expansion of $N S[\bm k]$. Since the leading power in $U$ is cubic in $(\bm k- \bm k^*) = N^{-1/2} \bm\delta$, the leading behavior of this remainder is $U = \mathcal{O}(N^{-1/2})$ so it can be regarded as a perturbation to the Gibbs distribution. Taylor expanding the exponential $\exp\left( - \frac{1}{2} \bm\delta \nabla^2 S[\bm k^*] \bm\delta - U(\bm\delta) \right) = \exp\left(  - \frac{1}{2} \bm\delta \nabla^2 S[\bm k^*] \bm\delta  \right)[ 1 - U + \frac{1}{2} U^2 + ...]$ in both numerator and denominator, we eliminate the presence of the higher order terms in the Gibbs measure. Lastly, we let $\left< \cdot \right>_0$ represent an average over the unperturbed Gaussian potential $\bm\delta \sim \mathcal{N}(0,[\nabla^2 S[\bm k^*]]^{-1})$. For notational simplicity, we let $\epsilon = N^{-1/2}$ and obtain
\begin{align}
    \left< O(\bm k) \right> &= \frac{\left< O(\bm k^* + \epsilon \bm\delta) \right>_0 - \left< O(\bm k^* + \epsilon \bm\delta) U(\bm\delta) \right> +\frac{1}{2} \left< O(\bm k^* + \epsilon \bm\delta) U(\bm\delta)^2 \right>_0 +... }{1 - \left< U(\bm\delta) \right>_0 + \frac{1}{2} \left< U(\bm\delta)^2 \right>_0  + ... } \nonumber
    \\
    &= \left< O(\bm k^* + \epsilon \bm\delta) \right>_0 - \left[\left< O(\bm k^* + \epsilon \bm\delta) U(\bm\delta)  \right> - \left< O(\bm k^* + \epsilon \bm\delta)\right> \left< U(\bm\delta)  \right> \right]  \nonumber
    \\
    &+ \frac{1}{2} \left[ \left< O(\bm k^* + \epsilon \bm\delta) U(\bm\delta)^2  \right> - \left< O(\bm k^* + \epsilon \bm\delta)\right> \left< U(\bm\delta)^2   \right> \right] \nonumber
    \\
    &- \left[\left< O(\bm k^* + \epsilon \bm\delta) U(\bm\delta)  \right> \left< U(\bm\delta) \right>_0 - \left< O(\bm k^* + \epsilon \bm\delta)\right> \left< U(\bm\delta)  \right>_0^2 \right] \nonumber
    \\
    &+ ... \nonumber
    \\
    &= \sum_{n=0}^{\infty} \frac{(-1)^n}{n!} \left< O(\bm k^* + \epsilon \bm\delta) U(\bm \delta)^n \right>_0^c
\end{align}
where $\left< \right>_0^c$ represents a \textit{connected cumulant} \cite{kardar2007statistical}. The first two connected correlations have the form
\begin{align}
    \left< O U \right>_0^{c} &= \left< O U \right>_0- \left< O \right>_0 \left< U \right>_0 \nonumber
    \\
    \left< O U^2 \right>_0^{c} &= \left< O U^2 \right>_0 - 2\left< O U \right>_0 \left< U\right>_0  - \left< O \right>_0 \left< U^2 \right>_0 + 2 \left< O \right>_0 \left< U \right>_0^2
\end{align}
If one is interested only in the leading order correction to the observable $\left< O(\bm k) \right>$, this can be obtained with the following correction
\begin{align}
    \left< O(\bm k) \right> = O(\bm k^*) +\frac{1}{2 N} \text{Tr} \left[\nabla^2 S[\bm k^*] \right]^{-1} \nabla_{\bm k}^2 O(\bm k^*) + \frac{1}{\sqrt N} \nabla_{\bm k}O(\bm k^*) \cdot \left< \bm\delta U(\bm\delta) \right>_0 + \mathcal{O}(N^{-2})
\end{align}
Since $U = \mathcal{O}(N^{-1/2})$ both corrections are of order $1/N$. This analysis shows that the leading order correction of the kernel distributions is $\mathcal{O}(N^{-1})$ and can be approximated by performing averages over a Gaussian distribution for $\bm k$ determined by the saddle point solution $\bm k^*$ and covariance given by $\frac{1}{N} \left[\nabla^2_{\bm k} S[\bm k]|_{\bm k=\bm k^*} \right]^{-1}$. We derive expressions for the components of this Hessian in \ref{app:action_hessian}. These fluctuations have standard deviation $\mathcal{O}(N^{-1/2})$. This technique is a common approach to identifying finite size effects \cite{Helias_2020} and was recently employed in Bayesian inference setting for networks in the lazy regime \cite{segadlo_nngp_field}. 

Before computing the Hessian terms, we can compare finite size effects under NTK scaling $\gamma = \mathcal{O}_N(1)$, and the mean field scaling $\gamma = \mathcal{O}(\sqrt{N})$. Concretely, we are interested in the feature learning component of the kernel change which is $\mathcal{O}( \frac{\gamma^2 }{N})$. Let $\left< \Delta \bm k \right>$ represent the change in the kernel through training, which we showed in \ref{app:pert_nonlin} is of size $\mathcal{O}(\gamma_0^2) = \mathcal{O}\left( \frac{\gamma^2}{N} \right)$. We will now define the signal to noise ratio of feature learning as
\begin{align}
    \text{SNR} = \frac{\left< \Delta \bm k \right>}{\sqrt{\text{Var}(\bm k)}} = \mathcal{O}\left( \frac{\gamma^2}{\sqrt{N}} \right)
\end{align}
For NTK regime, this is vanishing as $N \to \infty$, while for the DMFT regime, this is goes as $\mathcal{O}(\sqrt N)$ since kernel evolution is always $\mathcal{O}(1)$ but variance is $\mathcal{O}(N^{-1})$.

\subsubsection{DMFT Action Hessian}\label{app:action_hessian}
We now compute the various blocks of the Hessian of the DMFT action $-\nabla_{\bm k}^2 S[\bm k^*]$. The various necessary derivatives are given below. We will utilize the the factorization of the single site MGF $\sum_\ell \ln \mathcal{Z}_\ell$ to simplify many of the expressions. Below we provide exhaustive expressions for each of the relevant terms. These expressions are included for completeness but we did not yet attempt computing all of them as we did the saddle point equations which define $\bm k^*$. First, we list the collection of Hessian terms which do not involve $A,B$ below.
\begin{align}
    -\frac{\partial^2 S}{\partial \hat{\Phi}^\ell_{\mu\nu}(t,s) \partial \hat \Phi^{\ell'}_{\alpha\beta}(t',s') } &= \delta_{\ell\ell'} \left[ \left< \phi(h^\ell_\mu(t)) \phi(h^\ell_\nu(s)) \phi(h^{\ell}_\alpha(t')) \phi(h^{\ell}_\beta(s')) \right> - \Phi^{\ell}_{\mu\nu}(t,s) \Phi_{\alpha\beta}^\ell(t',s') \right]\nonumber
    \\
     -\frac{\partial^2 S}{\partial {\Phi}^\ell_{\mu\nu}(t,s) \partial \Phi^{\ell'}_{\alpha\beta}(t',s') } &= 0 \nonumber
     \\
     -\frac{\partial^2 S}{\partial \hat\Phi^{\ell}_{\mu\nu}(t,s) \partial \Phi^{\ell'}_{\alpha\beta}(t',s')} &= - \delta_{\ell \ell'} \delta_{\mu\alpha} \delta_{\nu\beta} \delta(t-t')\delta(s-s') + \frac{\partial}{\partial \Phi^{\ell'}_{\alpha\beta}(t',s')} \Phi^{\ell}_{\mu\nu}(t,s) \nonumber
     \\
     -\frac{\partial^2 S}{\partial \hat{G}^\ell_{\mu\nu}(t,s) \partial \hat {G}^{\ell'}_{\alpha\beta}(t',s') } &= \delta_{\ell,\ell'} \left[ \left< g^\ell_\mu(t) g^\ell_\nu(s) g^\ell_\alpha(t') g^\ell_\beta(s') \right> - G^{\ell}_{\mu\nu}(t,s) G_{\alpha\beta}^\ell(t',s') \right] \nonumber
     \\
     -\frac{\partial^2 S}{\partial {G}^\ell_{\mu\nu}(t,s) \partial G^{\ell'}_{\alpha\beta}(t',s') } &= 0 \nonumber
     \\
     -\frac{\partial^2 S}{\partial \hat{G}^{\ell}_{\mu\nu}(t,s) \partial G^{\ell'}_{\alpha\beta}(t',s')} &= - \delta_{\ell \ell'} \delta_{\mu\alpha} \delta_{\nu\beta} \delta(t-t')\delta(s-s') + \frac{\partial}{\partial G^{\ell'}_{\alpha\beta}(t',s')} G^{\ell}_{\mu\nu}(t,s) \nonumber
     \\
     -\frac{\partial^2 S}{\partial \hat{\Phi}^\ell_{\mu\nu}(t,s) \partial \hat G^{\ell'}_{\alpha\beta}(t',s') } &= \delta_{\ell,\ell'}\left[\left< \phi(h^\ell_\mu(t)) \phi(h^\ell_\nu(s)) g^{\ell}_\alpha(t') g^{\ell}_\beta(s') \right> - \Phi^{\ell}_{\mu\nu}(t,s) G^\ell_{\alpha\beta}(t',s')\right] \nonumber 
     \\
     - \frac{\partial^2 S}{\partial \hat{\Phi}^\ell_{\mu\nu}(t,s) \partial G^{\ell}_{\alpha\beta}(t',s') } &= \frac{\partial}{ \partial G^{\ell}_{\alpha\beta}(t',s')} \Phi^{\ell}_{\mu\nu}(t,s) \nonumber
     \\
     - \frac{\partial^2 S}{\partial \hat{G}^\ell_{\mu\nu}(t,s) \partial \Phi^{\ell}_{\alpha\beta}(t',s') } &= \frac{\partial }{\partial \Phi^{\ell}_{\alpha\beta}(t',s') } G^\ell_{\mu\nu}(t,s) \nonumber
     \\
     - \frac{\partial^2 S}{\partial G^\ell_{\mu\nu}(t,s) \partial \Phi^{\ell}_{\alpha\beta}(t',s') } &= 0 \nonumber
\end{align}
Now for the terms involving $A^\ell, B^\ell$, we find the following expressions.
\begin{align}
     -\frac{\partial^2 S}{\partial A^{\ell}_{\mu\nu}(t,s) \partial A^{\ell'}_{\alpha\beta}(t',s')} &= \left< \frac{\partial^2 }{\partial u^{\ell+1}_{\nu}(s) \partial u^{\ell'+1}_\beta(s')} [ g^{\ell+1}_{\mu}(t) g^{\ell'+1}_\alpha(t') ]\right> - B^{\ell}_{\mu\nu}(t,s) B^{\ell}_{\alpha\beta}(t',s')\nonumber \\
     -\frac{\partial^2 S}{\partial B^{\ell}_{\mu\nu}(t,s) \partial B^{\ell'}_{\alpha\beta}(t',s') } &= \left< \frac{\partial^2 }{\partial r^{\ell}_{\nu}(s) \partial r^{\ell'}_\beta(s')} [ \phi(h^{\ell}_{\mu}(t)) \phi(h^{\ell'}_\alpha(t')) ]\right> \nonumber
     \\
      -\frac{\partial^2 S}{\partial A^{\ell}_{\mu\nu}(t,s) \partial B^{\ell'}_{\alpha\beta}(t',s')} &= \delta_{\ell\ell'}\delta_{\mu\alpha}\delta_{\nu\beta}\delta(t-t')\delta(s-s') \nonumber
      \\
      &+ \left< \frac{\partial^2}{\partial u_\nu^{\ell+1}(s) \partial r^{\ell'}_\beta(s')} [g^{\ell+1}_{\mu}(t) \phi(h^{\ell'}_{\alpha}(s))] \right> - B^{\ell}_{\mu\nu}(t,s) A^{\ell'}_{\alpha\beta}(t',s') \nonumber
      \\
      -\frac{\partial^2 S}{\partial A^{\ell}_{\mu\nu}(t,s) \partial \Phi^{\ell'}_{\alpha\beta}(t',s')} &= \delta_{\ell,\ell' }\left< \frac{\partial^3}{\partial u^{\ell+1}_\nu(t) \partial u^{\ell+1}_\alpha(t') \partial u^{\ell+1}_\beta(s')} g^{\ell+1}_\mu(t) \right> \nonumber
      \\
      -\frac{\partial^2 S}{\partial A^{\ell}_{\mu\nu}(t,s) \partial \hat \Phi^{\ell'}_{\alpha\beta}(t',s')} &= \delta_{\ell+1,\ell'}\left[ \left< \frac{\partial}{\partial u^{\ell+1}_\nu(t)} [ g^{\ell+1}_\mu(t) \phi(h^{\ell+1}_\alpha(t')) \phi(h^{\ell+1}_\beta(s')) ] \right> - B^{\ell}_{\mu\nu}(t,s) \Phi^{\ell+1}_{\alpha\beta}(t',s') \right] \nonumber
      \\
      -\frac{\partial^2 S}{\partial A^{\ell}_{\mu\nu}(t,s) \partial G^{\ell'}_{\alpha\beta}(t',s')} &= \delta_{\ell+2,\ell'}\left< \frac{\partial^3}{\partial u^{\ell+1}_\nu(t) \partial r^{\ell+1}_\alpha(t') \partial r^{\ell+1}_\beta(s')} g^{\ell+1}_\mu(t) \right> \nonumber
      \\
      -\frac{\partial^2 S}{\partial A^{\ell}_{\mu\nu}(t,s) \partial \hat G^{\ell'}_{\alpha\beta}(t',s')} &= \delta_{\ell+1,\ell'}\left[ \left< \frac{\partial}{\partial u^{\ell+1}_\nu(t)} [ g^{\ell+1}_\mu(t) g^{\ell+1}_\alpha(t') g^{\ell+1}_\beta(s') ] \right> - B^{\ell}_{\mu\nu}(t,s) G^{\ell+1}_{\alpha\beta}(t',s') \right] \nonumber
      \\
       -\frac{\partial^2 S}{\partial B^{\ell}_{\mu\nu}(t,s) \partial \Phi^{\ell'}_{\alpha\beta}(t',s')} &= \delta_{\ell-1,\ell'} \left< \frac{\partial^3}{\partial r^{\ell}_\nu(t) \partial u^{\ell}_\alpha(t') \partial u^{\ell}_\beta(s')} \phi(h^{\ell}_\mu(t)) \right> \nonumber
      \\
      -\frac{\partial^2 S}{\partial B^{\ell}_{\mu\nu}(t,s) \partial \hat \Phi^{\ell'}_{\alpha\beta}(t',s')} &= \delta_{\ell,\ell'}\left[ \left< \frac{\partial}{\partial r^{\ell}_\nu(t)} [ \phi(h^{\ell}_\mu(t)) \phi(h^{\ell}_\alpha(t')) \phi(h^{\ell}_\beta(s')) ] \right> - A^{\ell}_{\mu\nu}(t,s) \Phi^{\ell}(t',s') \right] \nonumber
      \\
      -\frac{\partial^2 S}{\partial B^{\ell}_{\mu\nu}(t,s) \partial G^{\ell'}_{\alpha\beta}(t',s')} &= \delta_{\ell+1,\ell'}\left< \frac{\partial^3}{\partial r^{\ell}_\nu(t) \partial r^{\ell}_\alpha(t') \partial r^{\ell}_\beta(s')} \phi(h^{\ell}_\mu(t)) \right> \nonumber
      \\
      -\frac{\partial^2 S}{\partial 
      B^{\ell}_{\mu\nu}(t,s) \partial \hat G^{\ell'}_{\alpha\beta}(t',s')} &= \delta_{\ell,\ell'}\left[ \left< \frac{\partial}{\partial r^{\ell}_\nu(t)} [ \phi(h^{\ell}_\mu(t)) g^{\ell}_\alpha(t') g^{\ell}_\beta(s') ] \right> - A^{\ell}_{\mu\nu}(t,s) G_{\alpha\beta}^{\ell}(t',s') \right] \nonumber
\end{align}
From these block matrices which comprise the Hessian, we can obtain the finite width covariance structure in our order parameters $\bm k = \text{Vec}\{ \Phi^{\ell} ,G^{\ell} , A^\ell,B^\ell \}$ by computing the inverse $\bm C = \left( -\nabla^2_{\k} S[\k^*] \right)^{-1}$. In this approximation scheme, we have $\bm \k \sim \mathcal{N}(\k^*, \frac{1}{N}\left( -\nabla^2_{\k} S[\k^*] \right)^{-1})$. Many questions about this expansion remain including: can it be proven that these $\mathcal{O}(N^{-1/2})$ fluctuations always lead to higher expected test loss?

\end{document}